%% file: thesis.tex
\documentclass[12pt]{report}   

\usepackage[letterpaper, left=1in, right=1in, top=1in, bottom=1in]{geometry}
\usepackage{setspace}  
\usepackage{times}  
\usepackage[explicit]{titlesec}  
\usepackage[titles]{tocloft}  
\usepackage[utf8]{inputenc} 
\usepackage[backend=bibtex, sorting=nyt, style=authoryear, maxcitenames=2, maxbibnames=500]{biblatex}  
\usepackage{appendix}  
\usepackage{rotating}  
\usepackage[normalem]{ulem}  
\usepackage{textcomp} 
\usepackage{indentfirst} 
\usepackage{booktabs,array} 
\usepackage{amsmath} 
\usepackage[T1]{fontenc} 
\usepackage[hidelinks]{hyperref}
\usepackage{cleveref}
\usepackage{multirow}
\usepackage{subcaption}
\usepackage{microtype}
\usepackage{amssymb}
\usepackage{xspace}
\usepackage{latexsym}
\usepackage{caption}
\usepackage{graphics}
\usepackage{epsfig}
\usepackage{wrapfig}
\usepackage{xcolor}
\definecolor{Green}{rgb}{0.0, 0.5, 0.0}
\definecolor{Amethyst}{rgb}{0.6, 0.4, 0.8}
\definecolor{Green}{rgb}{0.0, 0.5, 0.0}
\definecolor{amethyst}{rgb}{0.6, 0.4, 0.8}
\definecolor{cadet}{rgb}{0.33, 0.41, 0.47}
\definecolor{airforceblue}{rgb}{0.36, 0.54, 0.66}
\definecolor{cadmiumgreen}{rgb}{0.0, 0.42, 0.24}
\usepackage{colortbl}
\usepackage{color}
\usepackage{bm,array}
\usepackage{tablefootnote}
\usepackage[usestackEOL]{stackengine}
\usepackage{pifont}

\usepackage{graphics}
\usepackage{epsfig}
\usepackage{times}
\usepackage{amsmath,bm,amssymb}

\usepackage{float}
\usepackage{multicol}
\usepackage{multirow}
\usepackage{graphicx}
\usepackage{makecell}
\usepackage{fancyvrb,fvextra}
\usepackage{url}
\usepackage{verbatim}
\usepackage{colortbl}
\usepackage{times}
\usepackage{csquotes}
\usepackage{braket}

\definecolor{green}{rgb}{0,0.6,0}

\bibliography{references,anthology}

\AtEveryBibitem{\clearfield{issn}}
\AtEveryBibitem{\clearlist{issn}}

\AtEveryBibitem{\clearfield{language}}
\AtEveryBibitem{\clearlist{language}}

\AtEveryBibitem{\clearfield{doi}}
\AtEveryBibitem{\clearlist{doi}}

\AtEveryBibitem{\clearfield{url}}
\AtEveryBibitem{\clearlist{url}}

\AtEveryBibitem{%
  \ifentrytype{online}
    {}
    {\clearfield{urlyear}\clearfield{urlmonth}\clearfield{urlday}}}

\begin{document}
\doublespacing  

\input{titlePage.tex}


\input{CopyrightPage.tex}

\pagenumbering{gobble}
\input{1-abstract.tex}

\pagenumbering{roman}
\setcounter{page}{1} 
\renewcommand{\cftchapdotsep}{\cftdotsep}  
\renewcommand{\cftchapfont}{\normalfont}  
\renewcommand{\cftchappagefont}{}  
\renewcommand{\cftchappresnum}{Chapter }
\renewcommand{\cftchapaftersnum}{:}
\renewcommand{\cftchapnumwidth}{5em}
\newcommand{\dataset}{\textsc{clinsum}\xspace}
\renewcommand{\cftchapafterpnum}{\vskip\baselineskip} 
\renewcommand{\cftsecafterpnum}{\vskip\baselineskip}  
\renewcommand{\cftsubsecafterpnum}{\vskip\baselineskip} 
\renewcommand{\cftsubsubsecafterpnum}{\vskip\baselineskip} 

\newcommand{\cmark}{\ding{51}}%
\newcommand{\xmark}{\ding{55}}%


\renewcommand\contentsname{Table of Contents}

\begin{singlespace}
\tableofcontents
\setlength{\cftparskip}{\baselineskip}
\listoffigures
\listoftables
\end{singlespace}

\clearpage

\phantomsection
\addcontentsline{toc}{chapter}{Acknowledgments}
\input{acknowledgements.tex}



\clearpage
\pagenumbering{arabic}
\setcounter{page}{1}

\titleformat{\chapter}[display]
{\normalfont\bfseries\filcenter}{}{0pt}{\large\chaptertitlename\ \large\thechapter : \large\bfseries\filcenter{#1}}  
\titlespacing*{\chapter}
  {0pt}{0pt}{30pt}	
  
\titleformat{\section}{\normalfont\bfseries}{\thesection}{1em}{#1}

\titleformat{\subsection}{\normalfont}{\thesubsection}{0em}{\hspace{1em}#1}



\input{2-introduction.tex}

\input{3-dataset.tex}


\input{4-modeling-faithfulness.tex}


\input{5-measuring-faithfulness.tex}

\input{6-llms.tex}


\clearpage
\phantomsection
\addcontentsline{toc}{chapter}{Conclusion or Epilogue}
\input{7-conclusion.tex}

\clearpage
\phantomsection 
\titleformat{\chapter}[display]
{\normalfont\bfseries\filcenter}{}{0pt}{\large\bfseries\filcenter{#1}}  
\titlespacing*{\chapter}
  {0pt}{0pt}{30pt}

\begin{singlespace}  
    \setlength\bibitemsep{\baselineskip}  
    \addcontentsline{toc}{chapter}{References}  
\end{singlespace}


\titleformat{\chapter}[display]
{\normalfont\bfseries\filcenter}{}{0pt}{\large\chaptertitlename\ \large\thechapter : \large\bfseries\filcenter{#1}}  
\titlespacing*{\chapter}
  {0pt}{0pt}{30pt}	
  
\titleformat{\section}{\normalfont\bfseries}{\thesection}{1em}{#1}

\titleformat{\subsection}{\normalfont}{\thesubsection}{0em}{\hspace{1em}#1}

\titleformat{\subsubsection}{\normalfont}{\thesubsubsection}{0em}{\hspace{1em}#1}

\input{appendix.tex}

\end{document}

%% file: titlePage.tex

\begin{titlepage}
\begin{center}

\begin{singlespacing}
\vspace*{6\baselineskip}

Generating Faithful and Complete \\
Hospital-Course Summaries \\
from the Electronic Health Record

\vspace{3\baselineskip}
Griffin Adams\\
\vspace{18\baselineskip}
Submitted in partial fulfillment of the\\
requirements for the degree of\\
Doctor of Philosophy\\
under the Executive Committee\\
of the Graduate School of Arts and Sciences\\
\vspace{3\baselineskip}
COLUMBIA UNIVERSITY\\
\vspace{3\baselineskip}
\the\year
\vfill

\end{singlespacing}

\end{center}
\end{titlepage}

%% file: CopyrightPage.tex
\begin{titlepage}
\begin{singlespacing}
\begin{center}

\vspace*{35\baselineskip}

\textcopyright  \,  \the\year\\
\vspace{\baselineskip}	
Griffin Adams\\
\vspace{\baselineskip}	
All Rights Reserved
\end{center}
\vfill

\end{singlespacing}
\end{titlepage}

%% file: 1-abstract.tex

\begin{titlepage}
\begin{center}

\vspace*{5\baselineskip}
\textbf{\large Abstract}

Generating Faithful and Complete \\
Hospital-Course Summaries \\
from the Electronic Health Record

Griffin Adams

\end{center}
\begin{flushleft}
\hspace{10mm}

The rapid adoption of Electronic Health Records (EHRs)--electronic versions of a patient's medical history--has been instrumental in streamlining administrative tasks, increasing transparency, and enabling continuity of care across providers. An unintended consequence of the increased documentation burden, however, has been reduced face-time with patients and, concomitantly, a dramatic rise in clinician burnout. Time spent maintaining and making sense of a patient's electronic record is a leading cause of burnout. In this thesis, we pinpoint a particularly time-intensive, yet critical, documentation task: generating a summary of a patient's hospital admissions, and propose and evaluate automated solutions. In particular, we focus on faithfulness, i.e., accurately representing the patient record, and completeness, i.e., representing the full context, as the sine qua non for safe deployment of a hospital-course summarization tool in a clinical setting.

\vspace{5mm}

The bulk of this thesis is broken up into four chapters: \textbf{\S \ref{ch:dataset} Creating and Analyzing the Data}, \textbf{\S \ref{ch:improving} Improving the Faithfulness of Summaries}, \textbf{\S \ref{ch:measuring} Measuring the Faithfulness of Summaries}, and, finally, \textbf{\S \ref{ch:llm} Generating Grounded, Complete Summaries with LLMs}. Each chapter links back to the core themes of faithfulness and completeness, while the chapters are linked to each other in that the findings from each chapter shape the direction of subsequent chapters.

\vspace{5mm}

Given the documentation authored throughout a patient's hospitalization, hospital-course summarization requires generating a lengthy paragraph that tells the story of the patient admission. In \textbf{\S \ref{ch:dataset}}, we construct a dataset based on 109,000 hospitalizations (2M source notes) and perform exploratory analyses to motivate future work on modeling and evaluation \parencite{adams-etal-2021-whats} [NAACL 2021]. The presence of highly abstractive, entity dense references, coupled with the high stakes nature of text generation in a clinical setting, motivates us to focus on faithfulness and adequate coverage of salient medical entities.

\vspace{5mm}

In \textbf{\S \ref{ch:improving}}, we address faithfulness from a modeling perspective by revising noisy references ~\parencite{adams-etal-2022-learning} [EMNLP 2022] and, to reduce the reliance on references, directly calibrating model outputs to metrics ~\parencite{adams-etal-2023-desired} [ACL 2023].

\vspace{5mm}

These works relied heavily on automatic metrics as human annotations were limited. To fill this gap, in \textbf{\S \ref{ch:measuring}}, we conduct a fine-grained expert annotation of system errors in order to meta-evaluate existing metrics and better understand task-specific issues of domain adaptation and source-summary alignments. We find that automatically generated summaries can exhibit many errors, including incorrect claims and critical omissions, despite being highly extractive. These errors are missed by existing metrics. To learn a metric which is less correlated to extractiveness (copy-and-paste), we derive noisy faithfulness labels from an ensemble of existing metrics ~\parencite{adams2023meta} and train a faithfulness classifier on these pseudo labels [MLHC 2023].

\vspace{5mm}

Finally, in \textbf{\S \ref{ch:llm}}, we demonstrate that fine-tuned LLMs (Mistral \parencite{jiang2023mistral} and Zephyr \parencite{tunstall2023zephyr}) are highly prone to entity hallucinations and cover fewer salient entities. We improve both coverage and faithfulness by performing sentence-level entity planning based on a set of pre-computed salient entities from the source text, which extends our work on entity-guided news summarization (\parencite{adams-etal-2023-generating} [ACL, 2023], \parencite{cod} [EMNLP, 2023].

\end{flushleft}
\vspace*{\fill}
\end{titlepage}

%% file: acknowledgements.tex

\clearpage
\begin{center}

\vspace*{5\baselineskip}
\textbf{\large Acknowledgements}
\end{center} 

\begin{flushleft}
\hspace{10mm}
I would first like to thank my thesis committee: No\'emie Elhadad, Kathleen McKeown, Smaranda Muresan, Jason Zucker, and Byron Wallace. I am indebted to each one of you, and your unwavering support has made the long PhD journey exciting, energizing, and ever entertaining.

I would like to specially single out my advisor No\'emie Elhadad. Every interaction I've shared with you--from the moment I first entered your office on PH20 thinking I would solve AI for healthcare in 4 years to now, just happy that faithfulness metrics are improving, has left an indelible mark on me as a student and as a person. You've faced every up and down in my early research career with steadiness, humor, actionable insights, compassion, and most importantly, a lot of laughs. I could not have done this without you and I hope I wasn't too difficult myself! I would also like to extend a special thank you to Jason, whose clinical insights and laser quick annotation efforts has been an invaluable resource to the projects described in this thesis.

I would also also like to thank my wonderful wife Caroline, patient parents Joe and Hilary, sensational sisters Natalie and Genevieve, and precocious nephews Ollie and Leo. I have always been incredibly close to my family and have leaned on them a great deal. It took a little extra leaning during the PhD and, for that, I am undeniably lucky to have them and I am eternally grateful. I would be remiss without including our beloved dog Remy who would have preferred I spent less time coding and more time playing with him these past 4 years. Your patience, or lack thereof, is duly noted!

I would also like to extend a thank you to Alex Fabbri and Faisal Ladhak for being invaluable friends and collaborators throughout my PhD.

\end{flushleft}
\clearpage


%% file: 2-introduction.tex
\chapter{Introduction and Background}

The electronic health record (EHR) contains critical information for clinicians to assess a patient's medical history (e.g., conditions, laboratory tests, procedures, treatments) and healthcare interactions (e.g., primary care and specialist visits, emergency department visits, and hospitalizations). While medications, labs, and diagnoses are documented through structured data elements and flowsheets, clinical notes contain rich narratives describing the patient's medical condition and interventions. A single hospital visit for a patient with a lengthy hospital stay, or complex illness, can consist of hundreds of notes. At the point of care, clinicians already pressed for time, face a steep challenge of making sense of their patient's documentation and synthesizing it either for their own decision making process or to ensure coordination of care~\parencite{hall2004information,ash2004some}. In fact, a leading factor for clinician burnout is the EHR, the information overload it produces, and the documentation burden it requires \parencite{shanafelt2016relationship, moy2021measurement}. A study of US physicians revealed that doctors spent 27\% of working hours with patients and nearly 50\% of their time on EHR and desk work, in addition to 1-2 hours at night, spent mostly on documentation \parencite{sinsky2016allocation}. Clinician burnout can have damaging consequences not only for clinicians \parencite{national2019taking}, due to, among other factors, increased rates of depression \parencite{maslach2016understanding} and interrupted work-life balance \parencite{kroth2019association}), but also patients, due to an increased risk of errors \parencite{salvagioni2017physical, panagioti2018association}. 

\vspace{5mm}

Automatic summarization has been proposed to support clinicians in multiple scenarios, from making sense of a patient's longitudinal record over long periods of time and multiple interactions with the healthcare system, to synthesizing a specific visit's documentation. In this thesis, we focus on \textit{hospital-course summarization}: faithfully and concisely summarizing the EHR documentation for a patient's specific inpatient visit, from admission to discharge.  A good hospital-course summary, which can be included into a patient's discharge summary, often serves as the primary document reviewed in subsequent encounters with different providers \parencite{kind2011documentation}. Not only crucial for continuity of care and patient safety after discharge~\parencite{kripalani2007deficits,van2002effect}, hospital-course summarization also represents an incredibly challenging multi-document summarization task with diverse knowledge requirements. To properly synthesize an admission, one must not only identify relevant problems, but link them to symptoms, procedures, medications, and observations while adhering to temporal, problem-specific constraints, which is a cognitively difficult task even for trained clinicians. They must review a high number of clinical notes and reports entered during the patient stay and synthesize them into a long paragraph. It is even more challenging when an admission is complex---the case for patients with comorbidities or chronic conditions.

Since we cannot share patient data, to illustrate the information dense nature of the summaries we prompted GPT-4 to generate 2 different Brief Hospital Course summaries. The instruction template was: ``Create a mock Brief Hospital Course summary (a mandatory section of the discharge summary) for \{\{ seed \}\}. The Brief Hospital Course should be a single paragraph and be dense with entities. It should cover why the patient was admitted to the hospital in the first place, the treatments, procedures, and medications administered during the hospital stay, as well as a plan for what needs to happen next.'' We used two different seeds: We prompted with two different seeds: ``an elderly patient with dizzy spells'' and ``an obese patient with diabetes scheduled for a coronary artery bypass graft (CABG)'' to elicit diverse outputs.

\paragraph*{GPT-4 Summary for Patient with Dizzy Spells.} \textit{The patient, an 82-year-old individual with a history of hypertension and type 2 diabetes, was admitted to the hospital presenting with recurrent episodes of dizziness and an episode of syncope. Upon admission, vital signs were stable, but orthostatic hypotension was noted. Initial laboratory tests, including a complete blood count (CBC), comprehensive metabolic panel (CMP), and electrocardiogram (EKG), were conducted, revealing no acute abnormalities. The patient was started on intravenous fluids to address dehydration and was closely monitored for cardiac arrhythmias. A Holter monitor was applied for 24-hour cardiac monitoring, which did not demonstrate significant arrhythmias. A head CT scan ruled out acute intracranial events. The patient's antihypertensive medication regimen was adjusted to manage orthostatic hypotension better. Physical therapy consultations were made to assess fall risk and implement preventive strategies. During the hospital stay, the patient's symptoms improved with the adjustments in medication and hydration therapy. Upon discharge, it was recommended that the patient follow up with their primary care physician within one week, continue with the adjusted antihypertensive medication, and participate in outpatient physical therapy to improve balance and prevent future falls. Additionally, the patient was advised to monitor blood pressure at home and maintain adequate hydration.}

\paragraph{GPT-4 Summary for Obese Patient with Diabetes scheduld for a CABG.} \textit{The patient, a 58-year-old male with a history of obesity and type 2 diabetes mellitus, was admitted to the hospital for elective coronary artery bypass grafting (CABG) due to significant coronary artery disease identified on cardiac catheterization. Upon admission, the patient underwent preoperative evaluations, including echocardiography and laboratory tests, which confirmed stable hemoglobin A1c levels and adequate renal function. The CABG procedure was performed without complications, utilizing off-pump techniques to bypass three major coronary arteries. Postoperatively, the patient was managed in the intensive care unit for 24 hours for close monitoring, followed by transfer to the step-down unit. The patient's postoperative course included initiation of low-dose aspirin, beta-blockers, and statin therapy, alongside adjustments to his insulin regimen to optimize glycemic control. Wound care was managed with standard protocols, and the patient participated in early mobilization and cardiac rehabilitation exercises. Nutrition consultation was provided to address obesity management and diabetic diet planning. Discharge planning emphasized the importance of outpatient follow-up with the cardiology and endocrinology teams within two weeks, continuation of cardiac rehabilitation, and adherence to a heart-healthy diabetic diet. The patient was educated on signs of wound infection, importance of medication adherence, and lifestyle modifications critical for long-term management of his cardiac and diabetic conditions.}

An automatically generated hospital course summary can act as a first draft for a clinician and ensure that the critical elements of the patient stay are not missed in the potentially overwhelmingly large amount of notes produced during the patient stay. Maintaining fidelity to the actual hospital course is both vital and highly nuanced. In this thesis, we propose training modifications (\textbf{\S \ref{ch:improving}}), develop protocols for manually identifying fine-grained errors in order to benchmark metrics (\textbf{\S \ref{ch:measuring}}), and, in \textbf{\S \ref{ch:llm}}, demonstrate the importance of grounded generation for reducing hallucinations when transitioning from smaller encoder-decoder models (e.g., LongFormer) to Large Language Models (LLMs).

\vspace{5mm}

The tendency of abstractive models to produce unfaithful summaries is well-studied in the general domain \parencite{maynez-etal-2020-faithfulness} in terms of identifying faithfulness errors with metrics \parencite{durmus-etal-2020-feqa,fabbri-etal-2022-qafacteval,ladhak-etal-2022-faithful,liu2023g}, developing annotation protocols for human labeling \parencite{pagnoni-etal-2021-understanding,goyal-durrett-2021-annotating,krishna-etal-2023-longeval}, training more faithful models \parencite{cao-wang-2021-cliff, narayan-etal-2021-planning,zhao2023slic}, as well as removing errors in existing summaries \parencite{cao-etal-2020-factual,dong-etal-2020-factcorrect,zhu-etal-2021-enhancing}. The literature on faithfulness of clinical summaries is more limited \parencite{xie2023faithful}, largely focused on developing and meta-evaluating faithfulness metrics for more self-contained, single-document tasks: including radiology report summarization \parencite{zhang-etal-2020-optimizing,xie2023factreranker}, echocardiogram conclusions \parencite{tang-etal-2022-echogen}, and doctor-patient conversations \parencite{ben-abacha-etal-2023-investigation}.

\vspace{5mm}

This thesis seeks to address this gap: to determine whether or not metrics, methods, and annotation protocols designed for single document news documents are equally as effective for long-form, multi-document clinical summarization. Where they are ill suited, or when they fall short, we devise new solutions which are tailored to the idiosyncrasies of the hospital-course task.

\section{Overview of Thesis} \label{sec:overview}

In this thesis, we explore a novel task for clinical NLP: hospital-course summarization. We first create and analyze large-scale dataset based on real-world clinical records from Columbia University Irving Medical Center (CUIMC) (\textbf{Chapter \ref{ch:dataset}}). Then, we dive deeply into the critical issue of faithfulness along three fronts: algorithms to improve faithfulness (\textbf{Chapter \ref{ch:improving}}), annotation protocols for fine-grained identification of errors which are used to meta-evaluate both existing and novel metrics (\textbf{Chapter \ref{ch:measuring}}), and a novel method for training and prompting LLMs to generate grounded, complete summaries (\textbf{Chapter \ref{ch:llm}}).

Most existing datasets for clinical summarization relate to more narrow, single-document tasks than hospital-course summarization. These datasets and tasks, which include radiology report synthesis \parencite{zhang2018learning} and extracting a patient's primary diagnoses from a single progress note \parencite{gao2023overview}, are typically based on an open-source dataset of clinical notes (MIMIC-III). While interesting benchmarks for clinical summarization, they are narrow in scope (often related to a single note type) and, when based on MIMIC, only relate to patients in a single setting: the Intensive Care Unit (ICU). In contrast, as defined and constructed in this thesis, hospital-course summarization is a longituindal summarization task which should be trained and evaluated on any in-patient hospital admissions across all patient populations. In \textbf{Chapter \ref{ch:dataset}}, we formally introduce the task. Given the documentation authored throughout a patient's hospitalization, generate a paragraph that tells the story of the patient admission. We construct an English, text-to-text dataset of 109,000 hospitalizations (2M source notes) and their corresponding summary proxy: the clinician-authored "Brief Hospital Course" paragraph written as part of a discharge note. Exploratory analyses reveal that the BHC paragraphs are highly abstractive with some long extracted fragments; are concise yet comprehensive; differ in style and content organization from the source notes; exhibit minimal lexical cohesion; and represent silver-standard references. Our analysis identifies multiple implications for modeling this complex, multi-document summarization task which inform the subsequent chapters.

The tendency of abstractive systems to produce unfaithful summaries is well-studied \parencite{maynez-etal-2020-faithfulness}, yet less attention is paid to the role of the data on which the models are trained. Given the noisiness of naturally occurring hospital-course reference summaries, in the next Chapter (\textbf{Chapter \ref{ch:improving}}), we directly measure and address the impact of noisy references on the faithfulness of downstream summaries. In \S \ref{sec:ltr}, we provide a pre-processing solution while in \S \ref{sec:calibration-section}, we avoid over-reliance on reference summaries by directly learning from synthetic summaries. Pre-existing work into mitigating the negative impact of reference summary quality on downstream faithfulness of learned models had largely addressed it by filtering out noisy references \parencite{nan-etal-2021-entity}. Our thesis task: hospital-course summarization, does not lend itself to filtering given the pervasive level of noise in our references--almost every reference has some degree of unsupported content. As such, we propose a new approach: to revise--not remove--noisy reference content. At a high-level, we take each supported sentence, corrupt it to form a diverse set of unsupported alternatives, and use this mix of real and synthetic data to create examples of (un)faithful revisions for contrastive learning. Yet, training on synthetically edited references has its limitations: we found that the summaries were not very coherent and did not always follow a logical arc. To deal with the reference issue without touching references, in \S \ref{sec:calibration-section}, we learn to directly calibrate model likelihoods to prioritize the generation of ``faithful`` summaries over ``unfaithful'' summaries. We propose novel contributions to calibration by focusing not on novel algorithms, but focusing on the makeup and characteristics of the offline candidate sets which are the inputs to any calibration, or alignment, algorithm.

\textbf{Chapter \ref{ch:improving}} relies on high-quality metrics for faithfulness either for the purposes of identifying references for revision (\S \ref{sec:ltr}) or for providing "AI feedback", i.e., inducing rankings over candidate sets for calibration (\S \ref{sec:calibration-section}). In \textbf{Chapter \ref{ch:measuring}}, we do a deep dive into faithfulness metrics with a particular emphasis on adapting existing metrics, which were designed for, and tested on, single document news summarization, to long-form clinical summarization. 

\textbf{Chapter \ref{ch:improving}} relies on high-quality metrics for faithfulness either for the purposes of identifying references for revision (\S \ref{sec:ltr}) or for providing "AI feedback", i.e., inducing rankings over candidate sets for calibration (\S \ref{sec:calibration-section}). Existing metrics of faithfulness which were largely developed on single document, general domain corpora~\cite{factcc,durmus-etal-2020-feqa,wang-etal-2020-asking,deng-etal-2021-compression,yuan2021bartscore,laban-etal-2022-summac,ribeiro-etal-2022-factgraph}, such as CNN/Daily Mail and BBC news articles. The generalizability of these metrics to long-form hospital-course summarization is not guaranteed.  Additionally, some metrics rely on external components and tools, such as question generation \parencite{fabbri-etal-2022-qafacteval} and dependency parsing \parencite{goyal2020evaluating}, which do not work on clinical texts given the massive difference in syntactic patterns and discourse structures. There is a wealth of evidence to demonstrate the beneficial impact of in-domain pre-training on clinical \parencite{alsentzer2019publicly, lehman2023we} and biomedical \parencite{gu2021domain} downstream tasks. Yet, domain adaptation has largely been unexplored when it comes to the development of the metrics used to evaluate domain-specific models. To be confident we are actually improving models, given the expense of large-scale human evaluations, it is important to develop automated metrics which correlate well with human intuitions of quality and faithfulness. In short, we need to meta-evaluate our metrics. Meta-evaluating metrics requires conducting a human evaluation study and then measuring the correlation of different metrics to the ratings provided by experts. To this end, we benchmark faithfulness metrics against fine-grained human annotations for model-generated summaries of a patient's Brief Hospital Course. Rather than adapt metrics to clinical text by training on references, we find it advantageous to learn directly from system summaries. We use an ensemble of our baseline metrics to produce a pseudo faithfulness score on system summaries and distill a metric from these noisy ground-truth labels. Our distilled metric has a higher correlation than baseline metrics to expert annotation labels.

Finally, in \textbf{Chapter \ref{ch:llm}}, we extend our work on entity-guided summarization of news articles \parencite{adams-etal-2023-generating, cod} onto the hospital-course task. We perform full parameter fine-tuning on Mistral and Zephyr 7B parameter models and instruct them to generate hospital course summaries. We demonstrate that these LLMs can generate fluent and readable summaries which hallucinate and fail to cover many salient medical entities pertinent to the patient. To better ground the LLMs on salient source entities, we train a specialized, smaller classification model to perform explicit content selection. Given the entity-dense nature of task, we select groups of synonymous entities---medical concepts---as the appropriate unit for content selection. A LongFormer encoder \parencite{beltagy2020longformer} first produces contextualized representations of entity spans. Synonymous mentions are identified and fused to form a single representation per each unique concept. Finally, concepts are sorted by inverse frequency of mentions and modeled with two additional self-attention layers before producing a binary salience score for each unique concept. By measuring entity overlap with clinician-authored references, we demonstrate that explicit content selection with a smaller dedicated model outperforms implicit content selection with Mistral or Zephyr. Given these findings, we investigate methods for entity-guided summarization, by which an LLM is instructed to generate a hospital course summary based on a pre-selected set of salient entities. To ground the content plan onto the source notes, we propose \underline{\textbf{SPEER}}: \textbf{\underline{S}}entence-Level \textbf{\underline{P}}lanning via \textbf{\underline{E}}mbedded \textbf{\underline{E}}ntity \textbf{\underline{R}}etrieval. Specifically, we mark each salient entity span with special \textbf{\texttt{\{\{ \}\}}} boundary tags and instruct the LLM to retrieve marked spans before generating each sentence. Sentence-level planning acts as a form of state tracking in that the model explicitly records the entities it uses.

\section{Contributions} \label{sec:contributions}

We contribute the following to the field of automatic summarization of long clinical narratives:

\begin{itemize}
    \item We are the first to create and analyze a large-scale dataset for Hospital Course Summarization \parencite{adams-etal-2021-whats}. We also open source two biomedical and clinical summarization datasets: ChemSum \parencite{adams-etal-2023-desired} and MIMIC-Sum \parencite{adams-etal-2022-learning}, which can enable more work into an understudied area: the summarization of long, highly scientific, texts.
    \item We make a unique contribution to faithfulness modeling by focusing on the role of data quality on downstream faithfulness, rather than model architectures or training objectives. Further, we demonstrate that directly revising noisy references \parencite{adams-etal-2022-learning}--for which we present a novel unsupervised method--is preferable to other noisy data mitigation strategies, such as filtering \parencite{nan-etal-2021-entity} or Loss Truncation \parencite{loss-truncation}.
    \item We demonstrate that the MLE training objective does not guarantee faithful or relevant clinical summaries \parencite{adams-etal-2023-desired}. We demonstrate that through careful selection of offline contrast sets that we can directly calibrate models to metrics of interest (faithfulness and relevance). We also provide an easy to extend codebase by which researchers can better calibrate summarization models to non-differentiable metrics (e.g., faithfulness).
    \item To enable efficient fine-grained annotation of faithfulness, we create a custom EHR documentation navigation tool and collect a valuable dataset of annotations. We explore novel methods for adapting pre-existing faithfulness metrics to clinical text, which include the use of granular source-summary alignments and distilling a metrics from other metrics \parencite{adams2023meta}.
    \item Finally, we demonstrate that open-source LLMs (Mistral \parencite{jiang2023mistral} and Zephyr \parencite{tunstall2023zephyr}) can be fine-tuned to produce fluent yet highly unfaithful and incomplete summaries. We present a new technique (\texttt{SPEER}: Sentence-Level Planning via Embedded Entity Retrieval) for learning to generate each summary sentence conditioned on a pre-selected set of uncovered medical entities (diagnoses, treatments, and tests). Entity-based faithfulness and coverage metrics demonstrate that sentence-level entity-guidance leads to more faithful and complete downstream summaries than competitive baseline summaries.
\end{itemize}

%% file: 3-dataset.tex
\chapter{Creating and Analyzing the Data} \label{ch:dataset}

To design and evaluate the Brief Hospital Course summarization task, we need a dataset of clinical documentation for hospital admissions that are both complete and diverse with respect to clinical settings and clinical specialties.  Yet, there are relatively few publicly available datasets for clinical NLP. The lack of available clinical data for NLP research is a direct result of HIPAA regulations, in particularly the The Privacy Rule, which protects all "individually identifiable health information" (PHI) from being shared with any external parties\footnote{\url{https://www.hhs.gov/hipaa}}. The de-identified MIMIC database (MIMIC-III and MIMIC-IV \parencite{johnson2016mimic}) is one such exception and has proven to be a valuable resource to the research community for many relevant clinical NLP tasks, e.g., acronym disambiguation \parencite{adams2020zero}, entailment \parencite{romanov2018lessons}, semantic role labeling \parencite{gao2023dr}, and medical concept annotation \parencite{kraljevic2021multi}. Yet, the MIMIC datasets are limited to a single clinical setting: the Intensive Care Unit (ICU), which means that it is insufficient to evaluate the generalizability of any MIMIC-based models across the full spectrum of in-patient settings, such as rehabilitation, surgical units, childbirth, and/or management of chronic illnesses. Additionally, as we demonstrate in Chapter \ref{ch:improving} and in related work \textcite{shing2021towards}, the MIMIC database is not adequately suited for multi-document tasks which track a patient's journey over time. This is due to the fact that not all notes recorded for a patient admission are included in the dataset, which causes many severe gaps in information. As a direct consequence of this, our work in \textcite{adams-etal-2022-learning}--conducted during an Amazon internship, empirically demonstrates that models trained on noisy MIMIC-derived hospital-course reference summaries suffer from severe hallucinations.

As such, to be able to adequately train and evaluate hospital-course summarization models, we first needed to build a dataset ourselves. In \S \ref{sec:whats}, we outline the steps taken to construct and analyze a new large-scale multi-document summarization dataset drawn from a set of highly diverse set of patients with diverse problems. Specifically, we construct an English, text-to-text dataset of 109,000 hospitalizations (2M source notes) and their corresponding summary proxy: the clinician-authored ``Brief Hospital Course'' paragraph written as part of a discharge note. Exploratory analyses reveal that the BHC paragraphs are highly abstractive with some long extracted fragments; are concise yet comprehensive; differ in style and content organization from the source notes; exhibit minimal lexical cohesion; and represent silver-standard references. For each finding, we explicitly detail highlight how it informed specific decisions in future

Our main contributions are as follows: (1) We collect a dataset of inpatient documentation and corresponding "Brief Hospital Course" paragraphs extracted from discharge notes; and (2) We assess the characteristics of these summary paragraphs as a proxy for target summaries and discuss implications for the design and evaluation of a hospital-course summarization tool.

\newcommand{\bhc}{\textsc{Brief Hospital Course}\xspace}
\newcommand{\xhdr}[1]{\noindent{{\bf #1.}}}

\section{Creating and an Analyzing a Hospital-Course Summarization Dataset} \label{sec:whats}

\noindent \textit{This section focuses on \textcite{adams-etal-2021-whats}, a publication which I presented as first author at NAACL in 2021.}

\begin{table}[t]
    \small
    \centering
    \begin{tabular}{c|l|c|c}
    & \bfseries Variable & \bfseries Value & \bfseries STD \\
    \multirow{3}{*}{Global}
     & \# Patients & 68,936 & \multirow{3}{*}{N/A} \\
     & \# Admissions & 109,726 &  \\
     & \# Source Notes & 2,054,828 &  \\ 
    \hline
    \multirow{5}{*}{\makecell{Per \\ Adm.}}
    & Length of Stay & 5.8 days & 9.0 \\
    & \# Source Notes & 18.7 & 30.1 \\
    & \# Source Sentences & 1,061.2 & 1,853.6 \\
    & \# Source Tokens & 11,838.7 & 21,506.5 \\
    & \# Summary Sentences & 17.8 & 16.9 \\
    & \# Summary Tokens & 261.9 & 233.8 \\
    \hline
    \multirow{2}{*}{\makecell{Per \\ Sent.}}
    & \# Source Tokens & 10.9 & 12.4 \\
    & \# Summary Tokens & 14.5 & 11.5 \\
    \hline
    \multirow{1}{*}{Ratio}
    & Word Compression & 42.5 & 164.6 \\
    \end{tabular}
    \caption{Basic Statistics for \dataset.  Value is the total for Global, and average for `Per Admission' and `Per Sentence'. STD is standard deviation.}
    \label{tab:clinsum-stats}
\end{table}

\begin{table*}[t]
    \centering
    \small
    \setlength{\tabcolsep}{2pt}
    \begin{tabular}{c|l|c|c|c|c|c|c|c|c}
    & & & Comp. & \multicolumn{2}{c|}{Extractiveness} & \multicolumn{2}{c|}{Summary} & Source \\
     & \bfseries Dataset & \# Docs & Ratio & Coverage & Density & \# words & \# sents & \# words \\
    \hline
    \multirow{7}{*}{SDS}
    & Gigaword \parencite{Rush_2015} & 4mn & 3.8 & 0.58 & 1.1 & 8.3 & 1 & 31.4 \\
    & CNN/DM \parencite{nallapati-etal-2016-abstractive} & 312k & 13.0 & 0.80 & 3.0 & 55.6 & 3.8 & 789.9 \\
    & Newsroom \parencite{grusky2018newsroom} & 1.2mn & 43.0 & 0.82 & 9.6 & 30.4 & 1.4 & 750.9 \\
    & XSum \parencite{narayan-etal-2018-dont-v2} & 226k & 18.8 & 0.57 & 0.89 & 23.3 & 1.0 & 431.1 \\
    & Arxiv \parencite{cohan2018discourse} & 215k & 39.8 & 0.92 & 3.7 & 292.8 & 9.6 & 6,913.8 \\
    & PubMed \parencite{cohan2018discourse} & 133k & 16.2 & 0.90 & 5.9 & 214.4 & 6.9 & 3,224.4 \\
    & BigPatent \parencite{sharma2019bigpatent} & 1.3mn & 36.4 & 0.86 & 2.4 & 116.5 & 3.5 & 3,572.8 \\
    \hline
    \multirow{5}{*}{MDS}
    & WikiSum \parencite{liu2018generating} & 2.3mn & 264.0 & N/A & N/A & 139.4 & N/A & 36,802.5 \\
    & Multi-News \parencite{fabbri2019multi} & 56k & 8.0 & 0.68 & 3.0 & 263.7 & 10 & 2,103.5 \\
    & SOAP \parencite{krishna2020generating} & 7k & 4.7 & N/A & N/A & 320 & N/A & 1,500 \\
    & \textbf{ \dataset (ours) } & \textbf{ 110k } & \textbf{  45.2 } & \textbf{  0.83 } & \textbf{  13.1 } & \textbf{ 261.9 } & \textbf{ 17.7 } & \textbf{  11,838.7 } \\
    \end{tabular}
    \caption{Basic statistics for single-document (SDS) and multi-document (MDS) summarization datasets. For multi-document summarization (MDS), \# Source words are aggregated across documents. Compression ratio is the average ratio of source words to summary words.  Extractiveness metrics (coverage and density) come from \textcite{grusky2018newsroom} and, for consistency, are calculated using the official \href{https://github.com/lil-lab/newsroom}{code} across the validation set for each dataset.  Spacy tokenization is performed before extracting fragments.  Other corpus statistics are pulled from either the corresponding paper or Table 1 in \textcite{sharma2019bigpatent}.  Entries are filled with N/A because the dataset is private \parencite{krishna2020generating}, or too expensive to generate \parencite{liu2018generating}. The Gigaword SDS dataset comes from the annotated Gigaword dataset \parencite{graff2003english, napoles2012annotated}}.
    \label{tab:corpora}.
\end{table*}

\subsection{Hospital-Course Summarization Task}

Given the clinical documentation available for a patient hospitalization, our task of interest is to generate a text that synthesizes the hospital course in a faithful and concise fashion.  For our analysis, we rely on the ``Brief Hospital Course'' (BHC), a mandatory section of the discharge note, as a proxy reference.  The BHC tells the story of the patient's admission: \textit{what} was done to the patient during the hospital admission and \textit{why}, as well as the \textit{follow up} steps needed to occur post discharge, whenever needed. Nevertheless, it is recognized as a challenging and time consuming task for clinicians to write~\parencite{dodd07,ucirvine20}.
 
\subsection{Dataset}

To carry out our analysis, we construct a large-scale, multi-document summarization dataset, \dataset. Materials come from all hospitalizations between 2010 and 2014 at Columbia University Irving Medical Center. \textbf{Table~\ref{tab:clinsum-stats}} shows summary statistics for the corpus. There are a wide range of reasons for hospitalizations, from life-threatening situations (e.g., heart attack) to when management of a specific problem cannot be carried out effectively outside of the hospital (e.g., uncontrolled diabetes). This contributes to the high variance in documentation. For reference, Table~\ref{tab:corpora} provides a comparison of basic statistics to widely used summarization datasets.  Relatively speaking, \dataset is remarkable for having a very high compression ratio despite having long reference summaries.  Additionally, it appears highly extractive with respect to fragment density yet qualify this in \S \ref{sec:extractiveness}.

Based on advice from clinicians, we rely on the following subset of note types as source documents: ``Admission'', ``Progress'', and ``Consult'' notes. The dataset does not contain any structured data, documentation from past encounters, or other note types (e.g., nursing notes, social work, radiology reports)~\parencite{reichert2010cognitive}.

\subsection{Tools for Analysis}

\paragraph**{Entity Extraction \& Linking.} 
We use the MedCAT toolkit \parencite{kraljevic2020multi} to extract medical entity mentions and normalize to concepts from the UMLS (Unified Medical Language System) terminology~\parencite{bodenreider2004unified}.  To exclude less relevant entities, we only keep entities from the Disorders, Chemicals \& Drugs, and Procedures semantic groups, or the Lab Results semantic type.

\paragraph**{Local Coherence.} We examine inter-sentential coherence in two ways. \textbf{Next-Sentence Prediction (NSP)}. Since we compare across a few datasets representing different domains, we use domain-specific pre-trained BERT models via HuggingFace \parencite{wolf2019huggingface}: 

\begin{itemize}
    \item \texttt{bert-base-cased} for CNN/DM and Arxiv.
    \item \texttt{monologg/biobert\_v1.1\_pubmed} for Pubmed.
    \item \texttt{emilyalsentzer/Bio\_ClinicalBERT} for \dataset.
\end{itemize}

\textbf{Entity-grids.} Entity-grids model local coherence by considering the distribution of discourse entities \parencite{barzilay2008modeling}. An entity grid is a 2-D representation of a text whose entries represent the presence or absence of a discourse entity in a sentence.  For our analyses, we treat UMLS concepts as entities and train a neural model, similar to \textcite{nguyen2017neural, mohiuddin2018coherence}, which learns to rank the entity grid of a text more highly than the same entity grid whose rows (sentences) have been randomly shuffled.

\paragraph*{Lexical Overlap Metric.} We use ROUGE-1 (R1) \& ROUGE-2 (R2) F-1~\parencite{lin2004rouge} to measure lexical overlap, while ignoring higher order variants based on analysis from other work \parencite{krishna2021}. We denote the average of R1 \& R2 scores as $R_{12}$.

\paragraph*{Extractive Summarization Baselines.} We rely on a diverse set of sentence extraction methods, whose performance on a held-out portion of \dataset is reported in Table \ref{tab:big-table}. \textbf{Oracle models} have access to the ground-truth reference and represent upper bounds for extraction.  Here, we define the sentence selection criteria for each oracle variant, leaving more in-depth discussion to the subsequent analysis. \textbf{\textsc{Oracle Top-K}}: Take sentences with highest $R_{12}$ vis-a-vis the reference until a target token count is reached; \textbf{\textsc{Oracle Gain}}: Greedily take source sentence with highest relative $R_{12}$ gain conditioned on existing summary\footnote{This is the Neusum model's objective \parencite{zhou2018neural}}. Extract sentences until the change in $R_{12}$ is negative;
\textbf{\textsc{Oracle Sent-Align}}: For each sentence in reference, take source sentence with highest $R_{12}$ score;
\textbf{\textsc{Oracle Retrieval}}: For each sentence in reference, take reference sentence from train set with largest BM25 score \parencite{robertson1994some}; and 
\textbf{\textsc{Oracle Sent-Align + Retrieval}}: For each sentence in reference, take sentence with highest $R_{12}$ between \textsc{Oracle Sent-Align} and \textsc{Oracle Retrieval}.  We provide two \textbf{unsupervised methods} as well. \textbf{\textsc{Random}}: extracts random sentences until summary reaches target word count (average summary length); \textbf{\textsc{LexRank}}: selects the top-k sentences with largest LexRank \parencite{erkan2004lexrank} score until target word count is reached.  For a supervised baseline, we present \textbf{\textsc{ClinNeusum}}: a variant of the Neusum model adapted to the clinical genre~\parencite{zhou2018neural}. \textsc{ClinNeusum} is a hierarchical LSTM network trained on ground-truth labels derived from \textsc{Oracle Gain}.

\begin{table*}[ht]
    \centering
    \begin{tabular}{l|ccc|ccc}
    \bfseries Extractive Baseline & \multicolumn{3}{c}{\textbf{ROUGE-1}} & \multicolumn{3}{c}{\textbf{ROUGE-2}} \\
    \hline
     & Recall & Precision & F1 & Recall & Precision & F1  \\
     \hline
    \textsc{Random} & 0.16 & 0.24 & 0.17 & 0.04 & 0.03 & 0.03 \\
    \textsc{LexRank} & 0.18 & 0.21 & 0.18 & 0.05 & 0.05 & 0.05 \\
    \textsc{ClinNeusum} & 0.36 & 0.25 & 0.27 & 0.14 & 0.1 & 0.11 \\
    \textsc{Oracle Top-K} & 0.28 & 0.52 & 0.32 & 0.16 & 0.32 & 0.19 \\
    \textsc{Oracle Gain} & 0.43 & 0.63 & 0.5 & 0.26 & 0.42 & 0.3 \\
    \textsc{Oracle Sent-Align (SA)} & 0.48 & 0.61 & 0.52 & 0.3 & 0.33 & 0.31 \\
    \textsc{Oracle Retrieval} & 0.51 & 0.70 & 0.58 & 0.25 & 0.28 & 0.29 \\
    \textsc{Oracle SA + Retrieval} & \textbf{0.6} & \textbf{0.76} & \textbf{0.66} & \textbf{0.4} & \textbf{0.49} & \textbf{0.43} \\
    \end{tabular}
    \caption{Performance of different sentence selection strategies on \dataset.}
    \label{tab:big-table}
\end{table*}

\subsection{Dataset Analysis \& Implications}

We distill task-specific characteristics to a few standalone takeaways. For each, we provide evidence in the data before pinpointing its impact on subsequent chapters of this thesis.

\subsubsection*{Summaries are mostly abstractive with a few long segments of copy-pasted text}
\label{sec:extractiveness}

\paragraph*{tl;dr.} \dataset summaries appear extractive according to widely used metrics.  Yet, there is large variance within summaries.  This directly affects the performance of a supervised extractive model, whose selection capability degrades as summary content transitions from copy-paste to abstractive.  In turn, we need models which can handle abrupt transitions between extractive and abstractive text.

\paragraph*{Background.} Clinicians copy forward information from previous notes to save time and ensure that each note includes sufficient evidence for billing and insurance purposes \parencite{wrenn2010quantifying}.  Copy-paste is both widely used (66-90\% of clinicians according to a recent literature review \parencite{tsou2017safe}) and widely applied (a recent study concluded that in a typical note, 18\% of the text was manually entered; 46\%, copied; and 36\% imported\footnote{Imported refers to text typically pulled in from structured data, such as a medication or problem list.} \parencite{wang2017characterizing}).

\paragraph*{Analysis - extractiveness.}

\begin{figure}[ht]
\begin{center}
\centerline{\includegraphics[width=0.5 \columnwidth]{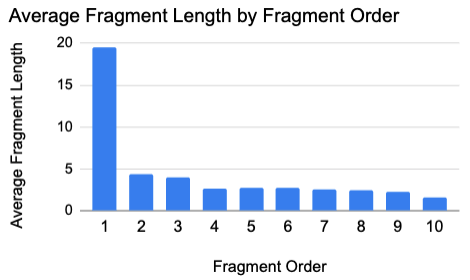}}
\caption{Average extractive fragment lengths according to their relative order within the summary.}
\label{fig:frag-len-rank}
\end{center}
\end{figure}

\dataset appears very extractive: a high coverage (0.83 avg / 0.13 std) and a very high density (13.1 avg / 38.0 std) (See \textcite{grusky2018newsroom} for a description of the statistics).  However, we find that 64\% of the extractive fragments are unigrams, and 25\% are bigrams, which indicate a high level of re-writing.  The density measure is large because the remaining 11\% of extractive fragments are very long. Yet, there is a strong positional bias within summaries for long fragments. \textbf{Figure \ref{fig:frag-len-rank}}, groups fragments according to their relative order within each summary.  The longest fragments are usually first.  Qualitative analysis confirms that the beginning of the BHC is typically copied from a previous note and conveys the ``one-liner'' (e.g., \textit{pt is a 50yo male with history of CHF who presents with edema.})

This abrupt shift in extractiveness should affect content selection.  In particular, when looking at oracle extractive strategies, we should see clear-cut evidence of \textbf{(1)} 1-2 sentences which are easy to identify as salient (i.e., high lexical overlap with source due to copy-paste), \textbf{(2)} a murkier signal thereafter.  To confirm this, we analyze the sentences selected by the \textsc{Oracle Gain} method, which builds a summary by iteratively maximizing the $R_{12}$ score of the existing summary vis-a-vis the reference.

\begin{figure}[ht]
\begin{center}
\centerline{\includegraphics[width=0.75 \columnwidth]{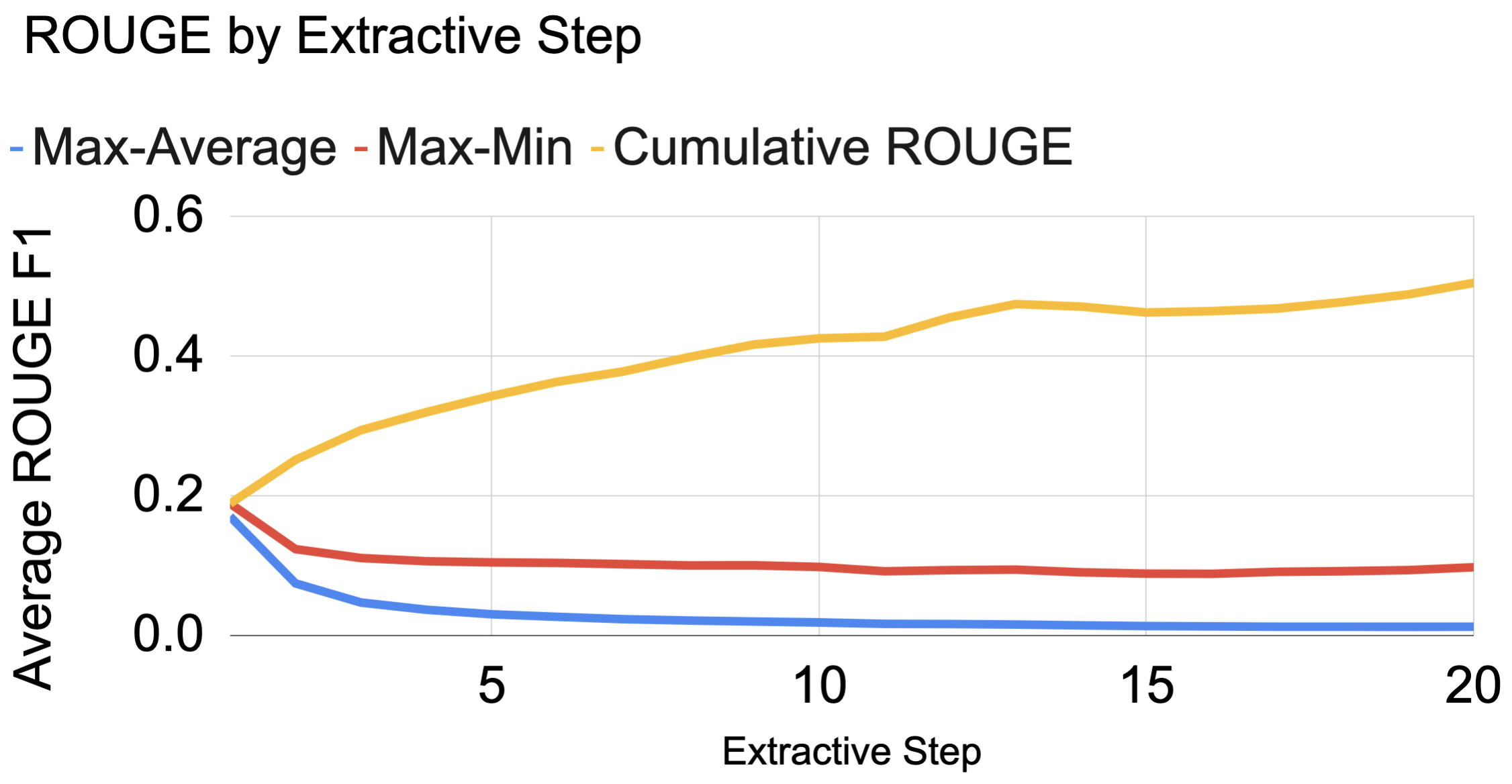}}
\caption{We plot average ROUGE score as summaries are greedily built by adding the sentence with the highest relative ROUGE gain vis-a-vis the current summary, until the gain is no longer positive (\textsc{Oracle Gain}).  We also include the difference between the highest scoring sentence and the average / minimum to demonstrate a weakening sentence selection signal after the top 1-2. }
\label{fig:rouge-by-step}
\end{center}
\end{figure}

In Figure \ref{fig:rouge-by-step}, two supporting trends emerge.  \textbf{(1)} On average, one sentence accounts for roughly 50\%\footnote{From Table \ref{tab:big-table}, the average $R_{12}$ score is 0.39 for \textsc{Oracle Gain}.  To reconcile this number with respect to Figure \ref{fig:rouge-by-step}, we note that the average oracle summary is far less than the 20 sentence upper bound shown in the chart.} of the overall $R_{12}$ score. \textbf{(2)} Afterwards, the marginal contribution of the next shrinks, as well as the $R_{12}$ gap between the best sentence and the minimum / average, according to the oracle.

There should also be evidence of the copy-paste positional bias impacting content selection.  Table \ref{tab:pos-rank-by-step} reveals that the order in which the \textsc{Oracle Gain} summary is built--by maximal lexical overlap with the partially built summary--roughly corresponds to the true ordering of the summary.  More simply, the summary transitions from extractive to abstractive.

\begin{table}[ht]
    \centering
    \begin{tabular}{c|c}
    \bfseries Extractive & \bfseries Average Rank of Closest \\
    \bfseries Step & \bfseries Reference Sentence \\
    1 & 4.7 \\
    2 & 6.0 \\
    3 & 6.3 \\
    4 & 6.7 \\
    5 & 7.3 \\
    > 5 & 10.1 \\
    \end{tabular}
    \caption{\textsc{Oracle Gain} greedily builds summaries by repeatedly selecting the sentence which maximizes the $R_{12}$ score of the partially built summary.  By linking each extracted sentence to its closest in the reference, we show that this oracle order is very similar to the true ordering of the summary.}
    \label{tab:pos-rank-by-step}
\end{table}

Unsurprisingly, a model (\textsc{ClinNeusum}) trained on \textsc{Oracle Gain} extractions gets progressively worse at mimicking it. Specifically, for each extractive step, there exists a ground-truth ranking of candidate sentences by relative $R_{12}$ gain.  As the relevance gap between source sentences shrinks (from Figure \ref{fig:rouge-by-step}), \textsc{ClinNeusum}'s predictions deviate further from the oracle rank (Table \ref{tab:rank-by-step}).

\begin{table}[ht]
    \centering
    \begin{tabular}{c|c|c}
    \bfseries Extractive & \multicolumn{2}{c}{\bfseries Ground Truth Rank} \\
    \bfseries Step & \bfseries Average & \bfseries Median  \\
    1 & 28 & 7 \\
    2 & 69 & 22 \\
    3 & 74 & 31 \\
    4 & 79 & 39 \\
    5 & 76 & 42 \\
    > 5 & 80 & 60 \\
    \end{tabular}
    \caption{Rank of selected sentence vis-a-vis oracle rank at each extraction step.  A perfectly trained system would have a ground-truth of 1 at each step.}
    \label{tab:rank-by-step}
\end{table}

\paragraph*{Analysis - Redundancy.} Even though we prevent all baseline methods from generating duplicate sentences (23\% of source sentences have exact match antecedents), there is still a great deal of redundancy in the source notes (i.e., modifications to copy-pasted text).  This causes two issues related to content selection. The first is fairly intuitive - that local sentence extraction propagates severe redundancy from the source notes into the summary and, as a result, produces summaries with low lexical coverage.  We confirm this by examining the performance between the \textsc{Oracle Top-K} and \textsc{Oracle Gain}, which represent summary-unaware and summary-aware variants of the same selection method.  While both extract sentences with the highest $R_{12}$ score, \textsc{Oracle Gain} outperforms because it incorporates redundancy by considering the relative $R_{12}$ gain from an additional sentence.

The second side effect is perhaps more surprising, and divergent from findings in summarization literature.  For most corpora, repetition is indicative of salience.  In fact, methods based on lexical centrality, i.e., TextRank \parencite{mihalcea2004textrank} and LexRank \parencite{erkan2004lexrank}, still perform very competitively for most datasets. Yet, for \dataset, LexRank barely outperforms a random baseline.  Poor performance is not only due to redundance, but also a weak link between lexical centrality and salience. The Pearson correlation coefficient between a sentence's LexRank score and its $R_{12}$ overlap with the reference is statistically significant ($p < 0.05$) yet weak ($r=0.29$).

\begin{figure}
\centering
\begin{center}
\includegraphics[width=0.5 \columnwidth]{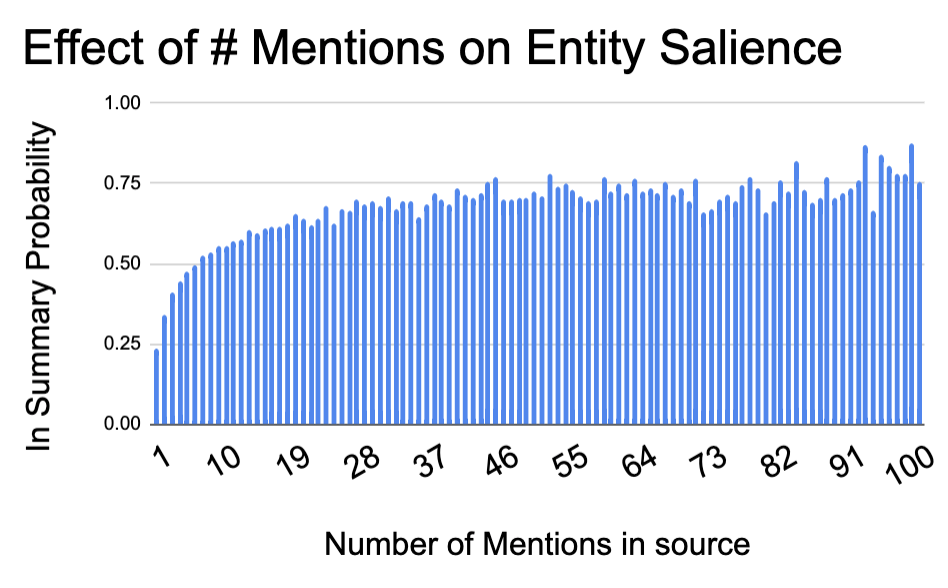}
\caption{Relationship between source entity mentions and probability of inclusion in the summary.
}
\label{fig:source-freq-recall}
\end{center}
\end{figure}

Qualitative analysis reveals two principal reasons, both related to copy-paste and/or imported data.  The first relates to the propagation of frequently repeated text which may not be useful for summaries: administrative (names, dates), imported structured data, etc.  The second relates to sentence segmentation. Even though we use a custom sentence splitter, our notes still contain some very long sentences due to imported lists and semi-structured text--a well-documented issue in clinical NLP \parencite{leaman2015challenges}. LexRank summaries have a bias toward these long sentences (26.2 tokens versus source average of 10.9), which have a greater chance of containing lexical centroid(s). To bypass some of these issues, however, one can examine the link between centrality and salience at the more granular level of entities. \textbf{Figure \ref{fig:source-freq-recall}} shows a clear-cut positive correlation between source note mention frequency of UMLS concepts and the probability of being included in the summary.

\subsubsection*{Summaries are concise yet comprehensive} \label{sec:concise}

\paragraph*{tl;dr.} BHC summaries are packed with medical entities, which are well-distributed across the source notes. As such, relations are often not explicit.  Collectively, this difficult task calls for a domain-specific approach to assessing faithfulness.

\paragraph*{Analysis - concise}


We find that summaries are extremely dense with medical entities: $20.9\%$ of summary words are medical UMLS entities, compared to $14.1\%$ in the source notes.  On average, summaries contain $26$ unique entities whereas the source notes contain $265$ --- an entity compression ratio of 10 (versus token-level compression of 43). 


\paragraph*{Analysis - comprehensive.}

\begin{figure}[!ht]
\centering
\begin{center}
\includegraphics[width=\columnwidth]{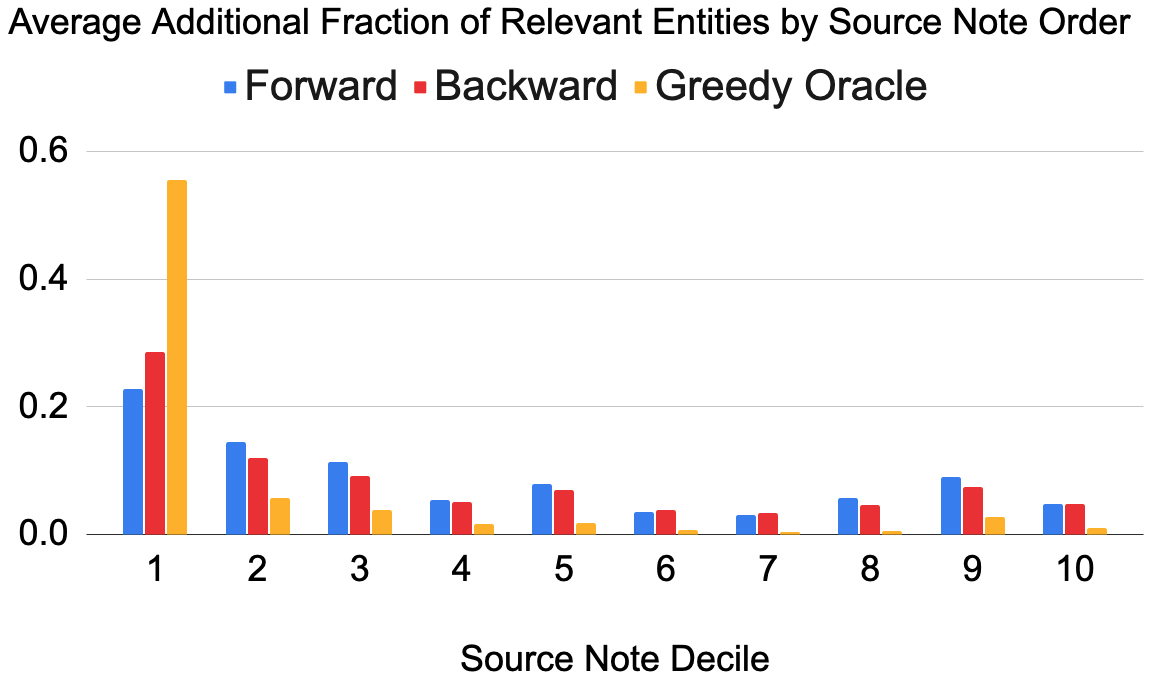}
\caption{The average fraction of additional relevant UMLS entities---present in the summary---from reading a patient's visit notes.  
\textsc{Forward} orders the notes chronologically, \textsc{Backward} the reverse, and \textsc{Greedy Oracle} in order of decreasing entity overlap.}
\label{fig:chronology}
\end{center}
\end{figure}

\begin{figure}[!ht]
\centering
\begin{center}
\includegraphics[width=0.75 \columnwidth]{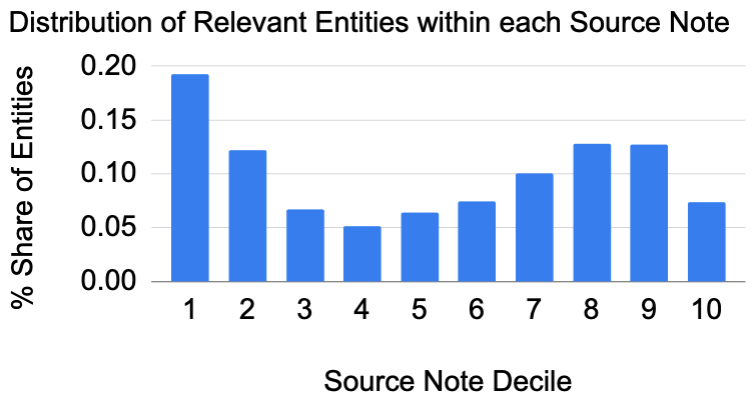}
\caption{The distribution of relevant entities---present in the summary---within an average source note. Source Note Decile refers to the relative position of each mention within a note. Relevant entities appear throughout an average note, with a slight lead bias.}
\label{fig:lead-bias}
\end{center}
\end{figure}

Many summarization corpora exhibit systematic biases regarding where summary content can be found within source document(s) \parencite{dey2020corpora}. On \dataset, we examine the distribution of entities along two dimensions: \textit{macro} considers the differences in entity share across notes, and \textit{micro} considers the differences within each note (i.e., lead bias). \textbf{(1) Macro Ordering.} When looking at the source notes one by one, how much \textit{additional} relevant information (as measured by entities present in the summary) do you get from each new note?  We explore three different orderings: (1) \textsc{Forward} orders the notes chronologically, (2) \textsc{backward} the reverse, and (3) \textsc{Greedy Oracle} examines notes in order of decreasing entity entity overlap with the target.  Given the large variation in number of notes per admission, we normalize by binning notes into deciles. \textbf{Figure \ref{fig:chronology}} shows that it is necessary to read the entire set of notes despite diminishing marginal returns.  One might expect the most recent notes to have the most information, considering present as well as copy-forwarded text. Surprisingly, \textsc{forward} and \textsc{backward} distributions are very similar.  \textsc{Greedy Oracle} gets at the level of information concentration.  On average, the top 10\% of most informative notes cover just over half of the entities found in the summary. We include absolute and percentage counts in Table \ref{tab:ordering}.  \textbf{(2) Micro Ordering.} We plot a normalized histogram of summary entities by relative position within the source documents. Figure \ref{fig:lead-bias} reveals a slight lead bias, followed by an uptick toward the end.  Clinical notes are organized by section: often starting with the past medical history and present illness, and typically ending with the plan for future care.  All are needed to write a complete BHC.

\begin{table}[ht]
    \centering
    \begin{tabular}{l|c|c|c}
    & \multicolumn{2}{c}{Avg Notes to Read} \\
    \bfseries Ordering & \bfseries Number & \bfseries Percent \\
    \textsc{Forward} & 8.5 & 0.80 \\
    \textsc{Backward} & 7.8 & 0.73 \\
    \textsc{Greedy Oracle} & 5.0 & 0.50 \\
    \end{tabular}
    \caption{Number of documents necessary to cover all relevant UMLS entities---present in the summary---according to three different ordering strategies. \textsc{Forward} orders the notes chronologically, \textsc{Backward} the reverse, and \textsc{Greedy Oracle} examines notes in order of decreasing entity overlap with the target.}
    \label{tab:ordering}
\end{table}

\subsubsection*{Summaries have different style and content organization than source notes} \label{misalignment}

\paragraph*{tl;dr.} Hospital course summarization involves not only massive compression, but a large style and organization transfer. Source notes are written chronologically yet the way clinicians digest the information, and write the discharge summary, is largely problem-oriented.  With simple oracle analysis, we argue that retrieve-edit frameworks are well-suited for hospital course generation.

\paragraph*{Analysis - Style.}  Clinical texts contain many, often obscure, abbreviations \parencite{finley2016towards, adams2020zero}, misspellings, and sentence fragments~\parencite{demner2009can}. Using a publicly available abbreviation inventory \parencite{moon2014sense}, we find that abbreviations are more common in the BHC.  Furthermore, summary sentences are actually longer on average than source sentences (15.8 versus 12.4 words).

\paragraph*{Analysis - Organization.}

\begin{figure}[ht]
\begin{center}
\centerline{\includegraphics[width=\columnwidth]{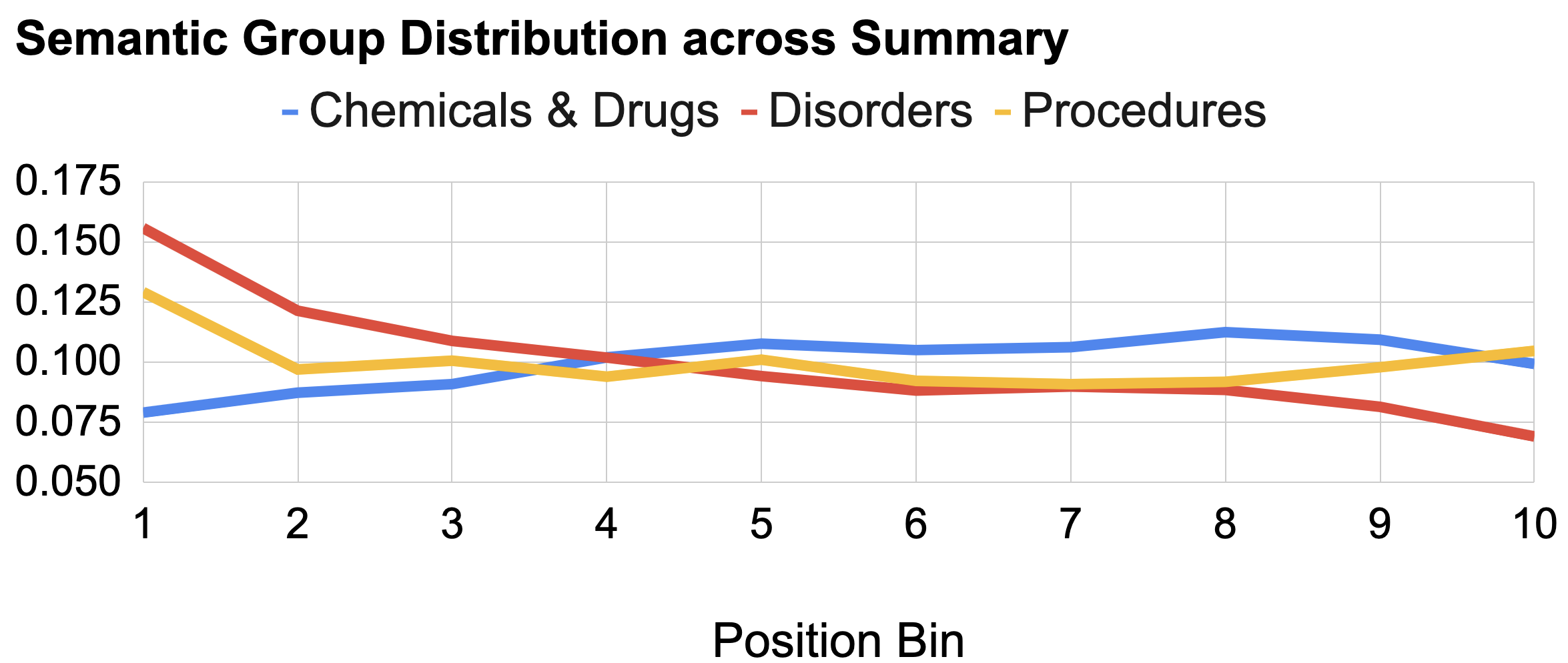}}
\caption{Position of entities within a summary.}
\label{fig:sem-group-positional}
\end{center}
\end{figure}

\begin{figure}
\centering
\begin{center}
\includegraphics[width=0.75 \columnwidth]{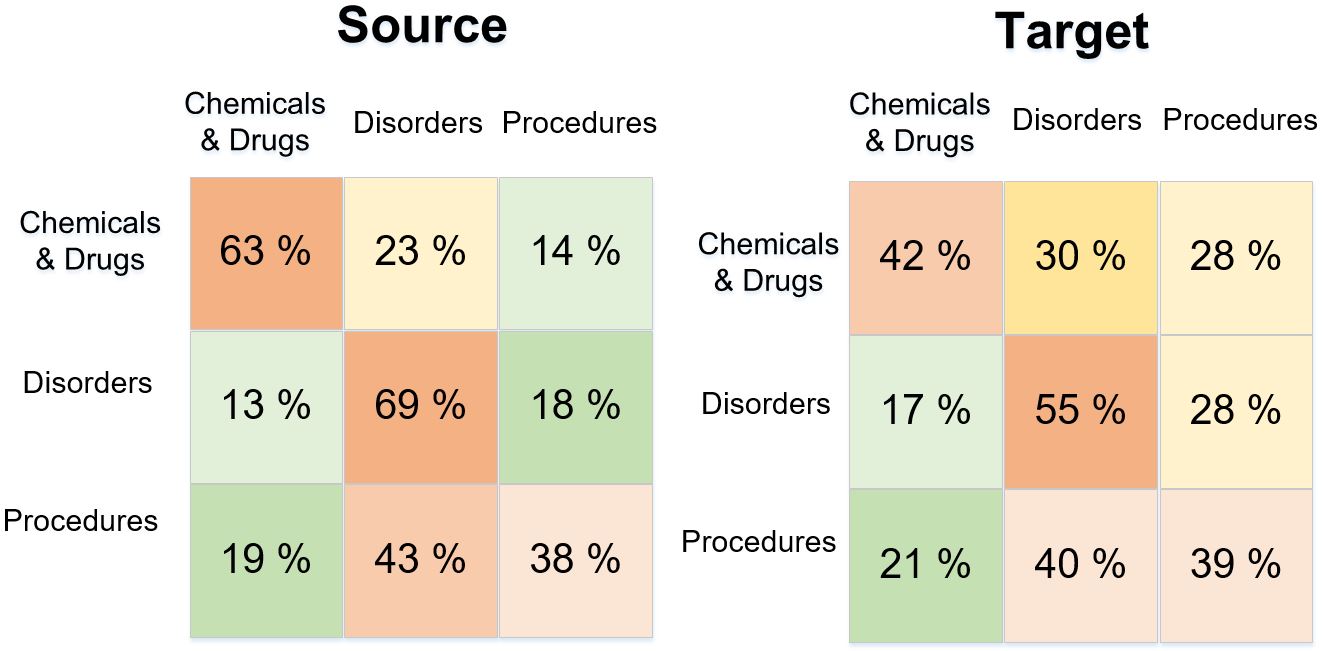}
\caption{Entity Transition Matrices for source notes and target summaries. Summaries have fewer clusters of semantically similar entities,  indicating that entity mentions are woven into a problem-oriented summary.}
\label{fig:ent-trans}
\end{center}
\end{figure}

Qualitative analysis confirms that most BHCs are written in a problem-oriented fashion \parencite{weed1968medical}, i.e., organized around a patient's disorders.  To more robustly analyze content structure, we compare linked UMLS entities at the semantic group level: \textsc{Drugs}, \textsc{Disorders}, and \textsc{Procedures} \parencite{mccray2001aggregating}. In particular, we compare \textbf{global} proportions of semantic groups, \textbf{transitions} between entities, as well as \textbf{positional} proportions within summaries.  \textbf{(1) Global.} Procedures are relatively more prevalent in summaries (31\% versus 24\%), maybe because of the emphasis on events happening during the hospitalization. In both summary and source notes, \textsc{Disorders} are the most prevalent (54\% and 46\%, respectively).  Drugs make up 23\% and 22\% of entity mentions in summary and source notes, respectively.  \textbf{(2) Transitions.} From both source and summary text, we extract sequences of entities and record adjacent transitions of their semantic groups in a $3\times3$ matrix. \textbf{Figure \ref{fig:ent-trans}} indicates that summaries have fewer clusters of semantically similar entities (diagonal of the transition matrix).  This transition matrix suggests a problem-oriented approach in which disorders are interleaved with associated medications and lab results.  \textbf{(3) Positional.}  Finally, within summaries, we examine the positional relative distribution of semantic groups and connect it to findings from Section \ref{sec:extractiveness}. In \textbf{Figure \ref{fig:sem-group-positional}}, we first compute the start index of each clinical entity, normalized by the total length, and then group into ten equally sized bins.  The early prevalence of disorders and late prevalence of medications is expected, yet the difference is not dramatic.  This suggests an HPI-like statement up front, followed by a problem oriented narrative.

If there is a material transfer in \textbf{style} and \textbf{content}, we would expect that summaries constructed from other summaries in the dataset would have similar or better lexical coverage than summaries constructed from sentences in the source notes.  To assess this, we compare two oracle baselines, \textsc{sent-align} and \textsc{retrieval}. For each sentence in the summary, we find its closest corollary either in the source text (\textsc{sent-align}) or in other summaries in the dataset (\textsc{retrieval}). While the retrieval method is at a distinct disadvantage because it does not contain patient-specific information and retrieval is performed with BM25 scores, we find both methods yield similar results (\textbf{Table \ref{tab:big-table}}).  An ensemble of \textsc{sent-align} and \textsc{retrieval} performs better than either alone, suggesting that the two types of sources may be complementary.  82\% of this oracle's summary sentences are retrievals. Summaries adapt the style and problem-oriented structure of other summaries, but contain patient-specific information from the source notes.

\begin{figure}[ht]
\begin{center}
\centerline{\includegraphics[width=0.75 \columnwidth]{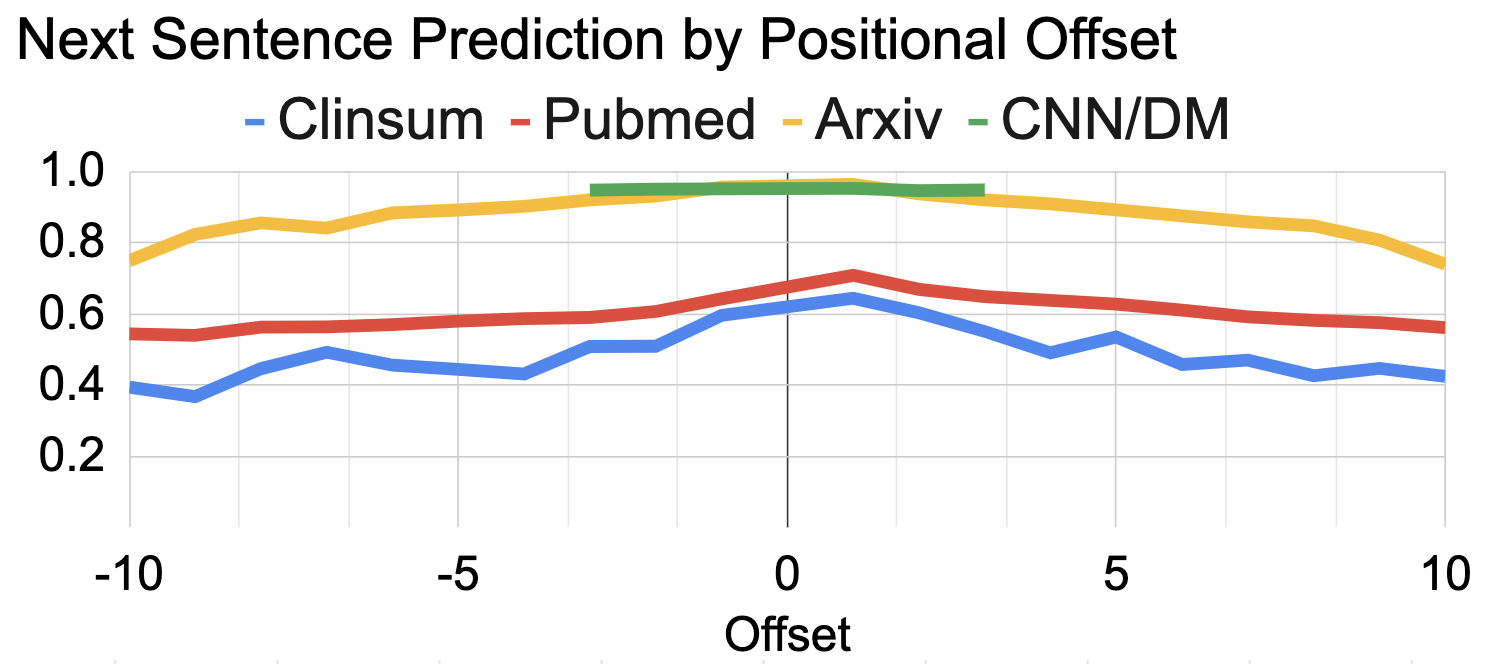}}
\caption{NSP logit by relative position of the next sentence across summaries for several datasets. An offset of 1 corresponds to the true next sentence. }
\label{fig:nsp}
\end{center}
\end{figure}

\paragraph*{Implications.}

Hospital-course summaries weave together disorders, medications, and procedures in a problem-oriented fashion.  It is clear that substantial re-writing and re-organization of source content is needed. One suitable approach is to use the retrieve-rerank-rewrite ($R^3$) framework proposed by \textcite{cao2018retrieve}.  To support this notion, more recent work demonstrates that retrieval augmented generation is effective for knowledge-intensive tasks \parencite{lewis2020retrieval}, enhances system interpretability \parencite{guu2020realm,krishna2020generating}, and can improve LM pre-training \parencite{lewis2020pre}\footnote{The related idea of template-based generation has gained traction within the probabilistic community \parencite{wiseman2018learning, guu2018generating, wu2019response, he2020learning}.}.  Also, efforts to bridge the gap between template-based and abstractive generation have been successful in the medical domain for image report generation \parencite{li2018hybrid}.

In this light, BHC generation could be truly problem-oriented.  The first step would involve selecting salient problems (i.e., disorders) from the source text--a well-defined problem with proven feasibility \parencite{van2010corpus}. The second step would involve separately using each problem to retrieve problem-specific sentences from other summaries. These sentences would provide clues to the problem's relevant medications, procedures, and labs. In turn, conceptual overlap could be used to re-rank and select key, problem-specific source sentences.  The extracted sentences would provide the patient-specific facts necessary to re-write the problem-oriented retrieved sentences.

\subsubsection*{Summaries exhibit low lexical cohesion} \label{cohesion}

\paragraph*{tl;dr.} Lexical cohesion is sub-optimal for evaluating hospital-course discourse because clinical summaries naturally exhibit frequent, abrupt topic shifts.  Also, low correlation exists between lexical overlap and local coherence metrics.

\paragraph*{Analysis.} Entity-based coherence research posits that "texts about the same discourse entity are perceived to be more coherent than texts fraught with abrupt switches from one topic to the next" \parencite{barzilay2008modeling}.  Yet, for \dataset summaries, coherence and abrupt topic shifts are not mutually exclusive. An analysis of the entity grids of summaries, presumably coherent, are sparse, with few lexical chains. In fact, over 66\% of the entities in the BHC appear only once.  Of those with multiple mentions, the percentage which appear in adjacent sentences is only 9.6\%.  As in \textcite{prabhumoye2020topological}, we also compare coherence with next-sentence prediction (NSP). \textbf{Figure \ref{fig:nsp}} plots the NSP logit by positional offset, where an offset of 1 corresponds to the next sentence, and -1 to the previous. NSP relies on word overlap and topic continuity \parencite{bommasani2020intrinsic}, so it makes sense it is lowest for \dataset.

To confirm the hypothesis that ROUGE does not adequately capture content structure, we use the \textit{pairwise ranking} approach to train and evaluate an entity-grid based neural coherence model~\parencite{barzilay2008modeling,nguyen2017neural}. 
\textbf{Table \ref{tab:pra}} shows ROUGE and coherence metrics side-by-side for \textsc{Oracle Gain}, which naively orders sentences according to document timestamp, then within-document position, and \textsc{Oracle Sent-Align}, which maintains the structure of the original summary. The poor coherence of \textsc{Oracle Gain} is obscured by comparable ROUGE scores.

\begin{table}[ht]
    \centering
    \begin{tabular}{l|c|c|c}
    \bfseries Summary & \bfseries Acc. & \bfseries R1 & \bfseries R2 \\
    Actual Summary & 0.86 & N/A & N/A \\
    \textsc{Oracle Sent-Align} & 0.75 & 0.52 & 0.30 \\
    \textsc{Oracle Gain} & 0.54 & 0.48 & 0.30 \\
    \end{tabular}
    \caption{Comparison of coherence and ROUGE. Acc. refers to pair-wise ranking accuracy from scoring summaries against random permutations of themselves. }
    \label{tab:pra}
\end{table}

\subsubsection*{BHC summaries are silver-standard} \label{silver-standard}
\paragraph*{tl;dr.}  Discharge summaries and their associated BHC sections are frequently missing critical information or contain excessive or erroneous content.  Modeling efforts should address sample quality.

\paragraph*{Analysis.} \textcite{kripalani2007deficits} find that discharge summaries often lack important information including diagnostic test results (33-63\% missing) treatment or hospital course (7-22\%), discharge medications (2-40\%), test results pending at discharge (65\%), patient/family counseling (90-92\%), and follow-up plans (2-43\%). The quality of the reporting decreases as the length of the discharge summary increases, likely due to copy-pasted information~\parencite{van1999necessary}.

These quality issues occur for a number of reasons: (1) limited EHR search functionality makes it difficult for clinicians to navigate through abundant patient data~\parencite{christensen2008instant}; (2) multiple clinicians contribute to incrementally documenting care throughout the patient's stay; (3) despite existing guidance for residents, clinicians receive little to no formal instruction in summarizing patient information~\parencite{ming2019discharge}; and (4) clinicians have little time for documenting care.

\section{Related Work}

Summarization of clinical data and documentation has been explored in a variety of use cases \parencite{pivovarov2015automated}. For longitudinal records, graphical representations of structured EHR data elements (i.e., diagnosis codes, laboratory test measurements, and medications) have been proposed~\parencite{powsner1997summarizing,plaisant1996lifelines}. Interactive visualizations of clinical problems' salience, whether extracted from notes~\parencite{hirsch2015harvest} or inferred from clinical documentation~\parencite{levy2020towards} have shown promise~\parencite{pivovarov2016can,2020levyfixthesis}.

Most work in this area, however, has focused on clinical documentation of a fine temporal resolution. Traditional text generation techniques have been proposed to synthesize structured data like ICU physiological data streams ~\parencite{hunter2008summarising,goldstein2016automated}. \textcite{liu2018learning} use a transformer model to write EHR notes from the prior 24 hours, while \textcite{liang2019novel} perform disease-specific summarization from individual progress notes. 
\textcite{mcinerney2020query} develop a distant supervision approach to generate extractive summaries to aid radiologists when interpreting images.
\textcite{zhang2018learning, zhang2019optimizing, macavaney2019ontology, sotudeh2020attend} generate the ``Impression'' section of the Radiology report from the more detailed ``Findings'' section.  Finally, several recent works aim to generate EHR notes from doctor-patient conversations~\parencite{krishna2020generating,joshi-etal-2020-dr,MSR2020}. Work on summarizing hospital admissions has tended to focus on extractive methods \parencite{moen2014evaluation,moen2016comparison, liu2018unsupervised, alsentzer2018extractive}.

\section{Conclusions and Limitations}

Based on a comprehensive analysis of clinical notes, we identify a set of implications for hospital-course summarization which we address in subsequent chapters of this thesis.

Noisy references can harm model performance, yet there is a rich body of literature to show that simple heuristics can identify good references \parencite{bommasani2020intrinsic} and/or filter noisy training samples \parencite{rush2015neural, akama2020filtering, matsumaru2020improving, nan-etal-2021-entity}. In \textbf{Chapter \ref{ch:improving}}, we demonstrate the negative impact of noisy data on downstream faithfulness. We redress this in two ways. Rather than filter out bad training examples and shrink the size of the training data, we devise a method to edit reference summaries such that they are fully grounded onto the \emph{available} source notes (\textbf{\S \ref{sec:ltr}}). The second strategy involves ignoring reference summaries and instead, learning directly from faithfulness metrics applied to generated contrast sets (\textbf{\S \ref{sec:calibration}}).

Regarding within-summary variation in reference summary \textbf{extractiveness}, we find that models learn to mimic this behavior. Many sentences are copied-and-pasted early on in the summary, while more abstraction and fusion happens towards the end. In \textbf{Chapter \ref{ch:measuring}}, we demonstrate that existing faithfulness metrics tend to over rely on this large variation in extractiveness when measuring faithfulness, which confirms findings for news summarization \parencite{ladhak-etal-2022-faithful}. Yet, as we demonstrate in human annotations, and is also noted in the general summarization literature \parencite{zhang-etal-2023-extractive}, extractive does not guarantee faithfulness. Given the strong role of extractivness, we learn a metric which is relatively de-correlated to extractiveness and find that, when combined with a simple measure of extractiveness (fragment density), we can show attain improved correlation to human assessments of faithfulness (\textbf{Chapter \ref{ch:measuring}}). 

In \textbf{Chapter \ref{ch:llm}}, regarding \textbf{entity density} and the \textbf{positive correlation of entity mentions to salience}, in contrast to the weak signal of lexical centrality to salience, we build off our work in general domain news summarization \parencite{adams-etal-2023-generating, cod} and explicitly perform content selection at the entity-level. In \textbf{Chapter \ref{ch:llm}} , we demonstrate that grounding summaries on a pre-defined set of entities from the source notes--which are produced by a separate model--improves both the coverage and fidelity of salient entities. Regarding \textbf{redundancy}, it is clear that, in contrast to some summarization tasks \parencite{kedzie2018content}, summary-aware content selection is essential for hospital-course summarization for both extractive and abstractive approaches. We operationalize this insight by choosing to condition generation of summary sentences on a set of salient entities which are \emph{not} covered by previous sentences. In this way, we discourage repetition of the same concepts across a summary, which tends to occur when decoding with vanilla greedy, sampling, or beam-search methods.

\paragraph*{Limitations.} The most glaring limitation to the work described in this Chapter stems from transparency and reproducibility. We are unable to share the dataset used for our analysis due to HIPAA regulations, which restricts its usefulness to the clinical NLP community to the confines of the paper itself. Given the zero-shot capabilities of LLMs on clinical tasks \parencite{nori2023capabilities}, and their well-documented abilities as high-quality generators of synthetic data \parencite{gunasekar2023textbooks}, future work, in our opinion, must explore synthetic creation of high-quality hospital-course summarization datasets. Augmentation methods, such as Self-Instruct \parencite{wang2022self}, provide a blue-print for how to increase the quality of synthetic data with rejection sampling. Yet, in the next chapter, we create and experiment with a publicly accessible dataset (based on the MIMIC-III clinical records database \parencite{johnson2016mimic}) to enhance the reach and transparency of this thesis.

%% file: 4-modeling-faithfulness.tex
\chapter{Improving Faithfulness}\label{ch:improving}

In the previous Chapter (\ref{ch:dataset}), we demonstrated that hospital-course reference summaries, as extracted from Electronic Health Records, often contain information which is not covered in any of the available source notes\footnote{Since we did not a large-scale human analysis to confirm this phenomenon, we also must highlight that it could be due to the use of shallower measures of lexical overlap rather than deeper semantic analysis.}. As naturally occurring reference summaries, this can happen for any number of reasons, including, but not limited to, the incomplete extraction of notes from the EHR, events that occurred on the day of discharge, imperfect data capture from clinicians regarding events that took place that were never documented, and the exclusive presence of data in structured tables.

 Faithfulness in summarization, and NLP more generally, is a loaded term. In this thesis, we focus on ``grounded'' faithfulness -- a hospital-course is faithful to its source notes only if it is logically entailed by some subset of the source notes \parencite{bos2005recognising}. This narrow formulation leads us to exclude two alternative aspects of faithfulness. The first is world knowledge. Introduction of significant external knowledge, even if true, is considered unfaithful if the patient-specific source notes do not entail it. Since most of our models produce highly extractive summaries, and most metrics correlate strongly with extractiveness, much of the work in this thesis focuses on direct lexical overlap between source and summary. More nuanced  pragmatic and semantics are often not necessary but should be covered in future work with LLMs. Secondly, we do not concern ourselves with inconsistencies in the source notes. Most models at the time of this thesis are not well equipped to reason about contradictions across longitudinal records. We view source note resolution and re-phrasing as a pre-processing step for summarization, since it will be necessary to any patient-centric NLP task.

Research in general domain summarization has found that training on unsupported reference summaries, unsurprisingly, leads to unfaithful downstream models \parencite{nan-etal-2021-entity,narayan-etal-2021-planning}. The colloquial phrase that best captures this phenomenon is: ``garbage in, garbage out''. A common approach to deal with training noise is \emph{filtering}: to identify and ignore low quality text at the reference \parencite{kang-hashimoto-2020-improved, nan2021entity, narayan2021planning} or span level \parencite{goyal2021annotating}. Yet, these methods largely work because they are applied to clean, large-scale corpora. For instance, after removing references with ``entity hallucinations'', \textcite{nan2021entity} still have 855k (92\% of the original) training examples for Newsroom, 286k (99\%) for CNN/DM, 135k (66\%) for Xsum. Different quality metrics are used: missing entities \parencite{nan2021entity, narayan2021planning}, entailment predictions \parencite{matsumaru-etal-2020-improving,goyal2021annotating}, or online indicators, such as log loss \parencite{kang-hashimoto-2020-improved}.

\begin{table}[t]
\centering
\footnotesize
\begin{tabular}{clc}
\hline
& \textbf{Statistic} & \textbf{Value}\\
\hline
\multirow{2}{*}{Global}
& Notes & 1.38M \\
& Unique Patients & 47,553 \\
\hline
\multirow{6}{*}{\makecell{Per \\ Admission\\ (Avg. \#s)}}
& Notes & 29 \\
& Note types & 3.3 \\
& Source sentences & 703 \\
& Source tokens & 7,553 \\
& Reference sentences & 23.5 \\
& Reference tokens & 370 \\
\multirow{3}{*}{\makecell{Extractive \\ Analysis}}
& Coverage & 46.0 \\
& Density & 1.2 \\
& Compression Ratio & 20 \\\hline
\end{tabular}
\caption{Hospital-Admission Summarization Dataset.}\label{tab:mimic-data-again}
\end{table}

\begin{table}[t]
\footnotesize 
\centering
\begin{tabular}{ccclc}
\hline
& \textbf{Type} & \textbf{Source} & \textbf{Ref.} & \textbf{Supported} \\
\hline
\multirow{1}{*}{\makecell{Per Visit}}
&  Mentions & 1,181 & 50 & 30 \\\hline
\multirow{5}{*}{\makecell{Mention \\ Share \\ By Type}}
&  Diagnosis & 61\% & 42\% & 49\% \\
&  Test & 23\% & 16\% & 14\% \\
&  Medication & 8\% & 15\% & 11\%  \\
&  Procedure & 5\% & 10\% & 13\% \\
&  Treatment & 3\% & 8\% & 5\%  \\ \hline
+ & & 100\% & 100\% & 100\% \\ \hline
\end{tabular}
\caption{Entity-level breakdown by entity type in source and reference (Ref.). Supported is the intersection of reference and source entities. The average entity novelty fraction is high $40\% = \frac{50 - 30}{50}$.}\label{tab:ent-break}
\end{table}

Yet, some datasets--such as ours--are so pervasively noisy that filtering would be too draconian. In this Chapter, we create and experiment on a hospital-course summarization dataset from the only viable publicly available dataset: MIMIC-III, which contains de-identified patient records from ICU patients at Beth Israel Deaconess Medical Center in Boston, Massachusetts. This dataset contains high-levels of unsupported content yet is critical for the advancement of hospital-course summarization research given that it is the only publicly available dataset. This chapter does not present results for our proprietary dataset from Columbia University Medical Center (CUIMC) due to the fact that the research conducted for the two papers discussed in this chapter were completed during internships with Amazon (\S \ref{sec:ltr}) and Microsoft Research (\S \ref{sec:calibration-section}).

The publicly available MIMIC-III dataset contains de-identified clinical records from patients admitted to Beth Israel Deaconess Medical Center \parencite{johnson2016mimic}. We follow the same protocol for creating the MIMIC dataset as we did with our proprietary dataset from CUIMC. Specifically, we treat the Brief Hospital Course section of the discharge summary as a reference and all notes authored during the patient's stay as the source. Table \ref{tab:mimic-data-again} shows that source documents are long (on average, 7.5K tokens and 703 sentences) and, while references are also long (370 tokens on average and 23.5 sentences), there is a high degree of word-level compression ($\sim$20x). Coverage and density metrics reveal high levels of abstraction\footnote{Please refer to \textcite{grusky2018newsroom} for details on extractive analysis, including formulas for density and coverage.}. Similar to previous work \parencite{shing2021towards, nan2021entity}, we motivate the work in this chapter by computing and analyzing (un)supported content at the entity level. \href{https://aws.amazon.com/comprehend/medical/}{Amazon Comprehend Medical} is used to extract entities of the following semantic types: diagnoses (using the ICD-10 classification), medications (RxNorm), procedures, treatments, and tests. To determine whether or not an entity is supported, we compute a similarity score for all mention pairs based on a combination of ontological (RxNorm/ICD-10 codes) and lexical overlap (embedding and exact match).

In \textbf{\S \ref{sec:ltr}}, to improve reference quality while retaining all data, we propose a new approach: to selectively re-write \emph{un}supported reference sentences to better reflect source data. We automatically generate a synthetic dataset of positive and negative revisions by corrupting supported sentences and learn to revise reference sentences with contrastive learning. The intensity of revisions is treated as a controllable attribute so that, at inference, diverse candidates can be over-generated-then-rescored to balance faithfulness and abstraction. According to metrics and human evaluation, models trained on revised clinical references are much more faithful, informative, and fluent than models trained on original or filtered data. In \textbf{\S \ref{sec:calibration}}, we take it a step further by moving away entirely from MLE-based training on human-written references.  We demonstrate that through careful construction of calibration sets-- summaries of varying quality formed offline--we can directly learn to generate more faithful and relevant summaries. In short, we explore a large set of methods for constructing these sets and directly calibrate to faithfulness and relevance-defined rankings. A human evaluation demonstrates that meaningful gains in both relevance and faithfulness are achievable through careful construction of calibration sets.

 The contributions of this chapter are to:
 
 \begin{itemize}
     \item Propose a new method to address variable reference quality: reference revision, which, as a data pre-processing step, is model agnostic and complementary to other faithfulness approaches.
     \item Show that training on revised references can improve faithfulness while also improving informativeness and fluency.
     \item Benchmark calibration models on three scientific long-form datasets, including a new, chemistry-focused corpus, for which we collect fine-grained faithfulness annotations and relevance rankings from experts.
     \item Conduct extensive experiments to better understand the underlying characteristics and dynamics of effective calibration tuning sets. We release easily extensible code for forming and optimizing calibration sets in the scientific domain.
     \item Release a new abstract generation corpus which was scraped from the Open Access subsets of eight chemical journals.
 \end{itemize}

\section{Learning to Revise References} \label{sec:ltr}

\textit{This section focuses on \textcite{adams-etal-2022-learning}, a first-author publication which appeared at EMNLP Findings in 2022.}

\input{4a-revision}

\section{Calibrating Models to Metrics}\label{sec:calibration}

\textit{This section focuses on ``What are the Desired Characteristics of Calibration Sets? Identifying Correlates on Long Form Scientific Summarization'' \parencite{adams-etal-2023-desired}, for which I was first-author and presented as an oral spotlight at ACL in 2023.}

\input{4b-calibration} \label{sec:calibration-section}

\section{Related Work}

\paragraph*{Improving Faithfulness.} Efforts to improve summary faithfulness can be roughly delineated into three categories based on where they land in the training pipeline: \textbf{dataset pre-processing}, \textbf{faithfulness-aware training}, and \textbf{post-hoc editing}.

We discussed the first in \ref{sec:ltr} and proposed a novel solution: to address noise in reference summaries by re-writing, or revising them. The most common proposal in the literature prior to our work was to identify low quality samples--with entailment \parencite{matsumaru-etal-2020-improving, goyal2021annotating} or entity overlap \parencite{nan2021entity, narayan2021planning}--and drop them. These filtering methods tend to improve faithfulness yet can degrade informativeness. \textcite{kang-hashimoto-2020-improved} address the data hunger issue by first training on all samples before implementing Loss Truncation--ignoring high log loss datapoints. This is effective on a relatively clean Gigaword corpus yet untested on noisier corpora in which hallucination behavior from the unfiltered data may be difficult to unlearn. Filtering can be ``insufficient to train a competitive model'' when high-quality data is limited \parencite{filippova-2020-controlled}. Our proposed method takes advantage of all available data while seeking to redress the underlying issue.

On the training side, given the recent success of aligning LLMs to human preferences via RLHF \parencite{stiennon2020learning} or DPO \parencite{rafailov2023direct}, improving faithfulness via alignment is a highly promising area of research.  The typical RLHF setup for pre-training is highly expensive, as a large dataset of pairwise human preferences is required to train a high-quality reward model for RLHF.  Yet, LLMs such as GPT-4 are becoming increasingly effective as evaluators, and, can be used to replace humans as preference labelers (RL-AI-Feedback -- \parencite{lee2023rlaif}).

Typically, when summarization models are aligned to be more faithful, little to no human annotation is used. Synthetic errors are directly inserted into summaries to directly create negative contrast sets \parencite{cao-wang-2021-cliff}. Contrastive learning for faithfulness is well-established \parencite{cao-wang-2021-cliff,nan-etal-2021-improving,tang-etal-2022-confit} and pre-dates work on RLHF. Contrastive learning, or calibration, is typically also used to increase the relevance of downstream summaries. For relevance calibration, models are typically calibrated to the ROUGE scores of their own outputs after an initial fine-tuning step \parencite{liu-liu-2021-simcls, liu-etal-2022-brio}. \textcite{zhao2022calibrating} extend the work of \textcite{liu-etal-2022-brio} and run a broad sweep of loss functions and candidate generation methods for two-step relevance calibration while establishing state of the art performance (ROUGE) across single document corpora. As opposed to contrasting positives and negatives in a latent space, these models are instructed to calibrate decoder likelihoods to ROUGE or BERTScore-defined rankings. While best results are seen for continued fine-tuning with a calibration objective, a second model can be used either to re-rank \parencite{liu-etal-2021-refsum, ravaut-etal-2022-summareranker} or fuse \parencite{ravaut-etal-2022-towards} candidate summaries.

Our work in \S \ref{sec:calibration} is distinct along three key dimensions: \textbf{(1)} we consider long-document scientific summarization, rather than single-document; \textbf{(2)} we consider both faithfulness and relevance calibration and analyze the interactions between the two, often competing, quality objectives; \textbf{(3)} we uncover relationships between key set statistics and downstream performance by systematically varying how calibration sets are formed from candidate pools.

Finally, post-hoc interventions involve re-ranking \parencite{falke2019ranking} over-generated summaries or editing, which is typically weakly supervised with synthetic datasets \parencite{cao-etal-2020-factual,dong-etal-2020-factcorrect,zhu-etal-2021-enhancing}. In the world of LLMs, however, post-hoc editing does not require supervision. It can be done quite effectively with prompting as a self-critical, or reflection, step \parencite{gou2023critic}. In \S \ref{sec:calibration}, we demonstrate how our reference revision model can be used without modification to fix model-generated summaries in the same way as its used to revise human-generated reference summaries.

\paragraph*{Faithful Clinical Summarization.} Faithfulness is less studied for clinical summarization. Abstractive approaches tend to focus on finer temporal granularities, e.g., synthesizing a single radiology report \parencite{macavaney2019ontology, sotudeh2020attend, zhang-etal-2020-optimizing} and doctor-patient conversations \parencite{krishna2020generating,joshi-etal-2020-dr, zhang2021leveraging, karn2022differentiable,chintagunta2021medically}.  \textcite{shing2021towards} survey extract-then-abstract approaches to section-specific discharge summarization on MIMIC-III. They measure factuality with entity overlap and remove poorly supported references.

\section{Conclusions and Limitations}

In this section, we demonstrate two methods for improving faithfulness: reference revision and metric-based calibration, and measure improved faithfulness of hospital-course summaries with metrics and human evaluation. A fundamental limitation of both studies is that they do not use Large Language Models (LLMs), which have demonstrated remarkable performance as off-the-shelf summarizers in the general domain \parencite{zhang2023benchmarking}, as well as in the clinical domain \parencite{waisberg2023gpt, abacha2023overview}. Beyond summarization, generalist LLMs have even surpassed specialized, fine-tuned models on medical knowledge and reasoning tasks \parencite{fleming2023medalign, nori2023can}. Where does this leave us with respect to the future of hospital-course summarization research and, specifically, ensuring faithfulness of AI-generated summaries? We list out the pros and cons of focusing on prompting generalist models versus fine-tuning / alignment. We note that medium sized LLMs can be fine-tuned with modest computing budgets yet the largest and most powerful models are often only usable in the zero-shot setting due to compute budgets and/or the closed nature of some models (GPT-4, Claude).

\paragraph*{In Favor of Generalist Zero-Shot.} We have established that clinical references are noisy and that simple fine-tuning of models on these references leads to undesirable behaviors, such as, increased hallucination. Prompt-based solutions are simple and require no costly fine-tuning.  Given the ability of LLMs to diagnose errors, many faithfulness errors can be reasonably avoided by encouraging grounded generation (e.g., with explicit citations \parencite{gao2023enabling}) and can be addressed after the fact with self-correction.

\paragraph*{In Favor of Specialized Models.} The hospital-course summarization task, as is the case with many EHR documentation tasks, is highly esoteric--both in terms of content selection, as well as style. The content and presentation is highly specialized and learning these patterns is most easily achieved with some degree of fine-tuning. Additionally, fine-tuning allows one to align models to be safe, harmless, accurate, and useful, which are all factors not guaranteed by zero-shot prompting. The most performant generalist models are closed-source (GPT-4, Claude) and are only accessible through APIs. Using API-based models on patient sensitive data is only possible through the use of specialized HIPAA-compliant offerings.

\paragraph*{Future Research.} Given the nature of the task, and the relative improvements seen from domain adaptation \parencite{lehman2023we}, we believe the most performant and faithful hospital-course summarization models will require some level of fine-tuning. Given fast adaptation to new tasks for instruction-tuned base models (Flan-T5 \parencite{longpre2023flan}), and the quality issues with existing references (\textbf{\S \ref{sec:ltr}}), we believe that only a limited number of high-quality human written examples will be necessary for LLMs to adapt to the style and content requirements of hospital-course summarization. This will require advanced metrics to identify high coverage, high fidelity reference summaries. Given an adapted, lightly tuned model, alignment will be necessary to ensure a model that is not only safe for deployment, but tailored to the precise needs of clinicians. This will likely require an expensive, yet necessary, exercise to procure pairwise preference annotations from clinicians, who will be the end users of any deployed system. High-quality preference data can serve as a meta-evaluation benchmark to test how well LLMs are at assessing quality themselves. More and more, "AI" is seen as a valid replacement for the "H" in HF. Anthropic's RL\emph{AI}F approach--which uses self-critical feedback, or Constitutional AI, from a strong base LLM, has led to a proliferation of smaller open-source models, e.g., Zepyr \footnote{\url{https://huggingface.co/HuggingFaceH4/zephyr-7b-beta}.} and Tulu \parencite{wang2023far}), which rely heavily on synthetic data. Either way, our work in \textbf{\S \ref{sec:calibration}} has shown that alignment, or calibration, can be accomplished with simple objectives and can be greatly improved with careful selection of summary candidates for comparison. Our work is complementary to state of the art work in RLHF, as it focuses on the data acquisition side. It can be useful for any method requiring multiple candidates for RLAIF or RLHF, and paired with any alignment objective, e.g., BRIO, SLiC, DPO, PPO.

\paragraph*{Marrying Both Papers.} An interesting angle to explore to combine our work in \S \ref{sec:ltr} and \S \ref{sec:calibration-section} would be to use good and bad clinician-authored references as the dataset for soliciting preferences. It remains to be seen whether or not the candidates need to come from a strong base model or can be slightly out-of-distribution, e.g., clinician-authored. The push for synthetic data is borne out of a need to cover the long-tail of knowledge. Yet, given the diversity of hospital settings, and the fact that our reference summaries are generated for \emph{each} hospital admission, it stands to reason that there will be sufficient data diversity in naturally occurring human-written reference summaries. Using what we have--many clinician authored summaries---is a good next step.

%% file: 4a-revision.tex
\begin{figure}[t]
\centering
\includegraphics[width=0.5\linewidth]{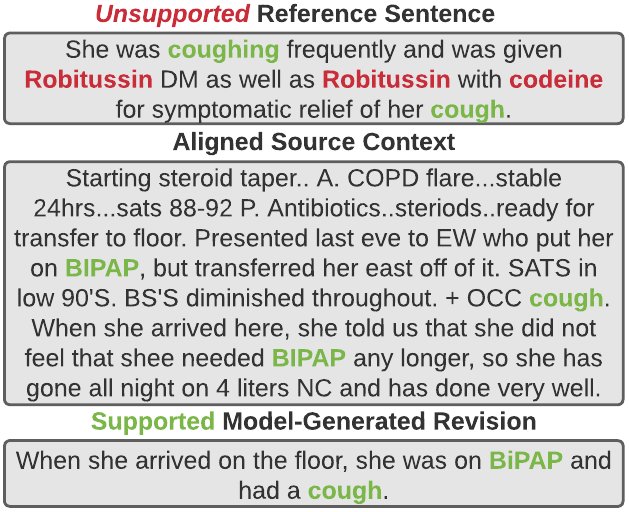}
\caption{Example of a revised reference sentence. \textbf{\textcolor{red}{Robitussin}} and \textbf{\textcolor{red}{codeine}} are edited out of the sentence, while \textbf{\textcolor{Green}{cough}} is correctly kept and a new supported entity \textbf{\textcolor{Green}{BiPAP}} is added. The model is trained on synthetic data to reconstruct well-supported sentences based on context and diverse model-generated hallucinations, which are over-generated to serve as both noisy inputs and negative outputs for contrastive learning.}
\label{fig:revision-example}
\centering
\end{figure}

The tendency of abstractive systems to produce unfaithful summaries is well-studied \parencite{maynez-etal-2020-faithfulness}, yet less attention has been paid to the role of the data on which the models are trained. This is problematic for two reasons: \textbf{(1)} many corpora are naturally occurring--not created for training models--and, as such, are noisy \parencite{kryscinski2019critical} and without ``inherent quality guarantees'' \parencite{bommasani-cardie-2020-intrinsic}; \textbf{(2)} noisy data is detrimental to training faithful models \parencite{duvsek2019semantic}; \textbf{(3}) simple methods to identify low quality references can be quite effective--often simpler than building smarter models \parencite{bommasani-cardie-2020-intrinsic}.

Our thesis task: hospital-course summarization, does not lend itself to filtering given the pervasive level of noise in our references--almost every reference has some degree of unsupported content. As such, we propose a new approach: to revise--not remove--noisy reference content. First, we align each reference sentence to 1-5 sentences in the source text and classify it as \emph{supported} (to be left alone) or \emph{unsupported} (to be revised). Our objective is to revise all unsupported reference sentences in such a way that retains faithful content, removes unfaithful content, and, as needed to preserve length, adds relevant context. An example output is shown in Figure \ref{fig:revision-example}. In a coherent sentence, the model removes unsupported entities (\textbf{\textcolor{red}{Robitussin}}, \textbf{\textcolor{red}{codeine}}) and introduces a twice mentioned concept from the context (\textbf{\textcolor{Green}{BiPAP}}). To learn this revision task, we need examples of supported and unsupported reference sentences for the same context. Without directly observing it, we generate synthetic data. At a high-level, we take each supported sentence, corrupt it to form a diverse set of unsupported alternatives, and use this mix of real and synthetic data to create examples of (un)faithful revisions for contrastive learning. To account for the abstractive-faithful tradeoff \parencite{durmus-etal-2020-feqa, ladhak2021faithful}, we treat the level of extractiveness as a controllable attribute with learned style codes \parencite{fan2017controllable}. During inference, we vary the control codes to over-generate candidate revisions before selecting one which balances abstraction and faithfulness.

\subsection{Building Source-Reference Alignments} \label{sec:alignment}

We link each reference sentence to a subset of source sentences to identify the minimal context necessary for revision and determine which sentences need to be revised. We select no more than five sentences because \textcite{lebanoff-etal-2019-scoring} find that reference sentences tend to reflect content from very few source sentences. We follow their approach to greedily select sentences with high ROUGE \parencite{lin2004rouge} overlap and minimize redundancy by removing covered tokens after each step. Yet, given high levels of abstraction (abbreviations \parencite{adams2020zero}, misspellings), we increase semantic coverage by replacing ROUGE with BERTScore precision \parencite{zhang2019bertscore} and adding additional sentences, as needed, to cover all supported entities. We compute BERTScore using the mean-pool of the last four layers of ClinicalBERT \parencite{alsentzer2019publicly}.

\begin{figure*}[t]
\centering
\includegraphics[width=0.75 \linewidth]{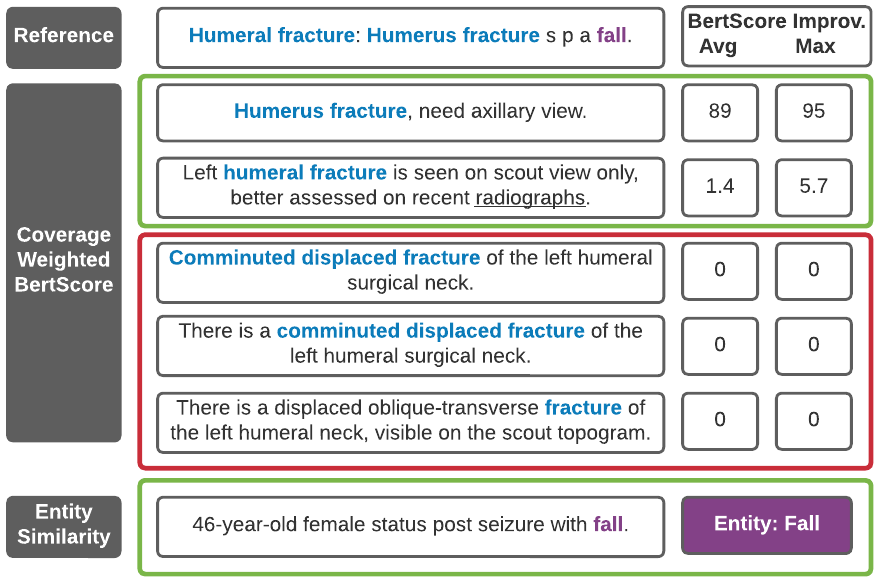}
\caption{Source-Reference Alignment. 5 source sentences are greedily extracted with a coverage-weighted BERTScore heuristic. Non-influential sentences are \textcolor{red}{discarded} (low BERTScore improvement). Finally, in the case of missing clinical concepts (\textbf{fall}), we find the source sentence whose contextualized representation of the entity span (\textbf{fall}) is closest to the reference usage (the last word of the top reference row). }
\label{fig:alignment}
\end{figure*}

\paragraph*{Alignment Algorithm.} \label{app:alignment}

Figure \ref{fig:alignment} provides an example alignment with improvement filtering and an extra extraction step to ensure full entity coverage.

\paragraph*{Notation.} Let $\braket{(\bm{S_1}, \bm{R_1}), ..., (\bm{S_N}, \bm{R_N})}$ represent the corpus, consisting of source-reference pairs for $N$ unique patient ICU admissions. Let $\bm{S_n} = \braket{\bm{s_1^{n}}, ..., \bm{s_{|S_n|}^{n}}}$ represent the sentences extracted from the source input for the $n^{th}$ example and, likewise, $\bm{R_n} = \braket{\bm{r_1^n}, ..., \bm{r_{|R_n|}^n}}$ the reference sentences. Similarly, $\bm{s_i^{n}} = \braket{x_1, ..., x_{|s_i^n|}}$ is the tokenized sequence for the $i^{th}$ source sentence from the $n^{th}$ example, and $\bm{r_j^{n}} = \braket{\hat{x}_1, ..., \hat{x}_{|r_j^{n}|}}$ the tokenization of the $j^{th}$ reference sentence.

Given very long inputs, we link each reference sentence $\bm{r_{j}^{n}}$ ($n \in N$, $j \in \bm{R_{n}}$), to a small subset ($\leq 5$) of source sentences corresponding to the same example. Due to the abstractiveness of the data (acronym usage, shorthand, etc.), as well as redundancy from copy-and-paste \parencite{hirschtick2006copy}, we align sentences using a new approach which combines BERTScore and subword-level coverage, rather than the conventional approach of lexical overlap with ROUGE \parencite{lebanoff2019analyzing, liu2019hierarchical}. Given a candidate alignment pair: a reference sentence $\bm{r_j^n}$ with $K$ tokens and a source sentence $\bm{s_i^n} \in \bm{S_n}$ with $L$ tokens, for each reference token $\hat{x}_{k}$, we find its closest match in $\bm{s_i^n}$:

$$
\text{align}(\hat{x}_k, \bm{s_i^n}) = \underset{1 \leq \ell \leq L}{\max}cos(h(\hat{x}_k), h(x_{\ell}))
$$

where $h(x)$ represents the contextualized BERT embedding\footnote{The mean-pool of the last four layers of ClinicalBERT.}. Based on these greedy alignments, we extract sentences for $T$ steps. At $t=0$, we initialize an importance vector $\bm{w}$ of length $K$, to all ones. Then, at each timestep, we select $\bm{s^*} \in \bm{S_n}$ which maximizes the importance-weighted BERTScore:

$$
\bm{s^*} = \underset{\bm{s^*} \in \bm{S_n} }{\mathrm{argmax}}({\frac{\sum_{k=1}^{K}{w_{tk} align(\hat{x}_k, \bm{s^*})}} {\sum_{k=1}^{K}{w_{tk}}})}
$$

After each timestep, we keep track of the best alignment score for each reference token via the importance vector. In particular, we update each token's importance by the inverse of its best coverage score: $w_{t+1,1} = min(w_{t1}, 1 - align(\hat{x}_1, \bm{s^*}))$ (formula shown for first element). Similarly to Maximal Marginal Relevance (MMR) \parencite{carbonell1998use}, the importance vector de-prioritizes well-covered tokens for future extractions. We also remove $\bm{s^*}$ from $\bm{S_n}$ to ensure it is only extracted once. After retrieving the top $K$ using this procedure, we only use sentences for the final alignment set for which the average coverage improvement $\geq 0.01$ or the max $\geq 0.05$, where improvement is defined as the reference token-level increase in coverage of the latest extraction over the previous max from prior extractions: $max(0, \bm{w^t} - \braket{align(\hat{x}_1, s^*), ..., align(\hat{x}_K, s^*)})$.

Infrequently, the medical concepts extracted from the aligned sentences do not cover all the concepts in the reference sentence. In this case, for each missing concept, we filter for the subset of source sentences containing that concept, and add the sentence with the highest pairwise similarity of contextualized concept embeddings--the mean of hidden states corresponding to the entity span.

\begin{figure*}[t]
\centering
\includegraphics[width=\linewidth]{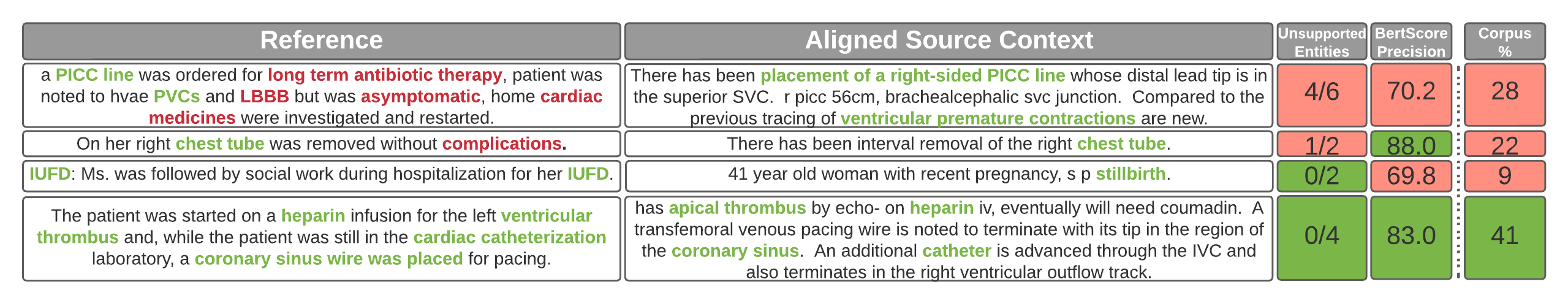}
\caption{From top to bottom: examples of source-reference alignments which fail to meet both criteria for supportedness (only supported entities and a high BERTScore precision) (28\% of corpus reference sentences), just entity overlap (22\%), just low BERTScore (9\%), and the remaining which meet both (41\%). Only the last reference sentence is classified as \emph{supported} and, for this paper, treated as a gold standard when training revisions. } 
\label{fig:quality}
\end{figure*}

\paragraph*{Merging Entities} \label{sec:entity-merging} For each pair of entity mentions in the source text and reference, we compute a code overlap score. Let entity $e_x$ have codes $\bm{c_x}$ and tokens $\bm{t_x}$ and $e_y$ have codes $\bm{c_y}$ and tokens $\bm{t_y}$, the pairwise code overlap score is:

$$
CodeOverlap(e_x, e_y) = \frac{|\bm{c_x} \cap \bm{c_y}|}{|\bm{c_x}| + |\bm{c_y}|}
$$

Then, we compute embedding cosine similarity between mention tokens with BioWordVec \parencite{biowordvec}, filtering out stopwords and punctuation. Let

$$
EmbedOverlap(e_x, e_y) = cosine(E(t_x), E(t_y))
$$

Finally, we compute the TF-IDF overlap ($TF\_IDF(t_x, t_y)$) to compensate for occasional noise in embedding space. We define the aggregate score as the average of embed, code, and TF-IDF overlaps. For entity types with no available codes (treatments, procedures, and tests), we only examine mention overlap. Based on validation on a manually labeled held-out set, we classify $e_x$ and $e_y$ as synonyms iff any of the following thresholds are met: $CodeOverlap(e_x, e_y) \geq 0.4$, $EmbedOverlap(e_x, e_y) \geq 0.75$, or $AggOverlap(e_x, e_y) \geq 0.4$.

\paragraph*{Classifying References.} We treat reference sentences with $0$ unsupported entities and a BERTScore precision with respect with its aligned source evidence of $\geq 0.75$ as supported. The remaining are unsupported. 417,318 (41\%) reference sentences qualify as supported and the remaining 595,300 (59\%) unsupported: 47\% fail both thresholds (280,839), 38\% have hallucination(s) with a high BERTScore (225,423), and 15\% have no hallucinations but a low BERTScore (88,189). Figure \ref{fig:quality} reveals why both BERTScore and entity overlap are needed to identify full coverage. The first sentence has unsupported entities and poor embedding-based scores. The second is semantically similar yet missing a critical concept (\textcolor{red}{complications}) which cannot be inferred from the context. The third has full entity coverage (\textcolor{Green}{IUFD} is a term for \textcolor{Green}{stillbirth}) yet BERTScore is low because there is no mention of social work. Only the final sentence is covered by both metrics and treated as supported.

\subsection{Learning to Revise Unsupported Text} \label{sec:revise-pipeline}

The goal is to re-write these \emph{un}supported reference sentences such that they are supported, i.e., covered by the source notes. To learn this revision task without a gold standard ground-truth, we take each supported reference sentence, inject noise to create unsupported, yet realistic, alternatives, and then use this mix of real and synthetic data to create a supervised set of positive and negative revisions to support a contrastive learning objective\footnote{We do not rely on \emph{un}supported sentences during training because they are set aside for inference. To use them, we would need to synthetically construct supported alternatives, which is not possible without first knowing how to revise.}.

\paragraph*{Generating \textit{Synthetic} Hallucinations}

\paragraph*{(D)esiderata.} Based on Figure \ref{fig:quality}, unsupported sentences look normal (unfaithful only in the context of available data) \textbf{(D1)}; contain many hallucinated entities \textbf{(D2)}; exhibit a wide range of semantic divergence from the aligned source context \textbf{(D3)}; and in spite of clear differences, are topically similar to aligned context \textbf{(D4)}. To illustrate \textbf{D3}, we note that the second sentence could be revised by simply removing the bigram ``without complications'', yet the first and third sentences require more substantial re-writing to remove unfaithful content while still remaining informative and coherent.

\paragraph*{High-Level.} The simplest way to construct, and control for, hallucinations is to perform entity swaps \parencite{kryscinski-etal-2020-evaluating,zhao2020reducing, zhang2021fine,chen2021improving}. Yet, this can produce disfluent text which is easily detectable \parencite{goyal2021annotating}. In contrast, generating from a LLM produces fluent \parencite{zhou2020detecting}, more diverse text \parencite{cao2021cliff}, yet without as much control over hallucinations. Given our desiderata, we combine entity swaps \textbf{(D2)} into a generative framework \textbf{(D1, D3)} and incorporate a set of topical entities to avoid excessive semantic drift \textbf{(D4)}. Our proposed method is called \textbf{ReDRESS}: \textbf{re}ference \textbf{d}ist\textbf{r}actor \textbf{e}ntity \textbf{s}et \textbf{s}wapping.

\paragraph*{Training Objective.} The \textbf{ReDRESS} backbone is a BART encoder-decoder model \parencite{lewis-etal-2020-bart}. BART is trained as a denoising autoencoder to reconstruct a corrupted sentence: $p(s|f(s))$, where $f$ is an arbitrary noise function(s). Across natural language tasks, the BART authors find span deletion to be the most effective noise. \textbf{ReDRESS} also uses span deletion but adds an extra noise function: entity swaps. Specifically, for each sentence $s$, we extract a set of topically related entities $\bm{e_s}$ and then exchange entities between $s$ and $\bm{e_s}$. Let us denote span deletion as $f$ and the swap transformation as $g(s,\bm{e_s}, k) \rightarrow {\bm{e_s}}_{-k}^{+k}, s_{-k}^{+k}$, where $k$ represents the number of entities exchanged. To vary the level of corruption (\textbf{D3}), we sample a different $k$ for each example such that, on average, half of the entity mentions in $s$ are swapped out for entities in $\bm{e_s}$. The final pre-training objective is $p(s|k,{\bm{e_s}}_{-k}^{+k}, f(s)_{-k}^{+k})$. $k$ is represented by a special token and added to the input to make entity swaps a controllable aspect of generation. Each component ($k$, ${\bm{e_s}}_{-k}^{+k}$, $f(s)_{-k}^{+k}$) is separated by a special \texttt{<sep>} token and passed to the BART encoder. During training, the decoder learns to reconstruct $s$ by re-writing $f(s)_{-k}^{+k}$ such that it reverses the $k$ entity swaps performed between $\bm{e_s}$ and $s$ and fills in a deleted text span\footnote{$\bm{e_s}$ is provided as input yet is not part of the target output. In other words, the output of the model is a natural sentence.}.

\paragraph*{Data to Train ReDRESS.} We pre-train on a large unlabeled sentence corpus: all sentences extracted from MIMIC-III discharge summaries, excluding notes related to patients in the summary test set. To minimize EHR-related noise, we filter out sentences without any clinical concepts and those found in non-narrative sections related to structured data, demographics, and/or administration (billing codes, dates, times, signatures, lab values).

\begin{figure*}[t]
\centering
\includegraphics[width=\linewidth]{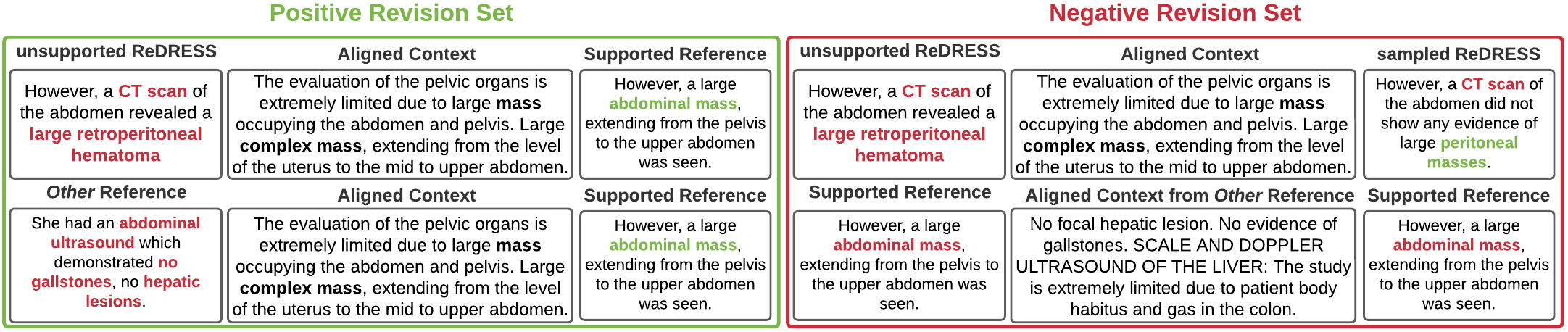}
\caption{Synthetic \textcolor{Green}{positive} and \textcolor{red}{negative} sets for revision training. The encoder input is the concatenation of the input (first box) and source context (second box), while the (un)faithful revision target is the third. Entities from inputs and targets are colored as \textcolor{red}{un}\textcolor{Green}{supported} \textbf{\emph{relative}} to the provided context. \textit{Sampled ReDRESS} is a randomly sampled synthetic hallucination, while \textit{Unsupported ReDRESS} is the sample most unsupported by \textit{Aligned Context}.}
\label{fig:revise-train-example}
\end{figure*}

\paragraph*{Inference.} To use \textbf{ReDRESS} to generate a plausible, corrupted version of a sentence $s$, we apply $f$ to $s$ and sample $k$ to apply $g$ to both $s$ and its distractor set $\bm{e_s}$. Two key \textbf{m}odifications are introduced to discourage the model from reconstructing $s$: \textbf{(m1)} entities removed from $s$ are not added to $\bm{e_s}$, and \textbf{(m2)} $k$ swaps are implemented, yet the model is provided $k+1$ as the swap code to trick the model into performing an additional swap than is required for reconstruction. Using the notation above, \textbf{ReDRESS} generates $(k+1, {\bm{e_s}}_{-k}, f(s)_{-k}^{+k}) \rightarrow \hat{s}$ using standard beam search decoding. Without access to the original entities from $s$ in $\bm{e_s}$ \textbf{(m1)}, the model looks for plausible, \textit{hallucinated} alternatives.

\paragraph*{Implementation Details.} We train \textbf{ReDRESS} from \texttt{bart-base} on a large set of unlabeled sentences extracted from MIMIC-III discharge summaries. To find the distractor set $\bm{e_s}$ specific to each $s$, we first retrieve the sentences most similar to $s$ (using BioSentVec \parencite{chen2019biosentvec} embeddings to create a Faiss index \parencite{JDH17}). The first 25 unique concepts extracted from the set of nearest neighbors form the distractor set $\bm{e_s}$ for $s$.

\subsection{Learning a Revision Model} \label{sec:reviser}

We apply \textbf{ReDRESS} to the set of supported reference summary sentences to generate positive and negative revision examples for contrastive learning.

\paragraph*{Notation.} Let $\bm{r}$ represent a reference sentence with aligned source sentences $\bm{S}$. $\bm{\hat{r}_n}$ is a corrupted version of $\bm{r}$ generated by \textbf{ReDRESS} with random seed $\bm{n}$. Given $N$ diverse outputs, each generated from its own sampled set of corruptions, $\bm{\hat{r}_u}$ is the most \textbf{u}nsupported (lowest BERTScore precision). 

\paragraph*{Training.} The input to the \textbf{reviser} model is the concatenation of a noisy input and aligned source context, while the output is a (un)faithful revision. We rely on \textbf{ReDRESS} to generate noisy inputs \textit{and} unfaithful outputs. \textbf{ReDRESS} hallucinations require moderate levels of revision to reconstruct the original supported input. This makes sense for most of the \textit{observed} unsupported sentences. Yet, sometimes, a reference sentence is almost entirely unsupported. In such cases, the model should effectively learn to ignore it and just summarize the aligned context. To simulate this more drastic scenario, we also retrieve a \textbf{random} reference sentence ($\bm{r^*}$), and its aligned source ($\bm{S^*}$), from the same example. Our ablation study in Table \ref{tab:results-ablation} shows that both sources of hallucinated content (\textbf{ReDRESS}-generated and random mis-alignments) are complementary, necessary to achieve the best downstream summary results.

Using the notation above, as tuples of format (input, context, target), the positive set is: $(\bm{\hat{r}_u}, \bm{S}, \bm{r})$ and $(\bm{r^*}, \bm{S}, \bm{r})$. The negative set is: $(\bm{\hat{r}_u}, \bm{S}, \bm{\hat{r}_{c \in N}})$ and $(\bm{r}, \bm{S^*}, \bm{r})$, where $\bm{\hat{r}_{n \in N}}$ is a randomly selected corruption. In other words, for the positive set, we learn to generate a supported reference $\bm{r}$ from its aligned context ($\bm{S}$) and either a \textbf{ReDRESS} output ($\bm{\hat{r}_u}$) or another reference sentence ($\bm{r^*}$) as the synthetic, unsupported input. For the negatives, we discourage the model from generating (1) a synthetic hallucination ($\bm{\hat{r}_{n \in N}}$), and (2) itself if unsupported. Figure \ref{fig:revise-train-example} shows an example from the training data.

As in \texttt{ConSeq} \parencite{nan2021improving}, we optimize the likelihood of positives ($\bm{Z^+}$) and unlikelihood \parencite{welleck2019neural} of negatives ($\bm{Z^-}$):

\begin{align}
\label{eq:revision}
\begin{split}
  \mathcal{L}_{contrast} = \mathbb{E}_{\bm{Z^+}} log (p_\theta(\bm{r_{out}}|\bm{r_{in}}, \bm{S})) -\\
  \mathbb{E}_{\bm{Z^-}} log (1 - p_\theta(\bm{r_{out}}|\bm{r_{in}}, \bm{S}))
  \end{split}
\end{align}

$\bm{r_{in}}$ stands for the noisy reference input and $\bm{r_{out}}$ the revision target (positive or negative). We concatenate $r_{in}$ and $S$ with a special \texttt{<SEP>} token as the input, in addition to two key revision codes ($input_{frac}$ and $source_{frac}$) which we discuss next.

\paragraph*{Controlling Revision Intensity.} Some sentences require minimal edits to be fully supported while others require major edits. This variance is difficult to learn without an explicit control for it. Qualitatively, we noticed a tendency of the revision model to over-edit mostly supported sentences and under-edit highly unsupported content. Given this, we introduce the notion of revision intensity, parameterized by the fraction of words in the revision copied from the input ($input_{frac} = \frac{|\bm{r_{out}} \cap \bm{r_{in}}|}{|\bm{r_{out}}|}$), and the fraction copied from the aligned context ($source_{frac} = \frac{|\bm{r_{out}} \cap \bm{S}|}{|\bm{r_{out}}|}$). Intense revisions tend to require a larger lexical shift from input to source: a low $input_{frac}$ and a high $source_{frac}$. During training, we bin the fractions into deciles and include them as style codes prefixed to the encoder. Our ablation study in Table \ref{tab:results-ablation} shows that controlling the intensity of revisions to support diverse candidate generation, followed by re-scoring, has a massive impact on downstream summaries.

\paragraph*{Inference.}  We apply the trained reviser model to all unsupported reference sentences in the summarization training dataset. In particular, we concatenate each unsupported sentence as $r_{in}$ to its aligned context $S$ for beam search decoding. For this set of sentences, the desired revision intensity codes are unknown because no ground-truth revision exists ($r_{out}$). As a proxy, we fix $input_{frac} = \frac{|\bm{r_{in}} \cap \bm{S}|}{|\bm{r_{in}}|}$, which instructs the model to remove words from the input proportional to its lexical overlap with $\bm{S}$. Then, we vary $source_{frac}$ and over-generate 10 revision candidates with different levels of copy-paste from $\bm{S}$ and re-rank each candidate to select a final revision. In this way, the codes are useful both as a control mechanism and as a prompt for diverse generation. We experiment with two different scoring functions for re-ranking, which are discussed below as part of the experimental setup. 

\paragraph*{Implementation Details.} The \textbf{reviser} is trained from \texttt{bart-base} on the subset of references sentences classified as supported (417k, from \S \ref{sec:alignment}), and then used to over-generate revisions for the 595k \emph{un}supported sentences. The top scoring revision replaces the original sentence in the training data.

\subsection{Experimental Setup} \label{sec:experiments}

We design experiments around our central hypothesis: \emph{for a setting (long form hospital-course summarization), in which high-quality reference data is limited, reference revision is the best data-centric intervention to improve model faithfulness.} As such, we restrict the set of comparison methods to model-agnostic methods which explicitly address data quality. Based on our literature review, we consider two classes of baselines: those which \textbf{filter} low quality data \parencite{kang-hashimoto-2020-improved, narayan2021planning, matsumaru-etal-2020-improving, nan2021entity,goyal2021annotating}, and those which \textbf{control} for it \parencite{filippova-2020-controlled}. Other approaches, such as optimizing factuality metrics \parencite{zhang2019optimizing}, incorporating facts into modeling \parencite{zhu-etal-2021-enhancing}, decoding-based constraints \parencite{zhao2020reducing}, or post-hoc editing \parencite{dong-etal-2020-factcorrect} are complementary to our method, not competing.

\paragraph*{Reference Revision Strategies.} We experiment with two different functions to re-score over-generated candidate revisions: \textbf{Less Abstractive} selects the one with the highest BERTScore precision, while \textbf{More Abstractive} adds a penalty, based on the extractive fragment density \parencite{grusky2018newsroom}, to encourage more abstraction. We also consider a baseline revision approach: \textbf{Fully Extractive}, which replaces each unsupported reference sentence with the source sentence with the highest BERTScore overlap. Even though our dataset is highly abstractive, this is necessary to justify the complexity of abstractive revision.

\paragraph*{Baselines.} \textbf{(1) Filtered.} We experiment with three heuristics for low quality: references where no Admission Note is available in the source documents (\textbf{No Admission}) \parencite{shing2021towards}, references where a significant portion of the content is unsupported by the source notes ($<0.75$ token coverage or entity hallucination rate\footnote{\textcite{nan2021entity, narayan2021planning} rely on entities.} of $>10\%$) (\textbf{Unsupported}), and \textbf{Halluc. Ents}, which masks the training loss over spans of hallucinated reference entities. \textbf{Halluc. Ents} is inspired by \textcite{goyal2021annotating} who use a factuality model to ignore negatively entailed dependency arcs\footnote{They use the dependency-arc entailment (DAE) model \parencite{goyal-durrett-2020-evaluating} to identify inconsistent spans. Without such a model for clinical text, we use unsupported entities.}. Given the poor performance of other filtering strategies, we did not implement entailment-based filtering \parencite{matsumaru-etal-2020-improving}. We also implement \textbf{Loss Truncation} \parencite{kang-hashimoto-2020-improved}, which involves training for a fixed number of steps on the full training data before skipping high log loss examples. We grid-searched for the optimal number of warmup steps (2k) and the fraction of examples to drop (0.6) during truncation. \textbf{(2) Control Hallucination.} We implement the method in \textcite{filippova-2020-controlled}: group training data into quality buckets based on token coverage, and control for hallucinations with encoder-prefixed style codes.

\input{revision_tables/results_table}
\input{revision_tables/results_bart}

\paragraph*{Training Details.} We fine-tune downstream summarization models from BART \parencite{lewis2019bart} and the encoder-decoder Longformer \parencite{beltagy2020longformer}. We train all models for 10,000 steps or until convergence on the validation set. Some methods use revised or filtered training data yet all use the same 1,195 validation examples and evaluation test set (1,190). For the revised and original datasets, these models are trained on 45,168 training examples, 1,195 validation examples, and 1,190 test examples. We compute all evaluation metrics on the same test set (1,190) regardless of training data used.

\paragraph*{Metrics.} To measure source faithfulness, we compute the entity \textbf{hallucination rate} (HR) using the soft matching heuristic described in \S \ref{sec:alignment}, \textbf{BERTScore} (BS) using in-domain weights from Clinical BERT \parencite{alsentzer2019publicly}, and the fraction of summary sentences predicted as \textbf{entailed} by SciFive \parencite{phan2021scifive} fine-tuned on MedNLI \parencite{romanov2018lessons}\footnote{MedNLI is a clinician-annotated entailment dataset whose premise sentences come from MIMIC-III. SciFive is a biomedical T5 model that achieves SOTA performance on MedNLI.}. To capture entity coverage, we record \textbf{faithful-adjusted recall} (FaR): the fraction of non-hallucinated reference entities included in the output~\parencite{shing2021towards}. As in \textcite{koto2020ffci, adams-etal-2021-whats}, we approximate \textbf{coherence} with the average next-sentence prediction (NSP) probability between adjacent sentences from an in-domain BERT model, ClinicalBERT \parencite{alsentzer2019publicly}.

\subsection{Results} \label{sec:sum-results}

\paragraph*{Impact of Revisions on Summarization.}
Table \ref{tab:results} and \ref{tab:results-bart} confirm that filtering improves faithfulness (\textbf{Filtered - Unsupported} lowers the HR from 36.8 to 18.4 and improves entailment from 48.4 to 61.6), yet degrades coverage (48.2 vs 46.9 FaR). Masking hallucinated entities from log loss training (\textbf{Filtered - Halluc. Ents}) only slightly improves faithfulness, which underscores the difficulty in assigning token-level credit to pervasively noisy references. \textbf{Loss Truncation} leads to worse performance except on entailment (48.4 vs 51.7), which can likely be attributed to: 1) learning from fewer examples (from truncation); 2) hallucination patterns learned from the full data warmup are not unlearned during truncation; and 3) log loss might not capture faithfulness as well as more direct measures, i.e., entity overlap (used by \textbf{Halluc. Ents}).

In comparison, all revision approaches yield dramatic improvements in faithfulness (e.g., for \textbf{Less Abstractive}, 3.8/94.6.3/83.7 vs 36.8/82.3/48.4). These precision-oriented gains do not come at the expense of coverage (as defined by FaR), which actually jumps from 48.2 to 54.0. Abstractive revision (\textbf{Less} and \textbf{More}) outperforms \textbf{Fully Extractive} on coverage (e.g., 54.0/57.1 vs 52.4). Surprisingly, despite Fully Extractive revision being perfectly faithful, Less Abstractive revision leads to more faithful models (e.g., 3.8/83.7 vs 5.4/78.6 for HR and entailment, respectively), which suggests the need to re-write, not replace, unsupported content. Out of the revised results, More Abstractive has the best coverage (56.3/57.1) while being competitive on faithfulness (5.6 vs 5.4/3.8 HR).

\input{revision_tables/human_eval_table}

\paragraph*{Human Evaluation.} We rely on the protocol from \textcite{fabbri2021summeval} to procure expert judgments on \emph{Consistency}, \emph{Relevance}, \emph{Fluency}, and \emph{Coherence}. An experienced clinician was shown the source notes for 10 examples and 3 randomly shuffled Longformer outputs side-by-side (10 x 3), and asked to rank each summary on a 1-5 Likert Scale. We include the most faithful baseline according to automatic metrics: \textbf{Filtered - Unsupported}, as well as \textbf{Original} and \textbf{Revised}. \subsection{Human Evaluation Setup} \label{app:human-eval}

We solicit summary feedback on 4 critical dimensions: Consistency, Relevance, Fluency, and Coherence. The annotator was provided the following guidance on each metric:

\begin{itemize}
    \item \textbf{Consistency}: The rating measures whether the facts in the summary are consistent with the facts in the original article. Consider whether the summary does reproduce all facts accurately and does not make up untrue information.
    \item \textbf{Relevance}: The rating measures how well the summary captures the key points of the article. Consider whether all and only the important aspects are contained in the summary.
    \item \textbf{Fluency}: The rating measures the quality of individual sentences: are they well-written and grammatically correct? Consider the quality of individual sentences.
    \item \textbf{Coherence}: The rating measures the quality of all sentences collectively: do they fit together and sound naturally?  Consider the quality of the summary as a whole.
\end{itemize}

The annotator was asked to assess summaries with independent rankings for each of the 4 metrics along a 1-5 Likert Scale. Given the complexity of the task and resource constraints, we sampled a set of 10 summaries from the test set. We first discard length outliers: the 10\% of examples with the fewest source tokens (insufficient input) as well as the 10\% of examples with most source tokens (too difficult to evaluate), before sampling at random. It took the expert 4 days to review 10 patient charts. Table \ref{tab:results-human} shows that the model trained on Abstractively revised references produces the most Consistent, Relevant, \emph{and} Fluent summaries. Training on Filtered references improves consistency and relevance but degrades fluency given the relative paucity of training data (a similar finding for data-to-text generation is noted by \textcite{filippova-2020-controlled}). Assessed coherence is lower for models trained on Abstractively Revised data, yet we note that for 3/10 summaries, it held the highest coherence rating. Additionally, there was large variance in coherence ratings within each system (1.34 standard deviation), as each had at least one summary rated as 1 and at least one as 5. Consistency had the lowest (1.04). \textcite{adams-etal-2021-whats} argue that discourse is hard to evaluate because ``clinical summaries naturally exhibit frequent, abrupt topic shifts''.

\begin{figure*}[t]
\centering
\includegraphics[width=\linewidth]{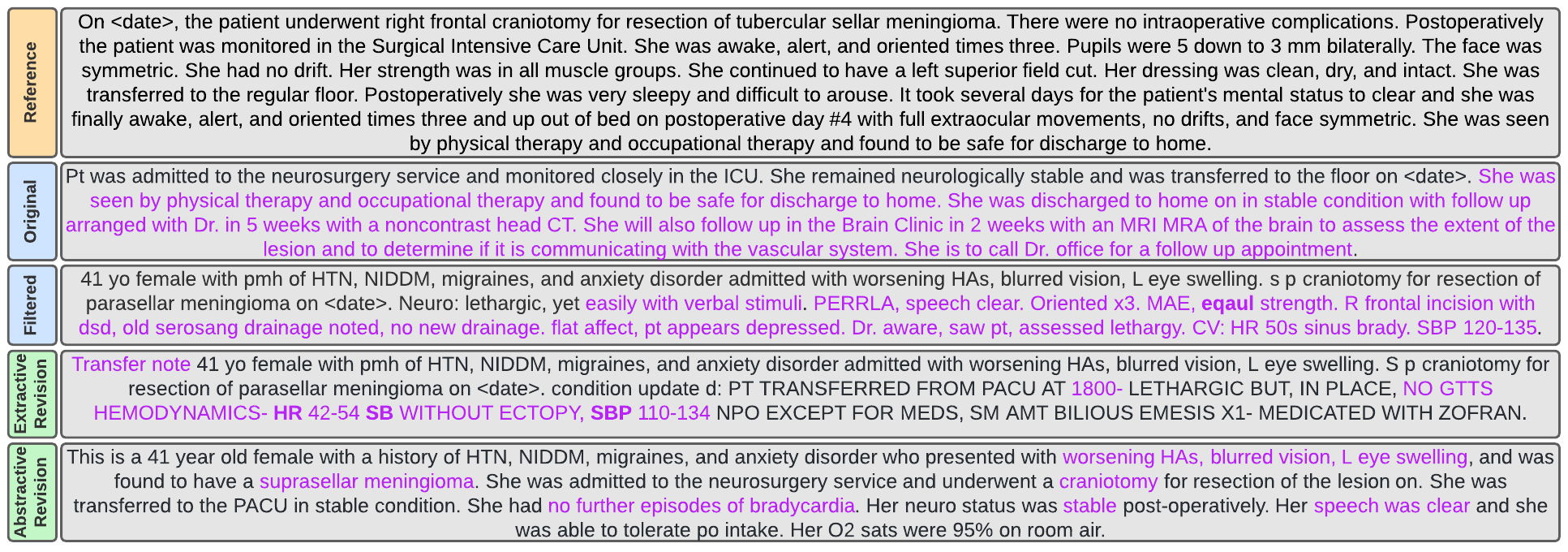}
\caption{An example \textcolor{orange}{Reference}, along with outputs from models trained on un-revised references (\textcolor{blue}{Original} and \textcolor{blue}{Filtered}), as well as \textcolor{Green}{Extractive} and \textcolor{Green}{Abstractive} revisions. Times and ages have been modified to ensure anonymity. Source notes are not shown for space reasons. \textbf{\textcolor{Amethyst}{Purple Snippets}} represent text mentioned in our analysis below.}
\label{fig:sum-output-example}
\end{figure*}

\paragraph*{Qualitative Analysis.} Figure \ref{fig:sum-output-example} shows an example from the human annotation test set for which we requested additional qualitative feedback. ``Original'' was inconsistent because it fabricates the last 4/6 sentences, operating almost as an unconditional LM. ``Filtered'' was less relevant than ``Abstractive'' because the second half of Filtered (starting with ``PERRLA'') contains non-essential status information. ``Filtered'' fluency was discounted for a misspelling (``eqaul'' rather than ``equal'') and a missing word (``easily \emph{roused} with verbal stimuli''). The annotator noted high fluency for Abstractive Revision (5) because the summary mimics the narrative style of the original references, rather than a sequence of choppy lists as in Filtered and Extractive summaries. Consistency is high (5) because the facts presented can be directly inferred. Coherence and consistency scored 5 because it follows a conventional arc from presenting illness (headache, vision, eye swelling), diagnosis (meningioma) and treatment (craniotomy), and ends with the patient's status (stable - no bradycardia, clear speech).

We also include ``Extractive'' to further emphasize the need to train on abstractive revisions. The ``Extractive'' summary does not match the target style of the Brief Hospital Course and includes many undesirable artefacts from source notes: note type metadata (``Transfer note''), long lists of non-essential data points (HR, SB, SBP), and unclear section boundaries (``1900- Lethargic...'').

\input{revision_tables/results_ablation}

\paragraph*{Ablation Analysis.} Which parts of the revision pipeline are necessary to improve \emph{downstream} summary performance? Separately, we remove ReDRESS hallucinations (top sequences from Figure \ref{fig:revise-train-example}), randomly sampled \emph{other} alignments (bottom sequences), and all negative examples (right side). We also train a model without control codes for revision intensity (\emph{w/o} Revision Codes) and, in turn, at inference time, generate a single revision rather than vary the codes to over-generate-then-rescore. Table \ref{tab:results-ablation} reveals that both sources of synthetic revision training data contribute to downstream performance (BS F1/Entailment of 80.4/72.1 and 80.7/73.8, individually, vs 81.3/76.3 combined). This 
aligns with \textcite{cao2021cliff} who also demonstrate "the importance of covering diverse types of errors in negative samples". Eliminating unlikelihood training of synthetic negatives reduces summary BS F1/Entailment from 81.3/76.3 to 80.7/72.4. Removing the ability to over-generate-then-rescore diverse revision candidates by varying style codes (\emph{w/o} Revision Codes) has the largest impact on downstream performance (78.7/62.5). A model trained without codes tends to insufficiently edit highly unsupported sentences.

\begin{figure}[t]
\centering
\includegraphics[width=0.75 \linewidth]{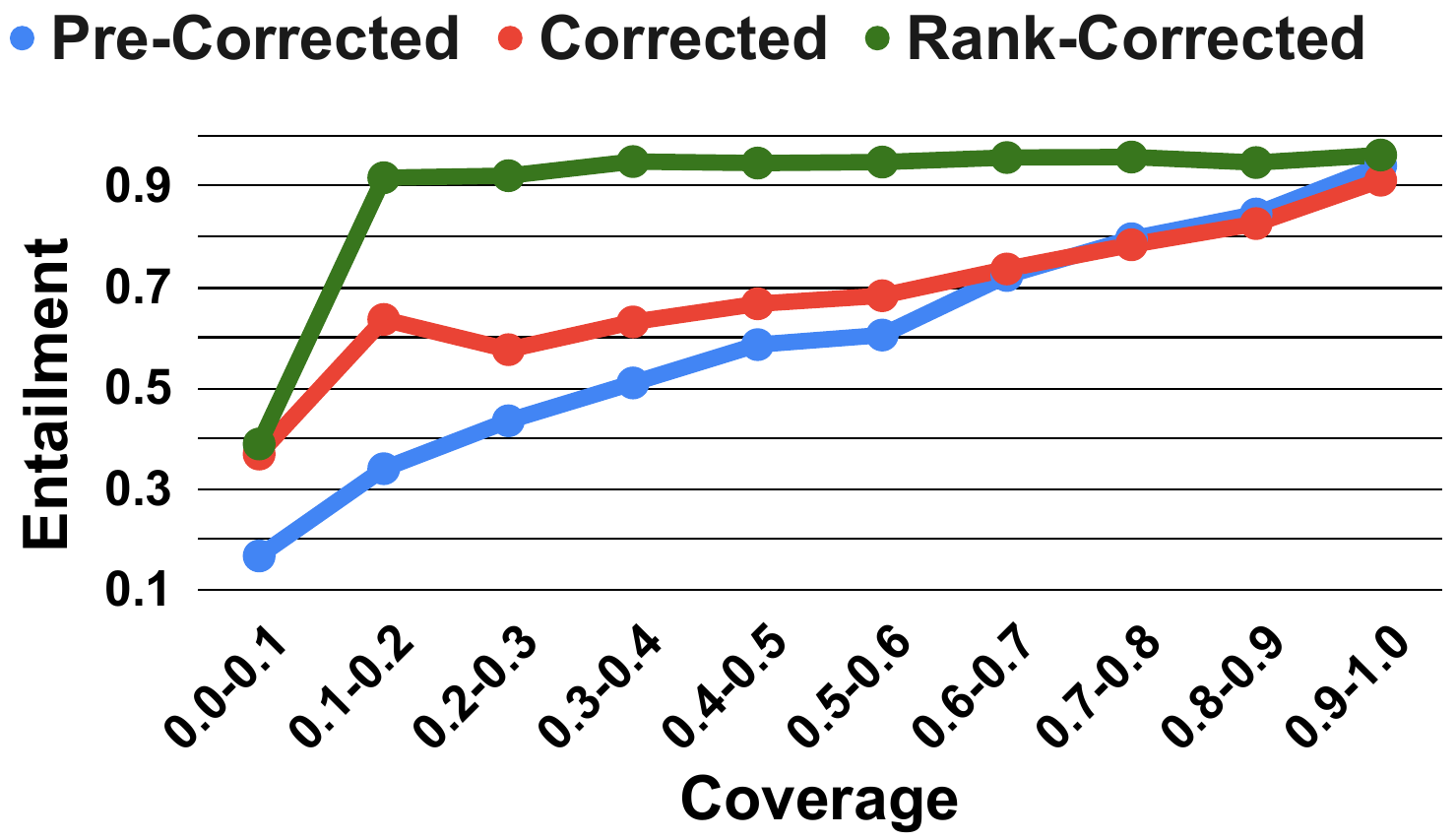}
\caption{Faithfulness before and after correcting summaries with the Reviser, controlling for extractiveness.}
\label{fig:effective-faithfulness}
\end{figure}

\paragraph*{Reviser as a Post-Hoc Editor.} As a natural extension, we experiment with using the revision model to correct system outputs. In particular, we take summaries from the model trained on \textbf{Original} references, and then separately feed each predicted sentence and aligned context to the reviser for generation. It would be easy to increase faithfulness by copying the context verbatim. Yet, we are more interested to see if the reviser can increases \emph{effective} faithfulness \parencite{ladhak2021faithful}, i.e., controlling for extractiveness. For each \emph{predicted} sentence, we over-generate revision candidates with different targeted levels of extraction (by varying the $source_{frac}$ code defined in Section \ref{sec:revise-pipeline}. Then, we bin sentences by extractiveness (coverage) and record faithfulness (fraction of sentences predicted as entailed) within each bin. We include separate plots for \textbf{Corrected}, which includes each over-generated candidate, and \textbf{Rank Corrected}, which selects the top candidate by entailment prediction (breaking ties according to the most abstractive). Figure \ref{fig:effective-faithfulness} demonstrates that reviser-corrected summaries are much more effectively faithful, although, naturally, the gap shrinks as all outputs become highly extractive and mostly entailed. The ability to re-rank and select from diverse candidates makes a huge difference, as evidenced by the large gap between \textbf{Corrected} and \textbf{Rank Corrected}.

\input{revision_tables/results_pretraining}

\paragraph*{ReDRESS/Revision as Pre-Training Objectives.} Beyond data cleaning, do our proposed methods to generate hallucinations (ReDRESS), and then edit them out (reviser), hold intrinsic value for the task as pre-training objectives? One can argue that both models are trained on faithfulness objectives: based on context, ReDRESS learns to add/remove/integrate entities, and the reviser to edit out synthetic hallucinations. Table \ref{tab:results-pretrain} shows that fine-tuning from ReDRESS and reviser--both trained from checkpoints of \texttt{bart-base}-- leads to an improvement in all evaluation metrics vis-a-vis \texttt{bart-base} (except slight entailment decrease from ReDRESS (56.0 to 55.2)). \emph{w/o} Entity Swapping is a denoising baseline in which corruptions are limited to span deletion and word re-ordering. Adding entity swaps into pre-training (ReDRESS), then, causes HR to drop (26.9 to 22.3) and NSP/FaR to rise (81.1/42.1 to 83.3/44.7).

\begin{figure*}[t]
\includegraphics[width=\linewidth]{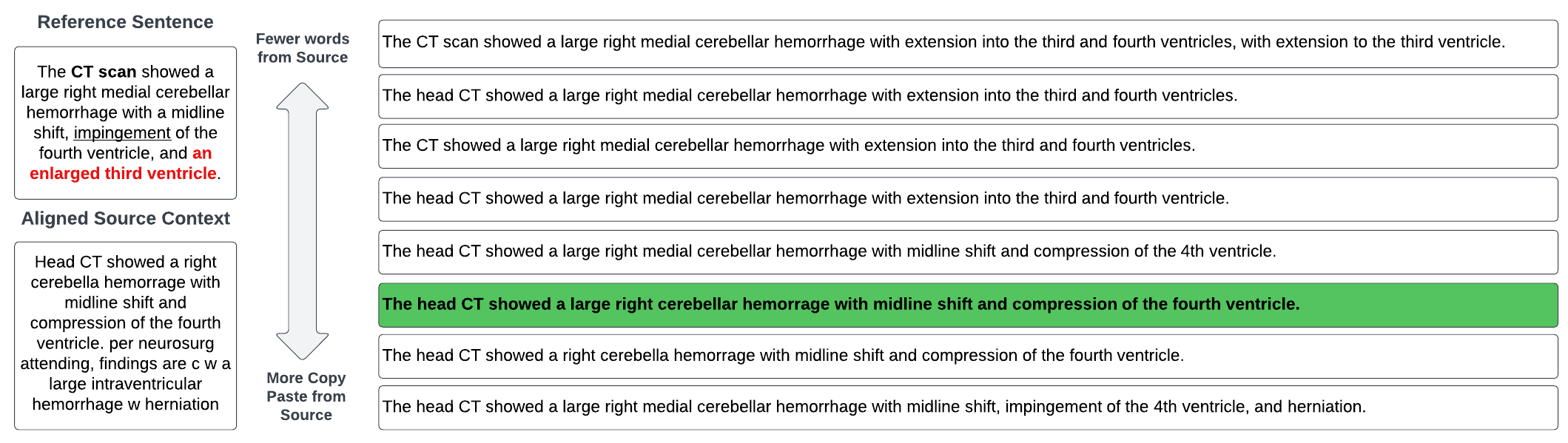}
\caption{Over-Generated Revisions from varying revision style codes. The \textcolor{Green}{highlighted} sentence is the one ultimately selected as the revision.}
\label{fig:over-generated}
\centering
\end{figure*}

\paragraph*{Qualitative Analysis.} We show an Abstractive example which received low assessed coherence (1) (other scores were all 3) to better understand its relative under-performance:

\begin{quote}
\small
A small right sub pectoral subcutaneous emphysema is mildly increased from the recent outside hospital exam. He was found to have a small right pneumothorax and multiple rib fractures on the right. A chest x-ray was obtained which showed a small anterior mediastinal hematoma likely tracking from the adjacent rib fractures. A CT of the chest was obtained to assess for progression of PTX contusions. There was no evidence of lung contusion or laceration on chest CT. He had no further episodes of bradycardia while in the Trauma ICU. A repeat chest CT was obtained on <date>.
\end{quote}

The Brief Hospital Course should roughly follow the same arc: presenting problem, investigation, diagnosis, treatment, (any complications and treatment), and, possibly, any follow-up plan. Deviation from this standard was penalized with lower coherence scores. The above summary does not follow this arc and the last sentence does not provide new information regarding the chest CT.

\paragraph*{Reviser Outputs at Different Intensities.} In Figure \ref{fig:over-generated}, we show diverse outputs from the reviser model, with each output conditioned on a different extractiveness code ($source_{frac}$ from \S \ref{sec:reviser}).  We see a relatively smooth interpolation from abstractive to extractive.  At low $source_{frac}$ codes, revisions include hallucinations regarding the third ventricle, but ultimately, the reviser edits them out, and, its place, introduces something that is only mentioned in the aligned source context: herniation. The green highlight is the sentence ultimately chosen by the More Abstractive strategy for its blend of abstraction and faithfulness (groundedness).

\subsection{Conclusion}

We propose a new approach to mitigating the downstream impact of noisy references on summarization models. We learn a model to improve the coverage of existing references from a large clinical corpus of EHR notes, and re-train models on revised references. Results show that reference revision is a more effective intervention for improving faithfulness than quality filtering.

%% file: revision_tables/results_table.tex
\begin{table*}[t!]
\small
\centering
\begin{tabular}{c|cc|cccccc}
& \textbf{\makecell{Reference \\ Version}} & \textbf{\makecell{Quality \\ Strategy}} & \textbf{\makecell{Hallucination\\Rate (HR) $\downarrow$}} & \multicolumn{3}{c}{\textbf{\makecell{BERTScore\\P / R / F1 (BS) $\uparrow$}}} & \textbf{\makecell{Entail. \\$\uparrow$}} & \textbf{\makecell{Faithful-Adjusted \\ Recall (FaR) $\uparrow$}} \\ \hline 
\hline
\parbox[t]{1ex}{\multirow{9}{*}{\rotatebox[origin=c]{90}{\sc{\textbf{{Longformer}}}}}} & 
\multirow{2}{*}{\textbf{Original}} & N/A & \textbf{36.8} & \textbf{82.3} & 69.5 & 75.2 & \textbf{48.4} & \textbf{48.2} \\ 
& & Control Halluc. & 36.5 & 83.3 & 70.2 & 76.0 & 51.5 & 49.0 \\ \cline{2-9}. 
& \multirow{4}{*}{\textbf{\makecell{Filtered \\ (Baselines)}}} & No Admission & 20.1 & 87.8 & 70.4 & 78.0 & 61.6 & 41.2 \\ 
& & Unsupported & \textbf{18.4} & 87.6 & 70.8 & 78.1 & \textbf{61.6} & \textbf{46.9} \\ 
& & Loss Truncation & 36.3 & 83.1 & 69.9 & 75.8 & \textbf{51.7} & 47.4 \\  
& & Halluc. Ents & 33.8 & 83.5 & 69.8 & 75.9 & 55.0 & 47.7 \\ \cline{2-9}. 
& \multirow{3}{*}{\textbf{\makecell{Revised\\(Ours)}}} & Fully Extractive & \textbf{5.4} & 94.5 & 73.2 & 82.3 & 78.6 & \textbf{52.4} \\ 
& & Less Abstractive & \textbf{3.8} & \textbf{94.6} & 73.1 & 82.3 & \textbf{83.7} & \textbf{54.0} \\ 
& & More Abstractive & \textbf{5.6} & 92.1 & 73.0 & 81.3 & 76.3 & \textbf{57.1} \\ \hline 
\end{tabular}
\caption{Summarization quality metrics across reference quality mitigation strategies (original, filtered, control, revised). The Longformer Encoder-Decoder (LED) model is used for fine-tuning. \textbf{Numbers} discussed below.}
\label{tab:results}
\end{table*}

%% file: revision_tables/results_bart.tex
\begin{table*}[t!]
\small
\centering
\begin{tabular}{c|cc|cccccc}
& \textbf{\makecell{Reference \\ Version}} & \textbf{\makecell{Quality \\ Strategy}} & \textbf{\makecell{Hallucination\\Rate (HR) $\downarrow$}} & \multicolumn{3}{c}{\textbf{\makecell{BERTScore\\P / R / F1 (BS) $\uparrow$}}} & \textbf{\makecell{Entail. \\$\uparrow$}} & \textbf{\makecell{Faithful-Adjusted \\ Recall (FaR) $\uparrow$}} \\ \hline 
\hline
\parbox[t]{1ex}{\multirow{9}{*}{\rotatebox[origin=c]{90}{\sc{\textbf{{Bart}}}}}} & \multirow{2}{*}{\textbf{Original}} & N/A & \textbf{38.9} & 81.3 & 69.2 & 74.7 & \textbf{43.6} & \textbf{47.7} \\ 
& & Control Halluc. & 40.3 & 81.7 & 69.2 & 74.8 & 43.2 & 46.6 \\ \cline{2-9}
& \multirow{4}{*}{\textbf{\makecell{Filtered \\ (Baselines)}}} & No Admission & 25.6 & 85.6 & 70.7 & 77.3 & 56.0 & 44.4 \\ 
& & Unsupported & \textbf{22.9} & 86.5 & 71.1 & 77.9 & \textbf{59.6} & \textbf{47.2} \\ 
& & Loss Truncation & 40.9 & 81.3 & 69.1 & 74.6 & \textbf{51.6} & 45.1 \\ 
& & Halluc. Ents & 37.6 & 82.3 & 69.1 & 75.0 & 48.6 & 46.4 \\ \cline{2-9} 
& \multirow{3}{*}{\textbf{\makecell{Revised\\(Ours)}}} & Fully Extractive & \textbf{9.1} & 92.3 & 72.8 & 81.2 & 72.1 & 52.1 \\ 
& & Less Abstractive & 7.3 & 91.8 & 72.9 & 81.1 & 72.5 & \textbf{56.3} \\ 
& & More Abstractive & \textbf{7.4} & 90.7 & 72.2 & 80.3 & 69.2 & 56.3  \\ \hline 
\end{tabular}
\caption{Summarization quality metrics across reference quality mitigation strategies (original, filtered, control, revised) used for training BART summarization models.}
\label{tab:results-bart}
\end{table*}

%% file: revision_tables/human_eval_table.tex
\begin{table}[ht]
\small
\centering
\begin{tabular}{lc|cccc}
\textbf{\makecell{Reference \\ Version}} & \textbf{\makecell{Quality \\ Strategy}} & \textbf{Con.} & \textbf{Rel.} & \textbf{Fl.} & \textbf{Coh.} \\ \hline
\textbf{Original} & N/A & 1.5 & 2.2 & 3.9 & \textbf{3.5} \\
\textbf{Filtered} & Unsupported & 3.2 & 2.8 & 3.3 & 3.3 \\
\textbf{Revised} & More Abstractive & \textbf{3.5} & \textbf{3.0} & \textbf{4.0} & 2.7 \\
\end{tabular}
\caption{Average rating (1-5) assigned by a domain expert (clinician) according to Consistency (Con.), Relevance (Rel.), Fluency (Fl.), and Coherence (Coh.).}
\label{tab:results-human}
\end{table}

%% file: revision_tables/results_ablation.tex
\begin{table}[ht]
\small
\setlength\tabcolsep{1.5pt}
\centering
\begin{tabular}{l|cc}
\textbf{\makecell{Reviser Training Objective \\ (Each Ablation is Separate)}} & \textbf{\makecell{BERTScore\\P / R / F1 (BS) $\uparrow$}} & \textbf{\makecell{Entail. \\$\uparrow$}} \\ \hline
\textbf{Full} (last row of Table \ref{tab:results}) & \textbf{92.1} / \textbf{73.0} / \textbf{81.3} & \textbf{76.3} \\ \hline
\textbf{\emph{w/o}} ReDRESS hallucinations & 90.8 / 72.4 / 80.4 & 72.1 \\
\textbf{\emph{w/o}} Random \textit{other} alignments & 90.9 / 72.8 / 80.7 & 73.8 \\
\textbf{\emph{w/o}} All Negatives (no contrast) & 90.9 / 72.8 / 80.7 & 72.4 \\
\textbf{\emph{w/o}} Revision Codes & 87.6 / 71.7 / 78.7 & 62.5 \\
\end{tabular} 
\caption{Separately removing key components of the reviser training objective (from Equation \ref{eq:revision}) hurts the \emph{downstream} performance of Longformer summaries.}
\label{tab:results-ablation}
\end{table}

%% file: revision_tables/results_pretraining.tex
\begin{table*}[t]
\small
\centering
\begin{tabular}{l|cccccc}
 \textbf{\makecell{Pretrain\\Weights}} & \textbf{\makecell{Hallucination\\Rate (HR) $\downarrow$}} & \multicolumn{3}{c}{\textbf{\makecell{BERTScore\\P / R / F1 (BS) $\uparrow$}}} & \textbf{\makecell{Entail. \\$\uparrow$}} & \textbf{\makecell{Faithful-Adjusted \\ Recall (FaR) $\uparrow$}} \\ \hline 
\textbf{\texttt{bart-base}} & 25.6 & 85.6 & 70.7 & 77.3 & \textbf{56.0} & 44.4 \\ \hline 
\textbf{ReDRESS} & \textbf{22.3} & 86.0 & 70.7 & 77.5 & \textbf{55.2} & \textbf{44.7} \\ 
\textbf{\emph{w/o} Entity Swap} & \textbf{26.9} & 85.7 & 70.3 & 77.1 & 54.4 & \textbf{42.1} \\ \hline 
\textbf{Reviser} & 23.0 & 86.0 & 71.0 & 77.7 & 56.7 & 47.9 \\ 
\end{tabular}
\caption{Assessing the usefulness of ReDRESS \& reviser models for pre-training. We separately fine-tune models from \texttt{bart-base}, ReDRESS, and reviser checkpoints on \emph{Filtered - No Admission} data. \emph{w/o} Entity Swap restricts noise to span-deletion, similar to the optimal configuration in the original BART paper. \textbf{Numbers} discussed below.}
\label{tab:results-pretrain}
\end{table*}

%% file: 4b-calibration.tex
In the previous Section, we essentially treated faithfulness as a data problem and worked on offline methods to revise, i.e., clean, the training data. We demonstrated improved faithfulness of models trained on revised references. Yet, in addition to being complex, reference revision is imperfect. It involves self-training on synthetic data and, as such, can only ever be as informative, coherent, and faithful as the revision model. In this Section, we explore ways in which we can improve faithfulness without having to alter references. Specifically, we simply give the model the ability to discriminate between faithful and unfaithful summaries such that, at inference time, the model is relatively more likely to generate a faithful summary than an unfaithful one. When applied to faithfulness, this approach is described as contrastive learning \parencite{cao-wang-2021-cliff}. More broadly, when applied to any downstream metric of interest, the term ``calibration'' has grown in popularity \parencite{liu-etal-2022-brio, zhao2023slic} because it involves \emph{calibrating} model likelihoods to metrics. The recently proposed Direct Preference Optimization (DPO) \parencite{rafailov2023direct} algorithm for Reinforcement Learning from Human Feedback (RLHF) \parencite{stiennon2020learning} is essentially a calibration model--instead of using metric-defined rankings, pairwise human preferences are used. In this Chapter, we use the term contrastive learning to refer to both contrastive learning for faithfulness and contrastive learning over model likelihoods (i.e., calibration).

Yet, much of this work has addressed \emph{how}---how to generate candidates \parencite{cliff} and how to define effective contrastive objectives \parencite{nan-etal-2021-improving, zhao2022calibrating}. Work has largely been separated into relevance and faithfulness contrastive learning, with less study of the interaction between the two. Relevance, often measured with ROUGE, captures the content overlap with a human-written reference, whereas faithfulness is typically reference-free, and captures the fidelity of a summary to the source text(s). In this section, we examine both faithfulness and relevance as the target metrics for contrastive learning and seek to uncover the underlying characteristics of effective contrast sets for each separately, as well as analyze the interactions between them. To accomplish this, we implement a diverse set of existing methods for constructing candidate and corrupted summaries and combine them to form a large candidate pool. From this pool, we implement different filtering strategies for set selection, which target specific characteristics, such as the metric margin between negatives and positives, diversity, and the model likelihood of generating each candidate in the set. We run experiments that vary only in the training data selected for candidate sets. For each experiment, we extract a wide range of relevant statistics (e.g., diversity, length) on the candidate sets and show the relationship between these set statistics and downstream performance. To guide future research, we analyze the plots to provide insights into, and rationale for, optimal set construction. For robustness, in addition to the MIMIC-III hospital course summarization dataset we created in the last section, we experiment with two other long-form, scientific corpora.

\paragraph*{Datasets}

\begin{table}[h]
\centering
\small
\begin{tabular}{l|c|c|c}
\textbf{Statistic} & \textbf{Clinical} & \textbf{Chemical} & \textbf{Bio.} \\ \hline
Train Size & 41,705 & 115,956 & 119,924 \\ 
Validation Size & 940 & 1,000 & 6,633 \\
Test Size & 1,861 & 2,000 & 6,658 \\ \hline
Source Tokens & 8,175 & 5,364 & 3,092 \\
Reference Tokens & 416 & 216 & 205 \\  \hline
Extractive Coverage & 0.66 & 0.90 & 0.88 \\
Extractive Density & 1.97 & 3.53 & 5.87 \\
\hline
\end{tabular}
\caption{Statistics for long-form scientific summarization datasets. The biomedical dataset is from \textcite{cohan-etal-2018-discourse}, the recipe to recreate the clinical from \textcite{adams-etal-2022-learning}, and the chemical from this work. } \label{tab:dataset-stats}

\end{table}

Dataset statistics are shown in Table~\ref{tab:dataset-stats}.

\paragraph*{Clinical.} We use the MIMIC-based hospital course summarization dataset from the previous section \parencite{adams-etal-2022-learning}.

\paragraph*{Chemical.} We introduce a dataset with a pure chemistry focus by compiling a list of chemistry academic journals with Open-Access articles. For each journal, we downloaded full-text article PDFs from the Open-Access portion of the journal using available APIs, or scraping this content using \href{https://www.selenium.dev/documentation/webdriver/}{Selenium Chrome WebDriver}. 
Each PDF was processed with Grobid \parencite{lopez2009grobid} via a \href{https://pypi.org/project/grobid-client-python/}{client} to extract free-text paragraphs with sections. The inputs for the summarization models are section headers and associated paragraphs for all sections from Introduction through Conclusion, excluding references, tables, and image captions. The abstract is treated as the reference. While other scientific summarization datasets exist \parencite{lu-etal-2020-multi-xscience, gupta-etal-2021-sumpubmed, deyoung-etal-2021-ms}, ours is the first to exclusively contain chemistry-related papers. \begin{table}[h]
\centering
\small
\begin{tabular}{l|c}
\textbf{Source} & \textbf{\# Articles}  \\ \hline
Beilstein & 1,829 \\
Chem Cell & 546 \\
ChemRxiv & 12,231 \\
Chemistry Open & 398 \\
Nature Communications Chemistry & 572 \\
PubMed Author Manuscript & 57,680 \\
PubMed Open Access & 29,540 \\
Royal Society of Chemistry (RSC) & 9,334 \\
Scientific Reports - Nature & 6,826 \\
\hline
\end{tabular}
\caption{Journals accessed for Chemical papers.} \label{tab:chemistry-sources-again}
\end{table}

Table \ref{tab:chemistry-sources-again} shows the journals from which Open Access articles were sourced, as well as the number of papers processed. For all journals, we filtered for papers with the provided topic of Chemistry when papers from other disciplines were also available (e.g. PubMed). We randomly split the aggregated dataset into train-validation-test splits.

The dataset is available for download on the HuggingFace Datasets Hub under \href{https://huggingface.co/datasets/griffin/ChemSum}{griffin/ChemSum}.

\paragraph*{Biomedical.} We use the PubMed abstract generation dataset \parencite{cohan-etal-2018-discourse}, which pairs automatically extracted abstracts with full-text articles from the PubMed Open-Access Subset.

\paragraph*{Background \& Contrastive Learning Pipeline}

At a high-level, we fine-tune (\texttt{FT}) language models with standard maximum likelihood estimation (\texttt{MLE}) on each summarization corpus, and then \textit{contrast}-tune (\texttt{CT}) on a combined objective, which adds a contrastive loss (\texttt{CA}) to the MLE loss:

\begin{align}
\label{eq:pipeline}
\begin{split}
&\mathcal{L}_{FT} = \mathcal{L}_{MLE} \\
&\mathcal{L}_{CT} = \lambda_{MLE} * \mathcal{L}_{MLE} + \lambda_{CA} * \mathcal{L}_{CA}
\end{split}
\end{align}

\noindent $\lambda_{MLE}, \lambda_{CA}$ are scalars controlling the relative weight of objective. For $\mathcal{L}_{CT}$, $\mathcal{L}_{MLE}$ acts as a regularizer, as in \textcite{liu-etal-2022-brio, zhao2022calibrating}.



We describe the setup (objective, metrics, and candidate generation methods) for Relevance Contrastive Learning and Faithful Contrastive Learning, before jointly discussing statistics on each setup.

\paragraph*{Relevance Contrastive Learning} \label{sec:relevance-setup}

As in \parencite{liu-etal-2022-brio, zhao2022calibrating}, we calibrate for relevance by learning to rank model-generated summaries (post-\texttt{FT}, pre-\texttt{CT} weights).

\paragraph*{Objective.} Specifically, a set of model-generated summaries $\bm{\hat{S}}$ is ranked: $q(\hat{S}_i; S) \geq q(\hat{S}_j; S)$, $\forall i, j \in |\bm{\hat{S}}|, i < j$, where $S$ is the reference and $q$ represents $Rel_{Agg}$ (defined below). A score function $f$ is applied to each candidate and calibrated to the metric ranking via a pairwise margin:

\begin{align}
\small
\label{eq:pmr}
\begin{split}
max(0, f(D, \hat{S}_j) - f(D, \hat{S}_i) + (j - i) * \lambda_{margin}) \\
\forall i, j \in |\bm{\hat{S}}|, i < j
\end{split}
\end{align}

$f$ can be a similarity metric between $S$ and $D$ \parencite{liu-liu-2021-simcls, zhong-etal-2020-extractive} or, more directly, the log likelihood \parencite{liu-etal-2022-brio}. For our use experiments, $f$ represents for the length normalized log likelihood of generating a summary.




\paragraph*{Rank Metric.} To define a gold-standard ordering, we aggregate 3 relevance metrics which are normalized to be zero after fine-tuning \texttt{FT}. $Rel_{Agg}$, a combination of ROUGE 1/2 F-1 \parencite{lin2004rouge} and \textbf{BERTScore-Ref} \parencite{zhang2019bertscore}, represents the standard deviation change in the aggregated metric from \texttt{FT}. For BERTScore \parencite{zhang2019bertscore}, we use \textit{allenai/scibert\_scivocab\_uncased} weights and all default settings from HuggingFace \parencite{transformers}. We normalize by subtracting each metric by its mean and then dividing by the standard deviation to account for metrics with different scales. We use test set fine-tuning (\texttt{FT}) scores to compute mean and standard deviation so that $Rel_{Agg}$ is $0$ after \texttt{FT} and $>0$ values are standard deviation improvements from contrastive learning.

\paragraph*{Candidates.} We fine-tune (\texttt{FT}) two long-document encoder-decoder models: LongT5 \parencite{longt5} and PRIMERA \parencite{primera}, on each corpus before decoding 10 candidates with diverse beam search \parencite{vijayakumar2016diverse} with diversity penalty of $1.0$, as in \textcite{liu-etal-2022-brio, nan-etal-2021-improving, ravaut-etal-2022-summareranker}.

\begin{table*}[t]
\centering
\small
\begin{tabular}{cl|c|c|c|c|c|c}
& \textbf{Method} & \textbf{\textcolor{red}{$-$}} & \textbf{\textcolor{Green}{$+$}} & \textbf{Source} & \textbf{Ref.} & \textbf{External Components} & \textbf{Models Used} \\
\hline
\multirow{2}{*}{\texttt{\makecell{Relevance \\ CL}}} & Diverse Beam & \textcolor{red}{\checkmark} & \textcolor{Green}{\checkmark} & \checkmark & & \makecell{Summarization\\Model} & PRIMERA \\
& Diverse Beam & \textcolor{red}{\checkmark} & \textcolor{Green}{\checkmark} & \checkmark & & \makecell{Summarization\\Model} & LongT5 \\ \hline
\multirow{4}{*}{\texttt{\makecell{Faithful \\ CL}}} & Mask-And-Fill & \textcolor{red}{\checkmark} &  & & \checkmark & \makecell{Syntax Parser,\\Pretrained LM} & \makecell{Stanza,\\SciFive} \\
& Entity Swap & \textcolor{red}{\checkmark} & & & \checkmark & NER & \makecell{BERN2,\\ Quantulum} \\
& Paraphrase & & \textcolor{Green}{\checkmark} & & \checkmark & \makecell{Paraphrase\\Generator} & GPT-3 \\
& Reference & & \textcolor{Green}{\checkmark} & & \checkmark & N/A & N/A \\ \hline
\end{tabular}
\caption{Methods to create \textcolor{red}{negative} and \textcolor{Green}{positive} candidates in support of relevance and faithfulness CL, respectively. For each candidate generation method, we include whether it is used as a positive or negative example (both in the case of relevance ranking), what inputs it requires (the source document and/or the reference (ref.)), as well as the external components needed and, finally, the specific models used for the experiments in this paper.} \label{tab:method-overview}
\end{table*}

\paragraph*{faithfulness contrastive learning} \label{sec:faithful-setup}

\paragraph*{Objective.} As in \textcite{gunel2020supervised, khosla2020supervised, cliff}, we use contrastive learning to minimize the latent distance between pairs of positive summaries vis-a-vis negative ones:

\begin{equation}
\label{eq:contrast}
\small
- \frac{1}{{|\bm{\hat{S}^P}| \choose 2 }} \sum_{ \hat{S}_i, \hat{S}_j \in \bm{\hat{S}^P}  }{ log \frac{exp(sim(h_i, h_j) / \tau )}{ \sum_{\hat{S}_k \in \bm{\hat{S}^N}}{exp(sim(h_i, h_k) / \tau)} } }
\end{equation}

where $\tau$ is a temperature parameter. It pushes positive summaries closer to each in latent space ($h_i$ and $h_j$) and further away from negatives ($h_k$). We follow \textcite{cliff} and use cosine similarity as $sim$ and treat $h$ as the mean-pooled decoder states, followed by a linear projection.

\paragraph*{Faithfulness Metric.} Similar to $Rel_{Agg}$, we compute $Faith_{Agg}$ as an aggregation of normalized metrics. We combine \textbf{BARTScore} \parencite{yuan2021bartscore}, \textbf{BERTScore-Src} (vis-a-vis source), and a new metric \textbf{FactScore}, which is based on a scientific fact detection model (MultiVERS \parencite{wadden-etal-2022-multivers}). For BARTScore, we use a PEGASUS \parencite{zhang2020pegasus} model pretrained on the PubMed summarization corpus\footnote{\texttt{google/pegasus-pubmed } on the HuggingFace Transformers Hub \parencite{transformers}.} for the PubMed and Clinical datsets, and we use a Longformer Encoder-Decoder \parencite{beltagy2020longformer} trained on a more faithful, synthetic version of our clinical corpus from \textcite{adams-etal-2022-learning}. We report the average log-likelihood of each candidate summary $S$: $\frac{1}{|S|} \sum_{i \in |S|}{p(s_i|, s_{j < i}, D)}$. BARTScore and BERTScore are not explicitly trained to detect domain-specific errors. As such, we implement \textbf{FactScore}, which is based on the state of the art model (MultiVERS \parencite{wadden-etal-2022-multivers}) trained on the SciFact scientific claims dataset \parencite{wadden-etal-2020-fact}. SciFact is an expert-annotated dataset of 1,409 sentence-level scientific claims. We first align each summary sentence to a handful of sentences (1-5) from the source document, following the greedy algorithm from \textcite{lebanoff-etal-2019-analyzing}. Then we score each sentence based on its alignment and average the \texttt{SUPPORTED} label prediction probabilities.

\paragraph*{Negative Methods.} We use an in-domain LM (SciFive \parencite{phan2021scifive}) to \textbf{Mask-And-Fill} text to create hallucinations, as well as perform \textbf{Entity Swaps} of scientific concepts and numbers which separately target \texttt{intrinsic} and \texttt{extrinsic} hallucinations \parencite{maynez-etal-2020-faithfulness}. \textbf{Mask-And-Fill} involves masking portions of a reference summary, and using a pre-trained language model to fill in the blanks. It has been used for contrastive fine-tuning \parencite{cliff}, evaluation \parencite{ctc}, and fine-grained optimization of noisy references \parencite{zhou-etal-2021-detecting}. First, following \textcite{goyal-durrett-2021-annotating, lee-etal-2022-masked}, we identify all noun phrases\footnote{`NP' using the annotation scheme from the Penn Treebank \parencite{marcinkiewicz1994building}.} as candidates for masking using Stanza's constituency parser \parencite{qi2020stanza}. Then, we sample a subset of non overlapping phrases to mask and generate replacements with SciFive \parencite{phan2021scifive}. SciFive is a language model pre-trained on diverse biomedical tasks with T5-inspired \parencite{raffel2020exploring} prefixes. We perform a beam search of size 4 to generate in-filled text for each spans and set the minimum generated tokens to be equal to the number of masked tokens to preserve length. \textit{Hyper-Parameters of Significance}: the target token mask rate: $m$, which defines the percentage of noun phrases from the unmasked reference to mask. We vary $m$ to measure the impact of corruption `intensity' on the efficacy of contrastive fine-tuning. For \textbf{Entity swapping} \parencite{factcc}, we replace reference entities and numbers with entities and numbers from the source text (\texttt{intrinsic} hallucinations) or the corpus (\texttt{extrinsic}). \textit{Hyper-Parameters of Significance}: the swap rate: $s$, which defines the percentage of named entities and numbers in the reference, separately, to replace. Entity and number swapping was initially proposed for faithfulness evaluation (FactCC \parencite{factcc}) and has subsequently been used for contrastive fine-tuning \parencite{tang-etal-2022-confit} and post-hoc editing \parencite{cao-etal-2020-factual, chen-etal-2021-improving, zhu-etal-2021-enhancing}, etc. For each corpora, we extract numbers with numbers with \href{https://github.com/nielstron/quantulum3}{quantulum3}. Separately for each corpora, we extract named entities relevant to each domain. For chemistry, we extract chemicals and other types\footnote{The list of types includes genes, diseases, species, mutations, cell lines, and cell types.} with BERN2 \parencite{kim2019neural}. BERN2 is trained on PubMed articles to identify chemicals and diseases and link them to a unique identifier (CUI) in the Unified Medical Language System (UMLS) \parencite{bodenreider2004unified}. For the clinical corpus, we use the Stanza transformer model \parencite{qi2020stanza, zhang2021biomedical} trained on the i2b2 corpus \parencite{uzuner20112010}, which learns to identify patient problems, tests, and treatments. Finally, for biomedical, we use the Stanza model trained on the BioNLP13CG corpus \parencite{pyysalo2015overview}, which includes a diverse set of 13 categories. To simulate intrinsic errors, we perform swaps at random with entities of the same semantic category from the source document. For extrinsic, we also restrict the swap to be from the same semantic category, yet sample from the entire corpus. \linebreak

\paragraph*{Positive Methods.} We pool together the \textbf{Reference} with \textbf{Paraphrased} versions of it. General domain neural paraphrases performed poorly on scientific text. As such, we collect 10 paraphrases from relevant domain experts, and incorporate them as few-shot demonstrations for paraphrase generation by GPT-3 \parencite{gpt3}.

\begin{figure*}[t]
\centering
\includegraphics[width=\linewidth]{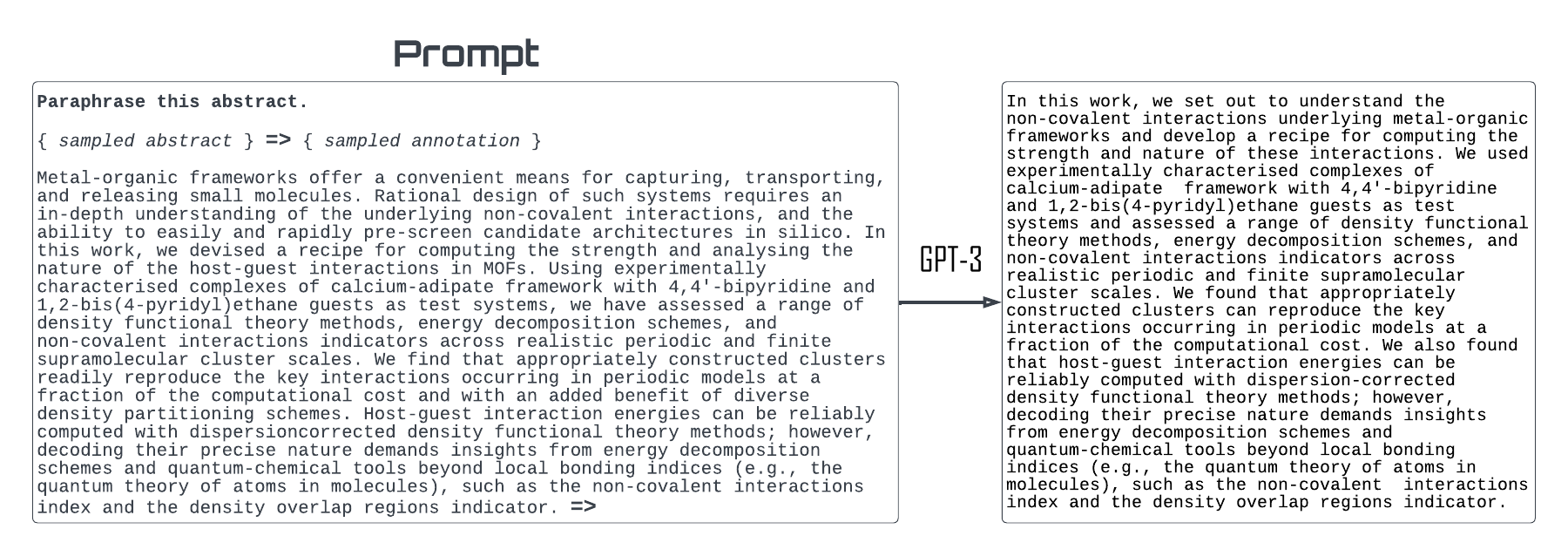}
\caption{An example prompt and paraphrase output from GPT-3. Words are changed but the meaning is preserved. } 
\label{fig:gpt3-paraphrase}
\end{figure*}

Paraphrasing is typically done with synonym substitution \parencite{zhou-bhat-2021-paraphrase}, neural models \parencite{goyal-durrett-2020-neural} trained on paraphrase corpora \parencite{wieting2017paranmt, zhang-etal-2019-paws}, or back-translation \parencite{factcc, fabbri-etal-2021-improving}. Yet, these methods performed very poorly on our long scientific texts, likely due to highly specialized lexicons and lack of large-scale, domain-specific paraphrase corpora. In Figure \ref{fig:gpt3-paraphrase}, we show an example prompt and sampled paraphrase from one-shot paraphrasing with GPT-3. A random sample of one annotation pair, as well as the abstract to be paraphrased, are then provided as prompts, which are both preceeded by a fixed instruction: \texttt{Paraphrase this abstract.} for abstract generation, and \texttt{Paraphrase this Summary.} for clinical summarization). We sample 1 due to token limits yet prompt sampling also increases diversity, as shown in \textcite{chintagunta-etal-2021-medically}. A softmax temperature $t$ of 0.7 is used to sample 5 unique outputs from GPT-3 (\texttt{text-davinci-002}). \linebreak

\begin{table}[h]
\centering
\small
\begin{tabular}{l|c|c}
\hline
\textbf{Method} & \textbf{Hyper-Param} & \textbf{Number} \\
\hline
Mask-And-Fill (\emph{Low}) & $m=0.25$ & 10 \\
Mask-And-Fill (\emph{High}) & $m=0.75$ & 10 \\
Swap Intrinsic (\emph{Low}) & $s=0.5$ & 10 \\
Swap Intrinsic (\emph{High}) & $s=1.0$ & 10 \\
Swap Extrinsic (\emph{Low}) & $s=0.5$ & 10 \\
Swap Extrinsic (\emph{High}) & $s=1.0$ & 10 \\ 
Paraphrase & $t=0.7$ & 5 \\
Reference & N/A & 1 \\ \hline
\textbf{Total For Faithfulness} & & \textbf{66} \\ \hline \hline
Diverse Beam (PRIMERA) & $p=1$ & 10 \\
Diverse Beam (LongT5) & $p=1$ & 10 \\ \hline
\textbf{Total For Relevance} & & \textbf{20} \\
\hline
\end{tabular}
\caption{\# of candidates pooled for each training instance. $m$ is \% of noun phrases masked, $s$ \% of entities swapped, and $t$ the softmax temperature for GPT-3. Contrast sets for training are formed by selecting a subset of 4 for relevance rank sets and 2 negatives / 2 positives for faithfulness contrast sets. } \label{tab:method-counts}
\end{table}

\paragraph*{Candidate Set Details} \label{sec:calibration-analysis}

Table \ref{tab:method-overview} displays the differences between candidate methods at a very basic level, as well as the particular models used for our experiments on long-form scientific summarization. In Table~\ref{tab:method-counts}, we show the number of distinct candidates we produce for each example in the training set by each method / hyper-parameter combination. When contrastive learning for faithfulness, we select 4 out of 66 possible candidates (2 positive and 2 negative), and for relevance, we select 4 out of 20 possible candidates. 4 is the maximum number which fits in GPU memory on an A100 40GB card, even with a device batch size of one (with gradient accumulation steps) and half precision (fp16). The idea behind generating candidates with different methods and parameters is twofold: (1) to better understand which candidate generation methods work best on our task of interest: long-form scientific summarization, and (2) to end up with a diverse candidate pool, which allows us to effectively control for certain characteristics when selecting final subsets for contrastive learning.

\begin{table*}[t]
\centering
\small
\setlength{\tabcolsep}{2pt}
\begin{tabular}{cl|ccc|ccc|ccc}
\hline
& \multirow{2}{*}{\texttt{\makecell{Candidate \\ Method}}} & \multicolumn{3}{c}{\texttt{Clinical}} & \multicolumn{3}{c}{\texttt{Chemical}} & \multicolumn{3}{c}{\texttt{Biomedical}} \\
& & Rel. & Faith. & Extract. & Rel. & Faith. & Extract. & Rel. & Faith. & Extract. \\
\hline
\multirow{8}{*}{\makecell{Faith. \\ Contrast}} & \textcolor{red}{Mask-And-Fill} (\emph{Low}) & 0.98 & 0.52 & 1.55 & 0.99 & 0.75 & 3.24 & 0.97 & 0.73 & 4.92 \\
& \textcolor{red}{Mask-And-Fill} (\emph{High}) & 0.97 & 0.52 & 1.44 & 0.97 & 0.73 & 2.90 & 0.95 & 0.71 & 4.05 \\
& \textcolor{red}{Swap Intrinsic} (\emph{Low}) & 0.94 & 0.52 & 1.64 & 0.97 & 0.70 & 2.92 & 0.98 & 0.71 & 4.70 \\
& \textcolor{red}{Swap Intrinsic} (\emph{High}) & 0.90 & 0.52 & 1.82 & 0.95 & 0.65 & 2.62 & 0.97 & 0.67 & 4.13 \\
& \textcolor{red}{Swap Extrinsic} (\emph{Low}) & 0.94 & 0.52 & 1.64 & 0.97 & 0.70 & 2.92 & 0.98 & 0.68 & 4.44 \\
& \textcolor{red}{Swap Extrinsic} (\emph{High}) & 0.90 & 0.52 & 1.82 & 0.95 & 0.65 & 2.62 & 0.97 & 0.64 & 3.79 \\ 
& \textcolor{Green}{Paraphrase} & 0.90 & 0.52 & 1.26 & 0.94 & 0.77 & 3.06 & 0.92 & 0.73 & 4.00 \\
& \textcolor{Green}{Reference} & 1.00 & 0.52 & 1.96 & 1.00 & 0.76 & 3.54 & 1.00 & 0.74 & 5.78\\ \hline
\multirow{2}{*}{\makecell{Rel. \\ Rank}} & Diverse Beam (PRIMERA) & 0.84 & 0.53 & 2.65 & 0.87 & 0.85 & 9.66 & 0.86 & 0.86 & 12.90 \\
& Diverse Beam (LongT5) & 0.83 & 0.52 & 2.06 & 0.86 & 0.83 & 7.46 & 0.85 & 0.82 & 8.39 \\ \hline
\end{tabular}
\caption{Statistics for each candidate generation method. Rel. stands for Relevance and is measured by BERTScore F1 overlap with the reference. Faith. stands for faithfulness and is measured by the FactScore (as defined in \S \ref{sec:faithful-setup}). Extract. stands for the extractive density (level of copy-and-paste) as defined by \textcite{grusky2018newsroom}. The first 6 rows (Mask-And-Fill and Swaps) construct negative examples for faithfulness contrastive learning. The next two rows form the positive candidate set for faithfulness. The last two (diverse beam) form candidates for relevance contrastive learning.} \label{tab:method-metrics}
\end{table*}

\paragraph*{Analysis.} In Table \ref{tab:method-metrics}, we show statistics (relevance, faithfulness, and extractive density) for each candidate generation method across the three datasets. As noted in \textcite{adams-etal-2022-learning}, the references for the clinical dataset are very abstractive (1.96 density) and unfaithful (0.52 FactScore), as compared to the chemical (3.54 / 0.76) and biomedical (5.78 / 0.74) data. The former is affected by missing clinical notes while the latter references are abstracts, which \emph{should} be mostly entailed by the claims made in the main paper. Interestingly, the reference is deemed less faithful than the model generations (0.52 vs 0.53/0.52, 0.76 vs 0.85/0.83, and 0.74 vs 0.86/0.82 for diverse beam search clinical, chemical, and biomedical). This likely has to do with the fact that the fine-tuned models (PRIMERA and LongT5) perform substantially more copy-and-pasting from the source input as the references (1.96 vs 2.65/2.06, 3.54 vs 9.66/7.46, and 5.78 vs 12.90/8.39, respectively).

The most unfaithful corruption method is Swap. When looking at (High) across Intrinsic and Extrinsic, its FactScores are 0.52/0.52, 0.65/0.65, and 0.67/0.64 versus 0.52, 0.73, 0.71 for Mask-And-Fill (High), respectively. This likely has to do with an in-domain LM (SciFive) making reasonably well-informed replacements for noun phrases, whereas entity swapping is indiscriminate and random. The (High) parameter settings for Mask-And-Fill and Swap create less faithful candidates vis-a-vis the (Low) settings (0.75/0.70/0.70 versus 0.73/0.65/0.65 for High and Low on Chemical, for example), as expected. Replacing more text from the references introduces more factual errors. The PRIMERA model produces more extractive summaries with diverse beam search (2.65/9.66/12.90 vs 2.06/7.46/8.39), which are scored as more relevant and faithful than LongT5.



\subsection{Identifying Possible Correlates} \label{sec:hypothesis}

We examine five basic aspects of contrast sets that \emph{should} have some impact on downstream performance. For each aspect, we provide intuition and some related work to guess the nature of the impact, which we investigate empirically in \S \ref{sec:results}.

\paragraph*{Overall Quality}

\paragraph*{Definition.} For the purposes of this analysis, for relevance-rank sets, we define quality as the average $Rel_{Agg}$ score of the candidates.

\paragraph*{Relevance Hypothesis.} For relevance, high-quality sets might be preferable to lower-quality sets for two reasons: (1) the model before contrastive learning tuning (pre-\texttt{CT}) has already been fine-tuned (post-\texttt{FT}) on the same training data used for \texttt{CT}, so it likely already assigns a high-probability mass to summaries which are close to the reference. Candidate summaries which deviate too much should already have a low probability of being generated and thus not provide much of a learning signal. In some ways, this hypothesis is supported by \textcite{zhao2022calibrating} who find that using a model's top beams produces consistently better results than diverse beam search or sampling-based methods (e.g., nucleus sampling \parencite{holtzman2019curious}). There is an inherent tension between the contrastive objective, which involves exploration, and the MLE, which assigns all probability mass to a single point.

\paragraph*{Margin}

Overall quality covers average metric values, while margin covers within-set variation in quality.

\paragraph*{Definition.} For relevance rank-based sets, we define the margin as the average relevance score between all adjacent pairs of ranked candidates: $Avg(Rel_{Agg}(\hat{S_i}, S) - Rel_{Agg}(\hat{S_{i+1}}, S)), i \in |\bm{\hat{S}}| - 1$. For faithfulness, we define it as the delta in average $Faith_{Agg}$ scores for summaries in 
 the positive and negative contrast sets, respectively.

\paragraph*{Relevance Hypothesis.} As noisy proxies for human judgments \parencite{peyrard-gurevych-2018-objective}, subtle differences in relevance metrics (e.g, ROUGE and BERTScore) might not be meaningful. As such, we hypothesize that, all else equal, sets with larger metric gaps will provide a clearer training signal during contrastive learning and superior downstream results.

\paragraph*{Faithfulness Hypothesis.} Trivially, one would want positive candidates which are fully faithful. For negatives, it is less clear. The emphasis in the literature has been on producing negative summaries which mimic model errors \parencite{goyal-durrett-2021-annotating}. Yet, less is discussed about the intensity of errors. \textcite{lee-etal-2022-masked} explore corruption intensity in the context of training a faithfulness evaluator, and the results suggest a concave relationship. Too few edits and the contrast sets are not easily separable, yet too dramatic, and the contrastive loss is ineffectual. We suspect a similar result for calibrating with a contrastive objective.

\paragraph*{Lexical Diversity}

The previous contrast set characteristic (Margin) covered metric-based comparisons. In this section, we perform comparisons solely at the word-level.

\paragraph*{Definition.} We define lexical diversity as the average pairwise self-BLEU score \parencite{zhu2018texygen, alihosseini-etal-2019-jointly} between all candidates in a relevance ranking set and separately, for positives and negative subsets in a faithfulness contrast set. 

\paragraph*{Relevance Hypothesis.} All else equal, high lexical diversity should improve the robustness of CL models as it somewhat dampens some of the noise from single-reference MLE training\footnote{We use the word \emph{somewhat} because we acknowledge that relevance metrics measure overlap to a single reference, so introducing diverse contrast set candidates does not necessarily encourage, or reward, more diverse outputs. Access to multiple references, or calibrating against human judgments, would better mitigate the single reference exposure bias problem.}.

\paragraph*{Faithfulness Hypothesis.} High lexical diversity within positive and negative sets should make the contrastive classifier less reliant on lexical overlap and focus more on the gap in faithfulness between positive and negatives. Lexical diversity likely means more coverage of error types, which has been shown to be beneficial for contrastive fine-tuning \parencite{cao-wang-2021-cliff, adams-etal-2022-learning}.

\paragraph*{Likelihood}

This section covers a model-specific aspect of contrast sets: the likelihood of the candidate summaries under the model post-\texttt{FT} and pre-\texttt{CT}.

\paragraph*{Definition.} For each candidate summary, we compute its length-normalized conditional log likelihood: $\frac{1}{L}\sum_{l=1}^{L}{log\textit{p}(s_{l}|D, S_{< l}; \theta_{FT})}$, where $\theta_{FT}$ denotes the model parameters after fine-tuning.

\paragraph*{Relevance Hypothesis.} One would suspect that likely contrast sets are preferable to unlikely since there is little need to calibrate a model to candidate summaries it was never likely to generate.

\paragraph*{Faithfulness Hypothesis.} In a similar vein, it makes sense that contrastive learning for faithulness will be most powerful when the model is most surprised. That is, the negatives are more likely to be generated than the positive. This relates to work by \textcite{goyal-durrett-2021-annotating}, who argue that negative sets should mimic observed errors.

\paragraph*{Spurious Correlates}

Automatic evaluation metrics have a tendency to reward outputs with characteristics which are spuriously correlated to quality \parencite{durmus-etal-2022-spurious}.

\paragraph*{Definitions.} While many possibilities exist \parencite{durmus-etal-2022-spurious}, 
for relevance, we focus on summary length, as defined by number of tokens. For faithfulness, we focus on extractiveness, which we measure with density \parencite{grusky2018newsroom}: the average squared length of extractive fragments. It approximates the level of copy-and-paste.

\paragraph*{Relevance Hypothesis.} \textcite{sun-etal-2019-compare} discover that ROUGE rewards longer summaries while humans prefer concise summaries. We hypothesize that exposing models to longer outputs during contrastive learning will lead to longer summaries, which will have higher relevance scores. By controlling for contrast set length, we can better understand whether or not some of the gains from contrastive learning simply come from length tuning\footnote{While length can be influenced during beam search with minimum/maximum length restrictions and length penalties, these measures do not expose a model to long summaries. }.

\paragraph*{Faithfulness Hypothesis.} \textcite{ladhak-etal-2022-faithful} note that faithfulness metrics tend to prefer summaries with high levels of extraction, all else equal. Yet, \textcite{zhang2022extractive} demonstrate that highly extractive does not always mean more faithful, so it is important to get a sense of how much faithfulness contrastive learning is driven by more copy-and-paste.

\begin{figure*}[t]
\centering
\includegraphics[width=\linewidth]{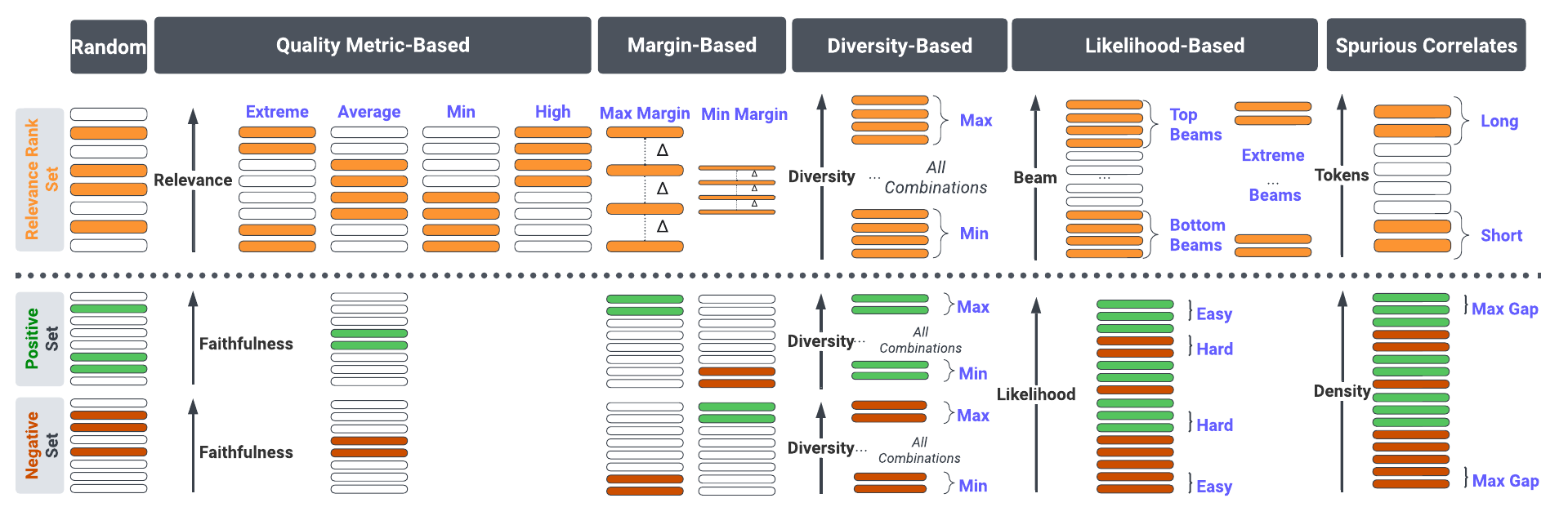}
\caption{Strategies for selecting \textcolor{orange}{rank sets} of size 4 from larger candidate pools for relevance contrastive learning (top half). The bottom half shows similar strategies to form binary contrast sets (2 \textcolor{Green}{positive}, 2 \textcolor{red}{negative}) for faithfulness. Each strategy for the top half of the Figure occupies a row in Table \ref{tab:relevance-results}, while the bottom corresponds to rows in Table \ref{tab:faithfulness-results}. } 
\label{fig:strategies}
\end{figure*}

\paragraph*{Selection Strategies.} \label{sec:strategies}

\paragraph*{Problem Statement.} From a large candidate pool, select a target number to be used for \texttt{CT} (2 positives and 2 negatives for faithfulness, and 4 for rank-based relevance). Figure \ref{fig:strategies} shows the different strategies implemented which are designed to target specific set characteristics. They do not represent optimal or recommended strategies, e.g., a minimum metric gap for faithfulness. In \S \ref{sec:hypothesis}, we hypothesized as to the specific nature and direction of the impact of the above characteristics on post-contrast tuning summaries.

\paragraph*{Random.} For random, for each training instance, we take a random sample without replacement.

\paragraph*{Quality-Based.} For quality-based, we rank all candidates by $Rel_{Agg}$ or $Faith_{Agg}$. Then, we select candidates at different extremes of these scales.

\paragraph*{Margin-Based.} For relevance ranking, we enumerate all possible subsets of size 4 and compute the average metric margin $Avg(Rel_{Agg}(\hat{S_i}, S) - Rel_{Agg}(\hat{S_{i+1}}, S)), i \in |\bm{\hat{S}}| - 1$. We implement both extremes: one which selects the set with the \texttt{Max Margin}, and its inverse, \texttt{Min Margin}. For faithfulness contrast sets, we either take the most faithful positives and least faithful negatives (\texttt{Max Margin}) or the inverse (\texttt{Min Margin}).

\paragraph*{Diversity.} For relevance ranking, we also enumerate all possible subsets of 4 and rank them by their average pairwise inverse self-BLEU score (1 - self-BLEU). We either take the set which has the most \texttt{Max} or \texttt{Min} lexical diversity. We do the same for Faithfulness, except that candidates are selected separately among positive and negative subsets.

\paragraph*{Likelihood.} For relevance ranking, we perform selections based on the model's own beam order. We either take the \texttt{Top Beams} (4), \texttt{Bottom Beams} (4), or top 2 and bottom 2 -- \texttt{Extreme Beams}. For faithfulness, we compute the average token-level log likelihood of generating each candidate in the positive and negative sets after \texttt{FT}. Then we either take the \emph{most} likely positives (2) and \emph{least} likely negatives (2) or the \emph{least} likely positives and the \emph{most} likely negatives. For the former, the model is already well-calibrated, which we call \texttt{Easy}. For the latter, confidence and faithfulness are in conflict, which, in comparison, is \texttt{Hard}.

\paragraph*{Spurious Correlates.} For relevance, we take the \texttt{Short}est and \texttt{Long}est summaries. For faithfulness, we filter for the \texttt{Max Extractive Gap}--the most \emph{extractive} positives and most \emph{abstractive} negatives (as measured by the extractive density).


\begin{table*}[t!]
\small
\centering
\begin{tabular}{cl|ccc|ccc|ccc}
&  \textbf{Model} & \multicolumn{3}{c}{\textbf{\texttt{Clinical}}} & \multicolumn{3}{c}{\textbf{\texttt{Chemical}}} & \multicolumn{3}{c}{\textbf{\texttt{Biomedical}}} \\ \hline
\multirow{3}{*}{\texttt{\makecell{Relevance \\ Metrics}}} & & \textbf{R1} & \textbf{R2} & \textbf{BS-Ref} & \textbf{R1} & \textbf{R2} & \textbf{BS-Ref} & \textbf{R1} & \textbf{R2} & \textbf{BS-Ref} \\
 & \textbf{PRIMERA} & \textbf{25.15} & \textbf{9.39} & \textbf{83.81} & \textbf{45.47} & \textbf{16.31} & \textbf{86.24} & \textbf{48.01} & \textbf{20.83} & \textbf{86.25} \\
 & \textbf{LongT5} & 24.22 & 8.57 & 83.15 & 42.51 & 14.46 & 85.74 & 44.32 & 17.91 & 85.02 \\ \hline
  \multirow{3}{*}{\texttt{\makecell{Faithful \\ Metrics}}} & & \textbf{Fact.} & \textbf{Bart.} & \textbf{BS-Src} & \textbf{Fact.} & \textbf{Bart.} & \textbf{BS-Src} & \textbf{Fact.} & \textbf{Bart.} & \textbf{BS-Src} \\
 & \textbf{PRIMERA} & 53.29 & -2.92 & \textbf{83.33} & \textbf{85.96} & \textbf{-6.29} & \textbf{88.89} & \textbf{86.91} & \textbf{-3.77} & \textbf{88.54} \\
 & \textbf{LongT5} & \textbf{53.71} & \textbf{-2.88} & 82.84 & 83.25 & -6.36 & 88.70 & 83.62 & -3.89 & 88.31 \\
\end{tabular}
\caption{ Benchmarking PRIMERA and LongT5 models after initial fine-tuning (\texttt{FT}) for relevance and faithfulness. R1, R2, and BS-Ref stand for Rouge-1/2 F1 and BERTScore F1 vis-a-vis reference, respectively. Fact., Bart., and BS-Src stand for FactScore, BARTScore, and BERTScore F1 vis-a-vis the source. Metrics defined in \S \ref{sec:relevance-setup} and \ref{sec:faithful-setup}.}
\label{tab:ft-results}
\end{table*}

\begin{table*}[t!]
\small
\setlength{\tabcolsep}{2pt}
\centering
\begin{tabular}{cl|cc|cc|cc||cc}
\multirow{2}{*}{\texttt{\makecell{Selection \\ Type}}} & \multirow{2}{*}{\texttt{\makecell{Selection \\ Strategy}}} & \multicolumn{2}{c}{\texttt{Clinical}} & \multicolumn{2}{c}{\texttt{Chemical}} & \multicolumn{2}{c}{\texttt{Biomedical}} & \multicolumn{2}{c}{\texttt{Dataset Avg.}} \\
& & $REL$ & $FAITH$ & $REL$ & $FAITH$ & $REL$ & $FAITH$ & $REL$ & $FAITH$ \\ \hline
\texttt{Random} & \texttt{-} & .220 & .180 & .081 & \textcolor{red}{-.038} & .028 & .061 & .110 & .068 \\ \hline
\multirow{4}{*}{\texttt{\makecell{Quality \\ Based}}} & \textit{Extreme} & .263 & .152 & .049 & \textcolor{red}{-.168} & .039 & .002 & .117 & \textcolor{red}{-.005} \\
& \textit{Average} & .028 & \textcolor{red}{-.080} & .015 & .056 & .030 & .025 & .024 & .000 \\
& \textit{Min} & .193 & \textcolor{red}{-.022} & .069 & -.049 & .039 & \textcolor{red}{-.012} & .100 & \textcolor{red}{-.027} \\
& \textit{High} & .218 & .095 & .056 & \textcolor{red}{-.029} & .019 & .004 & .098 & .023 \\ \hline
\multirow{2}{*}{\texttt{\makecell{Margin \\ Based}}} & \textit{Max} & .235 & .210 & .062 & .031 & .032 & \textcolor{red}{-.011} & .110 & .077 \\
& \textit{Min} & .158 & \textcolor{red}{-.115} & .028 & .080 & .014 & .015 & .067 & \textcolor{red}{-.007} \\ \hline
\multirow{2}{*}{\texttt{\makecell{Diversity \\ Based}}} & \textit{Max} & .274 & .151 & .054 & \textcolor{red}{-.166} & .015 & \textcolor{red}{-.011} & .114 & \textcolor{red}{-.009} \\
& \textit{Min} & .275 & .091 & \textcolor{red}{-.049} & \textcolor{red}{-.114} & .020 & \textcolor{red}{-.037} & .082 & \textcolor{red}{-.020} \\ \hline
\multirow{3}{*}{\texttt{\makecell{Likelihood \\ Based}}} & \textit{Extreme Beam} & .260 & .140 & .029 & \textcolor{red}{-.158} & .030 & \textcolor{red}{-.008} & .106 & \textcolor{red}{-.009} \\
& \textit{Top Beam} & .287 & .142 & .066 & \textcolor{red}{-.042} & .030 & \textcolor{red}{-.008} & .128 & .031 \\
& \textit{Bottom Beam} & .101 & .125 & .059 & .085 & .025 & \textcolor{red}{-.002} & .062 & .069 \\ \hline
\multirow{2}{*}{\texttt{\makecell{Spurious \\ Correlates}}} & \textit{Max Length} & .255 & .150 & .051 & \textcolor{red}{-.095} & .017 & \textcolor{red}{-.027} & .108 & .009 \\
& \textit{Min Length} & .181 & .243 & .042 & .052 & .033 & .022 & .085 & .106\\ \hline \hline
\multicolumn{2}{c}{\texttt{Avg. Across Strategies}} & .211 & .104 & .044 & \textcolor{red}{-.040} & .027 & .001 & .094 & .022 \\ \hline
\end{tabular}
\caption{ PRIMERA models calibrated to improve relevance. Calibration candidates are pooled from fine-tuned PRIMERA and LongT5 models. $REL$ stands for $Rel_{Agg}$ (from \S \ref{sec:relevance-setup}). $FAITH$ stands for $Faith_{Agg}$ (from \S \ref{sec:faithful-setup}).}
\label{tab:relevance-results}
\end{table*}

\begin{table*}[t!]
\small
\setlength{\tabcolsep}{2pt}
\centering
\begin{tabular}{cl|cc|cc|cc||cc}
\multirow{2}{*}{\texttt{\makecell{Selection \\ Type}}} & \multirow{2}{*}{\texttt{\makecell{Selection \\ Strategy}}} & \multicolumn{2}{c}{\texttt{Clinical}} & \multicolumn{2}{c}{\texttt{Chemical}} & \multicolumn{2}{c}{\texttt{Biomedical}} & \multicolumn{2}{c}{\texttt{Dataset Avg.}} \\
& & $REL$ & $FAITH$ & $REL$ & $FAITH$ & $REL$ & $FAITH$ & $REL$ & $FAITH$ \\ \hline
\texttt{Random} & \texttt{-} & \textcolor{red}{-.264} & .133 & \textcolor{red}{-.054} & .085 & .005 & .165 & \textcolor{red}{-.104} & .128 \\ \hline
\texttt{Quality} & \textit{Average} & \textcolor{red}{-.293} & .160 & \textcolor{red}{-.065} & .037 & .010 & .169 & \textcolor{red}{-.116} & .122 \\ \hline
\multirow{2}{*}{\texttt{\makecell{Margin \\ Based}}} & \textit{Max} & \textcolor{red}{-.326} & .313 & \textcolor{red}{-.139} & .011 & \textcolor{red}{-.033} & .018 & \textcolor{red}{-.166} & .114 \\
& \textit{Min} & \textcolor{red}{-.083} & .297 & \textcolor{red}{-.109} & .112 & \textcolor{red}{-.030} & .039 & \textcolor{red}{-.074} & .149 \\ \hline 
\multirow{2}{*}{\texttt{\makecell{Diversity \\ Based}}} & \textit{Max} & .002 & .290 & \textcolor{red}{-.124} & .043 & \textcolor{red}{-.052} & .029 & \textcolor{red}{-.058} & .121 \\
& \textit{Min} & \textcolor{red}{-.039} & .315 & \textcolor{red}{-.040} & .101 & \textcolor{red}{-.043} & .093 & \textcolor{red}{-.041} & .170 \\ \hline
\multirow{2}{*}{\texttt{\makecell{Likelihood \\ Based}}} & \textit{Easy} & .043 & .177 & -.058 & .002 & \textcolor{red}{-.024} & .071 & \textcolor{red}{-.013} & .083 \\
& \textit{Hard} & .071 & .174 & \textcolor{red}{-.233} & .215 & .013 & .147 & \textcolor{red}{-.050} & .179 \\ \hline
\texttt{Spurious} & \textit{Max Extract. Gap} & .044 & .278 & .058 & .046 & \textcolor{red}{-.051} & .067 & .017 & .131 \\ \hline \hline
\multicolumn{2}{c}{\texttt{Avg. Across Strategies}} & \textcolor{red}{-.094} & .237 & \textcolor{red}{-.085} & .072 & \textcolor{red}{-.023} & .089 & \textcolor{red}{-.067} & .133 \\ \hline
\end{tabular}
\caption{PRIMERA models trained to improve faithfulness with contrastive learning. Contrast sets are formed from the generation methods in \S \ref{sec:faithful-setup}. $REL$ stands for $Rel_{Agg}$ (from \S \ref{sec:relevance-setup}). $FAITH$ stands for $Faith_{Agg}$ (from \S \ref{sec:faithful-setup}). }
\label{tab:faithfulness-results}
\end{table*}

\subsection{Results} \label{sec:results}

\paragraph*{FT Training Details}

We fine-tune (\texttt{FT}) two state of the art long-document summarization models for 50,000 steps: PRIMERA \parencite{primera} (the backbone is a Longformer Encoder-Decoder (LED) \parencite{beltagy2020longformer} model) and LongT5 \parencite{longt5} (which incorporates the sparse attention of ETC \parencite{ainslie-etal-2020-etc} into PEGASUS \parencite{zhang2020pegasus}) on a single A100 40GB GPU with half precision (FP16)\footnote{Only for PRIMERA since LongT5 does not support half precision weights.}) and a batch a size of 1 (with 16 gradient accumulation steps). We set the  maximum learning rate to $3e-5$ with 2,000 warmup steps, followed by a linear decay. We set a maximum input sequence length of 4,096 for both models\footnote{Even though LongT5 has a maximum input sequence length of 16,384, we chose 4,096 to match PRIMERA and because of GPU memory constraints.}, and a maximum target length of 512 for training / inference for abstract generation (Chemical and Biomedical) and 256 for clinical summarization.

We select the better performing model (PRIMERA) as the model to be used for \texttt{CT} (See Table \ref{tab:ft-results}). As discussed in \S \ref{sec:relevance-setup}, LongT5 is still used to supply ten diverse summaries to the candidate pool for relevance contrastive learning.

\begin{table}[h]
\centering
\small
\begin{tabular}{cl|ccc}
& \textbf{Parameter} & \textbf{Clin} & \textbf{Chem} & \textbf{Bio} \\ \hline
\multirow{5}{*}{\makecell{Relevance \\ Ranking}} & $\lambda_{MLE}$ & 0.1 & 0.1 & 0.1  \\
 & $\lambda_{CA}$ & 1.0 & 1.0 & 1.0  \\
 & $\lambda_{margin}$ & .001 & .001 & .001  \\
 & $\alpha$ (length penalty) & 1.0 & 2.0 & 2.0  \\
 & $\tau$ (scale) & .01 & 0.1 & 0.1  \\ \hline
 \multirow{2}{*}{\makecell{Faithful \\ Contrast}} & $\lambda_{MLE}$ & 1.0 & 1.0 & 1.0  \\
 & $\lambda_{CA}$ & 1.0 & 10.0 & 1.0  \\ \hline
\end{tabular}
\caption{Hyper-Parameters for contrastive learning fine-tuning.} \label{tab:hparams}
\end{table}

\paragraph*{CT Training Details}

We run contrastive tuning (\texttt{CT}) for a maximum of 10,000 steps and select the checkpoint which maximizes either $Rel_{Agg}$ or $Faith_{Agg}$ (depending on the experiment) on the validation set in 1,000 step intervals.

We use the same hyper-parameters as $FT$ except the batch size is reduced from 16 to 8. Hyper-parameters related to the \texttt{CT} loss function were tuned separately for each dataset and quality metric (the values selected are shown in Table \ref{tab:hparams}). Each \texttt{CT} experiment took $\sim1$ day to train.

As in \textcite{longt5}, summaries are generated greedily, which we found to be significantly faster and even slightly outperformed beam search\footnote{This also means that a length penalty cannot be applied during decoding, which puts more emphasis on the significant role of length tuning during relevance contrastive learning.}.

\paragraph*{Fine-Tuning} \label{sec:ft-results}

Table \ref{tab:ft-results} shows that PRIMERA outperforms LongT5 across faithfulness and relevance and across datasets\footnote{We note that these our results from own runs. They do not represent results from the PRIMERA and LongT5 papers.}. Relevance and faithfulness are much higher for abstract generation (Chemical and Biomedical) than for clinical summarization, which has highly noisy references. Interestingly, the BARTScore results are lowest for the chemical dataset (-6.29/-6.36 versus -2.92/-2.88 and -3.77/-3.89). This underscores the difference in biomedical versus chemistry-specific papers because the BARTScore model used was trained on the PubMed dataset (\texttt{google/pegasus-pubmed}).

\paragraph*{Contrastive Tuning} \label{sec:calibration-results}

In Tables \ref{tab:relevance-results} and \ref{tab:faithfulness-results}, we report results for relevance, rank-based calibration (\S \ref{sec:relevance-setup}) and faithfulness contrastive learning (\S \ref{sec:faithful-setup}), respectively. $Rel_{Agg}$ and $Faith_{Agg}$ are normalized such that positive values represent standard deviation improvements over fine-tuning, while negative results show a decrease in performance from tuning (marked in \textcolor{red}{red}).

In the following sections, we break down analysis into a \textit{tl;dr}, \textit{evidence}, \textit{explanation}, and potential \textit{implications}, or takeaways, for future research.

\paragraph*{The Impact of Reference Quality} \label{sec:dataset-results}

\paragraph*{tl;dr.} Relevance and faithfulness contrastive learning offer the most upside when references are noisy.

\paragraph*{Evidence.} As detailed in Chapter \ref{ch:dataset} and \S \ref{sec:ltr}, clinical references are often unsupported by the source text. The average across strategies for both Tables \ref{tab:relevance-results} and \ref{tab:faithfulness-results} reveal the largest relative improvement in $Rel_{Agg}$ and $Faith_{Agg}$ for clinical, respectively ($.211$ / $.237$ versus $.044$ / $.072$ and $.027$ / $.089$ for chemical and biomedical abstracts).

\paragraph*{Explanation.} For relevance contrastive learning, it is likely that training on model outputs, especially highly extractive ones, dampens some of the noise from variable references. For faithfulness, the rationale is less clear because the reference (and paraphrases of it) form the positive set. Yet, there is an extensive body of work to suggest that training on unfaithful references leads to unfaithful outputs \parencite{loss-truncation}, which might make contrastive learning for faithfulness more impactful.

\paragraph*{Implications.} Calibration could be complementary to other methods which address noisy references, such as loss truncation \parencite{loss-truncation}, data filtering \parencite{narayan-etal-2021-planning, nan-etal-2021-entity}, and reference revision \parencite{wan-bansal-2022-factpegasus, adams-etal-2022-learning}.

\begin{figure}[ht]
\centering
\includegraphics[width=0.75 \linewidth]{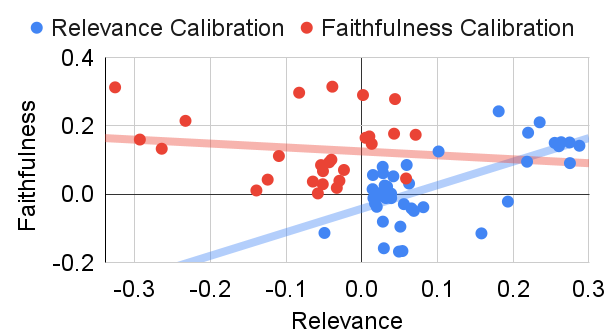}
\caption{A plot of average summary relevance and faithfulness across experiments, which are designed to either improve relevance (blue) or faithfulness (red).} 
\label{fig:rel-versus-faith}
\end{figure}

\paragraph*{Relevance and Faithfulness at Odds}

\paragraph*{tl;dr.} Relevance and faithfulness share an inverse relationship when contrastive learning for faithfulness. Research should focus on designing contrast sets that maximize their correlation for joint optimization.

\paragraph*{Evidence.} In Figure~\ref{fig:rel-versus-faith}, we plot $Rel_{Agg}$ versus $Faith_{Agg}$ across experiments to measure the tradeoff between relevance and faithfulness. On average, improving faithfulness comes at the cost of relevance, yet the trend is not conclusive. This is validated by previous work which shows a decrease in relevance when models are trained to be more faithful \parencite{filippova2020controlled, narayan-etal-2021-planning}. Faithfulness and relevance appear to be positively related when contrastive learning for relevance. This might be a spurious correlation, however. Model summaries are more extractive than references for each dataset. Including highly extractive summaries as candidates for contrastive learning, in turn, leads to to even more extractive models, as the extractive density of PRIMERA summaries rises from 3.1 / 9.2 / 13.0 after \texttt{FT} to an average of 3.5 / 11.4 / 14.0 for clinical / chemical / biomedical after a round of contrastive tuning.

\begin{table}[h]
\centering
\small
\begin{tabular}{l|ccc|c}
\hline
\textbf{System} & \textbf{Int.} & \textbf{Ext.} & \textbf{Total} & \textbf{Rel. Rank} \\
\texttt{FT} & 2.00 & 1.24 & 3.24 & 2.04 \\
\texttt{Most Relevant} & 1.67 & 1.43 & 3.10 & 1.85 \\
\texttt{Most Faithful} & 1.10 & 0.81 & 1.90 & 2.12 \\
\hline
\end{tabular}
\caption{Results from human evaluation on 75 total system summaries from the chemistry test set. Int. and Ext. stand for average intrinsic and extrinsic errors identified. Rel. Rank stands for the average rank assigned by annotators (1-3) with 1 being viewed as the most relevant.} \label{tab:human-eval}
\end{table}

To better understand whether or not our contra models are driving meaningful changes in quality, we conduct a human evaluation on the chemistry dataset. Specifically, we randomly select 25 papers from the test set and collect model generated abstracts from the \texttt{FT} checkpoint as well as most relevant (\texttt{Random} strategy) and most faithful (\texttt{Hard} strategy) \texttt{CT} weights. After randomly shuffling the order of abstracts, we ask each annotator (PhDs in chemistry-related fields) to first read the main paper and then, separately for each paper, highlight spans of abstracts containing errors (intrinsic or extrinsic), before ranking the summaries by Relevance \parencite{fabbri-etal-2021-summeval}. We defined relevance as in SummEval: \textit{how well does the summary captures the key points of the paper? Consider whether all and only the important aspects are contained in the summary.}. We collect fine-grained faithfulness annotations, rather than summary-level, due to the length of the summaries and prior work on inter-annotator agreement scores of fine-grained errors \parencite{pagnoni-etal-2021-understanding, goyal-durrett-2021-annotating}.

On a small sample, Table \ref{tab:human-eval} confirms what the metrics reveal: an inverse relationship between faithfulness (Int., Ext., Total error counts) and relevance (Rel. Rank). \texttt{Most Faithful} (according to $Faith_{Agg}$) summaries contain the fewest annotated total errors ($1.90$ versus $3.24$ and $3.10$) yet are ranked least relevant (average rank of $2.12$ versus $2.04$ and $1.85$). \texttt{Most Relevant} (according to metrics) achieves the highest relevance ranking from experts ($1.85$ versus $2.04$ / $2.12$) while slightly reducing the number of errors from $FT$: $3.10$ versus $3.10$. On average, there are more intrinsic errors versus extrinsic, which makes sense given how extractive the generated abstracts are. \texttt{Most Relevant} abstracts contain the highest average number of Extrinsic errors ($1.43$ versus $1.24$ and $0.81$), which could stem from the fact that abstracts, as naturally occurring summaries, may introduce external knowledge into the abstracts, for which the \texttt{Most Relevant} may be mimicking.

\paragraph*{Explanation.} From Table \ref{tab:method-metrics}, while references, from a metric perspective, are perfectly relevant, the GPT-3 paraphrases are seen as slightly less relevant (0.9 / 0.94 / 0.92), on average, than the negative methods (0.94 / 0.97 / 0.97) in aggregate). This is likely a by-product of the fact that the negative generation methods selected for this paper involve local corruptions to the reference. The meaning is changed but the word overlap is similar. The GPT-3 paraphrases are prompted with human paraphrases, which involve more substantial re-writing.

\paragraph*{Implications.} Most research is focused on either relevance or faithfulness. We advocate that more papers address them together, since both informativeness and faithfulness are important for real-world systems. Future research could explore joint optimization by intentionally introducing more errors into less relevant summaries.

\begin{table}[h]
\small
\centering
\begin{tabular}{l|ccc|ccc}
 & \multicolumn{3}{c}{\textbf{Average Strategy}} & \multicolumn{3}{c}{\textbf{Max Correlation}} \\
& Rel & Faith & Comb & Rel & Faith & Comb \\ \hline
\textbf{Clin.} & .211 & .104 & .158 & .090 & .325 & .208 \\
\textbf{Chem.} & .044 & -.040 & .007 & .040 & .104 & .158 \\
\textbf{Bio.} & .027 & .001 & .014 & .018 & .025 & .022 \\ \hline
\textbf{Avg.} & .094 & .022 & \textbf{.059} & .049 & .151 & \textbf{.100} \\  \hline
\end{tabular}
\caption{Relevance \texttt{CT} by forming sets which maximize rank correlation between Rel. and Faith. scores improves mean combined (comb.) Rel. and Faith. scores vis-a-vis an average of the strategies shown in Table \ref{tab:relevance-results}.}
\label{tab:both}
\end{table}

As a quick proof of concept, we define a hybrid selection strategy which maximizes the rank correlation between $Agg_{Rel}$ and $Agg_{Faith}$. Table \ref{tab:both} demonstrates that contrast tuning on these sets leads to positive (pareto) improvements for both metrics. The average improvement in combined metrics across datasets is $.1$, which is greater than an average of the strategies shown in Table \ref{tab:relevance-results} ($.059$).

\begin{figure}[h]
\centering
\includegraphics[width=0.75 \linewidth]{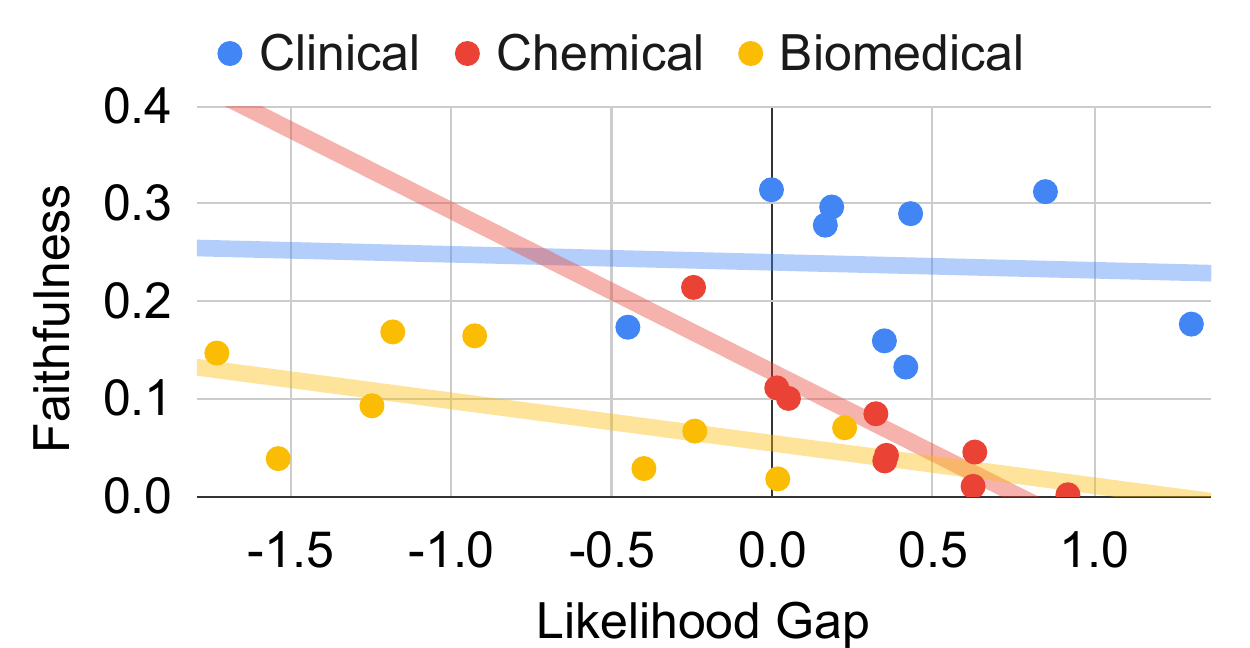}
\caption{A plot comparing the average likelihood gap (difference in log likelihood of generating a positive candidate over a negative pre-contrast tuning) against the average summary faithfulness after contrast tuning.} 
\label{fig:easy-hard}
\end{figure}

\begin{figure}[h]
\centering
\includegraphics[width=0.75 \linewidth]{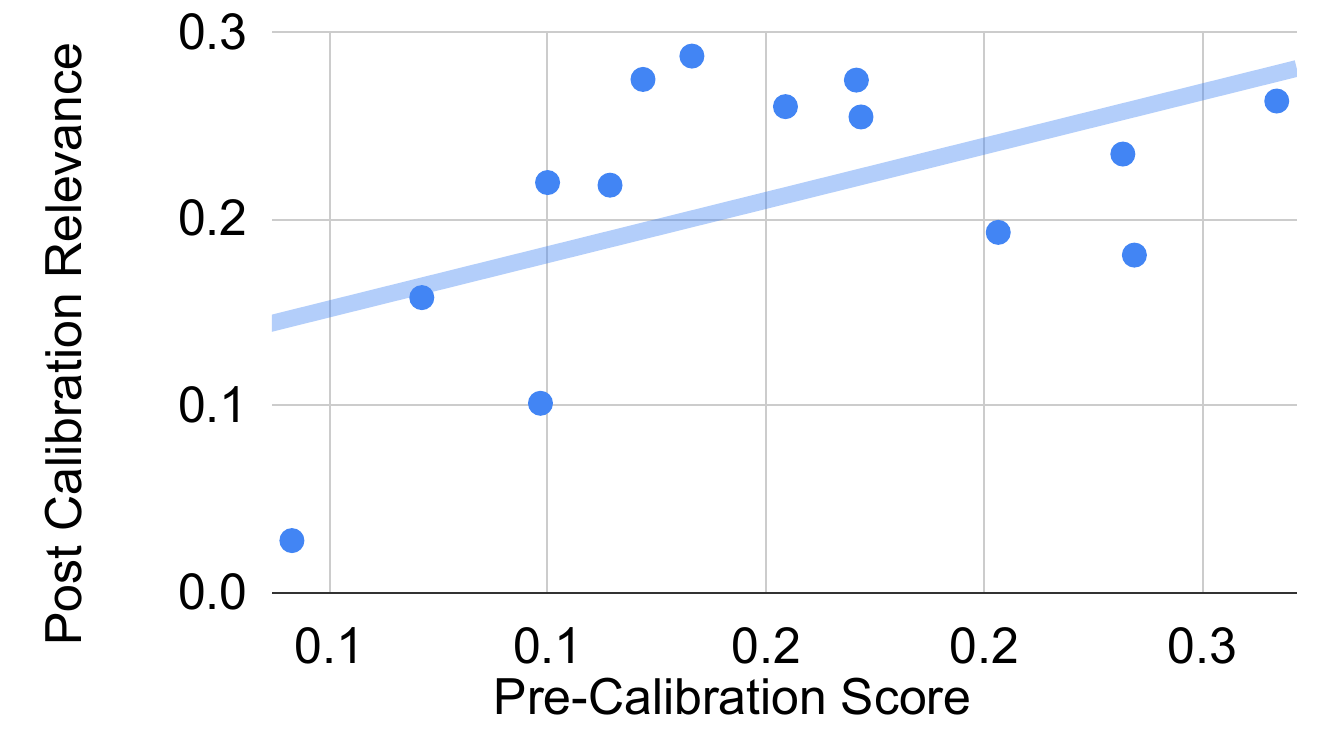}
\caption{A plot which shows average pre-tuning score for each clinical relevance experiment on the x-axis, and the post-tuning relevance on the y-axis.} 
\label{fig:pre-calibration-score}
\end{figure}

\paragraph*{On the Dual Role of Surprise} \label{sec:surprise}

\paragraph*{tl;dr.} Summaries in sets should be likely under the fine-tuned model. Yet, for relevance, this confidence should mostly already agree with the oracle ranking, while contrastive learning for faithfulness is most effective when the model is surprised.

\paragraph*{Evidence.} For relevance, we look at the \texttt{Likelihood} section of Table \ref{tab:relevance-results} and note that, of all strategies, taking the top 4 beams is the most effective (an average of $.128$ across datasets). Taking the bottom beams is one of the worst ($.062$) and taking some from each lies in the middle ($.106$). For faithfulness, we examine  the \texttt{Likelihood} section of Table \ref{tab:faithfulness-results} and note that \texttt{Hard} is the best strategy, on average, across datasets ($.179$ for $Faith_{Agg}$) and \texttt{Easy} is the worst ($-.083$). \texttt{Hard} selects negatives which are most likely under the model, which suggests that contrastive learning for faithfulness is most effective when the model is ``surprised'', i.e., the negative summaries are as likely, if not more, to be generated as the positives.

Across all selection strategies and datasets, we can compute the pre-tuning, average likelihood gap between positives and negatives and regress it against the post-tuning $Faith_{Agg}$ (Figure \ref{fig:easy-hard}). An inverse relationship emerges, especially for chemical dataset (a pearson correlation of $-.91$).

We can run a similar analysis for relevance contrastive learning by computing an average pre-tuning score for each selected set, which we define as the negative spearman correlation coefficient between the model beam and the $Rel_{Agg}$ ranking. It measures the extent to which the model is pre-calibrated from MLE \texttt{FT}. We plot this set statistic against the post-tuning $Agg_{Rel}$ score, as shown in Figure \ref{fig:pre-calibration-score}. The pearson correlation coefficient for the pre-tuning statistic to post-tuning relevance is .52, which is stronger than the correlation of average beam of candidates to relevance (.45).

\begin{figure}[h]
\centering
\includegraphics[width=0.75\linewidth]{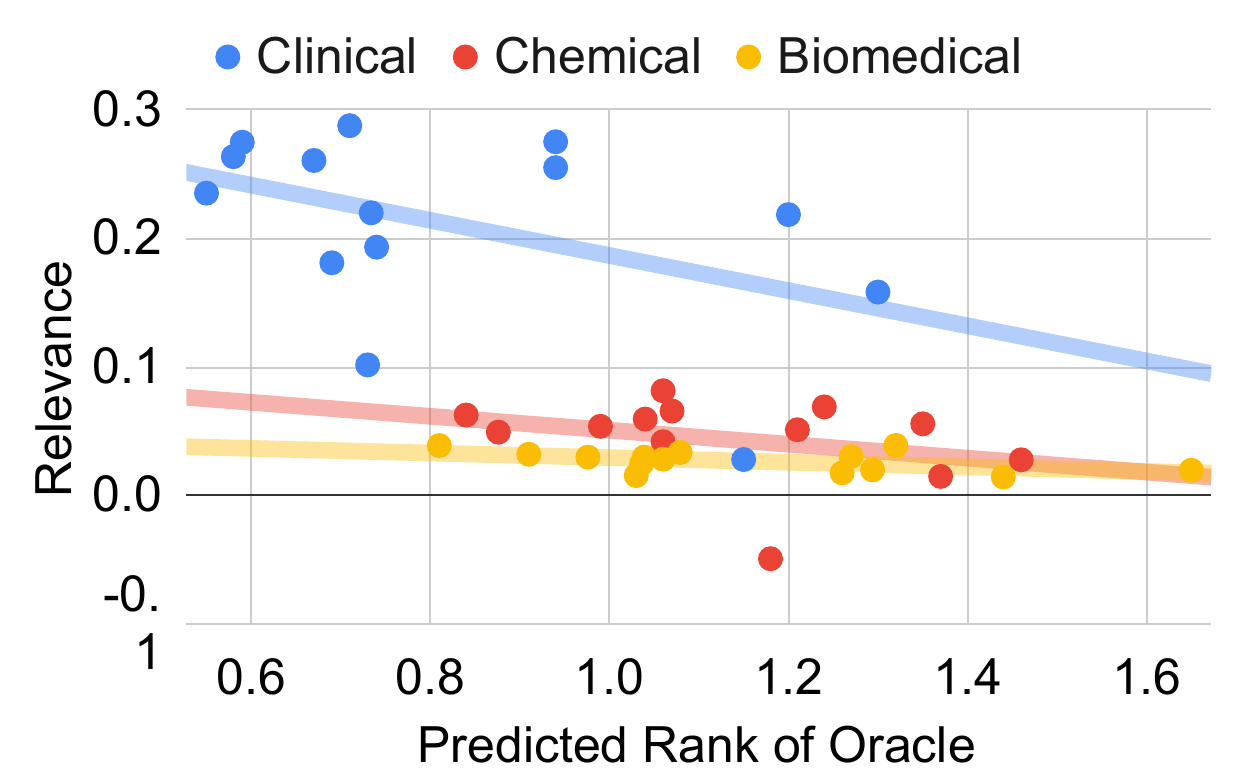}
\caption{A plot showing the impact of contrastive tuning on downstream performance (relevance). An average rank of 0 reveals a model which always identifies the most relevant summary. The worst score is 3.} 
\label{fig:predicted-rank}
\end{figure}

We can also link the model's ranking ability \emph{after} contrastive tuning to the post-contrastive relevance. In other words, does it matter how well the model can rank candidates given that, when used for inference, it generates a single candidate? Figure \ref{fig:predicted-rank} shows that a well calibrated model is a better generator due to an inverse relationship between the predicted rank of the top ranked candidate (x-axis) and the average post-tuning $Rel_{Agg}$ score (y-axis).

Taken together, the results suggest that an optimal rank set for relevance is one that is fairly calibrated before \texttt{CT} and well-calibrated after \texttt{CT}.

\paragraph*{Explanation.} A possible explanation for this conflicting evidence is a difference in objectives. As in \textcite{liu-etal-2022-brio}, the relevance ordering is directly calibrated to log likelihood of outputs, whereas for faithfulness, we contrast binary positives and negatives in latent space. For the former, large parameter updates from the ranking loss directly affect the generation behavior of the model, which \emph{may} push outputs further away from the MLE optimum.

\paragraph*{Implications.} The results suggest it might be preferable to \emph{surprise} for faithfulness contrastive learning yet \emph{confirm} for relevance contrastive learning. Yet, further work is necessary to assess whether this behavior is attributable to the objective or the metric. 


\begin{figure}[h]
\centering
\includegraphics[width=0.75\linewidth]{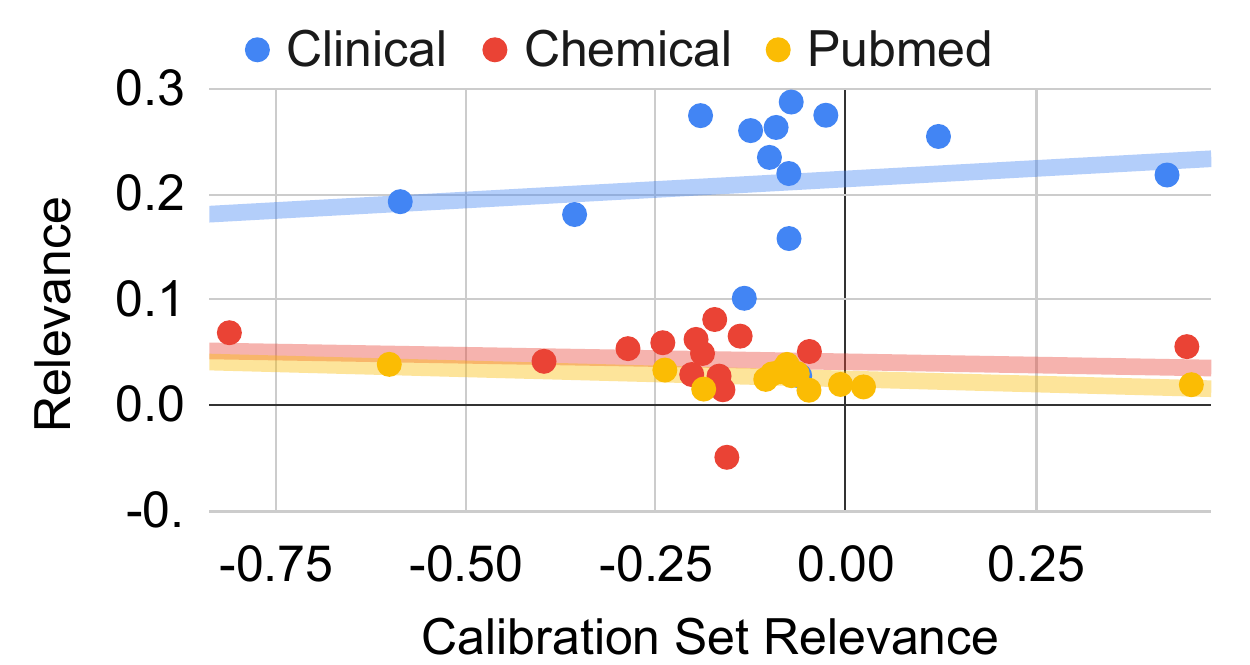}
\caption{The impact of the average relevance of contrast sets on downstream summary relevance.} 
\label{fig:absolute}
\end{figure}

\begin{figure}[h]
\centering
\includegraphics[width=0.75\linewidth]{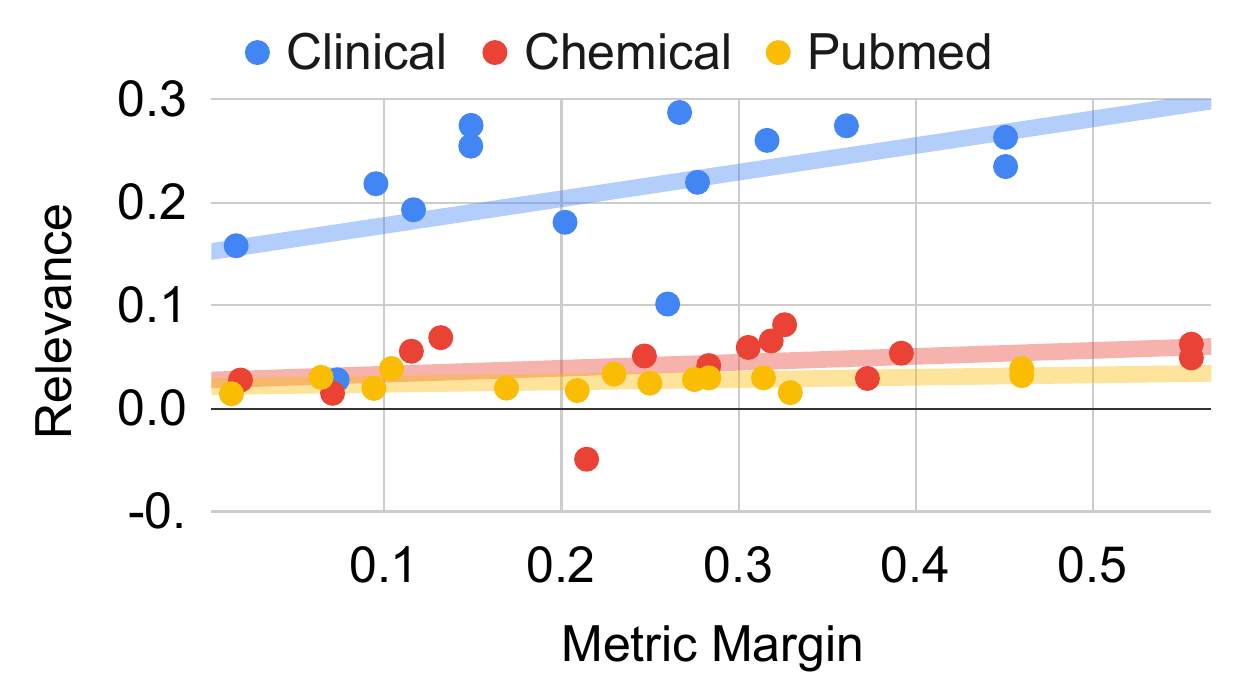}
\caption{The impact of the average metric-wise margin ($Rel_{Agg}$) between contrast \emph{candidates} on the relevance of downstream model outputs after contrastive learning.}
\label{fig:margin}
\end{figure}

\paragraph*{Margin over Absolute}

\paragraph*{tl;dr.} For relevance training, the presence of a large metric margin between candidate summaries appears to be more impactful to downstream performance than the overall relevance of the set.

\paragraph*{Evidence.} Based on Table~\ref{tab:relevance-results} for \texttt{Quality Based} Avg. Across Strategies, no clear-cut trend exists between $Rel_{Agg}$ and absolute relevance values: $.117 / .024 / .100 / .098 $ for \texttt{Extreme}, \texttt{Average}, \texttt{Min}, and \texttt{High}, respectively. For \texttt{Margin Based}, which targets the relative values, \texttt{Max} outperforms $.110$ over $.067$. To better uncover any trends, we separately plot the average set relevance (absolute value), and the Margin Gap (relative values), against downstream $Rel_{Agg}$ for each run (row in Table~\ref{tab:relevance-results}) in Figures~\ref{fig:absolute} and \ref{fig:margin}. Figure~\ref{fig:margin} shows a positive correlation between margin gap and downstream $Rel_{Agg}$ across datasets (pearson correlation of $.48$, $.29$, and $.38$ for clinical, chemical, and biomedical, respectively). The relationship in Figure \ref{fig:absolute} is less consistent, as it is positive for clinical ($.12$ correlation), yet negative for chemical ($-.10$) and biomedical ($-.51$). Larger margin gaps are related to diversity as lexically similar summaries will have similar metric values. In fact, we can examine the \texttt{Diversity} section of Table~\ref{tab:relevance-results} and note that average $Rel_{Agg}$ score across datasets is higher when lexical diversity is maximized ($.114$) than when it is minimized ($.082$). Yet, this trend only holds for the Chemical dataset. To get a more complete sense, we examine the impact of set diversity across runs and note a slightly more reassuring trend: a pearson correlation coefficient of $.21$, $.51$, and $.1$ for clinical, chemical, and biomedical. Interestingly, chemical has the strongest positive relationship between diversity and downstream relevance across runs, yet is negative when directly controlling for diversity.

\paragraph*{Implications.} Diversity may help contrastive learning with increased exploration and smooth out some noise from ROUGE / BERTScore defined rankings. Although \textcite{zhao2022calibrating} find consistently better performance using regular beam search over diverse beam search, the opposite may hold true for longer tasks with larger output search spaces.

\paragraph*{Analysis of Spurious Correlates}

\begin{figure}[h]
\centering
\includegraphics[width=0.5\linewidth]{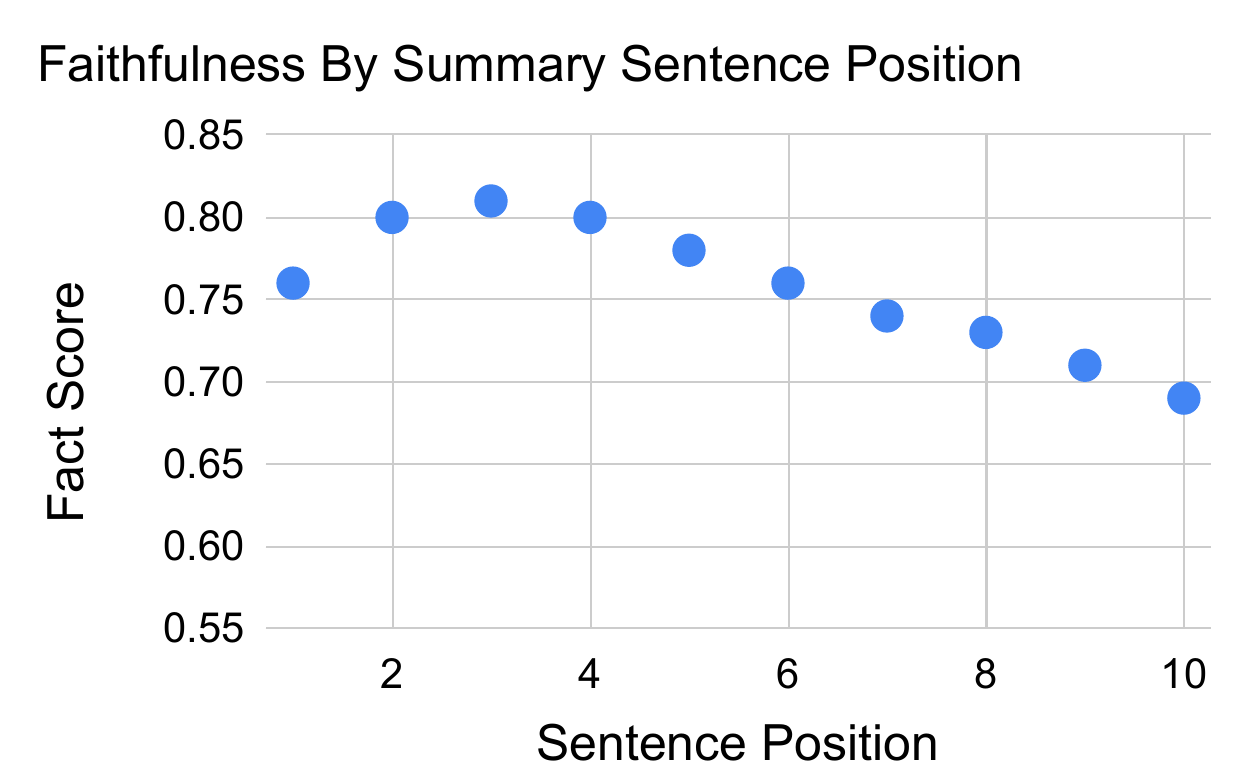}
\caption{Sentence-level faithfulness, as defined by FactScore in \S \ref{sec:faithful-setup}, declines as summaries grow longer.} 
\label{fig:curious-case}
\end{figure}

\paragraph*{The Outsized Role of Length}

\paragraph*{tl;dr.} The length of summaries is correlated with performance for both relevance and faithful conrtast tuning yet for different reasons. For relevance, it can help reduce discrepancies in token-level length between references and generated summaries after fine-tuning. For faithfulness, generated summaries become less faithful as average length increases.

\paragraph*{Evidence.} For relevance contrastive learning, the Table~\ref{tab:relevance-results} section on \texttt{Spurious Correlates} shows that selecting the longest summaries is preferable to the shortest for Clinical contrastive learning (.255 versus .181) yet the reverse is true for Biomedical (.017 for max length and .033 for min length). We can trace this to a gap, after fine-tuning, in model summary length and reference lengths. On average, PRIMERA summaries after \texttt{FT} are 119 tokens for clinical and 230 for biomedical. Yet, the clinical references are, on average, 416 tokens and only 205 for biomedical. The optimal length strategy seems contingent on the direction of the length gap.

For faithfulness, we simply compute the correlation between $Faith_{Agg}$ and summary tokens: $-.75$. For faithfulness, we can confirm the presence of text degeneration \parencite{holtzman2019curious} as a function of output length by measuring the average $FactScore$ at each sentence position in the summary. Figure \ref{fig:curious-case} confirms this story, despite an initial slight increase up to the third sentence.

\paragraph*{Implications.} For relevance, as argued by \textcite{sun-etal-2019-compare}, work should acknowledges changes in the lengths of summaries and address its role in impacting relevance metrics. Long-form summarization research which involves identifying and solving subproblems \parencite{krishna-etal-2021-generating} might mitigate some of the length-based degeneration.

\begin{table}[h]
\centering
\small
\begin{tabular}{l|ccc}
\textbf{Metric} & \textbf{Clinical} & \textbf{Chemical} & \textbf{Biomedical} \\ \hline
FactScore & .78 & .42 & .42 \\
BARTScore & .35 & .16 & .45 \\
BERTScore-Src & .52 & .47 & .60 \\
\end{tabular}
\caption{Correlation of faithfulness metrics to extractive density of summaries. Correlations computed on the test set of the PRIMERA models after fine-tuning.} \label{tab:extractive-is-faithful}
\end{table}

\paragraph*{Faithful or More Extractive?}

\paragraph*{tl;dr.} One would expect that training on contrast sets with a large difference in extractiveness (extractive positives, abstractive negatives) would lead to higher downstream faithfulness. Yet, we find the opposite to be true, which we connect to \S \ref{sec:surprise}.

\paragraph*{Evidence.} \textcite{ladhak-etal-2022-faithful} note a spurious correlation between the extractiveness of summaries and faithfulness metrics, which holds true for the metrics which make up $Faith_{Agg}$ (as shown in Table \ref{tab:extractive-is-faithful}). One would expect that reinforcing this correlation via contrastive learning (by targeting extractive positives and abstractive negatives) would lead to improved faithfulness metrics. Yet, this does not appear to be the case. Table \ref{tab:faithfulness-results} (\texttt{Spurious} selection type) shows that on average, controlling for a large extractiveness gap does not improve faithfulness ($.131$ versus an overall average improvement of $.133$). If anything, it leads to increased relevance ($.017$ versus $-.067$). While not definitive, a possible driver for this relationship relates to the analysis in \S \ref{sec:surprise}, for which we show that a low likelihood gap between positives and negatives is preferable (an adversarial setup). Since extractive summaries are more likely to be generated than abstractive ones (see Extractive density for Diverse Beam search in Table \ref{tab:method-metrics}), extractive negatives might be preferable to abstractive ones.

\paragraph*{Implications.} Given the extractiveness of long-form scientific summaries, more research should focus on subtle faithfulness errors, i.e., those which are less correlated to extractiveness. \textcite{zhang2022extractive} provide a helpful typology of errors in fully extractive systems, which can provide a blueprint for the design of more extractive synthetic errors.

\begin{figure*}[t]
\centering
\includegraphics[width=\linewidth]{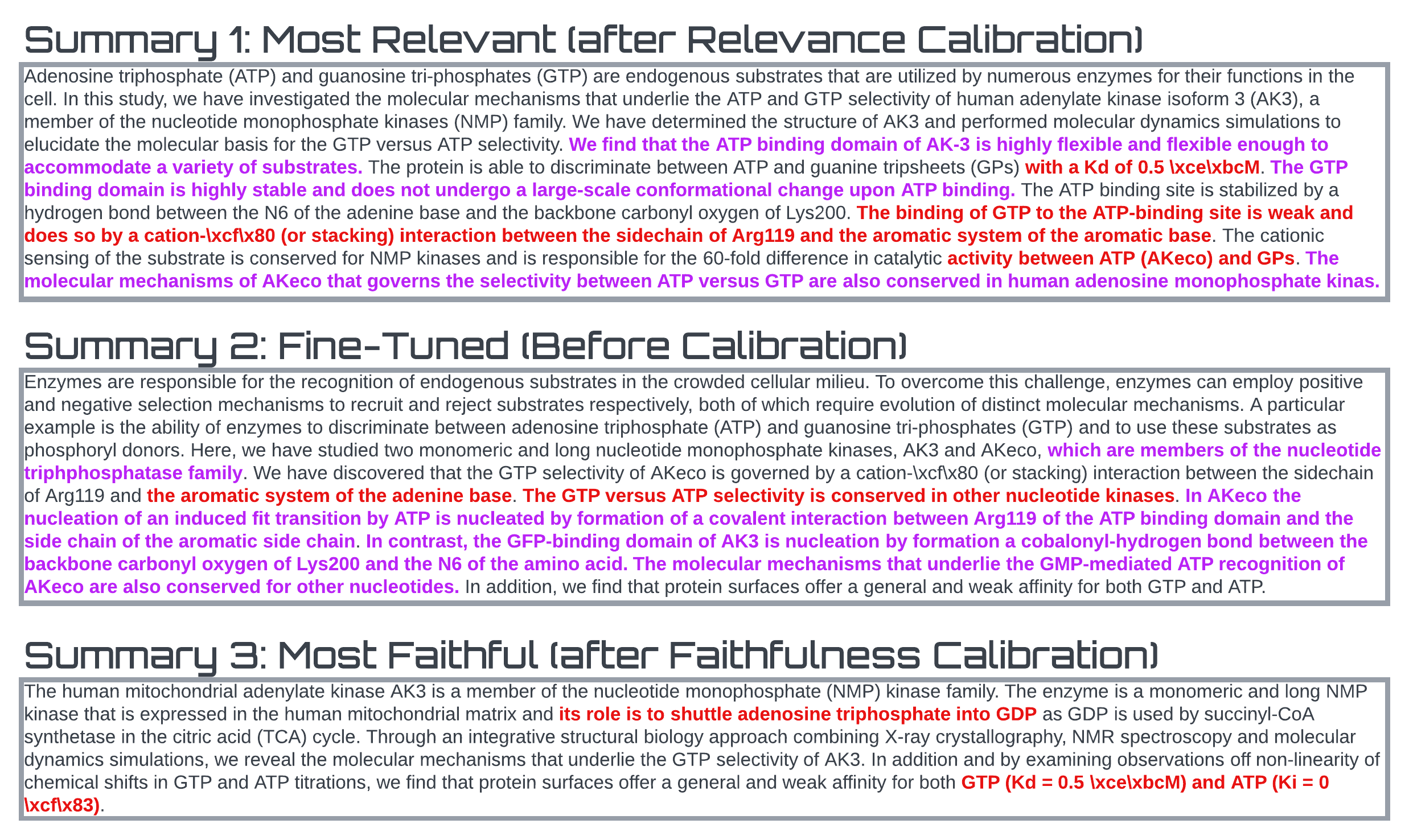}
\caption{Three abstracts generated from model checkpoints after relevance contrastive learning (Summary 1), Fine-Tuning (PRIMERA FT checkpoint, Summary 2), and after faithfulness contrastive learning (Summary 3). \textcolor{red}{Red Text} has been annotated as being part of an intrinsic error while \textcolor{purple}{Purple Text} is extrinsic.  The annotator rated Summary 1 as the most relevant and Summary 3 the least relevant. } 
\label{fig:human-example}
\end{figure*}

\paragraph*{Error Analysis} \label{sec:error-analysis}

In this section, we analyze the errors from an example in the human annotation set. The abstracts are shown in Figure \ref{fig:human-example}.

Abstract 1 takes the general form of an abstract, providing a reasonable motivation for the work then listing a number of key findings. It makes a number of errors in stating the key findings, however. First, the model seems to have had difficulty with abbreviations and measured values, misreporting a binding constant and confusing GTP and ATP on several occasions. Finally, the model includes several statements not supported in the text. Abstract 2 contains superior prose to Abstract 1, better enumerating the motivation for the work and providing a cleaner concluding statement. It suffers from similar shortcomings, however, confusing GTP and ATP on several occasions and making a number of unsupported claims. In some cases, the unsupported claims appear lifted whole-cloth from another publication. In total, we judge the errors in Abstract 2 to be more misleading than those made in Abstract 1 and thus find Abstract 1 to be more relevant. Abstract 3 is substantially shorter than either Abstract 1 or Abstract 2, minimizing the absolute number of errors it contains. Like the others, it has difficulty with both abbreviations and measured values, making errors due to both. Overall, Abstract 3 is not terribly written; however, its terseness leaves a highly limited description of the paper's contributions. For this reason, it is less relevant than either Abstract 1 or Abstract 2.

\subsection{Conclusion}

We explore what makes an effective contrast set for both relevance and faithfulness tuning. To do so, we create large candidate pools for contrastive learning and design strategies which systematically target set characterstics. We then analyze trends between these characteristics and downstream performance. Our analysis is intended to serve as a guide for subsequent research when designing methods to form synthetic candidates, as well as motivation to jointly consider relevance and faithfulness for contrastive learning, given their covariance and the importance of both to real-world systems.

%% file: 5-measuring-faithfulness.tex
\chapter{Measuring Faithfulness} \label{ch:measuring}

The work described in the previous chapter (Chapter \ref{ch:improving}) heavily relied on existing metrics of faithfulness which were largely developed on single document, general domain corpora~\cite{factcc,durmus-etal-2020-feqa,wang-etal-2020-asking,deng-etal-2021-compression,yuan2021bartscore,laban-etal-2022-summac,ribeiro-etal-2022-factgraph}, such as CNN/Daily Mail and BBC news articles. The generalizability of these metrics to long-form hospital-course summarization is not guaranteed.  Additionally, some metrics rely on external components and tools, such as question generation \parencite{fabbri-etal-2022-qafacteval} and dependency parsing \parencite{goyal2020evaluating}, which do not work on clinical texts given the massive difference in syntactic patterns and discourse structures. There is a wealth of evidence to demonstrate the beneficial impact of in-domain pre-training on clinical \parencite{alsentzer2019publicly, lehman2023we} and biomedical \parencite{gu2021domain} downstream tasks. Yet, domain adaptation has largely been unexplored when it comes to the development of the metrics used to evaluate domain-specific models. To confidently measure progress in the field, investigate the importance of domain adaptation, and build metrics that can be safely used for model alignment, we need to be able to trust our metrics. Meta-evaluating metrics requires conducting a human evaluation study and then measuring the correlation of different metrics to the ratings provided by experts. To this end, we benchmark faithfulness metrics against fine-grained human annotations for model-generated summaries of a patient's Brief Hospital Course.

Specifically, we fine-tune a long-range transformer (Longformer Encoder-Decoder (LED) \parencite{beltagy2020longformer} on our private dataset of Hospital Course summaries from Chapter \ref{ch:dataset}, as also described in \textcite{adams-etal-2021-whats}. It pertains to all in-patient hospital admissions at Columbia University Irving Medical Center from 2010-2014. On a \emph{held-out} set of admissions for patients from the HIV clinic \parencite{levy2020towards}, we rely on expert (clinicians) to collect fine-grained faithfulness annotations of LED summaries based on the clinical notes written before discharge. Annotators are presented with summaries and source notes, and asked to categorize manually highlighted summary elements (clinical entities like conditions and medications as well as actions like "following up") into one of three categories: ``Incorrect,'' ``Missing,'' and ``Not in Notes.'' To help annotators locate and identify relevant context, we built a custom user interface that allowed for quick search and visualization of all mentions of medical concepts across the source notes. We then meta-evaluate a large set of existing summarization evaluation metrics (including BARTScore \parencite{yuan2021bartscore}, BERTScore \parencite{zhang2019bertscore}, Entailment-based CTC \parencite{ctc} and SummaC \parencite{laban-etal-2022-summac}) by measuring their correlation to these fine-grained human annotations. 

Since these metrics were mostly developed on single document general-domain corpora, we identify three key dimensions pertinent to adaptation to long-form clinical summarization: domain adapation (pre-training and metric fine-tuning), length of inputs, and length of outputs. For length-based dimensions, we explore the impact of source-summary alignments and summary granularity (sentence-level versus summary-level). We find that metrics tend to correlate best with human annotations when provided summary sentences one at a time, and when only the most relevant content (high precision source-summary alignments) is provided. We see limited benefits from domain adaptation with respect to simple correlation analysis, yet we attribute much of this to the abstractiveness of the references on which metrics are tuned. When filtering for abstractive subsets of the annotation set, domain adaptation starts to outperform off-the-shelf variants. In-domain adaptation of metrics will likely be critical given the observed abstractiveness of summaries from LLMs \parencite{goyal2022news}. Rather than adapt metrics to clinical text by training on references, we find it advantageous to learn directly from system summaries. We use an ensemble of our baseline metrics to produce a pseudo faithfulness score on system summaries and distill a metric from these noisy ground-truth labels. Our distilled metric has a higher correlation than baseline metrics to expert annotation labels.

The contributions of this chapter are to:
 
\begin{itemize}
    \item We collect finegrained faithfulness annotations for the the task of hospital-course summarization, which contains substantially longer inputs than previous clinical annotation efforts.
    \item We benchmark existing faithfulness metrics against these annotations, as well as explore practical considerations of adapting general
    domain metrics to long-form clinical narratives.
    \item  We analyze the confounding role of extractiveness and show how a simple statistic (unigram coverage) can be complementary to other metrics, including a metric distilled from an ensemble of other metrics.
\end{itemize}

\begin{table}[h]
\centering
\small
\begin{tabular}{l|c|cc|cc}
\multirow{2}{*}{\texttt{Split}} & \multirow{2}{*}{\texttt{\#}} & \multicolumn{2}{c}{\texttt{Source}} & \multicolumn{2}{c}{\texttt{Reference}}  \\
& & Notes & Tokens & Sents & Tokens \\ \hline
Train - Full & 82k & 41 & 18.4k & 11.6 & 207 \\  
Train - HIV & 2.7k & 40 & 19.1k & 12.5 & 243 \\ 
Eval - HIV & 29 & 24 & 11.7k & 12.1 & 211 \\ 
\end{tabular}
\caption{Data Statistics for training the summarization LED model (Full Train), the subset used for in-domain \emph{evaluation metric} training, as well as the subset of the test set used for human evaluation (Annot.). } \label{tab:data-stats}
\end{table}

\subsection{Data}

The data is comprised of clinical notes from the Electronic Health Record (EHR) for in-patient admissions between 2010 and 2014 at a large metropolitan hospital \parencite{adams-etal-2021-whats}.

\paragraph*{Training Data.} We show training data statistics in the first row of Table \ref{tab:data-stats}. We delineate between the full training set, which is used to train the summarization models and the subset of the training set which is used for fine-tuning evaluation metrics in-domain. The subset filters for HIV patients which mirrors the filtering done to produce the human evaluation cohort (discussed directly below).

\paragraph*{Human Evaluation Cohort.} The training set comprises both HIV and non-HIV patients while the human annotation test set is solely HIV. We choose HIV patients as they typically have multiple co-morbidities and, concomitantly, complex hospital courses \parencite{gallant2017comorbidities}. We first filter the test set for patients admitted to the HIV clinic (10k to 339 admissions) \parencite{levy2020towards}. From this HIV-specific cohort, we remove outliers: the top and bottom ten percent by number of source notes, and do the same for the summary reference length. The admissions with the fewest notes tend to cover cases in which notes are missing and, as such, are difficult to annotate. Removing the longest examples (source notes and reference length) filters out a long tail of examples for which obtaining human annotations would be too time consuming. After filtering, we end up with 212 admissions. From this set, we bin the summaries by extractiveness (density) into deciles, similarly to \textcite{bhandari-etal-2020-evaluating}, and sample an equal number from each decile to ensure diversity in summaries for annotation. We sample from each bin and end up with 29 summaries for annotation (245 sentences), based on a total of 703 source notes.

\paragraph*{Generating Summaries for Annotation.} At a high-level, we fine-tune a Transformer Encoder-Decoder with sparse attention (Longformer Encoder-Decoder (LED) \parencite{beltagy2020longformer}). The LED handles inputs up to 16,384 tokens. To fit all inputs (the average input length from Table \ref{tab:data-stats} is $18.4k$), we train a simple bi-LSTM model to rank each sections and, during inference, retain the top 100 sections. \textbf{Coarse Filtering.} The average length of the inputs ($\sim$ 30,000 tokens) exceeds the maximum sequence length even for transformer models with sparse attention mechanisms designed for long input sequences \parencite{dai2019transformer, zaheer2020big, guo2021longt5}. Similarly to \textcite{liu-lapata-2019-hierarchical}, we learn a light-weight bi-LSTM model which learns the relevance of each section, to predict the average ROUGE-1 and ROUGE-2 recall of each section vis-a-vis the reference.  We choose a bi-LSTM over BERT models given the simplicity of the task iand the large number of sections (100s of millions). In particular, we pass a bi-LSTM over the tokens in each section and compute a soft cross-entropy loss between the gold-standard ROUGE-2 recall and the predicted logit (sigmoid(score)). Then, we score each section and filter for the top-K sections. The top 100 sections are provided by an oracle during training and by the model for evaluation. \textbf{Fine-Tuning.} We fine-tune the Longformer Encoder-Decoder (LED) for 10 epochs with a batch size of 1 and gradient accumulation steps of 16.  We set the maximum learning rate to $3e-5$ (tuned in range the range of $1e-6$ to $1e-3$) with a warmup of $200$ steps with linear decay. The maximum input size was set to 16,384 and outputs were produced with minimum length of 64, maximum length of 1,024, trigam-blocking, and a beam size of 4 with length penalty 4.0. Training took 8 days on 1 NVIDIA RTX 3090 GPU (24GB).

\begin{figure*}[t]
\centering
\includegraphics[width=\linewidth]{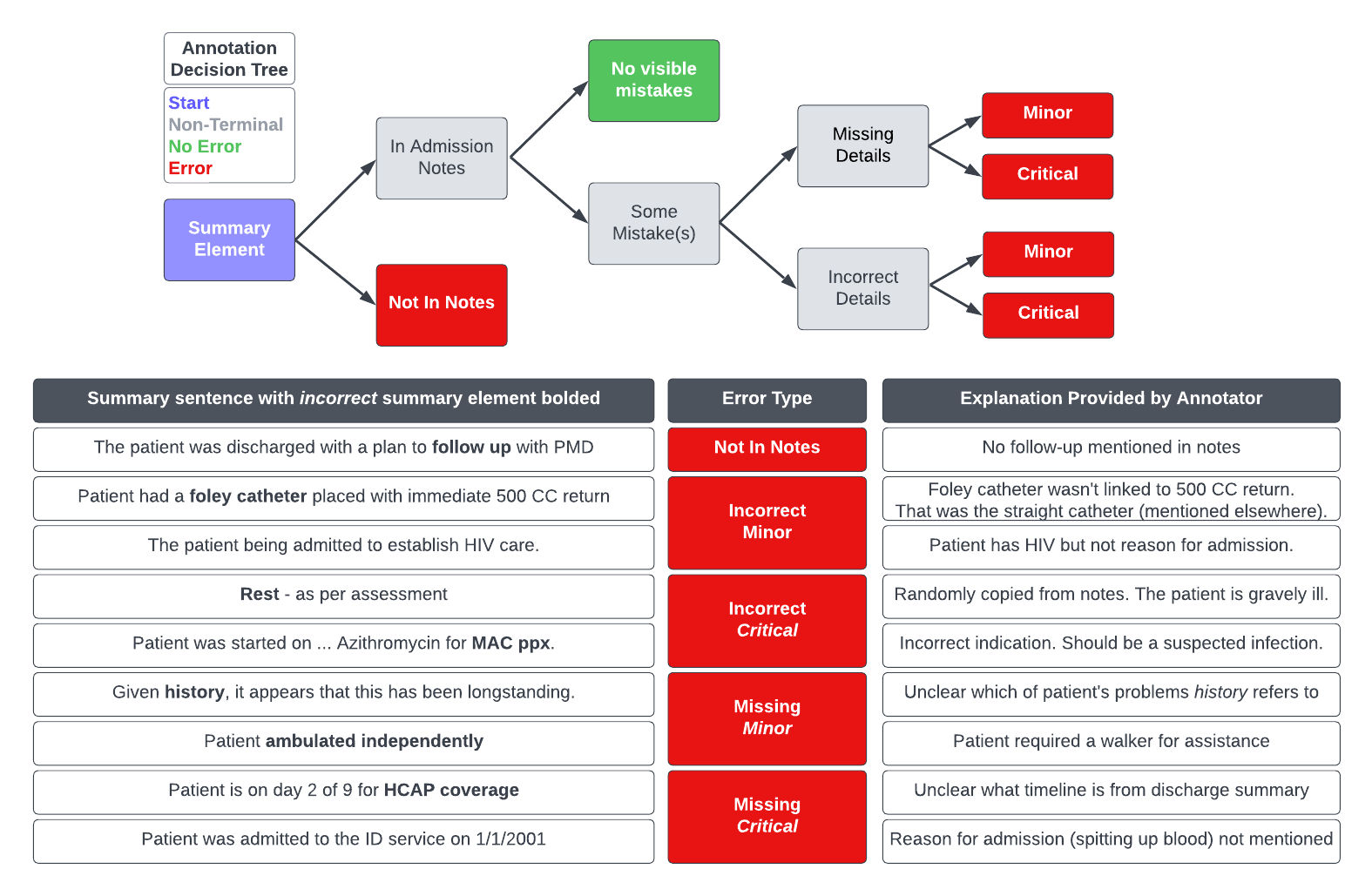}
\caption{Annotation Decision Tree with real, but modified, examples for each error type. Examples have been modified to removing any protected health information (PHI) and change all numbers (e.g., dates). } 
\label{fig:annotation-tree}
\end{figure*}

\subsection{Collecting Annotations} \label{sec:annotation-protocol}

At a high-level, the annotation task consisted of assigning an error category (or No Error) to each Summary Element (defined below) in a system output, based solely on clinical knowledge and all patient's clinical notes from the hospital admission.

\paragraph*{Summary Elements.} As in other faithfulness work \parencite{goyal-durrett-2021-annotating}, we decided to collect fine-grained annotations and experimented with different granularities while piloting the study. We found that entities (used in \textcite{cao-etal-2022-hallucinated}) were too granular, noisy, and incomplete on clinical notes. Syntactic parses were unreliable on our text as well. On the other hand, sentence-level annotation \parencite{wang-etal-2020-asking, durmus-etal-2020-feqa, pagnoni-etal-2021-understanding} was insufficiently fine-grained given the length and information density of many sentences. As such, the authors of the paper manually extracted \texttt{Summary Elements} (SE), which consist of standard medical concepts and actions, as well as compound concepts. Standard medical concepts included Disorders, Medications, Procedures, and Treatments, while actions encapsulate phrases such as ``discharged to home'' and ``plans to follow up''. When sensible, we merged compound entities into a single \textbf{SE}: ``alkanization of urine''.

\paragraph*{Error Categories.} For each SE, annotators were asked to identify and categorize errors. As represented as a decision tree in Figure \ref{fig:annotation-tree}, annotators were first asked to confirm whether or not the summary element is ``hallucinated'': \texttt{Not in Notes}. If the SE can be found in the notes, they either deem it correct: \texttt{No visible mistakes} or denote an inconsistency in its usage. For these intrinsic-focused errors, we delineate between \texttt{Incorrect Details} and \texttt{Missing Details}. A SE has \texttt{Incorrect Details} if it can be found in the source notes yet contains information that does not reflect what is written in the notes. This category encapsulates numerical errors (dosages, dates), mis-representations of symptoms (``afebrile'' is incorrect if patient had a fever), fusion errors (an incorrect indication for a drug), among others. An SE has a \texttt{Missing Details} error if the summary omits important information about the SE, which could lead to misleading conclusions being formed about the patient’s true hospital course. \texttt{Missing Details} is grounded on a specific SE and thus less open-ended than previously defined ``omission'' errors \parencite{huang-etal-2020-achieved, moramarco-etal-2022-human}.

\paragraph*{Severity of Errors.} For \texttt{Incorrect} and \texttt{Missing}, as in \textcite{moramarco-etal-2022-human}, we ask annotators to distinguish between Minor and Critical errors. We provide annotators with examples of both kinds of errors and define \texttt{Critical} as a mistake which could negatively impact the patient's present and future treatment. Minor is an exclusionary category defined as ``Not Critical''.

\paragraph*{Annotators.} We recruited $6$ clinical practitioners, with IRB-approved access to the patient data, to annotate the summaries in \texttt{Eval - HIV}. Each annotator was compensated at a rate of \$30 / hour. $4/6$ of the annotators self-identify as female, with the other two as male. $4/6$ self-identify as ``White'', and $1$ each as ``Black or African'' and ``Other''. $2$ annotators are attending physicians, $3$ are in medical residency, and $1$ is a fellow. They have a combined 25 years of medical practice. Each expert annotated summaries for a minimum of one hour at the same computer and location with the guidance of the authors of the paper, who were available for technical guidance and clarification questions regarding the annotation protocol. Collectively, the task was carried out over $\sim 10$ hours across 4 days.

\paragraph*{Description of Interface.} We develop a custom annotation interface within Prodigy \textcite{prodigy}. The interface presented each annotator with one summary at a time. For viewing ease, summaries were split such that one sentence was shown per line. Summary Elements (SE) were highlighted and annotation of non-SE spans prohibited.  For each SE, annotators would select the appropriate erorr category (or No Error) and then either double click or highlight the SE span. On a separate browser page, we displayed the source notes for the patient visit, which were hosted locally on a custom, light-weight app. The left-hand side of the full-text notes display showed section headers and free text for each note. Notes were sorted by date and annotators could search for a note by its title on a drop-down menu. Section headers were indexed and searchable to allow for efficient navigation of long notes. On the right hand side of the webpage, we enabled free-text search across notes. Each note was pre-indexed such that all mentions of matching search terms across notes could be quickly surfaced. We extracted all concepts with CLAMP NLP, highlighted them in the interface, and allowed for annotators to trigger a concept-based search query by double-clicking on the concept span in the note.

\section{Error Analysis} \label{sec:faithfulness-error-analysis}

\begin{table}[h]
\centering
\small
\setlength{\tabcolsep}{2pt}
\begin{tabular}{l|cc|c}
& \texttt{\textbf{\makecell{Per \\ Summary}}} & \texttt{\textbf{\makecell{Per \\ Sent}}} & \textbf{\texttt{\makecell{\% of \\ All SE}}} \\\hline
All Summary Elements (SE) & 27.10 &3.21 & - \\ \hline
Incorrect SE & 2.86 & 0.34 &11\% \\
Missing SE & 0.93 & 0.11 &3\% \\
Not In Notes SE & 1.03 & 0.12 & 4\% \\ \hline \hline
\emph{Any} Mistake SE & 4.83 &0.57 & 18\% \\
\end{tabular}
\caption{Statistics on Clinician-Annotated Summary Elements (SE), broken down across error categories. }
\label{tab:error-distribution}
\end{table}

\begin{figure}[h]
\centering
\includegraphics[width=0.5 \linewidth]{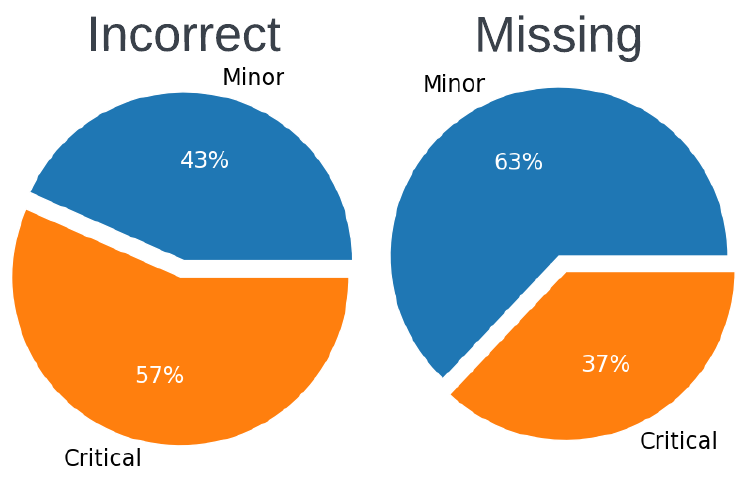}
\caption{Breakdown of errors deemed by clinicians as Minor versus Critical (potentially impacting patient care) for two error types: \texttt{Incorrect} and \texttt{Missing}. } 
\label{fig:critical-pie}
\end{figure}

\begin{figure}[h]
\centering
\includegraphics[width=0.75 \linewidth]{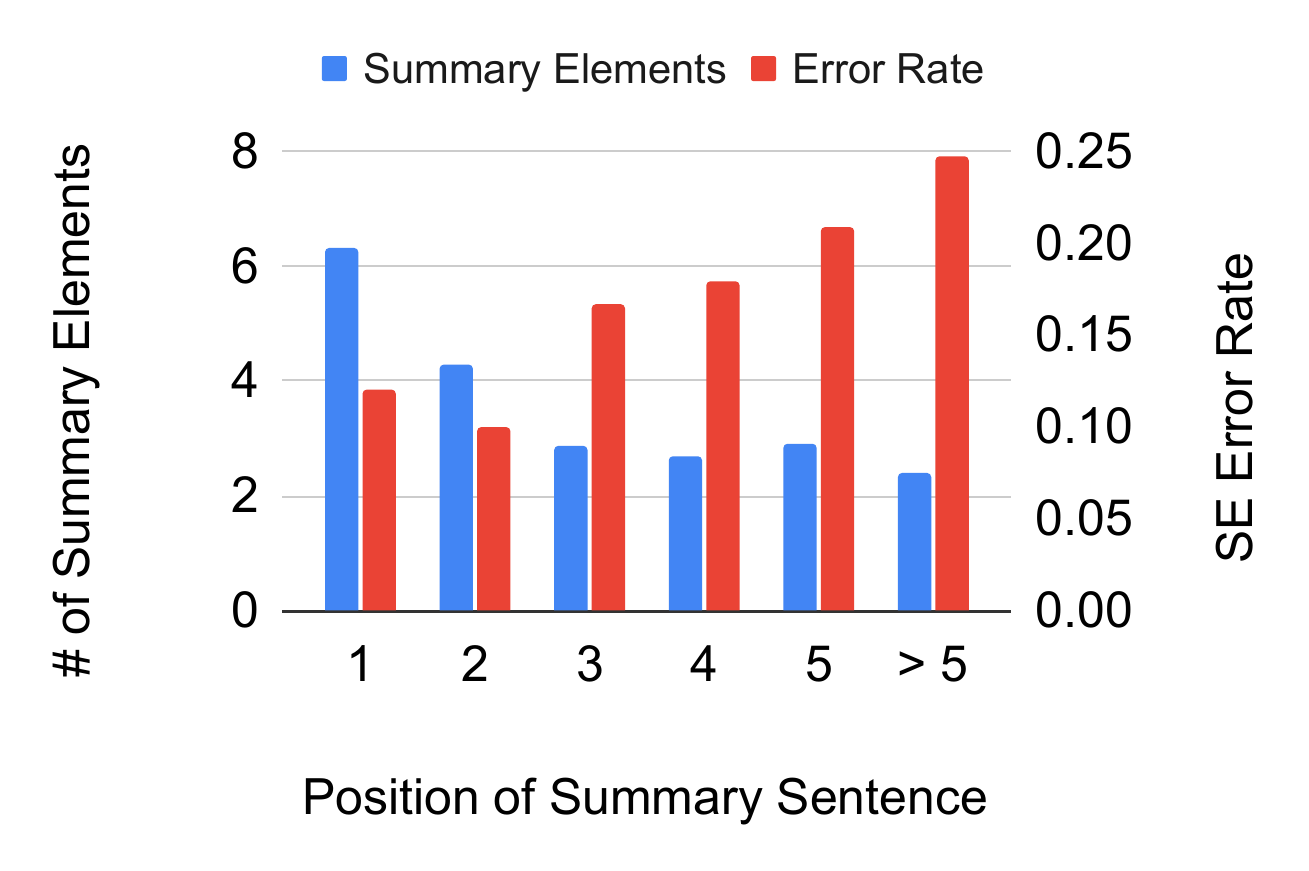}
\caption{Increasing error rate as summary length increases. There are more SEs at the beginning of summaries, which tend to involve longer sentences and many cover lists of diagnoses for the patients (HPI). } 
\label{fig:length}
\end{figure}

\paragraph*{Distribution of Errors.} Table \ref{tab:error-distribution} shows the number of SE per summary and per sentence, as well as the breakdown of SE into each error category. $18\%$ of SEs are marked as having \emph{Any mistake}, of which the predominant category is \texttt{Incorrect} ($11\%$ versus $3\%$ and $4\%$ for \texttt{Missing} and \texttt{Not in Notes}). In Table \ref{tab:error-distribution}, Minor and Critical are lumped together and contribute equally to the counts.

\paragraph*{Qualitative Analysis.} As shown in Figure \ref{fig:annotation-tree}, incorrect errors often result from improper fusion of concepts: (``foley catheter'' with ``500 CC return'', ``Azithromycin'' with ``MAC ppx'', and ``admitted'' with ``HIV care''). Incorrect errors can also be perfectly extractive. ``Rest - as per assessment'' is copied verbatim from a previous note, yet is incorrect because, at the time of discharge, the patient is actually gravely ill, which contradicts the recommendation. \texttt{Missing Errors} are also quite extractive (see analysis in \S \ref{sec:meta-by-type}) and tend to occur from the reverse problem: insufficient fusion. The model fails to fuse related concepts when they appear in different contexts. Specifically, the model fails to make the following links: use of a ``walker'' is relevant to his ``ambulat[ion]'', that the ``HCAP coverage'' duration should be related to the note timestamp, and that ``admitted to ID service'' should be linked to the reason for admission---``spitting up blood''.

\paragraph*{Severity of Errors.} Figure \ref{fig:critical-pie} breaks down error severity for \texttt{Incorrect} and \texttt{Missing}. The majority of \texttt{Incorrect} errors were marked as \textbf{Critical} ($57\%$), whereas a minority for \texttt{Missing} ($37\%$). As implicated by Figure \ref{fig:annotation-tree}, the difference between Critical and Minor errors is very subtle. Typically, the justifications for each, as provided by the annotators, were highly specific to the patient in question. This is interesting as it represents a non-standard definition of faithfulness, one which is more tightly connected to salience, and is grounded on a more holistic view of the patient journey.


\paragraph*{Impact of Position in Summary.} Similarly to degeneration in unconditional generation tasks \parencite{holtzman2019curious}. we can measure whether or not quality (as measured by faithfulness) declines at different summary positions. Figure \ref{fig:length} plots the percentage of SE marked with any error by the sentence position in the summary. A clear trend emerges of an increasing error rate as summaries grow longer. This may point to a task-agnostic factor: scaling limitations from full self-attention within the decoder, or task-specific factors: a shift in topics. Figure \ref{fig:length} shows the overall number of SEs decreasing by sentence position. From qualitative analysis, we, in fact, observe a topic shift: from dense history of present illness history recounting (diagnosis-heavy) to concise descriptions of procedures and, finally, any post-discharge instructions. 

\subsection{Evaluation Metrics} \label{sub:eval-metrics}


\paragraph*{Task-Specific Concerns.}

Broadly speaking, we identify three high-level challenges for evaluating long-form clinical summaries, which are distinct from those faced when evaluating single-document new summaries: \textbf{(1) Domain Adaptation}, \textbf{(1) Long Outputs}, \textbf{(3) Long Inputs}. 

\paragraph*{Domain Adaptation.} The first challenge relates to adapting metrics, typically trained and used on general domain data, to clinical text. We cannot adapt all metrics, especially metrics \parencite{sellam-etal-2020-bleurt, ribeiro-etal-2022-factgraph} which directly learn from news summary annotation benchmarks \parencite{wang-etal-2020-asking, pagnoni-etal-2021-understanding, fabbri-etal-2021-improving, laban-etal-2022-summac}. Domain-specific pre-training can improve performance of downstream models on many tasks \parencite{gururangan-etal-2020-dont}, including clinical \parencite{alsentzer-etal-2019-publicly}, yet the impact of in-domain exposure is less studied when meta-evaluating faithfulness metrics. As such, we implement three versions of each metric with increasing levels of domain adaptation: \texttt{Off-The-Shelf} (fully out-of-domain), \texttt{Tuned In-Domain} (pre-trained out-of-domain, tuned-in-domain), and \texttt{Double In-Domain} (pre-trained and tuned in-domain). For in-domain pre-training, we rely on existing models pre-trained on clinical or biomedical corpora, specific to each dataset.  For in-domain metric tuning, we use the \texttt{Train - HIV} data from Table \ref{tab:data-stats}. Training details are provided as part of each metric description in \S \ref{sec:metrics}.

\paragraph*{Output Lengths.} Given previous work \parencite{adams-etal-2021-whats} detailing the lack of inter-sentence discourse markers in clinical narratives, we evaluate each sentence independently. Performing meta-evaluation of metrics at the sentence-level also increases the level of support ($29$ vs $245$) when computing instance-level correlations. This choice also enables us to explore the impact of sentence-level partitioning of summaries on single metric, e.g., summary level, performance.


\paragraph*{Input Lengths.} Our inputs contain $\sim$30,000 tokens. Conditioning evaluation on the entire source is computationally expensive and often undesirable (e.g., entailment models are trained on short premises). Modern faithfulness metrics tend to struggle with long inputs \parencite{honovich-etal-2022-true-evaluating}, likely due to the fact that only a handful of sentences from the source text are relevant to a given summary sentence \parencite{lebanoff-etal-2019-analyzing}. Yet, computing source-summary alignments \parencite{ernst-etal-2021-summary} is particularly challenging for clinical text because 1) massive redundancy from copy-and-paste \parencite{hirschtick2006copy}; 2) lexical variation in discussing semantically identical concepts (abbreviations, acronyms, etc.) \parencite{adams2020zero}; 3) the need for complete context when assessing missing or misleading information. To explain 3), if a summary includes an outdated lab measurement, simply returning that single lab value as the alignment would provide a false sense of clinical correctness. The full chronology is needed.

Given this complexity, we separately evaluate the impact of alignment granularity (2-3 sentences to the whole input) on metric tuning and inference.

\begin{table}[h]
\centering
\small
\begin{tabular}{l|c}
\textbf{Alignment Method} & \textbf{Number of Source Sents} \\ \hline
\texttt{ROUGE-Gain} & 1.1 \\
\texttt{BS-Gain} & 1.8 \\
\texttt{ROUGE-TopK} & 5.0 \\
\texttt{BERT-TopK} & 5.0 \\
\texttt{Top Section} & 13.2 \\
\texttt{Entity Chain} & 15.3  \\
\texttt{Full} & 921.2* \\
\end{tabular}
\caption{The average number of source sentences aligned to each summary sentence for different alignment methods. $K$ is $5$. *\texttt{Full} differs for each metric based on token limits (pre-truncated lengths shown). } \label{tab:alignment-stats}
\end{table}

Each method aligns a summary sentence to a subset of sentences from the source. Duplicate source sentences are removed. Table \ref{tab:alignment-stats} shows the average number of aligned sentences by method.


\paragraph*{Alignments - Granular.} \texttt{ROUGE-TopK} takes the $k=5$ highest ROUGE-aligned sentences (average of R1, R2, RL F-1), while \texttt{ROUGE-Gain} follows \textcite{lebanoff-etal-2019-scoring} and maximizes the relative ROUGE gain of adding each additional sentence to the current set of aligned sentences. To account for lexical variation and noise, we also build alignments with BERTScore (BS) from in-domain weights (see description of BERTScore model used in \S \ref{sec:metrics}). \texttt{BS-TopK} selects the $k$ source sentences with the highest F-1 BS vis-a-vis the summary sentence. \texttt{BS-Gain} follows the approach in \parencite{adams-etal-2022-learning} in which a coverage weight is assigned to each token in the summary sentence and updated based on the maximal alignment so far.

\paragraph*{Alignments - Entity-Chain.} Given a summary sentence, we define an alignment method based on \texttt{Entity-Chains} \parencite{barzilay1997using, narayan-etal-2021-planning} as the set of sentences in the source with at least one medical concept (a CUI from the Unified Medical Language System (UMLS) aligned to any of the CUIs in the summary sentence. We extract and link entities to the Unified Medical Language System (UMLS \parencite{bodenreider2004unified}) with CLAMP \parencite{soysal2018clamp} and embed each entity mention with SapBERT \parencite{liu-etal-2021-self} and first merge all entity mentions which share the same CUI from the UMLS. Exact match of two entities by CUI is far too strict given the size of the UMLS vocabulary as well as extraction noise from abbreviations, acronyms, etc. \parencite{adams2020zero}. Then, we treat two distinct CUIs as synonyms based on a random forest classifier. The authors of this paper manually labeled 1,000 pairs of entities sampled from 10 different admissions, from a held-out set. The labels were \texttt{Unrelated}, \texttt{Related}, \texttt{Synonyms}. \textit{Ceftriaxone} is \texttt{Related} to \textit{antibiotics} since it is in the class of antibiotic, while it is a synonym of \textit{Rocephin}, its brand name. We split the 1,000 manually labeled examples into an 80-20 train-test split and compute features for all pairs of unique CUIs. They include similarity scores (cosine similarity) between CUIs, where CUI embeddings are provided by a pre-trained section-level CUI2Vec model on our corpus, as well as maximum pairwise alignments between mentions from different CUI sets: cosine similarity between SapBERT mention embeddings and lexical similarity (IDF overlap and string Levenshtein distance), and finally, binary indicators for TUI and semantic group status from the UMLS. Alignment is based on manually annotating pairs of mentions and learning a light-weight classifier on features which include mention similarity (using contextualized from SapBERT \parencite{liu-etal-2021-self}, TF-IDF overlap, and Levenshtein distance), CUI similarity (using a custom CUI2Vec model trained on MIMIC-III on our CUI vocabulary), and other UMLS-based features (TUI and semantic group).

\paragraph*{Alignments - Section-Level.} To avoid fragmented alignments pulled from different notes, we also consider the Top-1 most aligned section as its own alignment. In particular, we select the section with the highest average ROUGE-\{1, 2, L\} overlap vis-a-vis each sentence in the summary.

\paragraph*{Alignments - Full Input.} The conventional approach is to pass the whole source as input. Most of our inputs surpass both short and long transformer token limits. As needed for each metric, then, for \texttt{Full Input} alignments for each summary sentence, we select the source sentences with the highest ROUGE-\{1, 2\} overlap vis-a-vis summary sentence until a target token limit is reached.

\subsection{Metrics} \label{sec:metrics}

We describe each metric at a high-level and then detail domain adaptation.


\paragraph*{BERTScore.} \textbf{High-Level.} BERTScore \parencite{zhang2019bertscore} computes a greedy soft-alignment, based on BERT hidden state similarities, between a reference and a hypothesis text. As in \textcite{pagnoni-etal-2021-understanding}, we compute a \emph{reference-free} BERTScore: in our case, the hypothesis is a summary sentence and the reference its aligned source sentences. We only report BERTScore-Precision because it has the highest correlation with human judgment on our data. \textbf{Domain-Adaptation.} For \texttt{Off-The-Shelf}, we use \texttt{RoBERTA-Large}. There is no task-specific training for BERTScore so we report a single \texttt{In-Domain} variant. Specifically, we use a RoBERTA-Large model pre-trained from scratch with a custom BPE tokenizer on biomedical (PubMed and PubMed Central (PMC)), as well as clinical text (MIMIC-III) \parencite{lewis-etal-2020-pretrained}\footnote{The model weights (RoBERTa-large-PM-M3-Voc-large) can be downloaded from \href{https://github.com/facebookresearch/bio-lm}{GitHub} and used with HuggingFace.} For all variants, given that alignments can exceed the RoBERTA context window of 512, we separately encode sentences from the same section and concatenate them (similarly to the paragraph chunking method from \textcite{liu-lapata-2019-hierarchical}).

\paragraph*{BARTScore.} \textbf{High-Level.} BARTScore \parencite{yuan2021bartscore} computes the length-normalized log likelihood of a summary conditioned on the input. We measure BARTScore for each sentence based on its aligned source inputs. \textbf{Domain Adaptation.} For \texttt{Off-The-Shelf}, we use a BART-Large model fine-tuned on CNN/DailyMail news summaries\footnote{\texttt{facebook/bart-large-cnn} from HuggingFace.}. For \texttt{Tuned In-Domain} and \texttt{Double In-Domain}, we fine-tune BART-based models on \texttt{Train - HIV} corpus. The targets are single summary sentences and the inputs are their aligned source sentences. We fine-tune a separate model for each alignment method from \S \ref{sub:eval-metrics}. For \texttt{Double In-Domain}, we initialize fine-tuning on \texttt{Train - HIV} with the BART-based ReDRESS model from \textcite{adams-etal-2022-learning} \footnote{ReDRESS is pre-trained on a novel entity-based de-noising objective on unlabeled clinical text (MIMIC-III discharge summaries). The model weights are accessible on HuggingFace as ``griffin/redress-clinical-hallucination-generator''.}. For \texttt{Tuned In-Domain}, we initialize fine-tuning from \texttt{BART-Base} (to match ReDRESS). Using the Trainer from the Transformers library \parencite{wolf-etal-2020-transformers}, we fine-tune each model in batches of $16$ for $10,000$ steps with a learning rate of $3e-5$ ($200$ warmup steps followed by linear decay). We use a label smoothing factor of $0.1$.

\paragraph*{CTC.} \textbf{High-Level.} Compression, Transduction, Creation (CTC) \parencite{ctc} defines a unified series of weakly supervised methods to evaluate system outputs on several NLG tasks. For summary faithfulness, the \texttt{CTC Score} represents the average number of tokens predicted as ``fake'' given the source. To train the CTC model, spans from reference summaries are masked-and-filled with a separate language model: the generator. \textbf{Domain Adaptation.} For \texttt{Off-The-Shelf}, we use \texttt{D-cnndm}, a RoBERTA-Large model fine-tuned for CTC consistency \textbf{d}iscrimination on the CNN/Dailymail dataset. For domain adapation, we corrupt summary sentences from \texttt{Train - HIV} and learn to discriminate based on source alignments. As in BARTScore, we fine-tune a separate discriminator for each alignment method from \S \ref{sec:metrics}. To generate fake tokens (the generator), we first train a mask-infiller (BART-base) on all discharge summaries in MIMIC-III. We use the same span mask procedure from CTC (based on a dependency parse) to align the training objective with its usage. For \texttt{Double In-Domain}, we initialize the CTC Discriminator from the same biomedical RoBERTA model used for the \texttt{In-Domain} BERTScore \parencite{lewis-etal-2020-pretrained}. For \texttt{Tuned In-Domain}, we initialize tuning from \texttt{RoBERTA-Large} (to match the initialization for \texttt{Off-The-Shelf}). We use the CTC codebase\footnote{\url{https://github.com/tanyuqian/ctc-gen-eval}} to train the discriminator.


\paragraph*{Entailment.} \textbf{High-Level.} Faithful summaries should be entailed by the source text. \textbf{Domain Adaptation}. For \texttt{Off-The-Shelf}, we use a state-of-the-art entailment consistency model: SummaC \parencite{laban-etal-2022-summac}. SummaC computes a faithfulness score for a summary sentence by computing separate entailment scores for each source-summary pair and then aggregating (either with a greedy argmax--as in BERTScore--in a zero-shot setting, or with a learned 1D convolution\footnote{\textcite{falke-etal-2019-ranking} demonstrated that off the shelf NLI \parencite{bowman-etal-2015-large} models, trained on sentence-to-sentence data, do not transfer well to summary faithfulness task (document-sentence(s))}. We use the latter: SummaC-Conv, which is tuned using news summary human annotations. For \texttt{In-Domain}, we do not have enough annotations on which to tune a SummaC-Conv model. Rather, we rely on the zero-shot setting, in which an off-the-shelf in-domain entailment model is used to score summary sentences. Specifically, we use the SciFIVE Model\footnote{The weights can be downloaded from the HuggingFace Transformers library via the following model card: \texttt{razent/SciFive-large-Pubmed\_PMC-MedNLI}.} with SOTA performance on the MedNLI dataset \parencite{romanov2018lessons}--clinician-annotated entailment corpus whose premises come from MIMIC-III. SciFive is provided the summary sentence and its aligned source text as input, and generates a label: \texttt{\{contradiction, neutral, entailment\}}. For meta-evaluation, we convert each class label to an integer in the set $\{-1, 0, 1\}$. \\

\begin{figure*}[t]
\centering
\includegraphics[width=\linewidth]{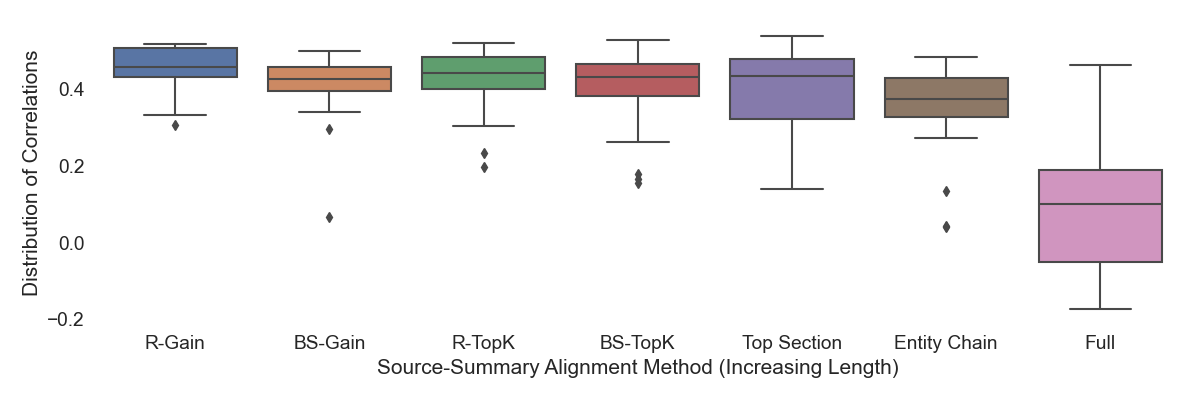}
\caption{The effect of alignment granularity on the distribution of instance-level Pearson correlations to human judgments across a wide range of metric variants (42). Correlations are more stable across metrics (higher average, higher minimum, and less overall variation) when the inputs (source-summary alignments) are shorter in length.} 
\label{fig:alignment-box}
\end{figure*}

\section{Meta-Evaluation of Metrics} \label{sec:meta}

Separately for each sentence of each summary in the human annotation set ($245$), we compute a human error rate \texttt{HErr}: defined as the fraction of summary elements (SE) in the sentence marked as either \texttt{Not In Notes}, \texttt{Incorrect}, or \texttt{Missing}. Unless explicitly stated, we do not distinguish between error type or severity (Minor, Critical) for the meta-evaluation. For the following analysis, we report the instance-level Pearson \parencite{cohen2009pearson} correlation coefficient between \texttt{HErr} and metric scores (two $245$ length vectors).

\begin{table}[h]
\centering
\small
\begin{tabular}{l|c|cccc}
\multirow{2}{*}{\textbf{\texttt{Method}}} & \multirow{2}{*}{\texttt{\textbf{\makecell{\# \\ Sent}}}} & \multicolumn{4}{c}{\textbf{\texttt{Correlations}}} \\
& & \texttt{Avg} & \texttt{Max} & \texttt{Min} & \texttt{Std} \\ \hline
\texttt{ROUGE-Gain} & 1.1 & .46 & .52 & 0.31 & .06 \\
\texttt{BS-Gain} & 1.8 & .42 & .50 & .07 & .07 \\
\texttt{ROUGE-TopK} & 5.0 & .43 & .52 & .20 & .07 \\
\texttt{BERT-TopK} & 5.0 & .41 & .53 & .16 & .09 \\
\texttt{Top Section} & 13.2 & .40 & .54 & .14 & .10 \\
\texttt{Entity Chain} & 15.3 & .36 & .48 & .04 & .10  \\
\texttt{Full} & 921.2* & .09 & .46 & -.17 & .16 \\
\end{tabular}
\caption{ Average of instance-level correlation of metric scores to human correlations at the summary sentence-level. Each row represents an alignment method, which provides inputs of varying lengths to each metric, and corresponds to a column in the box plot in Figure \ref{fig:alignment-box}. } \label{tab:alignment-correlations}
\end{table}

\subsection{Finding the Optimal Source Granularity} \label{sec:sa}

\paragraph*{Research Question.} How much of the source input (averaging $< 20k$ tokens across $> 40$ notes) is necessary to achieve high correlation with humans? 

\paragraph*{Experimental Setup.} To answer this question, we vary the number of source sentences provided to \emph{every} metric and variant from \S \ref{sec:metrics} and analyze its impact on performance (instance-level Pearson correlation with the Human Error Rate, \texttt{HErr}).

\paragraph*{Findings.} Figure \ref{fig:alignment-box} and Table \ref{tab:alignment-stats} reveal that, on average, metrics have higher correlations to human judgment when the inputs to the metric are shorter (with ROUGE-Gain being the shortest and having highest average Pearson Correlation of $.46$). The standard deviation of average instance-level correlations grows monotonically as alignments grow longer. Also, using the entire source is the most volatile (minimum of $-.17$) and the maximum correlation $.50$ is lower than the maximum correlation using a source-alignment (\texttt{Top Section}). These findings strongly suggest that scoring summaries based on the full source input is detrimental. 

\begin{table*}[ht]
\centering
\small
\setlength{\tabcolsep}{2pt}
\begin{tabular}{cl|cccccc|c}
& &\multicolumn{6}{c}{\textbf{\texttt{Usage Alignment}}} & \\
& & \textbf{\texttt{R-Gain}} & \textbf{\texttt{BS-Gain}} & \textbf{\texttt{R-TopK}} & \textbf{\texttt{BS-TopK}} & \textbf{\texttt{\makecell{Top \\ Section}}} & \textbf{\texttt{\makecell{Entity \\ Chain}}} & \textbf{\texttt{\makecell{Tune \\ Avg}}} \\ \hline
\multirow{6}{*}{\textbf{\texttt{\makecell{Tune \\ Alignment}}}} & \textbf{\texttt{R-Gain}} & \cellcolor[gray]{0.9} \textbf{.467} & .449 & .458 & \underline{.449} & .397 & .344 & .427 \\
& \textbf{\texttt{BS-Gain}} & \textbf{.458} & \cellcolor[gray]{0.9} .387 & .427 & .382 & .396 & .351 & .400 \\
& \textbf{\texttt{R-TopK}} & \textbf{.449} & .440 & \cellcolor[gray]{0.9} .442 & .446 & .408 & \underline{.387} & .428 \\
& \textbf{\texttt{BS-TopK}} & \textbf{.460} & .411 & .435 & \cellcolor[gray]{0.9} .407 & .416 & \underline{.387} & .419 \\
& \textbf{\texttt{Top Section}} & \underline{\textbf{.469}} & .440 & .463 & .446 & \cellcolor[gray]{0.9} \underline{.427} & .379 & .437 \\
& \textbf{\texttt{Entity Chain}} & .452 & \underline{.450} & \underline{\textbf{.469}} & .438 & .407 & \cellcolor[gray]{0.9} .379 & .432 \\ \hline
& \textbf{\texttt{Usage Avg}} & .459 & .429 & .449 & .428 & .408 & .371 & \\
\end{tabular}
\caption{ Each row represents the Source-Summary alignments computed for metric \emph{tuning}, whereas the columns denote the alignment method for inference (\emph{usage}). Each cell represents the instance-level metric correlation to the Human Error Rate, averaged across four metric variants (BARTScore and CTC, \texttt{Tuned In-Domain} and \texttt{Double Domain}). The row-wise max is \textbf{bolded} and column-wise is \underline{underlined}. The diagonal is shaded in gray. } \label{tab:alignment-matrix}
\end{table*}

\subsection{Optimal Alignments for Metric Tuning} \label{sec:tuning-alignments}

\paragraph*{Research Question.} \S \ref{sec:sa} reveals that shorter source alignments are preferable when \emph{using} metrics. Is the story the same when \emph{tuning} metrics? And should the alignment method used for metric tuning match the method used during inference?

\paragraph*{Experimental Setup.} To answer this question, we breakdown metric performance (correlation to \texttt{HErr}) by the alignment method used for metric \emph{tuning} and, separately, for \emph{usage}. We consider 4 metrics (\texttt{Tuned In-Domain} and \texttt{Double In-Domain} variants for BARTScore and CTC). Each training instance is a summary sentence from \texttt{Train - HIV} and its aligned source context.

\paragraph*{Findings.} Each cell in Table \ref{tab:alignment-matrix}\footnote{\texttt{Full} is not shown because it was not implemented for CTC due to token context restrictions for RoBERTA of 512.} represents an average of instance-level correlations to \texttt{HErr} across 4 metric variants (2 for BARTScore, 2 for CTC). Looking at the row-wise maximum values (\textbf{cells}), we notice that $5 / 6 $ involve using the shortest alignment (\texttt{R-Gain}) for metric \emph{usage}. This aligns with our analysis above in \S \ref{sec:sa}. Yet, the optimal alignment method for metric tuning is much less clear. If anything, we notice that $4 / 6$ of the column-wise maximum values (\underline{cells}) come from models tuned models from one of the two longest alignment methods (\texttt{Top Section} and \texttt{Entity Chain}). Additionally, on average, the diagonal values (shaded in gray) do not outperform the non-shaded regions. Taken together, at a high-level, this analysis suggests that additional context may be helpful when learning metrics (to make the task more difficult), yet, when using a metric, providing shorter, higher precision contexts are preferable.

\subsection{Effect of Summary Granularity} \label{sec:target-granularity}

\paragraph*{Research Question.} For our meta-analysis, we measure faithfulness at the summary sentence level. As such, we have been scoring summaries sentence-by-sentence (\texttt{Sentence-Level}). Yet, for some metrics with localized predictions, alternatively, we can process the entire summary and then post-hoc extract sentence-level scores (\texttt{Summary-Level}). Which method leads to higher metric correlations?

We separately consider BARTScore and BERTScore to answer this research question. 

\begin{figure}[h]
\centering
\includegraphics[width=\linewidth]{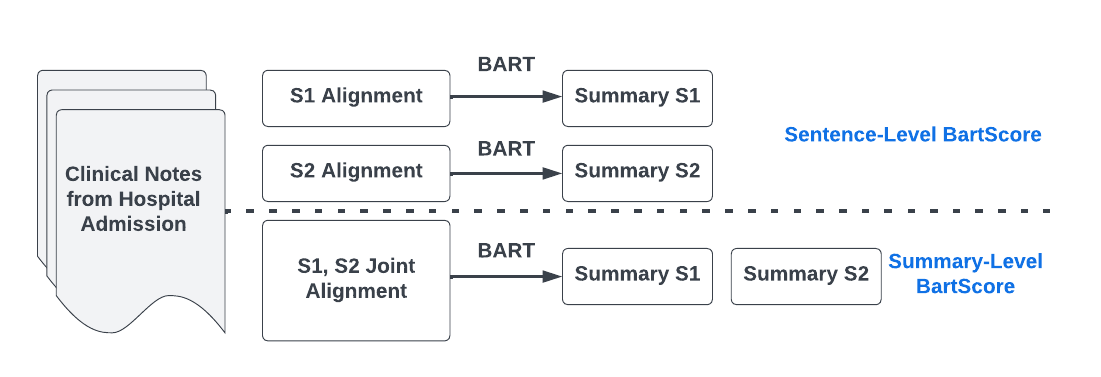}
\caption{ Sentence-Level BARTScore (BART-based) versus Summary-Level (LongFormer Encoder-Decoder (LED)). The LED scales BART to long inputs ($> 1024$ tokens). While Summary-Level generates a full summary, BARTScores are computed separately for each sentence by extracting logits from sentence boundaries.} 
\label{fig:bartscore-design}
\end{figure}

\paragraph*{BARTScore Experimental Setup.} \texttt{Sentence-Level} is the default approach  for all metrics, as detailed in \S \ref{sec:metrics}. \texttt{Summary-Level} BARTScore involves processing the full summary conditioned on aligned source sentences. For this setting, we simply treat the summary as a ``single sentence'' and align it to the source sentences. Yet, these source alignments often exceed the BART context window ($1,024$ tokens). To handle longer inputs, we replace BART with an LED model (which scales up BART with sparse attention). We fine-tune for 10,000 steps on \texttt{HIV - Train}) (as in \texttt{Sentence-Level}). For both sets of experiments, we consider two alignment methods: \texttt{ROUGE-Gain} and \texttt{ROUGE-TopK}. $K = 5$ for \texttt{Sentence-Level} and for aligning to full summaries, $K = 300$. During inference, we pass the same alignment granularity on which the model was fine-tuned. \texttt{Summary-Level} and \texttt{Sentence-Level} are contrasted in Figure \ref{fig:bartscore-design}.

\begin{table}[h]
\centering
\small
\begin{tabular}{lc|c}
\multirow{2}{*}{\textbf{\texttt{\makecell{Summary \\ Granularity}}}} & \multirow{2}{*}{\textbf{\texttt{\makecell{Source \\ Alignment}}}} & \multirow{2}{*}{\textbf{\texttt{\makecell{Pearson \\ Correlation}}}} \\
& & \\ \hline
\multirow{2}{*}{\textbf{\texttt{\makecell{Summary \\ Level}}}} & ROUGE-Gain & .438 \\
 & ROUGE-TopK & .424 \\ \hline
\multirow{2}{*}{\textbf{\texttt{\makecell{Sentence \\ Level}}}} & ROUGE-Gain & .516 \\
& ROUGE-TopK & .481 \\
\end{tabular}
\caption{ BARTScore correlation to human faithfulness labels by summary granularity (processing the full summary at once as opposed to sentence-by-sentence). } \label{tab:target-granularity-bartscore}
\end{table}

\paragraph*{BARTScore Findings.} Table \ref{tab:target-granularity-bartscore} reveals that \texttt{Sentence-Level} BARTScore (with separate alignments computed per sentence) is preferable to processing \texttt{Summary-Level} ($.516$ / $.481$ versus $.438 / .424$). This relates to the previous finding in \S \ref{sec:sa}. In both cases, tighter alignment between the inputs and outputs passed to a metric is preferable.

\begin{table}[h]
\centering
\small
\begin{tabular}{lc|c}
\multirow{2}{*}{\textbf{\texttt{\makecell{Summary \\ Granularity}}}} & \multirow{2}{*}{\textbf{\texttt{\makecell{Source \\ Alignment}}}} & \multirow{2}{*}{\textbf{\texttt{\makecell{Pearson \\ Correlation}}}} \\
& & \\ \hline
\textbf{\texttt{\makecell{Summary Level}}} & Full & .357 \\ \hline
\textbf{\texttt{\makecell{Sentence Level}}} & Full & .464 \\
\end{tabular}
\caption{ Correlation of BERTScore Precision to human labels by summary granularity (summary versus single-sentence). Both use the entire source (\texttt{Full} alignment). } \label{tab:target-granularity-bertscore}
\end{table}

\paragraph*{BERTScore Experimental Setup.} We evaluate \texttt{In-Domain} BERTScore variants \parencite{lewis-etal-2020-pretrained} which use the entire source (\texttt{Full} alignment method). Specifically, we compare our baseline BERTScore approach (\texttt{Sentence-Level}), which encodes each summary sentence independently, with a \texttt{Summary-Level} variant, which involves encoding the entire summary before computing a separate BERTScore for each sentence\footnote{Since we report BERTScore precision, we can compute the full similarity matrix before segmenting by sentence.}. The latter is typically how BERTScore is used.

\paragraph*{BERTScore Findings.} Table \ref{tab:target-granularity-bertscore} shows that encoding sentences independently (\texttt{Sentence-Level}) leads to higher correlation with human assessments ($.46$ versus $.36$). Given how choppy clinical notes are, including neighboring sentences can add substantial noise to contextual embeddings of summary sentences.

\subsection{Curious Case of In-Domain Training} \label{sec:in-domain-results}

\paragraph*{Research Question.} There is a wealth of evidence to demonstrate the beneficial impact of in-domain pre-training on clinical \parencite{alsentzer2019publicly, lehman2023we} and biomedical \parencite{gu2021domain} downstream tasks. Yet, to our knowledge, no previous work examines the benefits of in-domain pre-training on clinical evaluation metrics. Is domain adaptation: at the pre-training level, and at the task-specific fine-tuning level, necessary for developing clinical faithfulness metrics?

\paragraph*{Experimental Setup.} We breakdown instance-level metric correlations by the level of domain adaptation: \texttt{Off-The-Shelf}, \texttt{Tuned In-Domain}, and \texttt{Double In-Domain}. We consider \texttt{BARTScore}, \texttt{CTC}, and \texttt{Entailment}\footnote{We report correlations for best performing variants with respect to the alignment method used for tuning and inference.}.

\begin{table}[h]
\centering
\small
\begin{tabular}{cl|c}
\texttt{\textbf{\makecell{Domain \\ Adaptation}}} & \texttt{\textbf{Metric}} & \textbf{\texttt{\makecell{Pearson \\ Correlation}}} \\ \hline
\multirow{4}{*}{\textbf{\texttt{\makecell{Off The \\ Shelf}}}} & BARTScore & .539 \\
 & CTC & .507  \\
 & Entailment & .453 \\
 & \texttt{\textbf{Average}} & \textbf{.501} \\ \hline
 \multirow{4}{*}{\textbf{\texttt{\makecell{Tuned \\ In-Domain}}}} & BARTScore & .522 \\
 & CTC & .462  \\
 & Entailment* & .450 \\
 & \texttt{\textbf{Average}} & \textbf{.478} \\ \hline
\multirow{3}{*}{\textbf{\texttt{\makecell{Double \\ In-Domain}}}} & BARTScore & .516 \\
 & CTC & .439  \\
  & Entailment* & .450 \\
& \texttt{\textbf{Average}} & \textbf{.468 } \\ \hline
\end{tabular}
\caption{ The impact of domain adaptation of metrics on correlation to human assessments. For in-domain ``Entailment*'', we use a model pretrained on biomedical text and fine-tuned on the MedNLI dataset. It is not tuned on our clinical text, so it does not neatly fit into either \texttt{Tuned In-Domain} or \texttt{Double In-Domain}.} \label{tab:in-domain}
\end{table}

\paragraph*{Findings.} Table \ref{tab:in-domain} shows a curious trend: that increasing levels of metric domain adaptation is associated with lower correlation to faithfulness annotations at the metric-level and across systems (average declines $.501 \rightarrow .478 \rightarrow .468 $). Below, we link this outcome to summary extractiveness.

\begin{figure}[t]
\centering
\includegraphics[width=0.5 \linewidth]{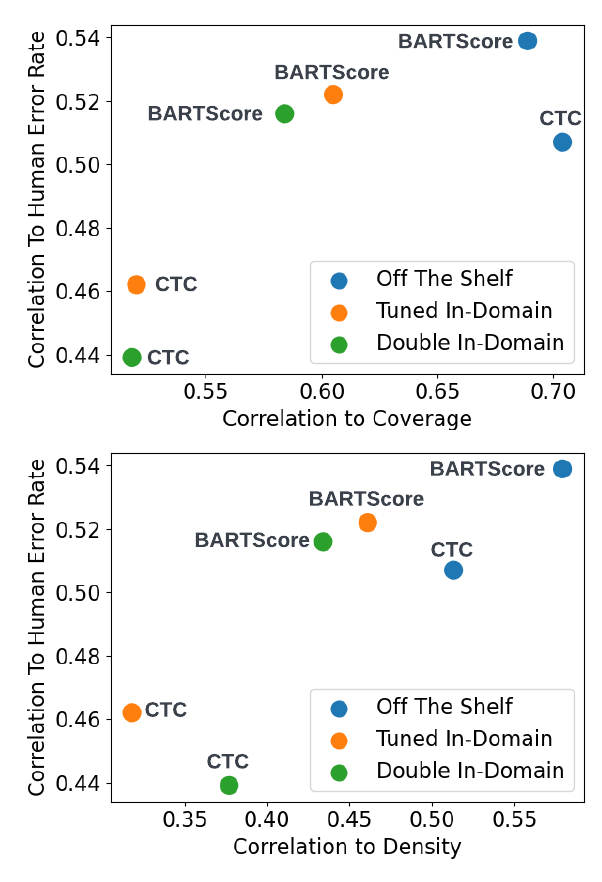}
\caption{Relationship between Correlation To Extractiveness and Correlation to Human Performance. Each dot represents the best performing (highest correlation) score across each source-summary alignment (see \S \ref{sec:sa}). }
\label{fig:spurious}
\end{figure}

\begin{figure*}[t]
\centering
\includegraphics[width=0.75 \linewidth]{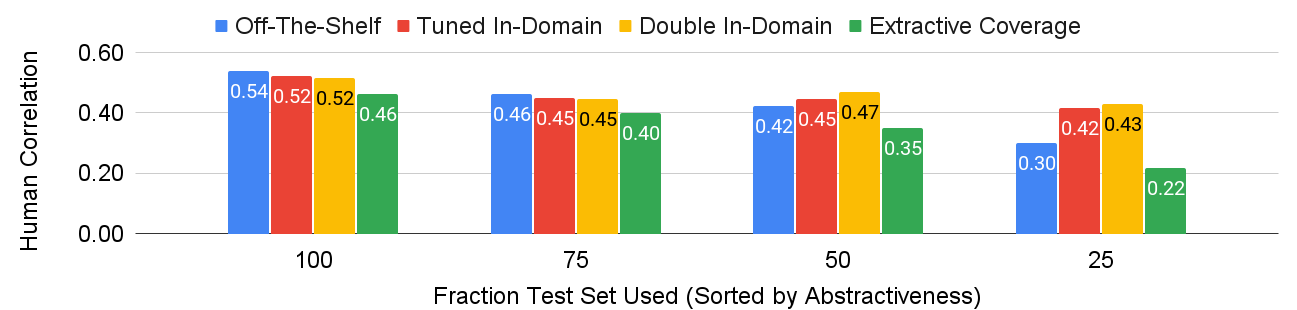}
\caption{Impact of summary extractiveness on metric correlation to human labels. BARTScore variants with different levels of in-domain training are shown, along with Extractiveness (Coverage). Coverage shows the steepest decline in correlation to human labels as average coverage declines, followed by the BARTScore variant most correlated to it (\texttt{Off-The-Shelf}). Metrics with in-domain training perform best on the more abstractive subsets.}
\label{fig:density}
\end{figure*}

\begin{figure*}[t]
\centering
\includegraphics[width=0.5 \linewidth]{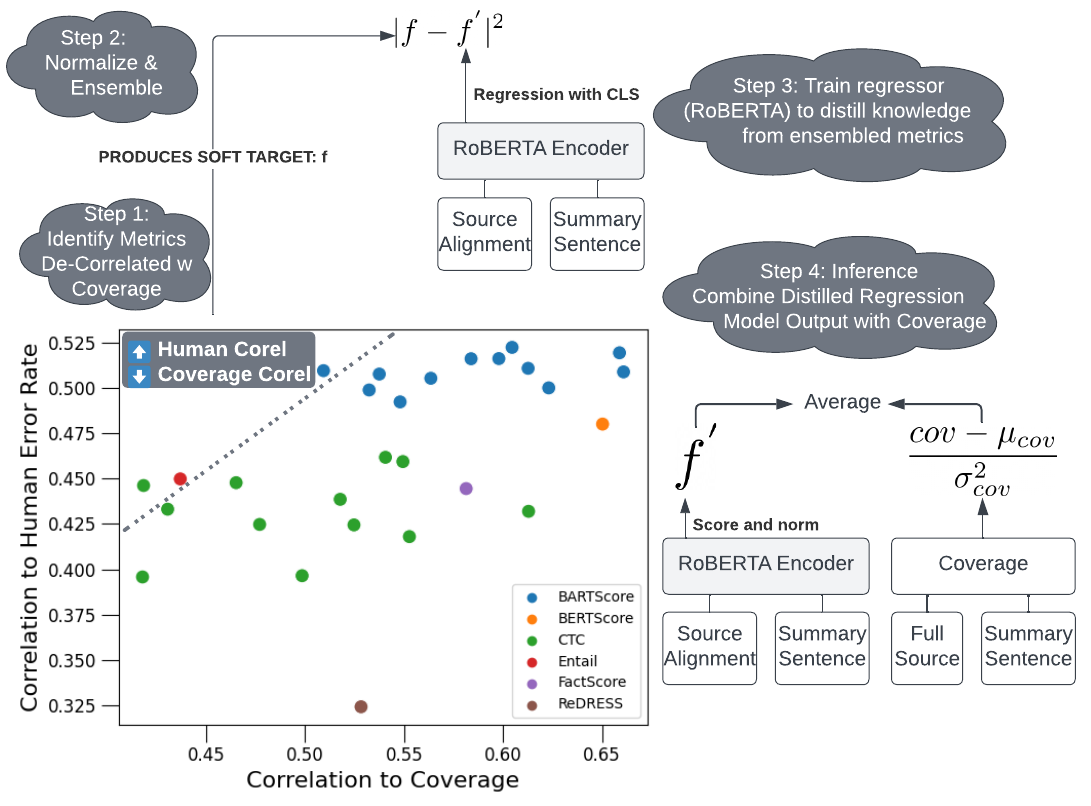}
\caption{\textbf{Step 1}: Identify Optimal Metrics for Knowledge Distillation: High Correlation to Human Labels and Low Correlation to Extractive Coverage. \textbf{Step 2}: Normalize and ensemble (average) to produce produce soft targets $f$ on the \texttt{Train - HIV} cohort. \textbf{Step 3}: Train a model (RoBERTA) as a regressor ($f^{'}$) against the ensembled soft targets $f$. \textbf{Step 4}: Create a combined metric: \texttt{\textbf{Distilled + Coverage}}, which combines the score from the RoBERTA model--distilled from metrics relatively less correlated with coverage--with a normalized coverage score. } 
\label{fig:distillation-metric}
\end{figure*}

\paragraph*{Spurious Correlates Hypothesis.} \textcite{durmus-etal-2022-spurious} find that reference-free metrics over rely on spurious correlates: variables which are highly correlated to human annotations on a biased test set, yet less correlated on a more realistic, diverse data distribution. Identifying such correlates is important because it suggests a metric brittleness which may not be captured by simple correlation analysis. As in their work, we focus on summary extractiveness \parencite{grusky-etal-2018-newsroom} as the potentially spurious correlate. In Figure \ref{fig:spurious}, we reveal a clear pattern between metric correlation to extractiveness and correlation to the human error rate. In particular, across Coverage (top) and Density (bottom), high correlations to extractiveness are positively related to the correlation with the human error rate. Additionally, we see that in-domain training de-correlates metrics to extractiveness (\texttt{Tuned-In-Domain} and \texttt{Double In-Domain}. To examine why this might be the case, we examine the extractiveness of reference versus system summaries and a clear bias emerges.

\begin{table}[h]
\centering
\small
\begin{tabular}{l|cc}
\textbf{\texttt{Summary}} & \textbf{\texttt{Coverage}} & \textbf{\texttt{Density}} \\ \hline
\textbf{Reference} & 0.88 & 12.04 \\
\textbf{Model-Generated} & 0.95 & 39.12 \\
\end{tabular}
\caption{ Model-Generated summaries are \emph{substantially} more extractive (Coverage, Density) than the references on which they are trained. This creates a train-test mismatch for metrics, which are fine-tuned on abstractive summaries and meta-evaluated on extractive ones. } \label{tab:extractiveness-mismatch}
\end{table}

Table \ref{tab:extractiveness-mismatch} shows that references are substantially more extractive in terms of both coverage (percentage of unigrams copied from the source) and density (average squared length of copied fragments) \parencite{grusky-etal-2018-newsroom}. In other words, clinicians write more abstractive summaries than the Longformer. To more closely approximate more abstractive, clinician-authored summaries, we examine changes in correlations to human judgments as we filter for more abstractive subsets of the test set.  We sort system summary sentences in the test set by coverage and filter for smaller and smaller subsets (making the average coverage lower). Figure \ref{fig:density} reveals that in-domain BARTScore metrics start to outperform when summaries are more abstractive ($.30 \rightarrow .42 \rightarrow .43$ for the smallest bucket, i.e., the top 25\% most abstractive sentences in the eval set).

\paragraph*{Domain-Adapted Metrics are Complementary to Coverage.} Recent work demonstrates the efficacy of ensembling de-correlated metrics \parencite{kasai-etal-2022-bidimensional, colombo2022glass}. In light of our previous analysis, we can normalize each metric variant from Figure \ref{fig:spurious} and ensemble it with a normalized score for extractiveness (e.g., coverage). To make this explicit, given raw metric score $f$ and raw coverage $cov$, we create a combined metric $g$

$$
g = \frac{1}{2} * \left( \frac{f - \mu_{f}}{\sigma_{f}^{2}} + \frac{cov - \mu_{cov}}{\sigma_{cov}^{2}} \right)
$$

\noindent where $\mu$ and $\sigma$ represent mean and standard deviations for $f$ and $cov$ across all summary sentences. We can then insert each metric in Figure \ref{fig:spurious} as $f$ into this equation and compare correlations to \texttt{HErr}.

\begin{table}[h]
\centering
\small
\begin{tabular}{cl|c}
\texttt{\textbf{\makecell{Domain \\ Adaptation}}} & \texttt{\textbf{Metric}} & \textbf{\texttt{\makecell{Pearson \\ Correlation}}} \\ \hline
 & Coverage (Cov) & .457 \\ \hline
\multirow{4}{*}{\textbf{\texttt{\makecell{Off The \\ Shelf}}}} & BARTScore + Cov & .542 \\
 & CTC + Cov & .522  \\
 & Entailment + Cov & .524 \\
 & \texttt{\textbf{Average}} & \textbf{.529} \\ \hline
 \multirow{4}{*}{\textbf{\texttt{\makecell{Tuned \\ In-Domain}}}} & BARTScore + Cov & .547 \\
 & CTC + Cov & .523  \\
 & Entailment + Cov & .535 \\
 & \texttt{\textbf{Average}} & \textbf{.535} \\ \hline
\multirow{4}{*}{\textbf{\texttt{\makecell{Double \\ In-Domain}}}} & BARTScore & .547 \\
 & CTC + Cov & .514 \\
  & Entailment + Cov & .535 \\
& \texttt{\textbf{Average}} & \textbf{ .532 } \\ \hline
\end{tabular}
\caption{ The impact of domain adaptation on metric correlation to human assessments when combining with an easy-to-compute extractiveness statistic (coverage).  } \label{tab:plus-coverage}
\end{table}

Table \ref{tab:plus-coverage} reveals that when combining metrics with coverage, In-Domain adaptation slightly helps. \texttt{Off-The-Shelf} averages across three metrics (+ Cov) are $.529$ versus $.535$ and $.532$ for \texttt{Tuned In-Domain} and \texttt{Double In-Domain}, respectively. Yet, the differences are still relatively minor. 

\paragraph*{Adapting to System Outputs with Knowledge Distillation.} Despite modest gains, domain adaptation does not help much, which may be due in part to differences in reference summaries versus system outputs. The above metrics are all trained solely on gold-standard references yet meta-evaluated on system outputs. To bridge this gap, we can learn a metric from system outputs. Yet, our annotation set is too small to use for this task. 

Instead, we leverage the fact that metrics, when ensembled, achieve relatively high correlation with human judgments, to create soft pseudo-targets on a larger set of system outputs (from \texttt{Train - HIV}). The goal, then, is to distill a single metric from the combined ``knowledge'' of multiple metrics\footnote{This kind of distillation is distinct yet related to conventional knowledge distillation \parencite{hinton2015distilling}, which typically involves using a large teacher to train a smaller student.}. To do this, we first generate summaries with our LED model on the \texttt{Train - HIV} subset and segment into sentences. To produce pseudo targets, as shown in Figure \ref{fig:distillation-metric}, we identify a subset of In-Domain metrics with desired attributes: high-correlation to human labels and relatively low correlation to coverage. We then score each summary sentence with each metric in the ensemble, normalize the scores on a per-metric basis, and then average them to produce 
 pseudo-target $f$ for each training instance. We then train a student model, which receives as input a concatenation of a model-generated summary sentence and its aligned source context, and outputs a scalar: $f'$ using the \texttt{[CLS]} hidden state. The student is trained with a standard MSE loss: $|f' - f|^{2}$ and is initialized from clinical/biomedical RoBERTA \parencite{lewis-etal-2020-pretrained}. We train in batches of $8$ for $10,000$ steps with a learning rate of $1e-5$ ($200$ warmup steps, followed by linear decay). For usage, we can \emph{optionally} combine the distilled score with the coverage score.

Via distillation of metrics which are relatively de-correlated with coverage, the goal is two-fold: to learn a single model that achieves a higher correlation on its own to other single-metric variants, and is complementary to coverage when combined.

\begin{table}[h]
\centering
\small
\begin{tabular}{l|c}
\textbf{\texttt{Metric}} & \textbf{\texttt{\makecell{Pearson \\ Correlation}}} \\ \hline
\textbf{Best Single Metric} & .539 \\
\textbf{Best Single Metric + Cov} & .547 \\
\textbf{Distilled Metric} & .564 \\ 
\textbf{Distilled + Cov} & \textbf{.573} \\
\end{tabular}
\caption{ Distilling a metric from the subset of metrics which are relatively less correlated to extractiveness (coverage) yields higher correlation with human labels than any other single metric. Additionally, combining the distilled metric with ($+$ Cov) obtains yields superior correlations to all single metric + coverage variants. } \label{tab:distillation}
\end{table}

Table \ref{tab:distillation} reveals that the Distilled metric outperforms the best baseline metric variant ($.564$ vs $.539$) and, because it is distilled from metrics which are relatively de-correlated with coverage, can be combined at inference with coverage to achieve an even higher correlation ($.573$). We ran a one-sided Williams Test \parencite{graham-baldwin-2014-testing} to estimate the significance of increase in correlation to human labels from \texttt{Best Single Metric + Cov} to \texttt{Distilled + Cov}. The p-value was $.081$. As such, we cannot state that the impact of distillation is statistically significant at $p < 0.05$. But, we note that the sample size is small ($245$).

\begin{table}[h]
\centering
\small
\begin{tabular}{l|cc} 
\multirow{2}{*}{\textbf{\texttt{Metric}}} & \multicolumn{2}{c}{\textbf{\texttt{Pearson Correlation}}} \\ & \textbf{\texttt{\makecell{Single}}} & \textbf{\texttt{\makecell{Avg In Ensemble}}} \\ \hline
\textbf{Coverage (Cov)} & .457 & .544 \\ \hline
\textbf{BARTScore} & .539 & .550 \\
\textbf{CTC} & .507 & .546 \\
\textbf{Entailment} & .453 & .539 \\
\textbf{BERTScore} & .482 & .535 \\
\textbf{Reviser} & .324 & .528 \\
\textbf{FactScore} & .444 & .536 \\ \hline
\textbf{Distilled} &  \bf .564 & .556 \\ \hline
\textbf{Best Ensemble} & N/A & \bf .583 \\
\end{tabular}
\caption{ Comparing the correlation to human annotations of single metrics, as well as the average correlation of ensembles of metrics that include a given metric. Lastly, we include the correlation of the best performing ensemble of metrics (Coverage, BARTScore, Distilled). } \label{tab:ensemble}
\end{table}

\paragraph*{Multi-Metric Ensembles.} Previously, we reported promising performance of our proposed Distilled metric--both on its own and combined with an extractiveness statistic. Yet, ideally, we would also want a metric that improves correlation when ensembled with other metrics. To this end, we enumerate all possible ensembles from a set which includes the coverage statistic and 7 metrics: our distilled model and our 6 implemented metrics (BARTScore, BERTScore, CTC, Entailment, FactScore, ReDRESS)\footnote{We report the best performing variant across in-domain pre-training / tuning and source-summary alignment methods.}. This provides us with $\sum_{n=1}^{N=8}{N \choose n} = 255$ unique ensembles, of which each metric takes part in $128$. Table \ref{tab:ensemble} shows correlation of metrics to \texttt{HErr} for metrics on their own (\texttt{Single}), as well as the average correlation to \texttt{HeRR} for metric ensembles which include a given metric (\texttt{In Ensemble}). Firstly, the metric rankings induced by \texttt{Single} and \texttt{In Ensemble} are mostly in agreement. Distilled outperforms all baselines on its own ($.564$) as well as its average correlation when used in an ensemble ($.556$). The last row of Table \ref{tab:ensemble} shows the correlation of the ensemble with the highest correlation to \texttt{HErr}: Coverage, BARTScore, and Distilled. To test significance of the \texttt{In Ensemble} results, we bootstrap $95\%$ confidence intervals (CI) for each metric's average \texttt{In Ensemble} correlation ($1000$ samples with replacement from vectors of size $128$) and find that the average correlation when \texttt{Distilled} is a part of an ensemble is significantly higher ($p < 0.05$) than the average correlation of any of the other 6 metrics (when part of an ensemble).

These results demonstrate that \texttt{Distilled} is useful on its own and is complementary to other metrics. More broadly speaking, the relative out-performance of ensembling (\texttt{In Ensemble} over \texttt{Single}) supports the notion that, when developing a metric, it is more useful to focus on its complementarity to existing metrics, rather than its performance in isolation~\parencite{colombo2022glass}.

\subsection{Correlation by Metric Type} \label{sec:meta-by-type}

Previously, we meta-evaluated metrics against the percentage of summary elements (SE) with \emph{any} error. In this section, we breakdown metric correlations separately by error category: \texttt{Incorrect}, \texttt{Missing}, and \texttt{Not in Notes}.  We analyze metrics at the sentence-level against the percentage of Summary Elements in the sentence marked with a certain error. To provide more granular insights, we breakdown error type correlations by Domain Adaptation, Source-Summary Alignment methods, and metric classes (BARTScore vs CTC, etc).

\begin{figure}[h]
\centering
\includegraphics[width=0.75 \linewidth]{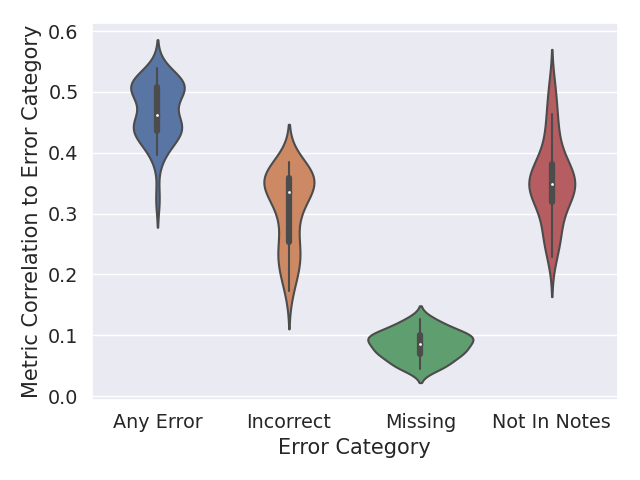}
\caption{Distribution of Metric Correlations to Human annotations by Category (includes Minor and Critical). } 
\label{fig:by-category}
\end{figure}

\begin{figure}[h]
\centering
\includegraphics[width=\linewidth]{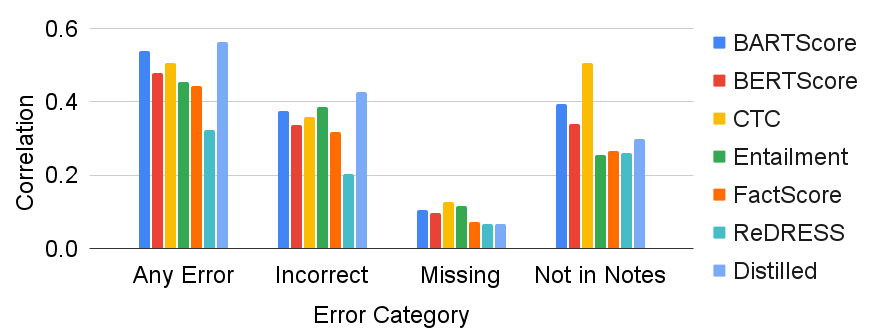}
\caption{Metric Correlations to Human Judgments by Error Category for each class of metrics from \S \ref{sec:metrics}. }
\label{fig:by-metric}
\end{figure}

Figure \ref{fig:by-category} shows that \texttt{Missing} is the hardest for metrics (the instance-level correlations of metrics to fraction Missing across metric variants), which makes sense given its negligible correlation with Coverage ($.021$). Not in Notes are the simplest as they tend to be most associated with lexical overlap: $.391$ Pearson correlation between coverage and fraction of SE's in a sentence identified as \texttt{Not in Notes}. \texttt{Incorrect} errors can be subtle and are less correlated to coverage than \texttt{Missing}: $.249$. More generally, the over-reliance of these metrics on the level of copy-and-paste obfuscates their actual ability to reason over clinical narratives.

\paragraph*{Metric-Wise.} Figure \ref{fig:by-metric} breaks down correlations to human judgments by metric and errory category. The primary take-away is that metric performance (here, correlation) does not exhibit monotonicity across error categories. Excluding Distilled, BARTScore is best at identifying \texttt{Any Error}, while Entailment outperforms on \texttt{Incorrect Errors}, and CTC performs best on \texttt{Not in Notes}. As discussed before, all metrics perform poorly on identifying missing content. CTC learns to identify extrinsic hallucinations so its strong performance on \texttt{Not in Notes} makes sense. Entailment metrics are trained on NLI datasets, which target the kinds of logic and inconsistency errors found in \texttt{Incorrect}. All metrics struggle with \texttt{Missing}. Taken together, these findings reveal that there is no one-size fits all solution to evaluation and we believe that metrics should be designed to fit the particular needs of a system and dataset \parencite{pagnoni-etal-2021-understanding}. Reporting a single score for meta-evaluation obscures important differences across categories, as well as ignores the potential complementarity of different metrics. Given the potential of ensembling, targeted metrics--which out-perform on one category--may be more valuable to real-world use cases than ``jack of all trades, master of none''-type metrics.

\section{Related Work} \label{sec:related-work}

\paragraph*{Faithfulness Metrics.} Metrics to assess faithfulness can be roughly distilled into the following categories: QA-based \parencite{wang-etal-2020-asking, fabbri-etal-2022-qafacteval, durmus-etal-2020-feqa}, entailment based metrics from NLI \parencite{falke-etal-2019-ranking} or synthetic data \parencite{kryscinski-etal-2020-evaluating, deng-etal-2021-compression, utama-etal-2022-falsesum}, fact-based, reference-free overlap \parencite{goodrich2019assessing}, and those which directly learn from human judgments \parencite{ribeiro-etal-2022-factgraph} (similar to BLEURT \parencite{sellam-etal-2020-bleurt} approach for machine translation). Most of these metrics have been developed on single document news summarization datasets, such as CNN / DailyMail \parencite{HermannKGEKSB15, see-etal-2017-get} and Xsum \parencite{narayan-etal-2018-dont-v2}. Faithfulness metrics proposed for clinical summary evaluation have typically come from the overlap category and focus on concept alignment between summaries and the source input \parencite{zhang-etal-2020-optimizing, tang-etal-2022-echogen}.

\paragraph*{Human Faithfulness Evaluation.} Assessing faithfulness is a challenging task to automate with metrics \parencite{bhandari-etal-2020-evaluating}, which underscores the importance of collecting high-quality human evaluation annotations \parencite{lux-etal-2020-truth, wang-etal-2020-asking, kryscinski-etal-2020-evaluating, maynez-etal-2020-faithfulness, huang-etal-2020-achieved, fabbri-etal-2021-summeval, pagnoni-etal-2021-understanding, goyal-durrett-2021-annotating, cao-wang-2021-cliff, cao-etal-2022-hallucinated}. Additionally, given the relatively small size of each separate evaluation, it can be useful for training and/or meta-evaluation to aggregate them into larger benchmark datasets \parencite{fabbri-etal-2021-summeval, laban-etal-2022-summac}.

Based on low inter-annotator agreements for summary-level faithfulness annotations \parencite{lebanoff-etal-2019-analyzing, factcc}, recent work has focused more on fine-grained annotations at the entity \parencite{cliff, cao-etal-2022-hallucinated}, sentence \parencite{pagnoni-etal-2021-understanding}, and span level \parencite{maynez-etal-2020-faithfulness}. These studies tend to have higher annotator agreement and allow for a better understanding of the typology of error distributions across datasets and systems. Sophisticated error taxonomies are generally formulated by examining system outputs (e.g., card-sorting exercises \parencite{lux-etal-2020-truth}) and tend to demarcate error types on two fronts: where the error is located (broken down by syntactic roles) and where it \emph{likely} comes from (intrinsic or extrinsic). \textcite{zhang2022extractive} challenges the notion that extractive summaries are consistent by analyzing inter-sentence discourse.

\paragraph*{Human Evaluation for Clinical Summarization.} \textcite{moen2014evaluation} evaluate extractively generated Discharge Summaries based on content criteria guidelines and benchmark ROUGE against these coverage-focused annotations. Much of the recent work on human evaluation of clinical summarization has focused on self-contained, single-document tasks: including radiology report summarization \parencite{macavaney2019ontology, zhang-etal-2020-optimizing} and echocardiogram conclusions \parencite{tang-etal-2022-echogen}. For these shorter tasks, summary-level assessments are collected, in the form of pairwise ranking \parencite{tang-etal-2022-echogen} or point-wise assessments \parencite{macavaney2019ontology} on a Likert Scale. \textcite{moramarco-etal-2021-towards} examine brief descriptions of SOAP notes for mock patient encounters (MTSamples\footnote{\url{https://mtsamples.com}}, and compare fact-based overlap between reference and system-generated summaries.

Most closely related to our work, \textcite{moramarco-etal-2022-human} perform a human evaluation on a more self-contained, conditional clinical note generation task: generating a SOAP note from consultation transcripts. They rely on a dataset of mock patient-doctor conversations and corresponding SOAP notes from \textcite{korfiatis2022primock57}. Annotators were asked to post-edit notes to correct errors, as well as manually highlight spans with incorrect or omitted information.  Automatic metrics were then benchmarked against post-editing time, as well as the number of incorrect and omitted spans. Our work differs as we define a typology of errors with more categories, consider more diverse faithfulness metrics, and, because our data includes much longer clinical narratives, explore the impact of using source-summary alignments and different summary granularities (sentence-level versus full).

\section{Conclusion and Limitations}

We collect fine-grained faithfulness annotations of Hospital Course summaries from clinicians and benchmark metrics against them. For each metric, we consider dimensions relevant to long-form clinical summarization: domain adaptation, input lengths, and output lengths. We find that metrics over-rely on the level of copy-and-paste in summaries. We can exploit this by computing a score which combines normalized extractiveness (coverage) with a new metric, which is distilled from a subset of the metrics most de-correlated with coverage. Moreover, metrics struggle with errors which require deep clinical knowledge (such as missingness, identification of mistakes from the source notes, etc.). 

Given the subtlety of the errors we see, and the strict faithfulness requirements necessary for safe deployment, we believe that scale will be necessary for the most effective development of clinical metrics. Recent advances in open-source models (Mistral, Zephyr, etc.), and compute-optimized inference (Flash Attention, model quantization), should enable us to fine-tune \emph{medium-to-large} LLMs ($\geq 70B$ parameters) for this task.

We have shown with small models (BERT, RoBERTA, LongFormer) that measuring faithfulness given pre-computed alignments based on word overlap is substantially more effective. Yet, for subtle error types, e.g., omission, identification of inconsistencies in the source notes, the entire patient history will be necessary context.  We believe that a hybrid approach: one which involves using the whole patient record as input, yet, requiring the LLM to produce the most relevant evidence from the source text when making its judgment, will be necessary. To capture diversity in error types, AlignScore \parencite{zha-etal-2023-alignscore} trains on a unified dataset of faithfulness-adjacent tasks: NLI, paraphrasing, etc.

In the pre-LLM, BART \parencite{lewis2019bart} and PEGASUS \parencite{zhang2020pegasus} era, summarization has been treated as a specialized task requiring large-scale supervision. As a side effect of this, research into evaluation has largely been split into reference-based and reference-free metrics. Reference-based metrics, such as ROUGE, are generally associated with the relevance and informativeness of a summary. Reference-free metrics, on the other hand, tend to predominantly be designed to identify faithfulness errors, including hallucinations. In the new era of LLMs, where models can summarize text without supervised adaptation to references, we argue the the lines between relevance and faithfulness are, and should be, more blurred, especially for our task. For instance, if a symptom is mentioned in a summary without mentioning that it was resolved, this should be treated as both a faithfulness error and a relevance issue. Relevant information was omitted that would lead a reader to draw incorrect, or unfaithful, conclusions about the current state of the patient. We believe that evaluations should no longer be restricted to one category or the other (relevance or faithfulness), but rather be considered jointly.

%% file: 6-llms.tex
\chapter{Generating Complete and Grounded Summaries with LLMs}\label{ch:llm}

Across domains and tasks, Large Language Models demonstrate tremendous zero-shot performance. This is in large part due to the fact that large companies have decided to push the boundaries of pre-training scaling laws (more data, more parameters) \parencite{hoffmann2022training}. Large-scale multi-task fine-tuning \parencite{longpre2023flan}\footnote{``Instruction tuning''} and alignment to human preferences and values \parencite{stiennon2020learning} have furthered enabled powerful base LLMs to be general purpose (strong zero-shot capabilities across diverse user instructions) and to serve, for the most part, as ``helpful and harmless assistants'' \parencite{bai2022constitutional}.

The LLM era has changed the way summarization is viewed within the NLP community. Pre-2020, summarization was exclusively seen as a specialized task which required large-scale datasets \parencite{see-etal-2017-get, hermann2015teaching} (input document(s), reference summary pairs) to achieve good performance. This was true of smaller encoder-decoder models like BART \parencite{lewis2019bart} and PEGASUS \parencite{zhang2020pegasus}. Yet, recent work has found that API-based closed source models, such as Claude, GPT-3, and GPT-4, can generate higher quality summaries of news articles than smaller specialized models \parencite{goyal2022news}. While reference-based metrics like ROUGE naturally favor specialized models, humans unequivocally prefer zero-shot LLM summaries of news articles \parencite{zhang2023benchmarking}. The zero and few-shot dominance of LLMs has extended into the clinical domain, with zero-shot GPT-4 matching or exceeding the previous, highly tuned, state of the art in radiology report summarization \parencite{liu-etal-2023-exploring-boundaries}, summarization of doctor-patient conversations \parencite{ben-abacha-etal-2023-overview}, and medical reasoning benchmarks \parencite{nori2023capabilities, nori2023can}.

With respect to our hospital-course summarization task, we believe that specialization is still needed for two reasons. The first relates to model sizes and access. The most powerful zero-shot models at clinical tasks are closed, e.g., GPT-4 and Med-Palm-2 \parencite{singhal2023towards}. Due to HIPAA regulations, patient data from Columbia University Irving Medical Center (CUIMC), as well as other institutions, cannot be shared with Open AI or other companies without a BAA in place. These are expensive and difficult to procure. So, for practical reasons, smaller, open-source models \parencite{touvron2023llama, jiang2023mistral, wang2023far} will remain the main tool for advancing the state of the art in hospital-course summarization for the foreseeable future. While open-source models still lag behind, and some of their success derives from distilling, or imitating, better closed models \parencite{gudibande2023false}, there is evidence that these models are catching up \parencite{chen2023chatgpt}. Additionally, for ethical and practical purposes, this has benefits. Closed source models, such as GPT-4, are best viewed as products, not models. They are black boxes whose behavior is not reproducible and not fully understood. This will affect the ability to control LLMs and, given instability and uncertainty \parencite{chen2023chatgpt}, could inhibit approval of certain tools as medical devices. Pre-training datasets may cover biomedical and some clinical language but it does not cover \emph{longitudinal patient data}. Reasoning over hundreds of pages of clinical notes and tracking temporal shifts in the progression of conditions, as well as dealing with source note inconsistencies and massive redundancies, is \emph{likely}\footnote{We cannot know for certain.} something that is not an emergent capability. Rather, we suspect that some degree of exposure is necessary before inference. Furthermore, most instruction tuning datasets cover short prompts \parencite{zellers2019hellaswag, weifinetuned, wang2022super} and require short answers. Prior non-clinical work has demonstrated that long-range understanding requires additional fine-tuning \parencite{peng2023yarn}.

In this Chapter, we explore fine-tuning two state-of-the-art open-source LLMs (\textit{Mistral-7B-Instruct} and \textit{Zephyr-7B-$\beta$}) on a large-scale instruction tuning dataset of source note and reference summary pairs. After a single epoch of training (9,000 steps across 165k individual examples), these models can generate fluent summaries and achieve high ROUGE scores ($\sim$ 50 ROUGE-1 F1 scores). Yet, we discover that they frequently do not do a good job at covering the salient entities from the admission and they frequently hallucinate. To ensure faithful coverage of salient entities, we explore entity-guided summarization. As we discovered for news summarization \parencite{adams-etal-2023-generating}, content selection is best treated as its own explicit step, rather than performed implicitly during generation.

Based on our prior work in Chapter \ref{ch:dataset}, we have already demonstrated that hospital-course summaries are entity dense and their clinical usefulness depends on comprehensive coverage of salient entities. As such, we treat groups of synonymous entities--Entity Synonym Groups (ESGs)--as the unit of granularity for content selection. We devise and train a small, LongFormer-based, encoder-only classifier which predicts the salience of each ESG. When adding a pre-selected set of salient ESGs to the LLM prompt, we find that downstream summaries are more faithful and coverage a higher percentage of salient entities, where salience is defined by the entities present in reference summaries. To encourage the LLM to focus on specific mentions in the source notes, we propose SPEER: Sentence-level Planning via Embedded Entity Retrieval. Specifically, we mark each salient entity span with special ``\{\{ \}\}'' boundary tags and instruct the LLM to retrieve marked spans before generating each sentence. Sentence-level planning acts as a form of state tracking in that the model is explicitly recording the entities it uses. We fine-tune Mistral and Zephyr variants on a large-scale, diverse dataset of \textasciitilde 167k in-patient hospital admissions and evaluate on 3 datasets. SPEER shows gains in both coverage and faithfulness metrics over non-guided and guided baselines.

Our primary contributions in this chapter are:

\begin{itemize}
    \item Fine-tune state-of-the-art open source LLMs (\texttt{Mistral-7B-Instruct} and \texttt{Zephyr-7B-$\beta$}) on a large-scale dataset for long-form clinical summarization, and test on three diverse datasets from different EHRs.
    \item Demonstrate that content selection should be thought of as its own classification task, even in the world of LLMs. Dedicated content selection, performed by a small encoder-only classifier, outperforms implicit content selection from auto-regressive LLM decoding.
    \item Introduce an easy-to-implement method---SPEER---which improves the coverage of salience entities and faithfulness over both non-guided and guided LLM baselines.
\end{itemize}

\section{SPEER: Sentence-Level Planning for Embedded Entity Retrieval} \label{sec:speer}

\textit{This section describes a ArXiV pre-print \parencite{adams2024speer}.}

\begin{table}[ht]
\centering
\begin{tabular}{lcc}
\setlength{\tabcolsep}{1pt}
\textbf{\texttt{Statistic}} & \textbf{\texttt{Source}} & \textbf{\texttt{Reference}} \\ \hline \hline
\textbf{\texttt{Entity Spans}} & 1666 & 36 \\
\textbf{\texttt{Unique ESGs}} & 473 & - \\ \hline
\textbf{\texttt{\% Problems}} & 0.364 & 0.492 \\
\textbf{\texttt{\% Treatments}} & 0.268 & 0.339 \\
\textbf{\texttt{\% Tests}} & 0.368 & 0.169 \\
\end{tabular}
\caption{ Entity and ESG statistics across source notes and reference summaries. ``Entity Spans'' refers to the total number of raw entity mentions, ``Unique ESGs'' to the number of synonym groups formed from the raw mentions across the source notes. We also report the fractional breakdown of entities by semantic type. }\label{tab:esg-stats}
\end{table}

\begin{figure*}[ht]
\begin{center}
\centerline{\includegraphics[width=\linewidth]{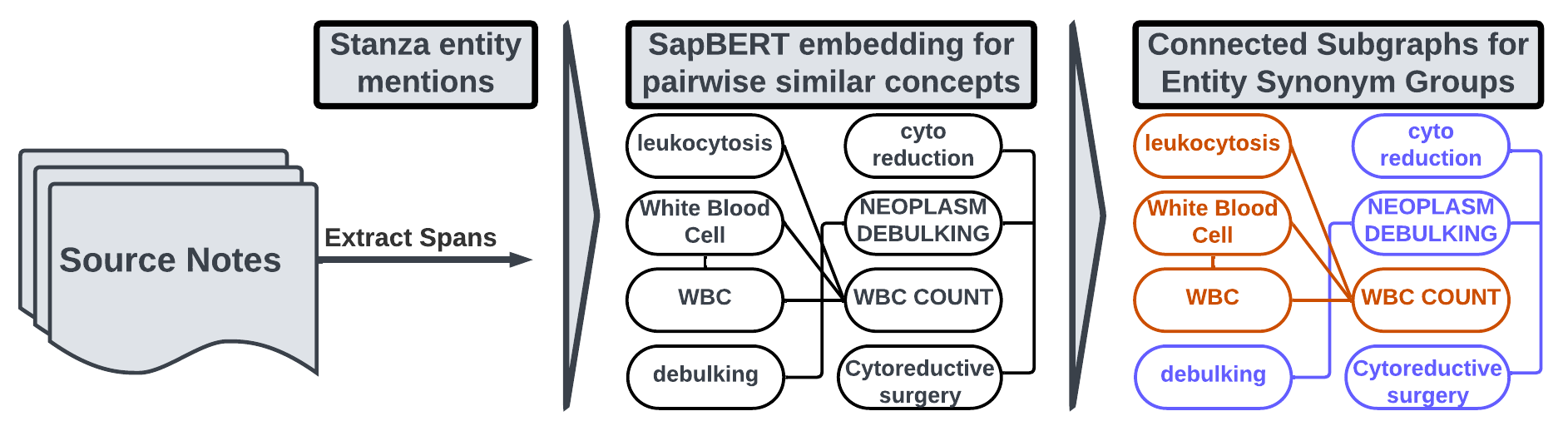}}
\caption{Extracting entities and forming groups of synonymous entities (ESGs). For each admission, we form a set of ESGs from the source notes and content selection is performed by classifying each ESG as salient or not. }
\label{fig:esg-formation}
\end{center}
\end{figure*}

\subsection{Selecting Salient Entities} \label{sec:esg}

The process of extracting entities, identifying synonym pairs, and then forming Entity Synonym Groups (ESGs) is graphically depicted in Figure \ref{fig:esg-formation}. In Table \ref{tab:esg-stats}, we plot entity-based statistics for source notes and reference summaries. We refer to both the Figure and Table in subsequent paragraphs.

\paragraph*{Extracting Entities.} We use Stanza \parencite{qi-etal-2020-stanza} for entity extraction. In particular, we use the clinical NER model \parencite{clin-stanza} which was trained on MIMIC-III notes \parencite{johnson2016mimic} for the i2b2-2010 clinical NER shared task \parencite{uzuner20112010}. The model extracts entity spans from three disjoint semantic types: \texttt{PROBLEM}, \texttt{TEST}, \texttt{TREATMENT}. Problems are diagnoses and symptoms, Tests cover lab tests and imaging, while Treatments span medications and procedures.

\paragraph*{Identifying Entity Synonym Pairs.} Clinicians frequently rely on acronyms and shorthand when documenting, which leads to large variance in how concepts are mentioned across notes \parencite{demner2016aspiring, adams2020zero}. We follow \parencite{adams2023meta} and use similarity in embedding space to identify synonymous clusters of entity spans. Specifically, we embed all mentions with SapBERT \parencite{liu-etal-2021-self}--which is trained to align synonymous clinical concepts--and use cosine similarity to identify all synonymous pairs. We manually assign binary labels (\texttt{unrelated}, \texttt{synonymous}) to 1,000 mention-pairs. We then select the threshold ($0.75$) for cosine similarity classification which maximizes the F1-score overlap with human labels. Figure \ref{fig:esg-formation} illustrates the need for semantic over lexical matching. Acronyms (\texttt{WBC} $\rightarrow$ \texttt{White Blood Cell}) have identical meanings and no lexical overlap. This also holds true for synonyms: \texttt{Cytoreductive surgery} and \texttt{NEOPLASM DEBULKING}, which both describe the resection of a tumorous growth.

\paragraph*{Forming Entity Synonym Groups (ESG).} For each hospital admission, we collect all entity mentions across the source notes and form a graph with one node for each unique entity mention. We assign an edge between two mentions \emph{iff} they are exact-match duplicates or have a pairwise SapBERT similarity of $\geq 0.75$. We then treat all fully-connected sub-graphs as Entity Synonym Groups (ESG). Entity content selection is then performed over ESGs, e.g., which represent a model a single medical concept, rather than over specific span mentions. As shown in Figure \ref{fig:esg-formation}, the process of forming ESGs by computing fully connected sub-graphs based on the pairwise similarity graph greatly reduces entity sparsity. Eight unique entity spans form two ESGs. The entities in the ESG all relate to the same precise topic--if not the exact same concept. For instance, \texttt{leukocytosis} is a condition characterized by a high white blood cell (WBC) count and thus is connected directly to \texttt{WBC COUNT} and, via the graph, to \texttt{WBC} and \texttt{White Blood Cell}.

\paragraph*{Defining ESG salience.} For each example (hospital admission), we extract the ESG from across the source notes. Based on the embedding similarity method for synonymity, given embedding similarity scores, we consider an ESG as ``salient'' if $\geq 1$ spans in the ESG is a synonym of $\geq 1$ entity span(s) extracted from the reference summary. Only $5.7\%$ of the source ESGs are ``salient'' by this definition, which underscores the difficulty of content selection for this task and dataset.

\paragraph*{Learning ESG salience.} We build a hierarchical token-to-ESG encoder model to perform binary classification over ESGs. This approach is inspired by previous hierarchical extractive methods from \textcite{liu-lapata-2019-text, bi-etal-2021-aredsum, adams-etal-2023-generating}. First, we demarcate each entity span with newly initialized \texttt{<e>} and \texttt{</e>} tokens. Then, we concatenate all source notes and encode tokens with a long-range encoder (LongFormer \parencite{beltagy2020longformer}), which can fit up to 16,384 tokens. We construct hidden-state representations of each entity span by mean-pooling the hidden states of each word-piece associated with the span (inclusive of \texttt{<e>} and \texttt{</e>} tokens). Next, we mean-pool the hidden states of all entity spans associated with the same ESG (e.g., ``Diabetes type 2", ``DM II'', ``DM2''). Then, we add an ESG modeling layer as a newly initialized, fully-connected BERT encoder layer. To exploit the fact that frequently mentioned concepts tend to be salient, we assign each ESG to a numerical range according to inverse frequency (most mentions first). We learn an embedding for relative frequency and add it to the ESG hidden state before passing through the modeling layer. A linear classification head is added to produce a single logit for each modeled ESG representation. We compute a logistic loss over each ESG logit.

\paragraph*{The Issue of Content Ordering.} It would be convenient if we could predict the proper \emph{ordering} of salient entities in the summary, rather than simply just salience. We attempted to learn a sentence prediction task over salient entities as part of the output--in a multi-task setup--yet we found it impossible to learn. There was no convergence. This is because human-written BHCs do not follow a consistent discourse structure and vary along different ordering strategies (e.g., temporal or problem-oriented). This supports our analysis in Chapter \ref{ch:dataset} that clinician-authors BHC summaries exhibit low lexical cohesion.

\paragraph*{ESG classification inference.} Concatenated source notes can sometimes exceed the LongFormer context window of $16,384$. During training, we simply truncate to $16,384$. Yet, during inference, to avoid information loss, we chunk notes into disjoint windows of at most $16,384$ and perform separate token-level encoding before concatenating hidden states\footnote{The concatenation of encoder hidden states is similar to Fusion-In-Decoder \parencite{izacard2020leveraging}.}. The ESG-modeling layer has a maximum length of $1024$ ESGs. For the rare case with $> 1024$ ESGs, we drop the ESGs whose term frequency of mentions across the source is lowest.

\begin{figure*}[t]
\begin{center}
\centerline{\includegraphics[width=\linewidth]{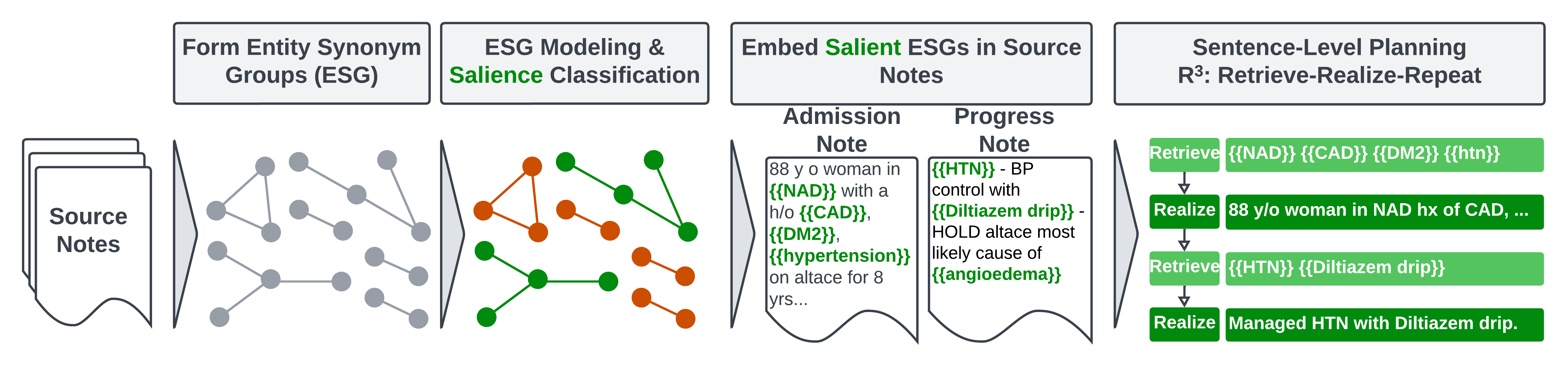}}
\caption{\textbf{\underline{\texttt{\emph{SPEER}}}}: \underline{\emph{S}}entence-Level \underline{\emph{P}}lanning via \underline{\emph{E}}mbedded \underline{\emph{E}}ntity \underline{\emph{R}}etrieval. The entire process of generating a hospital-course summary from a concatenated set of clinical notes is shown above. The first two steps relate to the formation and classification of Entity Synonym Groups (ESGs) from \S \ref{fig:esg-formation}. The next two steps visually describe the SPEER approach in \S \ref{sec:guided}. First, salient entity mentions are marked with special \textbf{\texttt{\{\{ \}\}}} boundary tags, which indicate that they are allowed to be retrieved during generation. Then, during generation, each summary sentence is generated on its own line. Above each sentence line, the model is instructed to first retrieve the entities it plans to use in the following sentence simply by generating entities within the \textbf{\texttt{\{\{ \}\}}} tags. This single-pass approach can best be captured with the acronym \textbf{$\bm R^3$: Retrieve-Realize-Repeat} because each sentence is a realization of a plan. }
\label{fig:speer-diagram}
\end{center}
\end{figure*}

\subsection{ESG-Guided Summarization} \label{sec:guided}

In this section, we explore methods for ESG-guided summarization: by which summary generation is conditioned on both the source documents and a pre-selected set of ESGs. We first present a simple prompt-based baseline before presenting our proposed approach, SPEER. As in \parencite{dou-etal-2021-gsum, adams-etal-2023-generating}, We train models on oracle provided ESGs as guidance while performing inference with the set of model-predicted ESGs (\S \ref{sec:esg}).

\paragraph*{Prompt Guidance.} A logical approach is to convert the set of salient ESGs into a natural language prompt and instruct the model to incorporate them into its summary. To compute the oracle prompt for training, we follow the same procedure to define the labels as in \S \ref{sec:esg}: ``Defining ESG Salience''. Then, we list out the salient ESGs by semantic type: ``PROBLEMS'', ``TREATMENTS'', and ``TESTS''. Each line contains each unique mention of an ESG across the source notes (delimited by ``;''). We randomly shuffle the order of the ESGs during training because, at inference time, the ESG classifier, as described in \S \ref{sec:esg}, produces an un-ordered set of entities. While the ESG classifier could theoretically learn the proper order, prior work on the hospital-course summarization task \parencite{adams-etal-2021-whats} has noted that clinical notes exhibit very low local coherence, which makes the ordering task noisy.




\paragraph*{SPEER.} The above approach is simple, intuitive, and helps ground the summaries onto a set of salient entities. Yet, two issues may arise. Firstly, the model may learn to focus more on the entities themselves and not their actual usage in the source notes. The source notes are lengthy and, as such, the ratio of relevant to irrelevant content is very high. The prompt guidance, however, is precise and only includes the entities which make it into the reference summary. Relatively speaking, the model may learn to over-rely on the list of entities themselves at the expense of their usage in the source notes, which can be difficult to identify. A consequence of this would be high coverage of salient entities on the surface, yet irrelevant, inconsequential, or erroneous context supplied for these entities. Secondly, the entity guidance is extensive--some reference summaries have 100+ unique ESGs. It would be difficult even for a clinician to keep track of which ESGs have been covered so far without a discrete state tracking mechanism. State tracking is necessary for the model to both determine which of the salient ESGs have yet to be covered, and, consequently, if all ESGs are covered, to break out.

To address both concerns: the lack of source note grounding and the lack of a discrete tracking mechanism, we propose \textbf{SPEER}: \textbf{S}entence-Level \textbf{P}lanning via \textbf{E}mbedded \textbf{E}ntity \textbf{R}etrieval. The \textbf{SPEER} process is shown in the last two steps of \ref{fig:speer-diagram}. To address the grounding concern, we first \textbf{E}mbed the salient \textbf{E}ntities in the source notes. To do this, we demarcate each entity span from a \textit{salient} ESG with \textbf{\texttt{\{\{ \}\}}} boundary tags. Before generating each summary sentence, the model generates a list of the entities it plans to use, in the order in which they should appear. We refer to this \textbf{S}entence-\textbf{L}evel planning step as ``\textbf{R}etrieval'' because the model is performing generative retrieval over a fixed set of embedded entities. To show the model that it must only use embedded entities to form its plan, the model is taught to generate the entities with their boundary tags \textbf{\texttt{\{\{ \}\}}}. As shown in Figure \ref{fig:speer-diagram}, this can be described as \textbf{$\bm R^3$: Retrieve-Realize-Repeat}. \textbf{$\bm R^3$} aims to address both the grounding and tracking concerns. Firstly, by retrieving the bracketed entities, we are encouraging the model to attend to--or focus on--a specific usage of the salient ESG in context. Secondly, the act of explicitly generating the entities to include in the next sentence makes it easier to keep track of which entities have already been included in the summary. The output template for this proposed $\bm R^3$ method of summarization is:

\begin{verbatim}
### Entities 1: {{span}} {{span}}
### Sentence 1: <sentence 1>
### Entities 2: {{span}} {{span}}
### Sentence 2: <sentence 2>
    ...
\end{verbatim}

\noindent The final summary is simply the concatenation of each line which begins with \texttt{\#\#\# Sentence}. We use oracle ESGs during training and those predicted by the ESG classifier at inference. For planning, during training, we extract the in-order entity mentions from each reference sentence and add them as spans to the corresponding \texttt{\#\#\# Entities} line.

\subsection{Experimental Setup} \label{sec:llm-setup}

\paragraph*{Coarse Filtering.} Source notes typically exceed the $8,192$ context window on which Mistral and Zephyr were trained. Additionally, clinical notes include many irrelevant sections, including administrative text, minutely detailed descriptions of surgical procedures, and patient disclosures. Such sections are typically easy to identify. To filter out irrelevant content and fit the maximum context window, we learn a coarse section filter. The inputs to the extraction model are individual sections (header and body), extracted from source notes with a custom toolkit based on Clarity NLP. A RoBERTA-classifier is trained with a logistic loss to predict the salience. The labels are continuous between 0 and 1 and represent the average of ROUGE-1 and ROUGE-2 F-1 scores between the reference summary and the section text. During inference, we preserve the original order of the text but remove sections--starting with the lowest scoring--until the total tokenizer token count is no greater than 8,192.

\begin{table*}[ht]
    \centering
    \footnotesize
    \setlength{\tabcolsep}{1pt}
    \begin{tabular}{ll|cc|cc|cc}
    \multirow{2}{*}{\texttt{\textbf{Dataset}}} & \multirow{2}{*}{\texttt{\textbf{Split}}}  & \multicolumn{2}{c}{\texttt{\textbf{Example-Level Stats}}} & \multicolumn{2}{c}{\texttt{\textbf{Source Stats}}} & \multicolumn{2}{c}{\texttt{\textbf{Reference Stats}}} \\
    & & \textbf{\# Admissions} & \textbf{Avg Length of Stay} & \textbf{\# Notes} & \textbf{\# Tokens} & \textbf{\# Sentences} & \textbf{\# Tokens} \\ \hline
    \texttt{\textbf{Columbia:2020-2023}} & \textbf{Train} & 167k & 6.3 days & 27.8 & 11k & 12.4 & 207.5 \\ \hline
    \texttt{\textbf{Columbia:2020-2023}} & \textbf{Test} & 1k & 5.6 days & 25.5 & 13k & 11.4 & 173.9 \\
    \texttt{\textbf{Columbia:2010-2014}} & \textbf{Test} & 1k & 5.2 days & 41.4 & 12k & 12.2 & 201.5 \\
    \texttt{\textbf{MIMIC}} & \textbf{Test} & 900 & 30.8 days & 162.7 & 44k & 37.0 & 542.9 \\
    \end{tabular}
    \caption{Statistics for data used for training and evaluating hospital-course summarization models. we use datasets from Columbia University Irving Medical (CUIMC) at two different points of time. We also report scores on MIMIC-III, despite MIMIC having a great deal of unsupported content in reference summaries \parencite{adams-etal-2022-learning}.}
    \label{tab:ehr-datasets}
\end{table*}

\begin{figure}[ht]
\begin{center}
\centerline{\includegraphics[width=0.6 \linewidth]{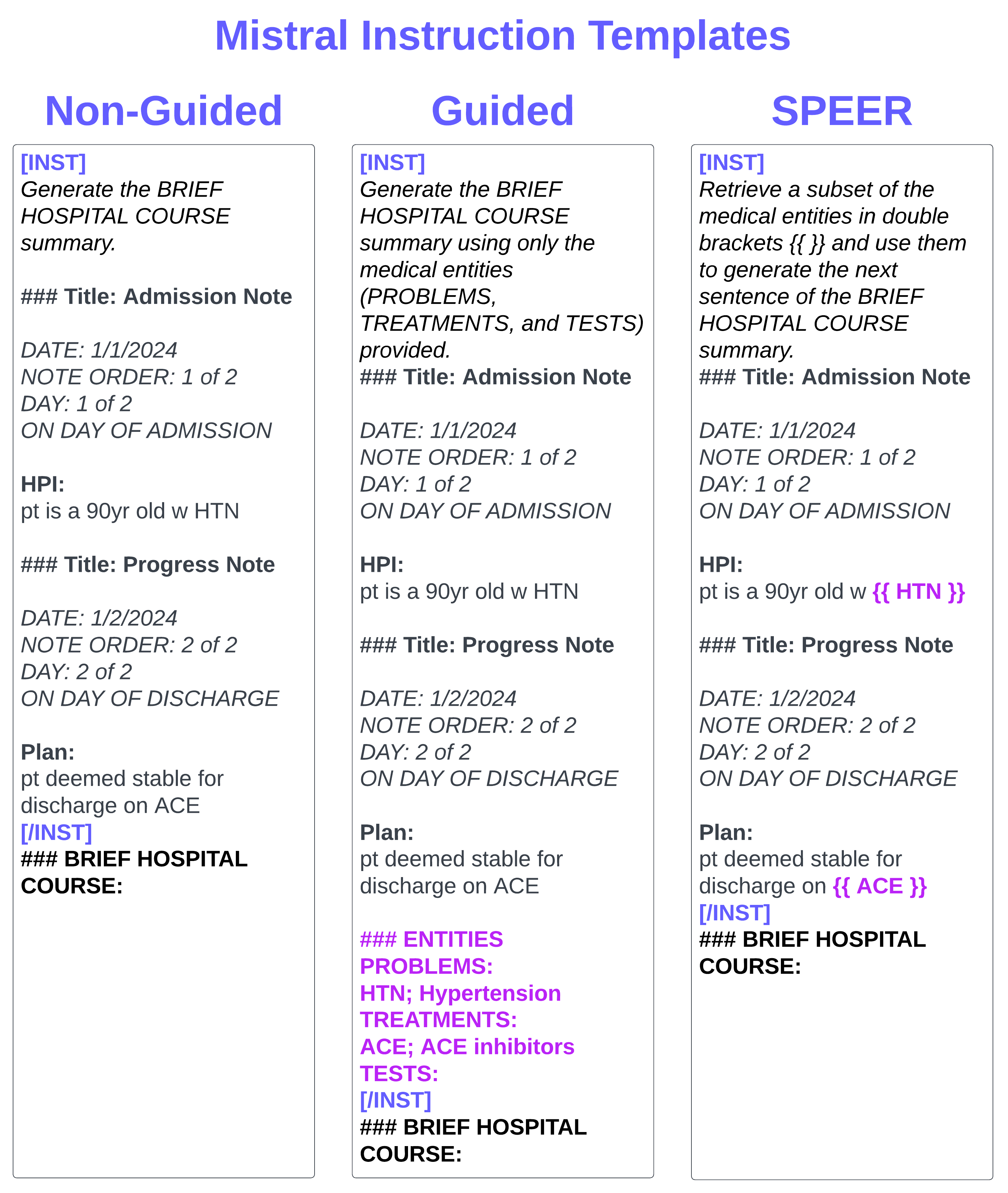}}
\caption{Instruction Template for Mistral, which follows the syntax used during the original instruction tuning for \texttt{Mistral-7B-v0.1}. }
\label{fig:mistral-template}
\end{center}
\end{figure}

\begin{figure}[ht]
\begin{center}
\centerline{\includegraphics[width=0.6 \linewidth]{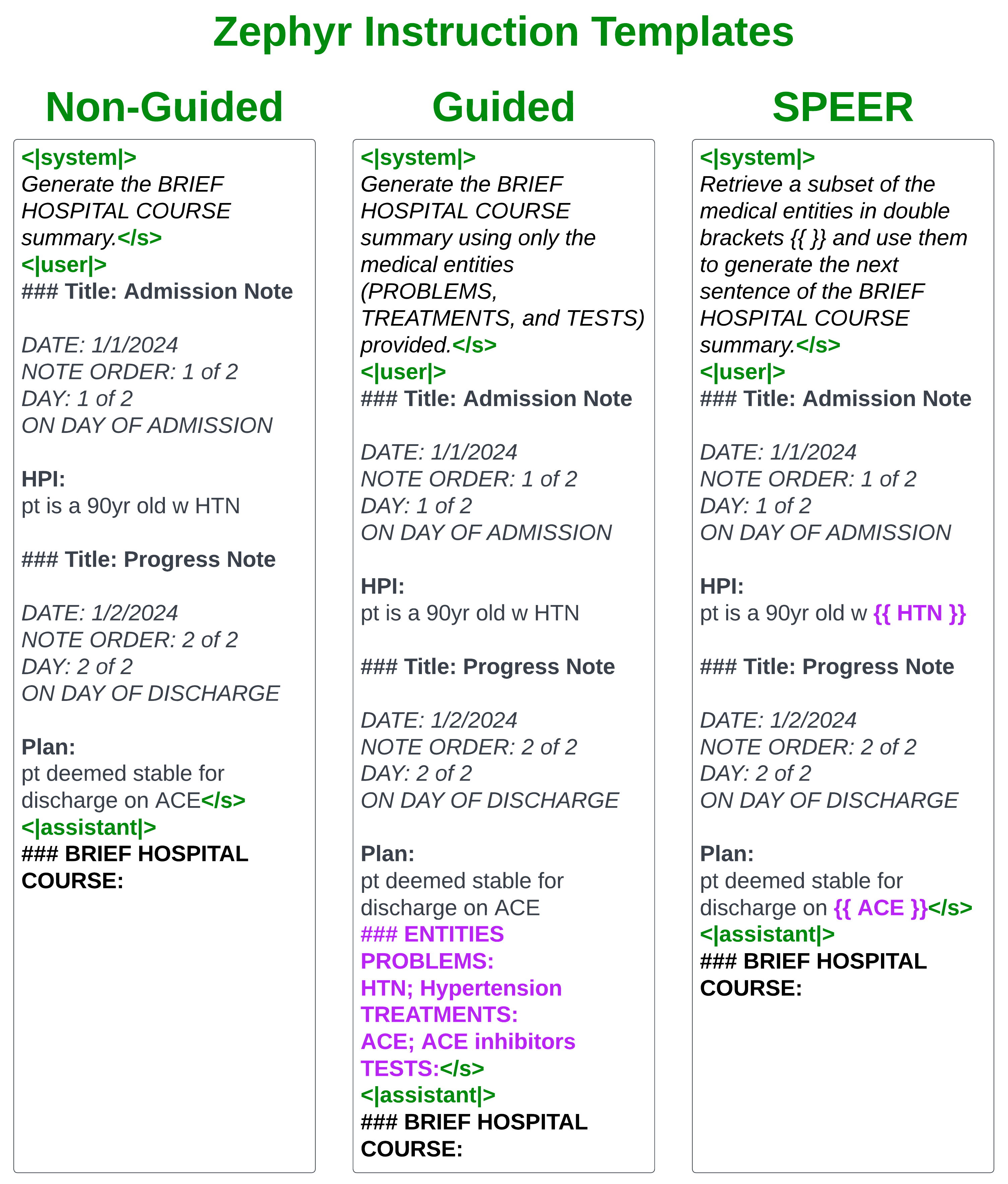}}
\caption{Instruction Template for Zephyr, which follows the syntax used during the original instruction tuning for \texttt{zephyr-7b-beta}. }
\label{fig:zephyr-template}
\end{center}
\end{figure}

\paragraph*{Instruction Templates.} For each model, source notes are simply concatenated chronologically. For each note, we generate a set of header lines which include the title of the note and the date of the note. We also explicitly specify where the note lands in relation to the rest of the admission, e.g., ``Day 1 of 4 (On Admission)''. Although we separately fine-tune each model, we still include custom instructions. The baseline non-guided instruction is: ``Generate the BRIEF HOSPITAL COURSE summary.'' The prompt guidance instruction is: ``Generate the BRIEF HOSPITAL COURSE summary using only the medical entities (PROBLEMS, TREATMENTS, and TESTS) provided.'' The guidance (list of ESGs grouped by semantic type) is appended to the source notes. SPEER's instruction is: ``Retrieve a subset of the medical entities in double brackets \textbf{\texttt{\{\{ \}\}}} and use them to generate the next sentence of the BRIEF HOSPITAL COURSE summary.'' The line: ``\#\#\# BRIEF HOSPITAL COURSE:$\backslash n$'' is appended to the end of the input and is an indicator to the model to start generating the summary. The full templates for Mistral and Zephyr are visualized in Figures \ref{fig:mistral-template} and \ref{fig:zephyr-template}.

\paragraph*{Datasets.} We train on a single dataset and evaluate on three diverse held-out sets. \textbf{Training.} We train on \textasciitilde 167k in-patient hospital admissions from a large metropolitan hospital (Columbia University Irving Medical Center (CUIMC)) from 2020-2023. It is highly diverse in terms of patient population and care setting: emergency, surgery, obstetrics, pediatrics, etc. \textbf{Testing.} We evaluate on a held out portion of 1,000 admissions in the same time-frame as the training set: \textbf{Columbia:2020-2023}, as well as admissions from an earlier time period: \textbf{Columbia:2010-2014}, in which the Electronic Health Record (EHR) system was different. As a result, note templates and titles may differ. By training on notes with one set of EHR templates and testing on both seen and unseen styles, we can test robustness of methods to subtle shifts in style and content organization. \textbf{MIMIC.} We also evaluate on a held-out set of 900 examples (source notes plus extracted reference summaries for evaluation) from publicly available MIMIC-III clinical notes with pre-processing from \textcite{adams-etal-2022-learning}. Table \ref{tab:ehr-datasets} shows high-level statistics for the train-test splits. In contrast to other commonly used clinical NLP benchmarks \parencite{chen2022toward, gao2023dr}, the hospital-course summarization stands out as longitudinal, multi-document, and lengthy. On average, the inputs contain from $25.5$ (Columbia) to $162.7$ (MIMIC) source notes, which are synthesized into lengthy (11-37 sentence) summaries. Identifying salient, non-redundant content amounts to finding many needles in many haystacks, and then de-duplicating to ensure that each needle is unique. MIMIC admissions, on average, contain a substantially higher number of source notes than Columbia, while also having very long reference summaries (37 sentences). Despite the long inputs, \textcite{adams-etal-2022-learning} reveal much of the content in MIMIC reference summaries is not mentioned anywhere in the source notes---due to incompleteness---which reduces scores on reference-based metrics.

\begin{figure}[h]
\begin{center}
\centerline{\includegraphics[width=0.5 \columnwidth]{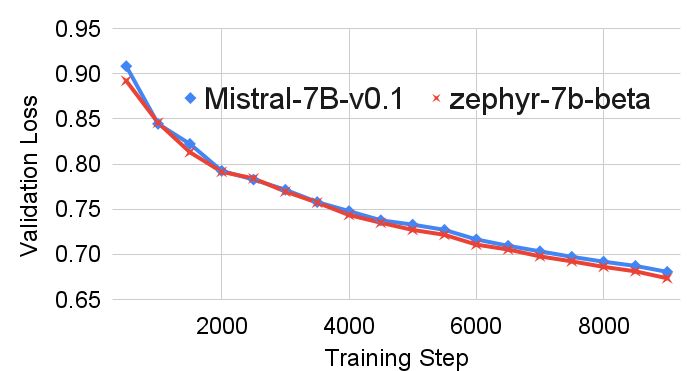}}
\caption{Validation Loss for \textbf{\texttt{Mistral-7B-v0.1}} and \textbf{\texttt{zephyr-7b-beta}} as a function of training steps across 1 epoch (covering \textasciitilde 167k hospital admissions). }
\label{fig:validation-loss}
\end{center}
\end{figure}

\paragraph*{Training Details.} \textbf{ESG Content Selection.}  We initialize the token-level encoder described in \S \ref{sec:esg} from a 279 million parameter encoder-only model: \textit{xlm-roberta-longformer-base-16384}, which is a Longformer initialized with weights from XLM-RoBERTa \parencite{conneau2019unsupervised} without any additional tuning. For modeling ESGs, we use a randomly initialize BERT encoder layer with the same configuration as XLM-RoBERTa. As described in \S \ref{sec:esg}, ESGs are sorted by inverse frequency and frequency rank embeddings are added to the representations of each ESG before passing to the modeling layer. We learn $1,024$ unique ranks and, if the source notes include more than $1,024$ ESGs, we truncate to $1,024$. We train on batches of 16 with AdamW optimizer for 100k using a scheduled learning rate (maximum $3e-5$ with linear warmup of $1000$ steps, followed by linear decay). Weight decay of $5e-5$ is used. The entity boundary tags (\texttt{<e>}, \texttt{</e>}) are added as special tokens to the tokenizer and the embeddings are updated during fine-tuning. \textbf{Abstractive Summarization.} We fine-tune in two stages. The first stage involves training the baseline non-guided model for 1 epoch. For the second stage, we further-finetune the baseline and entity-guided models for an extra 4,000 steps. We break up training into two stages to reduce total training time since the weights from first stage are re-used multiple times during the second stage. \textbf{\textit{Initial Fine-Tune.}} We fully fine-tune \texttt{Mistral-7B-Instruct} and \texttt{Zephyr-7B-$\beta$} using the baseline instructions for 1 epoch with a batch size of 16 and a learning rate of $5e-6$. We used the AdamW 8-bit optimizer with a cosine learning rate scheduler. To fit the model onto two Nvidia A6000 48GB GPUs, we use DeepSpeed Stage 2 \parencite{rasley2020deepspeed}, FlashAttention-2 \parencite{dao2023flashattention}, bfloat16 (BF16) precision, gradient checkpointing, and train on per device batches of size 1 with gradient accumulation. We used oracle section filtering to ensure that no training examples exceeded 8,192 tokens. 1 epoch took 10 days to complete. We computed a validation loss after every 500 steps of training and, in Figure \ref{fig:validation-loss}, show a smooth loss curve.  \textbf{\textit{Further Fine-Tune.}} We further fine-tune all variants: baseline and entity-guided models for an additional 4,000 steps from the initial fine-tuned weights. Performance plateaus after $2500$ additional fine-tuning steps. As such, we save checkpoints every 500 steps between steps $2500$ and $4000$ (inclusive), and report the average metric scores across this range of 4 checkpoints. We do this for robustness as there is considerable random variance for metrics across checkpoints.

\paragraph*{Generation Config.} We use greedy decoding and to mitigate the problem of repetition, set a repetition penalty hyper-parameter of $1.1$ \parencite{keskar2019ctrl}. We set the number of minimum tokens to be $4$ and maximum new tokens to be $1,024$ for the non-guided and prompted guided models. Since SPEER must generate sentence-level plans, we double the maximum new tokens from $1,024$ to $2,048$.

\paragraph*{Evaluation Metrics.} We rely on two entity-based overlap metrics: \emph{S}ource-\emph{G}rounded \emph{R}ecall (\textbf{SGR}) and the \emph{H}allucination \emph{R}ate (\textbf{HR}). Some concepts in clinical reference summaries are not present in source notes, as noted by \textcite{shing2021towards} and \textcite{adams-etal-2022-learning}. Unsupported--or hallucinated--reference content should not be included when computing entity overlap. As such, instead of directly computing overlap between reference and model-generated entities, we separately align summary entities to entities in the sources notes, and measure overlap between source-aligned entities. Specifically, we align reference and model-generated entities to a subset of ESGs from the source notes: $\{ESG_{ref \rightarrow src}\}$ and $\{ESG_{model \rightarrow src}\}$. Then we compute \emph{s}ource-\emph{g}rounded \emph{R}ecall (\textbf{SGR}) as:

$$
SGR = \frac{|\{ESG_{ref \Rightarrow src}\} \cap \{ESG_{model \Rightarrow src}\}|}{|\{ESG_{ref \Rightarrow src}\}|}
$$

\begin{table*}[ht]
    \centering
    \footnotesize
    \setlength{\tabcolsep}{1pt}
    \begin{tabular}{ll||cccccc|c||cccccc|c}

    & \multirow{3}{*}{\bfseries Model} & \multicolumn{7}{c}{\bfseries \texttt{\textcolor{cadmiumgreen}{MIMIC}}} & \multicolumn{7}{c}{\bfseries \texttt{\textcolor{cadet}{Average of Datasets}}} \\[5pt]
    & & \multicolumn{2}{c}{\bfseries Entity Overlap} & \multirow{2}{*}{\bfseries \makecell{BSP\\$\uparrow$}} & \multirow{2}{*}{\bfseries \makecell{Clin $\uparrow$\\Distill}} & \multicolumn{2}{c}{\bfseries ROUGE} & \multirow{2}{*}{\textit{ \makecell{\# of \\ Tokens}}} & \multicolumn{2}{c}{\bfseries Entity Overlap} 
    & \multirow{2}{*}{\bfseries \makecell{BSP\\$\uparrow$}} &  \multirow{2}{*}{\bfseries \makecell{Clin $\uparrow$\\Distill}} &  \multicolumn{2}{c}{\bfseries ROUGE} & \multirow{2}{*}{\textit{ \makecell{\# of \\ Tokens}}} \\
    & & \bfseries SGR $\uparrow$ & \bfseries HR $\downarrow$ & & & \bfseries R1 $\uparrow$ & \bfseries R2 $\uparrow$ & & \bfseries SGR $\uparrow$ & \bfseries HR $\downarrow$ & & & \bfseries R1 $\uparrow$ & \bfseries R2 $\uparrow$ &  \\ \hline

    \multirow{3}{*}{\bfseries \texttt{Mistral}} & \textbf{\texttt{Non-Guided}} & .230 & .116 & .664 & .971 & 24.3 & 6.7 & \textit{279} & .339 & .126 & .683 & .886 & 31.9 & 16.0 & \textit{197} \\
    & \textbf{\texttt{Guided}} & .236 & .171 & .648 & .541 & 23.5 & 6.2 & \textit{352} & .401 & .151	& .678 & .683 & \bfseries 33.9 &  \bfseries 18.1 & \textit{251} \\
    & \textbf{\texttt{SPEER}} & \bfseries .302 & \bfseries .040 & \bfseries .667 & \bfseries 1.240 & \bfseries 25.0 & \bfseries 7.0 & \textit{324} & \bfseries .430 & \bfseries .078 & \bfseries .686 & \bfseries .947 & 33.9 & 16.6 & \textit{234} \\ \hline

    \multirow{3}{*}{\bfseries \texttt{Zephyr}} & \textbf{\texttt{Non-Guided}} &  .245 & .121 & .653 & .899 & 25.0 & 6.8 & \textit{335} & .386 & .138 & .673 & .789 & 33.7 & \bfseries 16.4 & \textit{257}  \\
    & \textbf{\texttt{Guided}} & .247 & .136 & .651 & .593 & 24.0 & 6.3 & \textit{337} & .415 & .132 & .673 & .633 & 34.0 & 16.4 & \textit{267} \\
    & \textbf{\texttt{SPEER}} & \bfseries .306 & \bfseries .046 & \bfseries .662 & \bfseries 1.271 & \bfseries 25.9 & 7.1 & \textit{364} & \bfseries .439 & \bfseries .084 & \bfseries .682 & \bfseries .907 & \bfseries 34.1 & 16.2 & \textit{267} \\ \hline \hline & \\[-5pt]

    & \multirow{3}{*}{\bfseries Model} & \multicolumn{7}{c}{\bfseries \texttt{\textcolor{airforceblue}{Columbia: 2020-2023}}} & \multicolumn{7}{c}{\bfseries \texttt{\textcolor{amethyst}{Columbia: 2010-2014}}} \\[5pt]
    & & \multicolumn{2}{c}{\bfseries Entity Overlap} & \multirow{2}{*}{\bfseries \makecell{BSP\\$\uparrow$}} & \multirow{2}{*}{\bfseries \makecell{Clin $\uparrow$\\Distill}} & \multicolumn{2}{c}{\bfseries ROUGE} & \multirow{2}{*}{\textit{ \makecell{\# of \\ Tokens}}} & \multicolumn{2}{c}{\bfseries Entity Overlap} 
    & \multirow{2}{*}{\bfseries \makecell{BSP\\$\uparrow$}} &  \multirow{2}{*}{\bfseries \makecell{Clin $\uparrow$\\Distill}} &  \multicolumn{2}{c}{\bfseries ROUGE} & \multirow{2}{*}{\textit{ \makecell{\# of \\ Tokens}}} \\
    & & \bfseries SGR $\uparrow$ & \bfseries HR $\downarrow$ & & & \bfseries R1 $\uparrow$ & \bfseries R2 $\uparrow$ & & \bfseries SGR $\uparrow$ & \bfseries HR $\downarrow$ & & & \bfseries R1 $\uparrow$ & \bfseries R2 $\uparrow$ &  \\ \hline

    \multirow{3}{*}{\bfseries \texttt{Mistral}} & \textbf{\texttt{Non-Guided}} & .447 & .161 & .692 & .670 & 44.7 & 31.3 & \textit{117} & .341 & .099 & .695 & \bfseries 1.02 & 27.0 & 9.9 & \textit{195} \\
    & \textbf{\texttt{Guided}} & .568 &  .193 & .690 & .613 & \bfseries 49.5 & \bfseries 33.5 & \textit{180} & .399 & .091 & .696 & .903 & \bfseries 28.9 & \bfseries 14.8 & \textit{220} \\
    & \textbf{\texttt{SPEER}} & \bfseries .572 & \bfseries .117 & \bfseries .696 & \bfseries .741 & 48.4 & 32.7 & \textit{163} & \bfseries .417 & \bfseries .075 & \bfseries .696 & .872 & 28.4 & 10.0 & \textit{214} \\ \hline

    \multirow{3}{*}{\bfseries \texttt{Zephyr}} & \textbf{\texttt{Non-Guided}} & .516 & .176 & .682 & .570 & 48.1 & 32.8 & \textit{168} & .399 & .116 & .684 & \bfseries  .901 & 27.9 & 9.7 & \textit{269} \\
    & \textbf{\texttt{Guided}} & .582 & .152 & .684 & .554 & \bfseries 49.3 & \bfseries 33.1 & \textit{203} & .417 & .107 & .685 & .758 & \bfseries 28.7 & \bfseries 9.8 & \textit{260} \\
    & \textbf{\texttt{SPEER}} & \bfseries .588 & \bfseries .122 & \bfseries .692 & \bfseries .666 & 48.3 & 31.9 & \textit{188} &  \bfseries .424 & \bfseries  .084 & \bfseries .692 & .791 & 28.2 & 9.7 & \textit{249} \\ \hline

    \end{tabular}
    \caption{ Results from fine-tuning \texttt{Mistral-7B-Instruct-v1} and \texttt{Zephyr-7B-$\beta$} on \textbf{\texttt{Non-Guided}} and Entity-Guided (\textbf{\texttt{Guided}} and our proposed \textbf{\texttt{SPEER}}) hospital-course summarization. Metrics are defined in \S \ref{sec:llm-setup}. }
    \label{tab:llm-results}
\end{table*}

\noindent \textbf{SGR} does not explicitly capture entity-based faithfulness. For this, we define the hallucination rate (\textbf{HR}) as the fraction of model-generated entity mentions which do not have a source entity synonym:

$$
HR = \frac{|\{ENTITY_{model \nRightarrow src}\}}{|\{ENTITY_{model}\}|}
$$

\noindent where $|\{ENTITY_{model \nRightarrow src}\}|$ denotes the number of predicted entity mentions which do not have a corresponding source synonym. \textbf{HR} uses entity mentions and not ESGs (as in \textbf{SGR}) in order to penalize multiple hallucinations of the same synonym group. We report the number of tokens (\textbf{\# of Tokens}) to account for length biases in the metrics.

More broadly, we capture faithfulness at both the summary-level with BERTScore-Precision (\textbf{BSP}) \parencite{zhang2019bertscore} and, at the sentence level, with \textbf{ClinDistill} \parencite{adams2023meta}---a state of the art sentence-level faithfulness metric for hospital-course summarization. \textbf{ClinDistill} is a regression model which is distilled from an ensemble of several pre-existing state of the art faithfulness metrics. It predicts a raw, unnormalized score for each sentence, whose mean is roughly zero. We use BERTScore-Precision (\textbf{BSP}) \parencite{zhang2019bertscore}, which measures the degree to which summary tokens are well-aligned to at least one token in the source notes, rather than BERTScore-F1 because it was shown to correlate better to fine-grained expert annotations for the faithfulness of hospital course summaries \parencite{adams2023meta}. Specifically, we report the \textbf{BSP} between the model-generated summary and the source notes. We compute contextualized embeddings for each token with the encoder from \texttt{allenai/led-large-16384} and follow \textcite{adams2023meta} in using just the final hidden state as the representation for each token.

We also report \textbf{ROUGE-1} and \textbf{ROUGE-2} scores despite a known inverse relationship between ROUGE score and faithfulness for clinical summarization \parencite{adams-etal-2023-desired}. This negative correlation has to do with unsupported content in references \parencite{adams-etal-2022-learning}. Unfaithful models will mimic unsupported content in references, while faithful models tend to stick to what is explicitly stated and, as such, be penalized by ROUGE.

\subsection{Results} \label{sec:llm-results}

\begin{figure}[h]
\begin{center}
\centerline{\includegraphics[width=0.5 \columnwidth]{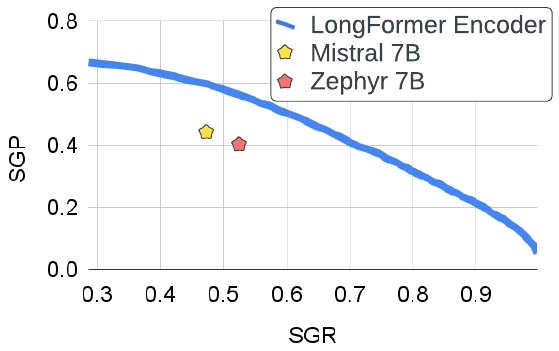}}
\caption{Comparing the entity-level performance (source-guided recall (SGR) and source-guided precision (SGP)) of \emph{explicit} content selection (classifying entities with a LongFormer Encoder) versus \emph{implicit} (autoregressive decoding with Mistral and Zephyr). }
\label{fig:precision-recall}
\end{center}
\end{figure}

\paragraph*{Implicit versus Explicit Content Selection.} One of the primary aims of this Chapter is to demonstrate that, even compared to fine-tuned LLMs, explicit content selection--using a dedicated content selector, is preferable to implicit content selection--performed during auto-regressive decoding of a summary. To demonstrate this, in Figure \ref{fig:precision-recall}, we vary the threshold for salience classification with the trained Longformer Encoder (from \S \ref{sec:esg}) and create a precision-recall curve, where recall is computed with SGR and precision with a similarly computed SGP(recision). On the same plot, we mark the SGP and SGR for fine-tuned \texttt{\textbf{Non-Guided}} Zephyr and Mistral models. Figure \ref{fig:precision-recall} demonstrates that Zephyr and Mistral point values fall well below the precision-recall curves of the classifier. The figure demonstrates that classification models can outperform auto-regressive models at hospital-course summary content selection, even when orders of magnitude smaller (279mn versus 7bn parameters).

\paragraph*{Models which rely on entity guidance achieve higher coverage of salient entities than those that do not.} As shown in Table \ref{tab:llm-results}, \texttt{\textbf{Guided}} and \texttt{\textbf{SPEER}}) have higher \textbf{SGR} (source-guided entity recall) fractions than \texttt{\textbf{Non-Guided}} across all dataset and base models. Looking at the average across datasets, \textbf{SGR} for models with guidance is $.401 / .430 $ for Mistral and $.415 / .439 $ for Zephyr. The \textbf{\texttt{Non-Guided}} model covers fewer salient entities: $.339$ and $.386$ \textbf{SGR} for Mistral and Zephyr, respectively. Summary length (\textbf{\# of tokens}) plays a role for Mistral ($197 < 251 / 234$) but less of a role for Zephyr ($257 < 267$). For Zephyr models on \textbf{Columbia:2010-2014}, \textbf{\texttt{Non-Guided}} produces the longest summaries ($269 > 260/249)$ while also covering fewer salient ESGs: \textbf{SGR} of $.399 < .417/424$. While length can be controlled, the proclivity toward longer, more complete summaries may stem from the fact that guided models are provided a clear stopping criteria: to stop only when all pre-selected entities are covered.

\begin{table*}[ht]
    \centering
    \footnotesize
    \setlength{\tabcolsep}{1pt}
    \begin{tabular}{lll||cccccc|c}
    & \multirow{3}{*}{\bfseries Model Name} & \multirow{3}{*}{\bfseries Change to Model} & \multicolumn{7}{c}{\bfseries \texttt{\textcolor{airforceblue}{Columbia: 2020-2023}}} \\
    & & & \multicolumn{2}{c}{\bfseries Entity Overlap} & \multirow{2}{*}{\bfseries \makecell{BSP\\$\uparrow$}} & \multirow{2}{*}{\bfseries \makecell{Clin $\uparrow$\\Distill}} & \multicolumn{2}{c}{\bfseries ROUGE} & \multirow{2}{*}{\textit{ \makecell{\# of \\ Tokens}}} \\
    & & & \bfseries SGR $\uparrow$ & \bfseries HR $\downarrow$ & & & \bfseries R1 $\uparrow$ & \bfseries R2 $\uparrow$ & \\ \hline
    
    \multirow{4}{*}{\bfseries \texttt{Zephyr}} & \textbf{\texttt{Non-Guided}} & - & .516 & .176 & .682 & -.430 & 48.1 & 32.8 & \textit{168}  \\
    & \textbf{\texttt{Guided}} & \textbf{\textcolor{Green}{+} Prompt Guidance} & .582 & .152 & .684 & -.446 & 49.3 & 33.1 & \textit{203} \\
    & \textbf{\texttt{Embedded}} & \textbf{Prompt \textcolor{Green}{$\rightarrow$} Embedded} & .574 & .147 & .688 & \bfseries -.327 & \bfseries 50.5 & \bfseries 34.7 & \textit{191} \\
    & \textbf{\texttt{SPEER}} & \textbf{\textcolor{Green}{+} Planning with Retrieval} & \bfseries .588  & \bfseries .122 & \bfseries .692 & -.334 & 48.3 & 31.9 & \textit{188} \\ \hline
    \end{tabular}
    \caption{ From \textbf{\texttt{Non-Guided}} to \textbf{\texttt{SPEER}}: a step-by-step transition with incremental improvements in faithfulness. }
    \label{tab:llm-ablation}
\end{table*}

\paragraph*{Prompt Guided is surprisingly less faithful than Non-Guided.} Across test sets and base models, prompt \texttt{\textbf{Guided}} summaries hallucinate more (higher \textbf{HR}) and score lower on faithfulness (\textbf{BSP}, \textbf{ClinDistill}) than \texttt{\textbf{Non-Guided}}. Looking at the average of datasets for Mistral, for example, \texttt{\textbf{Guided}} summaries have worse scores for \textbf{HR} / \textbf{BSP} / \textbf{ClinDistill} than \texttt{\textbf{Non-Guided}}: $.151 / .678 / -.317$ versus $ .126 / .683 / -.114 $, respectively. One would expect that instructing a model to stick to entities present in the source text would increase faithfulness. We suspect that \texttt{\textbf{Guided}} may learn to over-rely on the list of entities themselves at the expense of their usage in the source notes. The entity guidance, when not embedded onto the source notes, might be stealing attention away from the source notes themselves.

\paragraph*{SPEER improves \emph{both} coverage \emph{and} faithfulness.} While adding the guidance to the prompt (\texttt{\textbf{Guided}}) creates a faithfulness-coverage tradeoff, \texttt{\textbf{SPEER}} consistently improves on both fronts. When looking at the average across datasets, the coverage of salient entities (\textbf{SGR}) is the highest for \texttt{\textbf{SPEER}} for both Mistral and Zephyr: $.430 / .439$ versus $ .339 / .386 $ for \textbf{\texttt{Non-Guided}} and $ .401 / .415 $ for \textbf{\texttt{Guided}}. On faithfulness, \textbf{\texttt{SPEER}} hallucinates less: the average \textbf{HR} is $.078 / .084$ for Mistral / Zephyr versus $.126 / .132$ for \textbf{\texttt{Non-Guided}} and $.151 / .132$ for \textbf{\texttt{Guided}}. Additionally, BERTScore-Precision (\textbf{BSP}) and sentence-level average faithfulness (\textbf{ClinDistill}) are highest for \textbf{\texttt{SPEER}}. For \textbf{BSP}, \textbf{\texttt{SPEER}} Mistral and Zephyr score $.686 / .682$, more than $.683 / .673$ for \textbf{\texttt{Non-Guided}} and $.678 / .673$ for \textbf{\texttt{Guided}}. For \textbf{ClinDistill}, \textbf{\texttt{SPEER}} Mistral and Zephyr score $-.053 / -.093$, more than $-.114 / -.211$ for \textbf{\texttt{Non-Guided}} and  $-.317 / -.367$ for \textbf{\texttt{Guided}}\footnote{ClinDistill is a regression model trained on synthetic labels whose mean is 0. The model was trained on highly extractive summaries from a LongFormer model. This likely explains why we are observing many scores which are $< 0$.}.

\paragraph*{SPEER is more robust to unseen EHRs.} The model was trained on \textbf{\texttt{Columbia: 2020-2023}} data, so it is unsurprising that performance is best on a held-out set of admissions from the same date range. When switching datasets and EHRs, there is a noticeable performance drop across models, especially for \textbf{\texttt{MIMIC}}. As discussed in \textcite{adams-etal-2022-learning}, MIMIC-III notes are highly incomplete. As such, much of the reference content is not supported by the available source notes and reference-free metrics are understandably poor. Yet, it is notable that the largest advantage (coverage and recall) for \textbf{\texttt{SPEER}} comes from \textbf{\texttt{MIMIC}}, for which the data is the noisiest and the notes come from an unseen institution. \textbf{\texttt{SPEER}} might be more robust to this ``zero-shot'' setting because it requires the least effort on the part of the abstractive component. The LLM only needs to locate \textbf{\texttt{\{\{ \}\}}} tags, rather than needing to implicitly perform salience modeling (\textbf{\texttt{Non-Guided}}) or to link prompted guidance back onto specific parts of the source (\textbf{\texttt{Guided}}).

\paragraph*{SPEER Ablations.} Table \ref{tab:llm-ablation} demonstrates the mostly incremental improvements in faithfulness and coverage of salient entities as we transition from the baseline model (\textbf{\texttt{Non-Guided}}) to the fully loaded \texttt{\textbf{SPEER}} model. As discussed earlier, going from non-guided \textbf{\texttt{Non-Guided}}) to prompt \textbf{\texttt{Guided}}) increases coverage dramatically (\textbf{SGR} goes from $.516 \rightarrow .582$ while sentence-level faithfulness (\textbf{ClinDistill}) declines: $-.430 \rightarrow -.446$. If we replace prompt guidance with embedded guidance: \texttt{\textbf{Embedded}}, we achieve a slight decline in \textbf{SGR}: $.582 \rightarrow .574$ yet a decrease in hallucinations (\textbf{HR}): $.152 \rightarrow .147$ and an increase in sentence-level faithfulness: $-.446 \rightarrow -.327$. \texttt{\textbf{Embedded}} is \texttt{\textbf{SPEER}} without the sentence-level planning. The input is the same (notes with embedded salient ESGs) yet the target output is the summary without planning. We fine-tune \texttt{\textbf{Embedded}} using the procedure in \S \ref{sec:llm-setup} for \textbf{\textit{Further Fine-Tune}}. Adding in planning, we arrive at \textbf{\texttt{SPEER}}, which leads to an increase in coverage of salient entities: $.574 \rightarrow .588$ for \textbf{SGR} and a further decrease in hallucinations: $.147 \rightarrow .122$. We do note that ROUGE scores decline: ROUGE-1 from $50.5 \rightarrow 48.3$ but we believe that this is a necessary side effect of sentence-level planning, which encourages the model to stick to the entities in the source text and not hallucinate plausible, yet unsupported, content. Qualitatively, planning seems to cause a reduction in the number of sentences with no entities which occur in many reference summaries, yet do not contain important details. A paraphrased example is: ``Patient verbalized understanding of instructions and plans to follow up with his primary doctor in two weeks.'' These types of sentences often achieve high ROUGE scores as they are true for many patients, but more often than not, are never stated in the source notes. Including common, yet unsupported, content can artificially boost ROUGE at the expense of faithfulness and coverage.

\begin{table}[h]
    \centering
    \small
    \setlength{\tabcolsep}{2pt}
    \begin{tabular}{ll|ccc}
    & \multirow{3}{*}{\bfseries Model} & \multicolumn{3}{c}{\bfseries \texttt{\textcolor{airforceblue}{Columbia: 2020-2023}}} \\
    & & \multicolumn{3}{c}{\bfseries \textbf{Overlap w/ Guidance }} \\
    & & \textbf{Recall} & \textbf{Precision} & \textbf{F1} \\ \hline

    \multirow{3}{*}{\bfseries \texttt{Mistral}} & \textbf{\texttt{Non-Guided}} & .376 & .621 & .426 \\
    & \textbf{\texttt{Guided}} & .596 & .695 & .621 \\
    & \textbf{\texttt{SPEER}} & \bfseries .633 & \bfseries .749 & \bfseries .666 \\ \hline

    \multirow{4}{*}{\bfseries \texttt{Zephyr}} & \textbf{\texttt{Non-Guided}} & .443 & .564 & .462 \\
    & \textbf{\texttt{Guided}} & .620 & .681 & .629 \\
    & \textbf{\texttt{Embedded}} & .580 & .678 & .602 \\
    & \textbf{\texttt{SPEER}} & \bfseries .678 & \bfseries .745 & \bfseries .691 \\
    \end{tabular}
    \caption{ Model adherence to provided entity guidance. \textbf{\texttt{Embedded}} is an ablation of \texttt{\textbf{SPEER}} without sentence-level planning as described in the \textit{Ablations} paragraph.}
    \label{tab:adherence}
\end{table} 

\paragraph*{SPEER follows the instructions better than Guided Prompt.} We compute the adherence to the instructions--which are to write a summary with a given set of ESGs--in a similar way as we measure entity based overlap between model-generated and reference entities. Specifically, we extract entities from each generated summary and align them to a subset of the source ESGs. Then, we measure the overlap (recall, precision, F1) scores vis-a-vis the guidance itself (the set of ESGs predicted as salient by the ESG classifier from \S \ref{sec:esg}). Table \ref{tab:adherence} demonstrates that for both Mistral and Zephyr, \texttt{\textbf{SPEER}} adheres better to the provided guidance. \texttt{\textbf{SPEER}} Mistral and Zephyr F1 score is $.666 / .691$ versus $.621 / .629$ for \texttt{\textbf{Guided}}. Even though no guidance is given, we include \texttt{\textbf{Non-Guided}} to illustrate how different the entities \emph{explicitly} selected by the classifier is from the entities \emph{implicitly} chosen during summary generation. In other terms, auto-regressive \emph{implicit} content selection diverges from \emph{explicit} content selection. Taken together with the results in Table \ref{tab:llm-results}, we believe that content selection for long-form clinical summarization is best viewed as a separate task from realization, with its own set of models, architectures, and objectives.

\begin{figure}[h]
\begin{center}
\centerline{\includegraphics[width=\columnwidth]{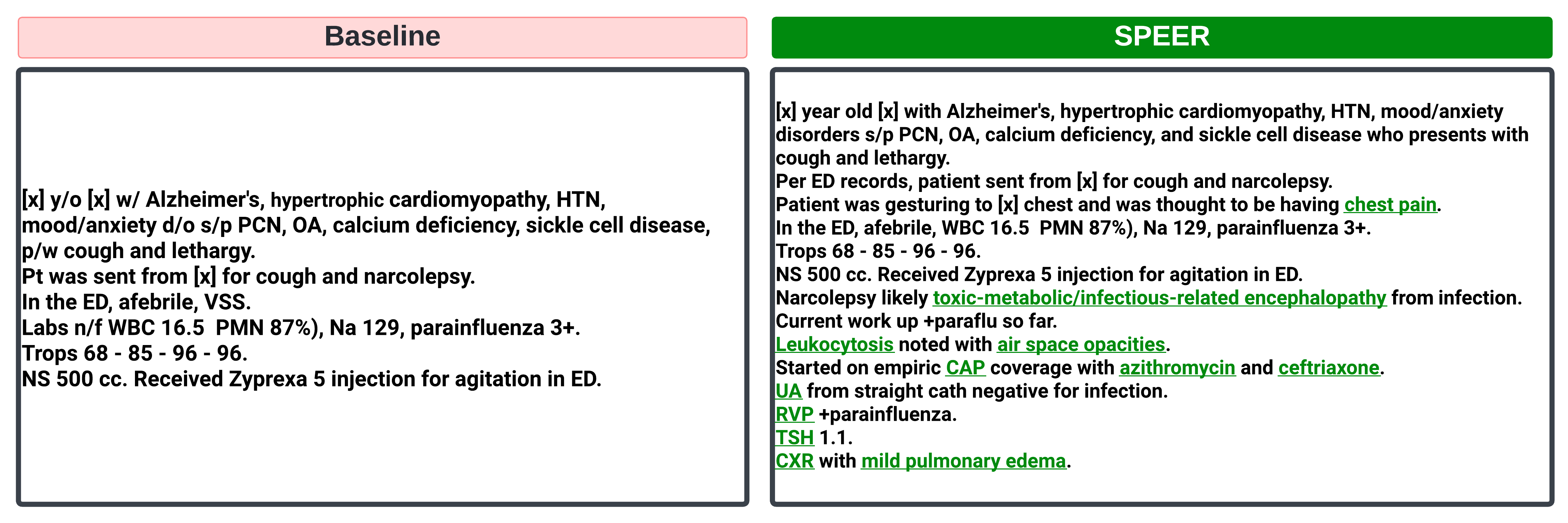}}
\caption{An example summary shown comparing the Baseline Non-Guided summary to a SPEER summary. The in-plan, additional salient entities covered by SPEER are colored in Green. Some disorders have been changed to related disorders and information masked in order to preserve patient anonymity. }
\label{fig:speer-example}
\end{center}
\end{figure}

\paragraph*{A Qualitative Example.}

Figure \ref{fig:speer-example} shows a side-by-side example of a baseline, non-Guided summary against a SPEER summary. Entities which were not included in Non-Guided but are in the reference summary, are highlighted in green in the summary. Adhering to a complete content plan improves the coverage of SPEER summaries.

\paragraph*{Comparison to Human Written Reference Summaries.}  We have demonstrated that human-written reference summaries often contain errors or, for various reasons, contain content that cannot be supported by the source notes. It is thus important to compare the faithfulness of SPEER summaries to those of human-written reference summaries. Using \texttt{Columbia: 2020-2023} test set as example we find that human written reference sumamries have \textbf{HR} and \textbf{ClinDistill} of $.218$ / $.226$.  The \textbf{HR} of $.218$ for human-written reference summaries is much higher (worse) than SPEER ($.117$ for Mistral and $.122$ for Zephyr). The \textbf{ClinDistill} faithfulness metric of $.226$ is much lower (worse) than SPEER ($.741$ for Mistral and $.666$ for Zephyr).

\section{Related Work} \label{llm-related-work}

\paragraph*{LLM Summarization.} Recent work has found that API-based closed source models, such as Claude, GPT-3, and GPT-4, can generate high quality summaries of news articles in the zero-shot setting \parencite{goyal2022news}. Humans prefer LLM-generated summaries from GPT-3 over summaries generated from the previous generation of smaller, fine-tuned models (e.g., BART, PEGASUS) \parencite{zhang2023benchmarking}. Human evaluation is critical to revealing the superiority of LLM-generated summaries given the limitations of reference-based metrics. In fact, \textcite{zhang2023benchmarking} find that annotators judge LLM-generated summaries on par with expert-level summaries carefully crafted by freelance writers. By iteratively fusing new entities into an existing summary draft, Chain-of-Density (CoD) \parencite{cod} enables LLMs, e.g., GPT-4, to generate entity-dense summaries, which are often favored by human annotators over earlier, less dense, drafts. Explicit planning can also be performed with Summary Chain-of-Thought (SumCoT), a technique that guides LLMs to generate summaries by focusing on core news elements in a step-by-step manner, leading to more coherent and comprehensive summaries \parencite{wang-etal-2023-element}. On the evaluation side, \textcite{chen2023evaluating} demonstrate that zero-shot prompted LLMs can outperform existing specialized classifiers on factuality detection.

\paragraph*{Guided Summarization.} Abstractive summarization requires three sequential tasks: content selection (extraction), content planning (organization), and surface realization (abstraction). With simple auto-regressive generation, the first two steps are generally performed implicitly with the last. Yet, prior work suggests that making content selection an explicit step, which is handled by a separate, dedicated model, can outperform the all-in-one approach \parencite{sharma-etal-2019-entity}. For instance, an extractive model can be used to enhance the performance of an abstractive model by treating the extract as an auxiliary input with its own encoder (GSum, \parencite{dou-etal-2021-gsum}). Other work simply prepends salient content to the source text as a form of control (CTRLsum \parencite{he-etal-2022-ctrlsum}). Relevant to our work, CTRLsum explores using entities as a form of control. The FROST model separates content selection and planning from realization with a single model, by having the model first generate an entity-based plan before generating the full abstractive summary \parencite{narayan-etal-2021-planning}. \textbf{\texttt{SPEER}}, on the other hand, interleaves planning and realization and relies on a separately trained classifier for content selection. \textcite{adams-etal-2023-generating} demonstrate that offloading content selection can be used to directly control the diversity of downstream summaries. Our PGA model can be used to enhance the performance of both small encoder-decoder (BART and PEGASUS) and large decoder-only (GPT-3.5) abstractors.

\paragraph*{Hospital-Course Summarization.} The multi-document summarization task of synthesizing the course of events during a patient's admission to a hospital is an area of active research \parencite{adams-etal-2021-whats}. Using a publicly available source of data (MIMIC-III), \textcite{adams-etal-2022-learning} tackle faithfulness by re-writing, or revising, reference summaries before training models on synthetic, grounded hospital-course reference summaries. Similarly to our work, \textcite{searle2023discharge} guide an abstractive model with clinical concepts to improve ROUGE scores of hospital-course summaries. They guide the model with all source concepts, while we perform content selection, or filtering, first. We embed the selected content back onto the source notes, whereas they guide BART with a separate encoder stream (as in GSum \parencite{dou-etal-2021-gsum}). Preliminary studies of LLMs into biomedical applications have largely focused on well-defined medical reasoning tasks \parencite{nori2023can, nori2023capabilities}, such as providing a differential diagnosis based on a patient record \parencite{mcduff2023towards}. Those that have explored clinical summarization have largely focused on single-document tasks of a finer temporal granularity \parencite{vanveen2023clinical}, such as generating the impressions section of a radiology report \parencite{liu2023tailoring, liu2023exploring, chuang2023spec, van-veen-etal-2023-radadapt}, ICD coding \parencite{boyle2023automated}, constructing a patient's problem list construction from a progress note \parencite{vanveen2023clinical}, and generating a note from a doctor-patient conversation \parencite{abacha2023overview, ionescu2023overview}.

\section{Conclusion and Limitations} \label{sec:llm-conclusion}

We are the first to explore fine-tuning LLMs (\texttt{Mistral-7B-Instruct} and \texttt{Zephyr-7B-$\beta$}) on the highly difficult, yet highly important, task of hospital-course summarization. We find that content selection--selecting which entities to include in the summary--is best performed by a dedicated salience classifier, which then guides the LLM in summary generation. We observe that simply appending the guidance to the prompt improves the coverage of salient entities yet harms faithfulness. To improve coverage while also improving faithfulness, we introduce \underline{\emph{\texttt{\textbf{SPEER}}}}: \\ \underline{\emph{S}}entence-Level \underline{\emph{P}}lanning via \underline{\emph{E}}mbedded \underline{\emph{E}}ntity \underline{\emph{R}}etrieval. By directly retrieving the entity guidance from the source notes, metrics suggest that \texttt{\textbf{SPEER}} summaries are more grounded and complete.

We present a few relevant takeaways from this Chapter:

\begin{itemize}
    \item Content selection for long-form clinical summarization should be treated as its own task, separate from planning and realization. It is highly challenging and, at smaller scale LLMs, it might require its own specialized models, architectures, and training objective.
    \item Simply adding guidance to the prompt is not enough and may be detrimental. It appears to distract the model from the actual source notes, which we observe in the form of reduced faithfulness metrics.
    \item Forcing the model to perform planning is similar to Chain-of-Thought reasoning \parencite{wei2022chain} in that it enables more computation before deciding on a final output. In our case, it also adds a degree of interpretability as we are retrieve explicit verbatim portions of the source notes as part of the generated output.
    \item We do not know if a prompting method like \texttt{\textbf{SPEER}} is equally as necessary for larger LLMs. It may be the case that larger models can faithfully fuse longitudinal source notes and a separated prompt content plan. This is a key limitation of this Chapter, a limitation that is shared with any work on LLMs with $< 70bn$ parameters.
    \item ROUGE is not the metric we should be targeting, yet we do not have any good reference-free methods for assessing the relevance of a summary. This is a problem although we believe that entity-based overlap (\textbf{SGR}) is a better alternative.
\end{itemize}

One last key limitation of this Chapter is that we rely solely on automatic metrics and do not perform a human assessment. As demonstrated in Chapter \ref{ch:measuring} and \textcite{adams2023meta}, evaluating hospital-course summaries for faithfulness is incredibly time-consuming as it requires looking over many clinical notes (often $> 100$) for each summary. It is difficult to recruit enough clinicians to achieve the scale necessary for meaningful model comparisons. Making the annotation process more efficient should be a key area of future focus for hospital-course summarization.

%% file: 7-conclusion.tex
\begin{center}
\pagebreak
\vspace*{5\baselineskip}
\textbf{\large Conclusion and limitations}
\end{center}

In this thesis, we investigate automating a complex, real-world clinical summarization task: hospital course-summarization.

In \textbf{Chapter \ref{ch:dataset}}, we lay the groundwork for future chapters by building and analyzing a large-scale dataset of hospital-course summaries from the Electronic Health Record at Columbia University Irving Medical Center (CUIMC) \parencite{adams-etal-2021-whats}. We are the first to build such a dataset and the first to analyze the properties of the task and contrast it to more commonly used benchmark summarization datasets in the general domain. We found that hospital-course summaries appear extractive according to widely used metrics. Yet, there is large variance within summaries. This mix of extractive content selection with substantial re-writing led us to explore hybrid extract-then-abstract solutions in \textbf{Chapter \ref{ch:llm}}. Hospital-course summaries are packed with medical entities, which are well-distributed across the source notes. As such, relations are often not explicit. This affects both modeling and evaluation.  On the evaluation side, in Chapter \ref{ch:measuring}, we perform fine-grained human annotation on the entity level. We find that a learned metric trained on actual clinical summaries outperforms those trained on general domain data, which is less entity dense in nature. From a modeling perspective in \textbf{Chapter \ref{ch:improving}}, we find that reference summaries include entities and relations which are not present in the source text. We propose a solution which involves first grounding reference summaries on the actual entities and relations present in the source notes. Hospital course summarization involves not only massive compression, but a large style and organization transfer. Source notes are written chronologically yet the way clinicians digest the information, and write the discharge summary, is largely problem-oriented. Also, lexical cohesion is sub-optimal for evaluating hospital-course discourse because clinical summaries naturally exhibit frequent, abrupt topic shifts. This means that predicting order is more difficult than predicting salience. We use this finding in \textbf{Chapter \ref{ch:llm}} by having our entity planner perform content selection, not content planning. Discharge summaries and their associated BHC sections are frequently missing critical information or contain excessive or erroneous content. We leverage this in \textbf{Chapter \ref{ch:improving}} by demonstrating that faithfulness can be drastically improved by training on edited, or revised, references. We take it a step further in \textcite{adams-etal-2023-desired} by showing that calibrating models directly to automatic metrics and thereby by-passing the standard teacher forcing on reference summary paradigm.

\section{Takeaways}

\paragraph*{Each Step in the LLM Pipeline is Important for Hospital-Course Summarization.} Hospital-course summarization is an incredibly challenging task and we believe the best efforts to automate it will, and should, involve a strong general base model, supervised fine-tuning, and alignment for clinical safety. \textbf{Strong Base Model.} In \textbf{Chapter \ref{ch:llm}}, we demonstrate that state of the art 7B models can achieve very high ROUGE scores on the task. More broadly, it is a well-established fact that on most tasks, including clinical \parencite{nori2023capabilities, nori2023can}, scale matters. \textbf{Supervised Fine-Tuning.} The hospital-course task is highly idiosyncratic in nature (\textbf{Chapter \ref{ch:dataset}}) and requires a kind of understanding: longitudinal reasoning over patient narratives, that is not typically called upon during pre-training. Thus, as shown in \textbf{Chapter \ref{ch:llm}}, we see consistent improvements (in validation loss) across an epic of supervised fine-tuning of strong chat models. Yet, as shown in \textbf{Chapter \ref{ch:dataset}} and \textbf{Chapter \ref{ch:improving}}, clinican-authored hospital-course summaries are flawed, inconsistent, and contain many errors (\textbf{Chapter \ref{ch:measuring}}). As such, models trained on clinician-authored reference summaries will hallucinate which can have harmful impacts on patient care. To mitigate this, some form of alignment, i.e., calibration (\textbf{Chapter \ref{ch:improving}}), is both unavoidable and critical. We demonstrated an offline reinforcement-learning solution that involved forming large summary candidates of variable quality and demonstrated that simple, stable objectives can help improve the faithfulness of downstream summaries. These findings should work similarly well when using state of the art alignment methods such as DPO.

\paragraph*{There is still a role for smaller models.} In \textbf{Chapter \ref{ch:llm}}, we showed that a small model (LongFormer) can outperform larger decoder models at entity-based content selection. This is a fact and finding that we also found holds true for news summarization \parencite{adams-etal-2023-generating}. More generally speaking, we have shown that for a multi-document longitudinal task as challenging as ours, content selection is best treated as its own separate problem. Content selection--when performed extractively--can be performed with classification models rather than implicitly as a function of auto-regressive decoding. Importantly, we are not saying that smaller models are better than LLMs at content selection (which could be similarly trained with classification heads), but that LLM-level performance can be surpassed with smaller models whose sole purpose is to identify salient and non-salient content across the sources notes.

\paragraph*{Evaluation, Not Modeling, is the Bottleneck for Advancing the SOTA in Hospital Course Summarization Research.} Open source models are powerful and easy to fine-tune with HuggingFace and other libraries. Given an existing EHR dataset, it is trivial to perform large-scale instruction tuning. We also have the tools (DPO, PPO, clinical calibration (ours)) to perform alignment. The issue is the lack of reliable metrics, and large-scale annotation, for the task. In \textbf{Chapter \ref{ch:measuring}}, we demonstrated that existing evaluation metrics over-rely on the level of copy-and-paste used for summaries, which represents a coarse, naive understanding of clinical factuality and safety. Ideally, we could turn solely to humans to provide the preference annotations necessary for training a reward model (PPO) or direct preference optimization (DPO). Yet, as showed in \textbf{Chapter \ref{ch:measuring}}, collecting annotations is incredibly time-consuming and challenging because 1) experts are needed; 2) experts (clinicians) are burnt out and strapped for time; 3) even with the time-saving measures we implemented in \textbf{Chapter \ref{ch:measuring}}, it takes about 30 minutes to thoroughly fact check a long hospital-course summary against the source notes for the admission. We demonstrated in our work ``Learning to Revise References for Faithful Summarization'' that there is even a role for metrics in dataset construction: even before supervised fine-tuning and alignment.

\paragraph*{Open-Source Datasets are not Suitable for Hospital-Course Summarization research.} Due to HIPAA regulations, health records cannot be publicly shared. The only large-scale longitudinal clinical notes database which can be shared is MIMIC (III and IV), which has been de-identified. It is the de facto dataset used for most clinical summarization tasks. Yet, we demonstrate unequivocally in \textbf{Chapter \ref{ch:improving}} that it is not a good source of data for longitudinal summarization tasks like ours due to its high degree of incompleteness. Self-contained single document tasks, such as summarizing a patient's problems from a single progress note, or generating the Impressions section of a single radiology report from the Findings section, can be explored using MIMIC. The absence of any open-source datasets for the task is a huge impediment to open, safe, and reproducible research in the field. As we discuss in \ref{sec:future}, synthetically created data from powerful base models should occupy a large portion of future research into clinical NLP tasks, especially for ones not suited for MIMIC. Synthetic data is having a moment and will continue to grow in importance as a way to transfer knowledge from powerful commercial products, such as Open AI's GPT-4 and Anthropic's Claude\footnote{\url{https://www.interconnects.ai/p/llm-synthetic-data}}.

\section{Limitations} \label{sec:limitations}

While our work makes several contributions to the nascent NLP field of hospital-course summarization, we note there are several critical limitations which deserve equal levels of attention.

\paragraph*{We do not experiment with \emph{large} LLMs.} There still remains a sizable gap in performance between large LLMs (\texttt{70B+}) and the smaller LLMs we explore in \textbf{Chapter \ref{ch:llm}} (\texttt{7B}). Scale is critical driving factor in improvements in language models over the last few years. Larger models have been shown to have emergent properties \parencite{wei2022emergent}, which correspond to skills obtained indirectly, e.g., without any training. Hence, in the LLM era, research which leverages smaller LLMs might not hold relevant insights for larger LLMs, which might possess more emergent capabilities than smaller models. Relevant to our work in \textbf{Chapter \ref{ch:llm}}, we demonstrate that 7B Mistral and Zephyr instruction-tuned models are outperformed in terms of content selection by a much smaller encoder-only classification model (LongFormer with ~400mn parameters).

Yet, a larger model (70B or GPT-4) might be able to seamlessly select, plan, and realize long clinical summaries all in one step with simple tweaks to prompting. We use smaller models solely due to limitations with respect to GPUs. As LLM training and inference is democratized more and more due to technologies like Flash Attention and vllm, larger models will be more feasible to train and run on academic budgets. A Business Associate Agreement (BAA) is a pre-requisite for being able to use closed-source API-based models on identifiable patient records. This is unfortunate given the strong performance of GPT-4 and GPT-sized models (MedPalm) on clinical tasks with little to no clinical expertise required (minimal prompting). Even with long context windows, the ability of these models to reason and select content distributed throughout the source notes is a fascinating and under-explored area of research.

With respect to retrieval, for instances, there is evidence of a "Lost in the Middle" problem where early and late positions have a disproportionately large influence on generated outputs \parencite{liu2023lost}. This might be haphazardly advantageous for our task since a patient admission is rich with history (what brought patient to hospital) and the end tends to cover the events throughout the admission (end plus repeated information from middle). But, the presence and potentially negative impact of this bias warrants a deep dive and thorough sweep of methods for extending context windows.

\paragraph*{Not all of our results are reproducible.} While all the work in Chapter \ref{ch:improving} is performed on open source datasets and is thus fully reproducible, much of our work is carried out on proprietary, closed dataset from Columbia University Irving Medical Center (CUIMC). This largely makes the impact of our work limited to what is described in papers which we have presented at conferences (*CL, MLHC, ML4H). This is largely unavoidable and should be a major focus of research going forward -- creating an open-source, fully de-identified hospital course summarization dataset.

\paragraph*{We focus on a single clinical summarization task.} There are other naturally occurring summarization tasks related to clinical workflows which we could evaluate the methods described in this thesis on, yet we focus on hospital-course summarization.  With strong zero-shot capabilities and instruction following with prompting, it would be interesting to explore how much instruction data is necessary to train a general-purpose clinical summarization model. We do view, however, hospital-course summarization as a great bell weather for generalist progress in clinical summarization task. It requires longitudinal reasoning, a keen understanding of what is clinically relevant, and an ability to structure and organize outputs according to clinical semantics.

\paragraph*{Human evaluations are small-scale.} In \textbf{Chapters \ref{ch:improving} and \ref{ch:measuring}}, we conduct 3 separate small human annotations ($<100$ summaries). On the plus side, this underscores how difficult the task is and how impactful an automated system which required little to no human validation would be on clinical practices. On the negative side, our results are not statistically significant with respect to human judgments. Our thesis findings must be asterisked, then, with the fact that we report strong automatic metrics yet, with respect to human judgments, we can only show positive trends.

\section{Future Work} \label{sec:future}

While mentioned briefly in the above Limitations section, we want to delve more deeply into the areas in which we believe are most promising for future research onto hospital course summarization.

\paragraph*{Synthetic Data.} Synthetic data has the potential to directly address several of the limitations discussed above around transparency, reproducibility, and scalability. In fact, in this thesis, we successfully demonstrated two applications in which synthetic data could improve the faithfulness of models (in Chapter \ref{ch:improving}) and how we can use synthetic data (faithfulness labels) to train a faithfulness detection model in Chapter \ref{ch:measuring}. Future work will likely involve extending the use of synthetic data for domain adaptation, instruction tuning, and alignment (RLHF $\rightarrow$ RLAIF). We are currently severely limited by being unable to use GPT-4 and powerful closed source models on protected clinical notes. To by-pass this, we can use these models to generate synthetic data and train open-source smaller models on this data. Student-teacher distillation is a popular approach in training open source LLMs, such as Orca and Orca-2 \parencite{mukherjee2023orca,mitra2023orca}. We believe that high-quality instruction tuning data for the task can be created by oversampling closed-source LLMs (GPT-4, Claude) with clinician-crafted prompts and then filtering out bad samples. This over-generate-then-reject pipeline was popularized with Self-Instruct \parencite{wang2022self}. The key bottlenecks will be data diversity and input lengths. Clinical data is diverse and without careful prompt engineering, synthetic data will likely not cover the long-tail of patient cases and histories.  Regarding input lengths, it will likely be too expensive to produce longitudinal patient records with closed-source LLMs.

Yet, work has shown that models can be trained on longer contexts with few to no examples (``train short, test long'' \parencite{press2021train}). The right curriculum distribution for hospital-course summarization should be determined beforehand. This will involve fine-tuning on a curriculum of short inputs (oracle-filtered to fit context windows of $\leq 2048$ or $\leq 4096$ tokens) before transitioning to training on fewer, longer inputs. Improving training token sample efficiency is a worthy research agenda on its own, but even more-so if the goal is to enable the scalable creation of synthetic data. On the alignment side, we've demonstrated that human annotation of patient records is not scalable. We argue for devoting a large portion of future research into figuring out how to improve the efficiency of human annotation on the task. The principle reason for this is to provide more statistical support for meta-evaluation efforts.  We have no reliable mechanism currently for determining just how good LLMs are at evaluating longitudinal patient records. As such, we cannot realistically expect better alignment from using RLAIF (Reinforcement Learning from AI Feedback). With strong, multi-dimensional evaluation metrics, RLAIF efforts can then be scaled up.

\paragraph*{Open-Sourced Dataset.} We need a better alternative to MIMIC-III for hospital-course summarization. This is related to the above paragraph on Synthetic Data. The best mechanism for advancing the state of the art in any field is to make high-quality datasets available to researchers around the world. Without this, progress will be fragmented and difficult to track. A PhD student should take on a thesis topic revolved around creating realistic, clinician-verified synthetic data.  This can be used initially as a test set for zero-shot evaluation of LLMs. Then, it can be scaled by distilling smaller models on it with sample-efficient PEFT methods, before sampling at scale.

\paragraph*{Diverse Instruction Tuning.} Above, we discussed how token inefficient training is for hospital-course summarization. Source notes span 10,000 or more tokens. Although we perform oracle filtering at the section level to fit tokens into the available context window of the LLM ($8192$ for Mistral and Zephyr), we are still maxing out the available context window for each example. Yet, much of the context is irrelevant.  While work is needed to explore even more aggressive oracle filter, we also believe that identification of relevant sub-tasks is an interesting area of research. Hospital-course summarization involves cross-note modeling to determine salience, inconsistencies, chronologies, etc. Sub-tasks might include: update summarization, tracking diagnoses over time, identification of source note conflicts, and determining order of events \parencite{agrawal2022leveraging}.

\textbf{Update Summarization.} For each note, this task involves updating an existing summary by inserting information from the note which is both novel and important. This is what clinicians do already, as they will often include updates to the hospital-course summary as part of their note taking. Synthetic data can be created for this task by aligning portions of the reference summaries to different notes.

\textbf{Problem Tracing.} Conditions, particularly chronic conditions, evolve over time. Synthesizing and tracking inflection points in the patient record can be seen as problem-specific form of a hospital-course summary. Each input would consist of sentences which contain a specific entity and the outputs would involve portions of reference summaries which relate to that entity. This would be substantially  more sample efficient and makes sense if we view entity content selection as a separate task to be performed before generation.

\textbf{Identification of Source Note Errors and Inconsistencies.} Clinicians make mistakes and since notes are often taken by different providers, there may be inconsistencies or simply uncertainties regarding differential diagnoses. Explicitly noting such inaccuracies or inconsistencies is an important implicit sub-task of the hospital-course summarization task and is not explicitly instructed during fine-tuning. Identification of such errors will likely require grouping text by topic and assessing inter-sentence entailment. A human evaluation will be necessary for meta-evaluating appropriate metrics.

\textbf{Chronology.} The timestamp of notes do not always correspond to the true timeline of events given the frequent use of copy-and-paste and history taking. Forming timelines is an interesting problem in and of itself and one that is generally not explicitly learned during next-word pre-training.

\paragraph*{Do not Assume Source Notes are Ground Truth.} This is a point that comes up when discussing automating hospital-course summaries with clinicians, who understand that errors and inconsistencies exist. This might be best viewed as a pre-processing step and its own separate task in which an LLM examines the source notes and flags, and possible resolves, errors and inconsistencies. Most realistically, this process will involve a human-in-the-loop who is responsible for addressing LLM-raised concerns about the source notes. The process of summarization is not simply to synthesize but to critically evaluate the source notes, and then synthesize. Right now, summarization research is largely concerned with blind-faith synthesis, yet this is not realistic for many settings where the data itself is highly flawed and propagating errors in input texts is potentially dangerous.

\begin{flushleft}
\hspace{10mm}
\end{flushleft}
